\let\mypdfximage\pdfximage
\def\pdfximage{\immediate\mypdfximage}

\documentclass[journal,onecolumn,10pt]{IEEEtran1}

\makeatletter
\def\markboth#1#2{\def\leftmark{\@IEEEcompsoconly{\sffamily}\MakeUppercase{\protect#1}}%
\def\rightmark{\@IEEEcompsoconly{\sffamily}\MakeUppercase{\protect#2}}}
\makeatother

\let\labelindent\relax

\usepackage[bigfiles]{pdfbase}
\usepackage{xparse}
\ExplSyntaxOn
\NewDocumentCommand\embedvideo{smm}{
  \group_begin:
  \leavevmode
  \tl_if_exist:cTF{file_\file_mdfive_hash:n{#3}}{
    \tl_set_eq:Nc\video{file_\file_mdfive_hash:n{#3}}
  }{
    \IfFileExists{#3}{}{\GenericError{}{File~`#3'~not~found}{}{}}
    \pbs_pdfobj:nnn{}{fstream}{{}{#3}}
    \pbs_pdfobj:nnn{}{dict}{
      /Type/Filespec/F~(#3)/UF~(#3)
      /EF~<</F~\pbs_pdflastobj:>>
    }
    \tl_set:Nx\video{\pbs_pdflastobj:}
    \tl_gset_eq:cN{file_\file_mdfive_hash:n{#3}}\video
  }
  \pbs_pdfobj:nnn{}{dict}{
    /Type/RichMediaInstance/Subtype/Video
    /Asset~\video
    /Params~<</FlashVars (
      source=#3&
      skin=SkinOverAllNoFullNoCaption.swf&
      skinAutoHide=true&
      skinBackgroundColor=0x5F5F5F&
      skinBackgroundAlpha=0.75
    )>>
  }
  \pbs_pdfobj:nnn{}{dict}{
    /Type/RichMediaConfiguration/Subtype/Video
    /Instances~[\pbs_pdflastobj:]
  }
  \pbs_pdfobj:nnn{}{dict}{
    /Type/RichMediaContent
    /Assets~<<
      /Names~[(#3)~\video]
    >>
    /Configurations~[\pbs_pdflastobj:]
  }
  \tl_set:Nx\rmcontent{\pbs_pdflastobj:}
  \pbs_pdfobj:nnn{}{dict}{
    /Activation~<<
      /Condition/\IfBooleanTF{#1}{PV}{XA}
      /Presentation~<</Style/Embedded>>
    >>
    /Deactivation~<</Condition/PI>>
  }
  \hbox_set:Nn\l_tmpa_box{#2}
  \tl_set:Nx\l_box_wd_tl{\dim_use:N\box_wd:N\l_tmpa_box}
  \tl_set:Nx\l_box_ht_tl{\dim_use:N\box_ht:N\l_tmpa_box}
  \tl_set:Nx\l_box_dp_tl{\dim_use:N\box_dp:N\l_tmpa_box}
  \pbs_pdfxform:nnnnn{1}{1}{}{}{\l_tmpa_box}
  \pbs_pdfannot:nnnn{\l_box_wd_tl}{\l_box_ht_tl}{\l_box_dp_tl}{
    /Subtype/RichMedia
    /BS~<</W~0/S/S>>
    /Contents~(embedded~video~file:#3)
    /NM~(rma:#3)
    /AP~<</N~\pbs_pdflastxform:>>
    /RichMediaSettings~\pbs_pdflastobj:
    /RichMediaContent~\rmcontent
  }
  \phantom{#2}
  \group_end:
}
\ExplSyntaxOff

\usepackage[T1]{fontenc}
\usepackage{tgtermes}
\usepackage{marvosym}
\usepackage{longtable}
\usepackage{tikz}
\usetikzlibrary{positioning,calc,arrows}
\usepackage{graphics}
\usepackage{graphicx}
\usepackage{caption}
\usepackage[english]{babel}
\usepackage{subfig}
\usepackage{epsfig}
\usepackage{color}
\usepackage{multirow}
\usepackage{mathrsfs}
\usepackage{textcomp}
\usepackage{amsfonts}
\usepackage{amsmath}
\usepackage{amssymb}
\usepackage{nccmath}
\usepackage[numbers,square,sort&compress,comma]{natbib}
\usepackage{chapterbib}
\usepackage[ruled,vlined,linesnumbered]{algorithm2e}
\usepackage{setspace}
\usepackage{verbatim}
\usepackage{wrapfig}
\usepackage{dblfloatfix}
\usepackage{comment}
\usepackage[strict]{changepage}
\usepackage{ragged2e}
\usepackage{microtype}
\usepackage{enumitem}
\usepackage{nccmath}
\usepackage{hyperref}
\usepackage{doi}
\usepackage{afterpage}
\usepackage{multirow,bigdelim}
\usepackage{booktabs}
\usepackage{color}
\usepackage[outline]{contour}
\usepackage{xcolor}
\usepackage[nameinlink,noabbrev]{cleveref}
\usepackage{adjustbox}
\usepackage{colortbl}%

\usepackage{multicol}
\usepackage{bibunits}
\usepackage[skins]{tcolorbox}

\def\imagetop#1{\vtop{\null\hbox{#1}}}

\setlist{parsep=0pt,listparindent=\parindent}

{\fontsize{9.75}{10}\selectfont \defaultbibliographystyle{Sledge-TNNLS-2020-1col-1} \defaultbibliographystyle{IEEEtran}}

\DeclareCaptionFont{singlespacing}{\setstretch{1}}
\numberwithin{figure}{section}
\renewcommand{\thefigure}{\arabic{section}.\arabic{figure}}

\renewcommand \thesection{\arabic{section}}
\numberwithin{equation}{section}

\hypersetup{bookmarks=false,bookmarksopen=false,pdfpagemode=UseNone,pdfstartview=}

\let\oldnl\nl
\newcommand{\nonl}{\renewcommand{\nl}{\let\nl\oldnl}}

\tikzset{add/.style n args={4}{
    minimum width=6mm,
    path picture={
        \draw[black] 
            (path picture bounding box.south east) -- (path picture bounding box.north west)
            (path picture bounding box.south west) -- (path picture bounding box.north east);
        \node at ($(path picture bounding box.south)+(0,0.13)$)     {\tiny #1};
        \node at ($(path picture bounding box.west)+(0.13,0)$)      {\tiny #2};
        \node at ($(path picture bounding box.north)+(0,-0.13)$)    {\tiny #3};
        \node at ($(path picture bounding box.east)+(-0.13,0)$)     {\tiny #4};
        }
    }
}

\title{\singlespacing\sf\huge Adapting the Exploration Rate for Value-of-Information-Based Reinforcement Learning}
\author{Isaac J. Sledge, \emph{Member, IEEE}\; and\; Jos\'{e} C. Pr\'{i}ncipe, \emph{Life Fellow, IEEE}%
\thanks{\fontdimen2\font=1.55pt\selectfont Isaac J. Sledge is the Senior Machine Learning Scientist with the Advanced Signal Processing and Automated Target Recognition Branch, Naval Surface Warfare Center, Panama City, FL, USA (email: isaac.j.sledge.civ@us.navy.mil).  He is also the Principal Machine Learning Scientist with the Machine Intelligence Defense (MIND) lab at the Naval Sea Systems Command.}
\thanks{\fontdimen2\font=1.55pt Jos\'{e} C. Pr\'{i}ncipe is the Don D. and Ruth S. Eckis Chair and Distinguished Professor with both the Department of Electrical and Computer Engineering and the Department of Biomedical Engineering, University of Florida, Gainesville, FL 32611, USA (email: principe@cnel.ufl.edu).  He is the director of the Computational NeuroEngineering Laboratory (CNEL) at the University of Florida.\vspace{0.1cm}}
\thanks{This work was funded by grants N00014-19-WX-00636 (Marc Steinberg), N00014-21-WX-00525 (Thomas McKenna), and N00014-21-WX-01348 (Marc Steinberg) from the US Office of Naval Research.  The first author was additionally supported by in-house laboratory independent research grant N00014-19-WX-00687 (Frank Crosby) from the US Office of Naval Research and a Naval Innovation in Science and Engineering grant from the US Naval Sea Systems Command.}%
}
\begin{document}
\bstctlcite{IEEEexample:BSTcontrol}

\RaggedRight\parindent=1.5em
\fontdimen2\font=2.1pt\selectfont

\begin{bibunit}
\bstctlcite{IEEEexample:BSTcontrol}

\maketitle%\doublespacing
\RaggedRight\parindent=1.5em
\fontdimen2\font=2.1pt
\vspace{-1.55cm}\begin{abstract}\normalsize\singlespacing
\vspace{-0.25cm}{\small{\sf{\textbf{Abstract}}}}---In this paper, we consider the problem of adjusting the exploration rate when using value-of-information-based exploration.  We do this by converting the value-of-information optimization into a problem of finding equilibria of a flow for a changing exploration rate.  We then develop an efficient path-following scheme for converging to these equilibria and hence uncovering optimal action-selection policies.  Under this scheme, the exploration rate is automatically adapted according to the agent's experiences.  Global convergence is theoretically assured.

We first evaluate our exploration-rate adaptation on the Nintendo GameBoy games \emph{Centipede} and \emph{Millipede}.  We demonstrate aspects of the search process, like that it yields a hierarchy of state abstractions.  We also show that our approach returns better policies in fewer episodes than conventional search strategies relying on heuristic, annealing-based exploration-rate adjustments.  We then illustrate that these trends hold for deep, value-of-information-based agents that learn to play ten simple games and over forty more complicated games for the Nintendo GameBoy system.  Performance either near or well above the level of human play is observed.
\end{abstract}%
\begin{IEEEkeywords}\normalsize\singlespacing
\vspace{-1.35cm}{{\small{\sf{\textbf{Index Terms}}}}---Value of information, rate-distortion theory, exploration, exploration rate, exploration-exploitation dilemma, reinforcement learning, information theory}
\end{IEEEkeywords}
\IEEEpeerreviewmaketitle
\allowdisplaybreaks
\singlespacing

%%%%%%%%%%%%%%%%%%%%%%%%%%%%%%%%%%%%%%%%%%%%%%%%%%%%%%%%%%%%%%%%%%%%%%%%%%%
%%%%%%%%%%%%%%%%%%%%%%%%%%%%%%%%%%%%%%%%%%%%%%%%%%%%%%%%%%%%%%%%%%%%%%%%%%%
\vspace{-0.4cm}\subsection*{\small{\sf{\textbf{1.$\;\;\;$Introduction}}}}\addtocounter{section}{1}

During reinforcement learning, two opposing objectives should be balanced \cite{SuttonRS-book1998a}, environment exploration and experience exploitation.  The fundamental trade-off between the two demands efficient search capabilities \cite{ThrunSB-coll1992b}.  A variety of such schemes have been proposed over the years.  Kaelbling et al. \cite{KaelblingLP-jour1996a} survey classical techniques.  More recent advances are discussed by Taylor and Stone \cite{TaylorME-jour2009a}, Garc\'{i}a and Fern\'{a}ndez \cite{GarciaJ-jour2015a}, among others. 

A shortcoming of many of these approaches is that they often do not directly quantify the effects of exploring a certain amount on obtainable reinforcements.  In \cite{SledgeIJ-jour2017a,SledgeIJ-jour2017b,SledgeIJ-jour2017c}, we provided a series of information-theoretic criteria with this functionality.  These criteria are based on Stratonovich's value of information \cite{StratonovichRL-book1975a}.  They describe the best obtainable benefit for a given state-action information rate and hence exploration amount.  They additionally permit optimal decision-making under uncertainty in a way that non-linearly generalizes utility theory \cite{vonNeumannJ-book2007a}.  Related approaches based on rate-distortion theory seek to implement similar functionality (see \hyperref[sec2]{Section 2}).

Optimizing the value of information yields a weighted-random exploration scheme for reinforcement learning.  The amount of exploration is driven by a single parameter that codifies the information bound amount.  The parameter's influence on performance is application dependent, so choosing good values is crucial.  In \cite{SledgeIJ-jour2017b}, we empirically analyzed the parameter's effect on both the state abstractions that formed and the riskiness of the agent's action-selection process \cite{BelavkinRV-coll2014a}.  We proposed a deterministic annealing schedule for online parameter updating.  This approach relies on prior knowledge of the environment to set the annealing rate.  In \cite{SledgeIJ-jour2017b,SledgeIJ-jour2017c}, we furnished an adaptive annealing schedule.  It is based on the action-selection policy cross-entropy, which is a bounded measure of how much the policy is being modified across episodes in response to the agent's experiences.  The update process relies on pre-specified cross-entropy thresholds.  Tuning these thresholds can be difficult for new environments.  Existing parameter updates for other exploration schemes either possess similar issues or are not suitable for complex environments (see \hyperref[sec2]{Section 2}).

We have yet to give a principled scheme for adjusting the value-of-information's exploration-rate parameter that adapts to the environment dynamics and provably converges to optimal policies.

Here, we address this shortcoming.  We analyze properties of the value-of-information's Lagrangian to determine when equilibria of an associated gradient flow occur for changing parameter values (see \hyperref[sec3]{Section 3}).  These equilibria correspond to optimal policies for a given exploration rate and the current set of agent experiences.  We then develop second-order path-following techniques to iteratively uncover equilibria for a changing exploration rate (see \hyperref[sec4]{Section 4} and \hyperref[secA]{Appendix A}).  A benefit of using path following is that the exploration-rate adjustment is discerned automatically from local properties of the gradient flow and hence what the agent has currently learned about the environment.  This avoids potentially poor empirical convergence rates that may be witnessed for deterministic-annealing parameter schedules.  It also ensures that an existing solution is mapped to a neighborhood around the next equilibrium, which facilitates quick convergence to good agent behaviors.  Another benefit is that there is little human involvement in the learning process.  Only a single hyperparameter, which controls the overall solution accuracy, must be set.  We specify a non-heuristic process for automatically choosing it. 

We evaluate the behavior of this path-following procedure on the arcade games \emph{Centipede} and \emph{Millipede}, where discrete state-action spaces are used (see \hyperref[sec5]{Section 5} and \hyperref[secB]{Appendix B}).  For these games, we illustrate how a value-of-information-based search with a deterministic parameter annealing schedule investigates the domain.  We then quantify the search improvements when utilizing parameter path-following and our path-following approach.  We also show that pseudo-arc-length path-following yields meaningful state abstractions.  Additionally, we highlight the disadvantages of conventional search heuristics for large-scale state-action spaces.  Neither epsilon-greedy nor soft-max-based selection can explore the policy space as well as the value of information with path-following.   This occurs regardless of whether deterministic or variable annealing schedules are used.  We demonstrate these trends hold for deep, curious agents that learn to play ten simple Nintendo GameBoy games, like \emph{Defender}, \emph{Joust}, \emph{Galaga} and \emph{Galaxian}, along with over forty complicated games for this system, like \emph{Super\! Mario\! Land}, \emph{Double\! Dragon}, \emph{Castlevania}, and \emph{Street\! Fighter\! 2} (see \hyperref[secC]{Appendix C}).  We consider continuous state and discrete action spaces for these environments.

\phantomsection\label{sec2}
%%%%%%%%%%%%%%%%%%%%%%%%%%%%%%%%%%%%%%%%%%%%%%%%%%%%%%%%%%%%%%%%%%%%%%%%%%%
%%%%%%%%%%%%%%%%%%%%%%%%%%%%%%%%%%%%%%%%%%%%%%%%%%%%%%%%%%%%%%%%%%%%%%%%%%%
\subsection*{\small{\sf{\textbf{2.$\;\;\;$Literature Review}}}}\addtocounter{section}{1}

%%%%%%%%%%%%%%%%%%%%%%%%%%%%%%%%%%%%%%%%%%%%%%%%%%%%%%%%%%%%%%%%%%%%%%%%%%%
%%%%%%%%%%%%%%%%%%%%%%%%%%%%%%%%%%%%%%%%%%%%%%%%%%%%%%%%%%%%%%%%%%%%%%%%%%%
\subsection*{\small{\sf{\textbf{2.1.$\;\;\;$Rate-Distortion-Based Exploration}}}}

There have been a few successful uses of rate-distortion-like concepts for determining what and how agents should learn.  In \cite{SledgeIJ-jour2017a}, we considered the first application of this theory for single-state, multi-action Markov decision processes, which are referred to as multi-armed bandits.  We demonstrated that optimal regret bounds could be obtained for an expectation-maximization-style, Blahut-Arimoto algorithm \cite{ArimotoS-jour1972a,BlahutR-jour1972a} when using an augmented exploration factor.  Our bounds guarantee that the optimal actions, specified by the value of information, will be chosen earlier and more consistently than many other optimal-regret exploration strategies.

In \cite{ArumugamD-conf2021a}, Arumugam and Van Roy showed that, like us, rate-distortion formulations of the multi-armed bandit problem could be addressed via expectation-maximization approaches.  They augmented these approaches with an information-directed-sampling scheme \cite{RussoD-coll2014a,RussoD-jour2016a} to perform probability matching \cite{ThompsonWR-jour1933a}.  Other authors have similarly relied on information-directed sampling \cite{DongS-coll2018a} for multi-armed bandits \cite{KirschnerJ-conf2018a,KirschnerJ-conf2020a,HaoB-coll2021a}.  However, it is known that probability matching yields sub-optimal action exploration \cite{BubekS-conf2019a}.  In a later work \cite{ArumugamD-coll2021a}, Arumugam and Van Roy partly addressed this concern by modifying their sampling process so that the fundamental information ratio characterizes both expected regret and information gain \cite{LuX-coll2019a}.  Note that, in \cite{ArumugamD-conf2021a,ArumugamD-coll2021a}, the authors specify regret bounds that depend on a history of chosen agent actions.  It is thus not easy to compare directly with our bounds in \cite{SledgeIJ-jour2017a}.  Blahut-Arimoto-type algorithms will likely not yield optimal regret bounds, though, without modifications like those in \cite{SledgeIJ-jour2017a}.

% Rubin, Shamir, and Tishby, "Trading value and information in MDPs"
In \cite{SledgeIJ-jour2017b,SledgeIJ-jour2017c}, we considered an early application of rate-distortion theory to multi-state, multi-action Markov decision processes.  As in the single-state case, we proposed Blahut-Arimoto-type updates for the agent action selection probabilities.  We coupled these updates with model-free reinforcement learning methods and showed that agents could learn policies for relatively simple environments.  Similar ideas were considered by Rubin, Shamir, and Tishby \cite{TishbyN-coll2012a}.  Several contributions \cite{AlemiAA-conf2017a,SaxeAM-conf2018a,WuT-conf2020a} related to Tishby's information bottleneck \cite{TishbyN-coll2010a} are relevant for reinforcement learning.  His criterion is a special case of the value of information with a penalty function that limits the mutual dependence between an internal representation of the states and an action response.  In \cite{SledgeIJ-jour2017c}, we modified the value of information to incorporate uncertainty-based search principles \cite{GhavamzadehM-conf2007a,KolterJZ-conf2009a,AchiamJ-conf2017a,ChentanezN-coll2005a,BellemareM-coll2016a}.  The corresponding action-selection probabilities are adjusted, based on a running estimate of a transition model, to force the agents into poorly investigated regions of the state-action space so that they can generalize well.

As Mohamed and Rezende note \cite{MohamedS-coll2015a}, Blahut-Arimoto-based updates do not natively scale well to large state-action spaces.  Experience generalization is often needed.  As an alternative, they proposed using variational approximation \cite{LopesM-coll2012a} for conducting exploration in deep-reinforcement-learning networks.  There are, however, some issues with this.  Parametric simplifying assumptions are often needed to create tractable inference procedures \cite{FardM-coll2010a,GuezA-coll2012a,DeisenrothMP-conf2011a}.  There is no guarantee, though, that densities for a given environment will adhere to these assumptions.  Additionally, variational approximations can introduce biases when updating the network parameters.  Biased gradients may destroy the bound \cite{DiengAB-coll2017a,DomkeJ-coll2018a}, leading to non-convergence \cite{RobbinsH-jour1951a} and hence repercussions for exploration.  Lastly, the convergence rate of variational approximation is provably dependent on the density dimensionality.

In this paper, we show that rate-distortion-like reinforcement learning is applicable to both simple and complicated state-action spaces.  In the former case, we utilize direct optimization strategies that work with tabular action-selection policies.  We leverage non-parametric estimators, which converge at a dimensionally-agnostic rate, for various environmental densities that are required by these strategies.  In the latter case, we eschew attempting to directly model the densities and instead utilize our group's unbiased, non-parametric information-theoretic measures, which converge at a dimensionally-agnostic rate.  We integrate these measures within a gradient-descent-based framework for training deep reinforcement-learning networks.  This addresses a major concern that we and other researchers have had---that existing optimization approaches may be ineffective for learning in interesting environments.

%%%%%%%%%%%%%%%%%%%%%%%%%%%%%%%%%%%%%%%%%%%%%%%%%%%%%%%%%%%%%%%%%%%%%%%%%%%
%%%%%%%%%%%%%%%%%%%%%%%%%%%%%%%%%%%%%%%%%%%%%%%%%%%%%%%%%%%%%%%%%%%%%%%%%%%
\subsection*{\small{\sf{\textbf{2.2.$\;\;\;$Exploration-Rate Adjustments}}}}

Most of the work on adapting learning rates has been for single-state, multi-action Markov decision processes.  Classical approaches have focused on the discrete-action, stochastic-reward case \citep{SalganicoffM-conf1995a}.  Other variants of the bandit problem exist, including adversarial bandits \citep{AuerP-jour2002a,AuerP-jour2002c}, non-stationary bandits \citep{AuerP-jour2002a,AuerP-jour2002c}, associative bandits \citep{AuerP-jour2002a,StrehlAL-conf2006c}, and budgeted bandits \citep{MadaniO-conf2004a}, each of which has distinct exploration-exploitation and parameter-update strategies.  Extensions for the continuous-action case have also been made \cite{KleinbergRD-coll2004a,WangY-coll2008a,BubeckS-jour2011a}.  % that make different assumptions about the decision space and obtainable rewards or costs

Several exploration strategies are available for the discrete, stochastic bandit problem.  One of the most widely employed is epsilon-greedy \citep{SuttonRS-book1998a,VermorelJ-coll2005a}, which involves taking random actions at a rate defined by the hyperparameter epsilon \cite{EvendarE-conf2002a,MannerS-jour2004a,CesaBianchiN-conf1998a,AuerP-jour2002b}.  Another popular exploration mechanism is soft-max selection, which entails assessing action expected returns and choosing actions in a weighted-random manner via a Gibbs distribution.  A single hyperparameter dictates the selection randomness \citep{CesaBianchiN-conf1998a,McMahanHB-conf2009a,AuerP-jour2002c,BeygelzimerA-conf2011a}.  Other schemes include the upper-confidence-bound method \citep{AuerP-jour2002a,AuerP-jour2010a} and its extensions \citep{GarivierA-conf2011a,CappeO-jour2013a}, Thompson sampling \cite{ThompsonWR-jour1933a,AgarwalS-jour2012a}, and the minimum-empirical-divergence algorithm \cite{HondaJ-jour2011a,HondaJ-jour2015a}.  Associated parameter-update processes are provided for each to achieve (near-)optimal asymptotic performance.

Single-state, multi-action algorithms are appealing because they are formally justified.  Unfortunately, they are largely ineffective for the multi-state, multi-action case, as they cannot capture multi-state dependencies.  Moreover, their parameter-update schedules would not necessarily facilitate optimal-rate convergence for the multi-state case.

An exception is the work of Meuleau and Bourgine \cite{MeuleauN-jour1999a}.  They advocated using multi-armed bandit algorithms to define local measures of action uncertainty \cite{MooreAW-jour1993a}.  The exploration bonuses would be scaled, added to the accrued rewards, and then back-propagated both using temporal-difference mechanisms \cite{SuttonRS-conf1995a,SuttonRS-jour1988a}.  A related uncertainty-propagation idea was implemented in Sutton's Dyna-$Q$ \cite{SuttonRS-conf1990a} directed exploration framework.  In propagating local uncertainty details, Meuleau and Bourgine argued that their approach would better avoid being misled by coupled-state-dependent environment dynamics than simply solving a series of independent bandit algorithms for each state.  The authors demonstrated promising results for simple problems.  However, they did not furnish convergence assurances.  It is therefore unknown as to if some of the theoretical guarantees of bandit algorithms would translate to multi-state, multi-action Markov problems.  Continuous state-action spaces also would likely pose difficulties.

There are few other exploration-rate adjustments for multi-state, multi-action Markov decision processes that are formally justified.  One instance is the explicit-explore-or-exploit algorithm \cite{KearnsM-conf1999a,KearnsM-jour2002a}.  It entails maintaining a list of how many times a state has been visited \cite{WhiteheadSD-conf1991a}.  If a state has been sufficiently encountered, then it is added to a so-called known-state list and either exploitation of the current policy or exploration is performed for that state.  If the agent transitions to a state that is not on the list, then the action chosen the fewest number of times at that state is taken.  This approach therefore modulates the exploration rate to emphasize either pure exploration or exploitation.  When following such a procedure, convergence to the goal state is possible a rate which is polynomial in the number of states and actions.  Brafman and Tennenholtz proposed one of the first practical implementations of this idea \cite{BrafmanRI-jour2000a,BrafmanRI-jour2002a}.  Rigorous analyses of this approach are provided by Strehl et al. \cite{StrehlAL-conf2006a,StrehlAL-conf2006b,StrehlAL-jour2009a}.  A downside of \cite{KearnsM-conf1999a,KearnsM-jour2002a} and similar methodologies is that the hyperparameter controlling the exploration-exploitation-rate modulation is typically not adapted.  Either too much or too little action-space search may be performed for practical domains if a good hyperparameter value is not selected.  It is also difficult to scale this work to discrete state-action spaces that are very large.  Continuous state-action spaces would significantly complicate matters.

The remaining exploration-rate adjustment strategies for the multi-state, multi-action case mostly target either epsilon-greedy-like \cite{WunderM-conf2010a,PriceB-conf1999a,OsbandI-coll2016a} or soft-max-like \cite{NachumO-coll2017a} searches.  They are largely heuristic and usually rely on either constant exploration rates or deterministic parameter annealings that can be ill-informed about the environment dynamics.  They hence may neither empirically nor theoretically converge to (near-)optimal policies.  Parameter values are typically manually supplied and guided by environment-specific knowledge that may be difficult to acquire.  Our previous work on the value of information for multi-state, multi-action reinforcement learning \cite{SledgeIJ-jour2017b,SledgeIJ-jour2017c} also has these issues.  We have found that its performance, and that of the remaining methods, is highly dependent on the chosen values.

In this paper, we use path following for altering the exploration rate when using value-of-information search.  Such an approach uses properties of local gradient flows to automatically determine exploration-rate adjustments for the current set of agent experiences.  There are several benefits of this.  Foremost, we prove that the chosen exploration-rate changes permit repeatedly converging to stationary points of the value-of-information criterion.  These stationary points are global-best policies that optimize the value of information for the current set of agent experiences.  If the agent can interact long enough with the environment, and some other mild assumptions are satisfied, then globally cost optimal policies will be uncovered.  This addresses the primary concern that we had about existing exploration adjustments---that they may be unlikely to converge.  Moreover, path following only has a single hyperparameter, which controls the solution accuracy of an intermediate optimization process.  We specify an automated procedure for adjusting this hyperparameter that ensures consistency of the intermediate solutions and without impacting convergence.  This behavior addresses our secondary concern---that currently available schemes may have difficult-to-set parameters and that improperly choosing their values can noticeably impede obtainable performance.  Lastly, our approach is amenable to both discrete and continuous state-action spaces.  This addresses a third concern---that existing parameter-update strategies are mainly suited for discrete spaces.

\phantomsection\label{sec3}
%%%%%%%%%%%%%%%%%%%%%%%%%%%%%%%%%%%%%%%%%%%%%%%%%%%%%%%%%%%%%%%%%%%%%%%%%%%
%%%%%%%%%%%%%%%%%%%%%%%%%%%%%%%%%%%%%%%%%%%%%%%%%%%%%%%%%%%%%%%%%%%%%%%%%%%
\subsection*{\small{\sf{\textbf{3.$\;\;\;$The Value of Information}}}}\addtocounter{section}{1}

In \cite{SledgeIJ-jour2017a,SledgeIJ-jour2017b,SledgeIJ-jour2017c}, we sought means to determine when it is appropriate to choose actions that deviate from the policy and when it is not.   This desire was realized by leveraging information that the states carry about the actions to determine which action should be taken.  Utilizing information in this way was developed into a rigorous theory by Stratonovich \cite{StratonovichRL-jour1965a,StratonovichRL-jour1966a}, which took the form of a value-of-information criterion.  This criterion describes the maximum benefit obtainable from a piece of information for either reducing average costs or increasing average rewards.  Expectation-maximization updates for this criterion can be formed, allowing it to be applied to reinforcement learning.  The updates provide action-selection probabilities in each state for a given exploration rate.  The value of information hence facilitates iteratively learning a stochastic policy.

In this section, we review the value of information (see \hyperref[sec3.1]{Section 3.1}).  We focus on the discrete-space case of the criterion so that tabular policies can be used.  This choice is for ease of presentation and to facilitate comparisons with classical methods.  The theory is easily extensible to continuous spaces, though, and we consider the case of continuous state spaces and discrete action spaces for experiments in an online appendix (see \hyperref[secC]{Appendix C}).  We then establish properties of the solutions for this criterion (see \hyperref[sec3.2]{Section 3.2}).  We we show that solutions for the value-of-information's Lagrangian correspond to policies where the Hessian of the Lagrangian is negative semi-definite on the nullspace of a Jacobian matrix.  This condition permits us to specify a second-order path-following process for simultaneously updating the action-selection policy and the exploration rate (see \hyperref[sec4]{Section 4}). 

A table of our notation is presented near the end of the paper (see \hyperref[tblB.1]{Table 1}).

\phantomsection\label{sec3.1}
%%%%%%%%%%%%%%%%%%%%%%%%%%%%%%%%%%%%%%%%%%%%%%%%%%%%%%%%%%%%%%%%%%%%%%%%%%%
%%%%%%%%%%%%%%%%%%%%%%%%%%%%%%%%%%%%%%%%%%%%%%%%%%%%%%%%%%%%%%%%%%%%%%%%%%%
\subsection*{\small{\sf{\textbf{3.1.$\;\;\;$Criterion Definition}}}}

Consider a composite system defined by a discrete state space $\mathcal{S}$ and discrete action space $\mathcal{A}$, both measurable.\\ \noindent  We assume that the state $s \!\in\! \mathcal{S}$ visited, at some discrete timestep, is a random variable.  After observing $s$, the agent\\ \noindent chooses an optimal action $a \!\in\! \mathcal{A}$ which minimizes a conditional expected penalty, assuming that 
reinforce-\\ \noindent ments are costs. That is, $\textnormal{inf}_{a \in \mathcal{A}}\, \mathbb{E}(Q(s,a)|p(s)) \!=\! \textnormal{inf}_{a \in \mathcal{A}}\, \sum_{s \in \mathcal{S}} p(s)Q(s,a)$.  Averaging the penalties yields the\\ \noindent total expected penalty, $\mathbb{E}(\textnormal{inf}_{a' \in \mathcal{A}}\, \mathbb{E}(Q(s,a')|p(s))|\pi(a|s)) \!=\! \sum_{s \in \mathcal{S}} \sum_{a \in \mathcal{A}} p(s)\pi(a|s)\textnormal{inf}_{a' \in \mathcal{A}}\,Q(s,a')$.  Here, $Q(s,a)$\\ \noindent is a penalty function, such as an action-state value-function associated with the agent costs.  The term $\pi(a|s) \!=\! p(a|s)$ represents the stochastic action-selection policy. 

There are two extreme cases to consider when finding an action-selection policy that solves the total expected penalty criterion.  The first case is when no information about the value of the random variable $s \!\in\! \mathcal{S}$ is available.\\ \noindent  That is, states carry no information about the actions that should be selected.  There is only one way to choose the optimal estimator $a \!\in\! \mathcal{A}$ when this occurs, which is to minimize the average penalties $\mathbb{E}(\textnormal{inf}_{a' \in \mathcal{A}}\, \mathbb{E}(Q(s,a')|p(s))|\pi(a|s)) =$\\ \noindent $\textnormal{inf}_{a \in \mathcal{A}}\, \mathbb{E}(Q(s,a)|p(s))$.  Only the prior, $p(s)$, can be leveraged to make an optimal decision in this case.  If the states carry total information about the actions, then $\mathbb{E}(\textnormal{inf}_{a' \in \mathcal{A}}\, \mathbb{E}(Q(s,a')|p(s))|\pi(a|s)) \!=\! \mathbb{E}(\textnormal{inf}_{a \in \mathcal{A}}\, Q(s,a)|p(s))$.  In this situation, the optimal action-selection policy is a delta function for the current state, as a single cost-optimal action will be chosen with unit probability.

The transition between no information to complete information, and hence a reduction of costs, is not immediate.  There is a smooth, non-linear transition \cite{BelavkinRV-coll2014a} between these two extremes for varying levels of information.  Stratonovich \cite{StratonovichRL-jour1965a,StratonovichRL-jour1966a} proposed an expression for these intermediate cases, which took the form of the value of information.  For Markov-decision-process reinforcement learning, an optimal estimator can be chosen by minimizing the difference in average costs for the no-information case with the total expected costs for the partial-information case,
\begin{equation}
f(\pi) = {\textnormal{inf}_{a \in \mathcal{A}}}\,\mathbb{E}\Bigg(\!Q(s,a)\,\Bigg|\,p(s)\!\Bigg) - {\textnormal{inf}_{\pi}}\,\mathbb{E}\Bigg(\underset{a' \in \mathcal{A}}{\textnormal{inf}}\,\mathbb{E}\Bigg(\! Q(s,a')\,\Bigg|\,p(s)\!\Bigg)\Bigg|\,\pi(a|s)\!\Bigg).
\label{eq:voi1}
\end{equation}
Here and in what follows, we use $\pi \!=\! \pi(a|s)$ $\forall a,s$ to represent the policy.  For the second term in (\ref{eq:voi1}), we have that\\ \noindent the conditional probabilities representing the policy are subject to an action-state mutual dependence constraint, like Boltzmann, Hartley, or R\'{e}nyi information.  Here, we use Shannon mutual information
\begin{equation}
\pi \textnormal{ such that} :  \displaystyle\mathbb{E}\Bigg( \!D_\textnormal{KL}(\pi(a|s)\|p(a))\,\Bigg|\, p(s) \!\Bigg) = \varphi_\textnormal{inf},\; \varphi_\textnormal{inf} \!\geq\! 0.
\label{eq:voi2}
\end{equation}
where $\mathbb{E}(D_\textnormal{KL}(\pi(a|s)\|p(a))|p(s)) \!=\! \sum_{s \in \mathcal{S}} p(s)\sum_{a \in \mathcal{A}} \pi(a|s) \textnormal{log}(\pi(a|s)/p(a))$.  This constraint is parameterized\\ \noindent by a positive, user-selectable value $\varphi_\textnormal{inf}$.  The value dictates how much information the states carry about what actions should be taken.  

More specifically, the value of information facilitates an optimal trade-off between the obtainable reinforcements and the uncertainty associated with the state-action random variables.  The amount of uncertainty is dictated by the information bound, which specifies the mutual dependence between states and actions.  The higher the bound, the greater the action-choice uncertainty.  This spurs a high degree of action exploration.  The potential for decreasing costs is great, since the agent should understand well the environment dynamics, given enough experience.  As the information bound is lowered, the agent becomes increasingly certain as to what actions should be taken for given states.  An exploitation-driven search of the action choices is realized.  The obtainable costs may be great or few, depending on the problem and the information-bound value.  Whenever a Markov-decision-process abstraction is used, the information-bound constraint has the effect of explicitly aggregating the state space \cite{SledgeIJ-jour2018a}.  That is, it provides a state abstraction \cite{MankowitzDJ-coll2016a,AbelD-conf2016a,AkrourR-jour2021a} and hence limits the complexity of the action-search problem during reinforcement learning.

\phantomsection\label{sec3.2}
%%%%%%%%%%%%%%%%%%%%%%%%%%%%%%%%%%%%%%%%%%%%%%%%%%%%%%%%%%%%%%%%%%%%%%%%%%%
%%%%%%%%%%%%%%%%%%%%%%%%%%%%%%%%%%%%%%%%%%%%%%%%%%%%%%%%%%%%%%%%%%%%%%%%%%%
\subsection*{\small{\sf{\textbf{3.2.$\;\;\;$Criterion Solution Properties}}}}

There are a variety of efficient ways to optimize the value of information.  The approach that we consider entails converting the constrained criterion (\ref{eq:voi1})--(\ref{eq:voi2}) into an unconstrained one using Lagrange multiplier theory: $\mathcal{L}((\pi,\beta),\vartheta) \!=\! F(\pi,\vartheta) \!+\! \sum_{s \in \mathcal{S}} \beta_{s}(\sum_{a \in \mathcal{A}} \pi(a|s) \!-\! 1)$, with $F(\pi,\vartheta) \!=\! f(\pi) \!+\! \mathbb{E}[D_\textnormal{KL}(\pi(s|a)\|p(a))]/\vartheta$; here\\ \noindent $\vartheta,\beta \!\in\! \mathbb{R}$ are Lagrange multipliers.  We can then differentiate the corresponding Lagrangian, set the expression to\\ \noindent zero, and solve for the conditional action-selection probabilities.  This yields soft-max-like expectation-maximization updates for the policy, where $1/\vartheta$ controls the action exploration rate \cite{SledgeIJ-jour2017a,SledgeIJ-jour2017b,SledgeIJ-jour2017c}.  

For what follows, it is important to characterize value-of-information solution properties.  Toward this end, we note that the gradient of the Lagrangian $\nabla_{\!\pi,\beta}\mathcal{L}((\pi,\beta),\vartheta)$ is given by 

\begin{equation}
\nabla_{\!\pi,\beta}\mathcal{L}((\pi,\beta),\vartheta) \!=\! \Bigg(\begin{matrix}[\nabla F(\pi,\vartheta) \!+\! (\beta^\top,\ldots,\beta^\top)^\top]_{mn \times 1}\vspace{0.05cm}\\ [\sum_{a \in \mathcal{A}} \pi(a|s_1) \!-\! 1,\ldots,\sum_{a \in \mathcal{A}} \pi(a|s_n) \!-\! 1]^\top_{1 \times m}\end{matrix}\Bigg) \!\in \mathbb{R}^{mn + m \times mn + m}
\label{eq:voi-jacobian}
\end{equation}
For the Lagrangian gradient, the first matrix row is given by $\nabla_{\!\pi}\mathcal{L}((\pi,\beta),\vartheta)$, while the second row is $\nabla_{\!\beta}\mathcal{L}((\pi,\beta),\vartheta)$.  We denote the number of states by $n$ and the number of actions by $m$.

The structure of the gradient for the Lagrangian $\nabla_{\!\pi,\beta}\mathcal{L}((\pi,\beta),\vartheta)$ can be exploited by various optimization techniques to find optima.  These optima adhere to the first-order necessary conditions \cite{BertsekasDP-book1995a}.  The gradient of the Lagrangian simultaneously satisfies $\nabla_{\!\pi,\beta}\mathcal{L}((\pi^*,\beta^*),\vartheta) \!=\! 0$, for optimal policies $\pi^*$ and corresponding Lagrange multipliers $\beta^*$, whenever the conditions are met.

We can also specify the Hessian of the Lagrangian $\nabla_{\!\pi,\beta}^2\mathcal{L}((\pi,\beta),\vartheta)$, which proves useful for classifying local solutions.  That is, we can use it to determine if a solution is merely a saddle point of the criterion or if it is global optimizer of the convex constrained criterion (\ref{eq:voi1})--(\ref{eq:voi2}).  The Hessian is given by the following block matrix
\begin{equation}
\nabla_{\!\pi,\beta}^2\mathcal{L}((\pi,\beta),\vartheta) = \Bigg(\begin{matrix} [\nabla^2_{\!\pi} F(\pi,\vartheta)]_{mn \times mn} & [\partial_\pi \nabla_{\!\beta}\mathcal{L}((\pi,\beta),\vartheta)]_{mn \times mn}\vspace{0.05cm}\\ [\partial_\pi \nabla_{\!\beta}\mathcal{L}((\pi,\beta),\vartheta)]_{mn \times mn} & [0]_{m \times m} \end{matrix}\Bigg) \!\in \mathbb{R}^{mn + m \times mn + m}
\label{eq:voi-hessian}
\end{equation}
where $\nabla^2_{\!\pi} F(\pi,\vartheta)$ is the Hessian of $F(\pi,\vartheta)$ and $J \!=\! \partial_\pi \nabla_{\!\beta}\mathcal{L}((\pi,\beta),\vartheta)$ is the Jacobian of $\nabla_{\!\beta}\mathcal{L}((\pi,\beta),\vartheta)$.  The Hessian\\ \noindent of $F(\pi,\vartheta)$ is itself a block matrix with zeros for the off-diagonal blocks.  Given this matrix, we can now quantify whether a given stationary point is a global solution of this criterion.  The proof is provided in \hyperref[secA.1]{Appendix A.1}.\vspace{0.15cm}

\phantomsection\label{prop3.3}
\begin{itemize}
\item[] \-\hspace{0.0cm}{\small{\sf{\textbf{Proposition 3.1.}}}} For a given optimal policy $\pi^* \!\in\! \mathbb{R}_{+}^{m \times n}$, we suppose that there is a vector of Lagrange\\ \noindent multipliers $\beta^* \!\in\! \mathbb{R}^n$ such that the Karush-Kuhn-Tucker conditions are satisfied.  If, for the Jacobian of the con-\\ \noindent straints $J$, we have that the Hessian $\psi^\top \nabla^2_{\!\pi,\beta}\, \mathcal{L}((\pi^*,\beta^*),\vartheta)\psi \!<\! 0$, then $\pi^*$ is a local solution of the value of\\ \noindent information.  Here, $\psi$ is an element of the Jacobian nullspace, $\psi \!\in\! \textnormal{ker}(J)$.  The converse is also true. \vspace{0.15cm}
\end{itemize}

\noindent Alternatively, we can relax the negative-definite property of $\nabla^2_{\pi,\beta}\, \mathcal{L}((\pi^*,\beta^*),\vartheta)$.  That is, let $\Gamma \!\in\! \mathbb{R}^{mn \times d}$ be a full-rank\\ \noindent column matrix whose columns span $\textnormal{ker}(J)$, where $d \!=\! \textnormal{dim}\, \textnormal{ker}(J)$.  The strict inequality condition in \hyperref[prop3.3]{Proposition 3.1},\\ \noindent $\psi^\top \nabla^2_{\!\pi,\beta}\, \mathcal{L}((\pi^*,\beta^*),\vartheta)\psi \!<\! 0$, can be replaced with $h^\top \Gamma^\top \nabla^2_{\!\pi,\beta}\, \mathcal{L}((\pi^*,\beta^*),\vartheta) \Gamma h \!\leq\! 0$, where $h \!\in\! \mathbb{R}^d$.  Hence, we have that the matrix $\Gamma^\top \nabla^2_{\!\pi,\beta}\, \mathcal{L}((\pi^*,\beta^*),\vartheta) \Gamma$ must be negative semi-definite.

In view of the preceding proposition, to find solutions for some given hyperparameter value, we need to construct a policy $\pi^*$ such that the gradient of the Lagrangian is equal to the zero vector, $\nabla_{\pi,\beta}\, \mathcal{L}((\pi^*,\beta^*),\vartheta) \!=\! 0$.  Likewise, we\\ \noindent need that the Hessian $\nabla^2_{\!\pi,\beta}\, \mathcal{L}((\pi^*,\beta^*),\vartheta)$ is negative semi-definite on the nullspace of the Jacobian, $\textnormal{ker}(J)$.  In what follows, we consider an approach that relies on these conditions to find such global solutions while simultaneously adjusting the exploration rate for the current set of agent experiences.

\addtocounter{equation}{-4}
\phantomsection\label{sec4}
%%%%%%%%%%%%%%%%%%%%%%%%%%%%%%%%%%%%%%%%%%%%%%%%%%%%%%%%%%%%%%%%%%%%%%%%%%%
%%%%%%%%%%%%%%%%%%%%%%%%%%%%%%%%%%%%%%%%%%%%%%%%%%%%%%%%%%%%%%%%%%%%%%%%%%%
\subsection*{\small{\sf{\textbf{4.$\;\;\;$Value-of-Information Path-Following for Reinforcement Learning}}}}\addtocounter{section}{1}

We want to iteratively uncover global solutions for (\ref{eq:voi1})--(\ref{eq:voi2}) while automatically tuning the exploration rate $\vartheta$ for a given reinforcement-learning environment.  A way to do this is by tracing solution branches \cite{ChowSN-book1982a,KuznetsovYA-book2004a} of a corresponding dynamical system $(\dot{\pi},\dot{\beta}) \!=\! \nabla_{\!\pi,\beta}\mathcal{L}((\pi,\beta),\vartheta)$ as the parameters $\vartheta,\beta$ are modified and as the agent accumulates\\ \noindent more experiences.  If each parameter, $\pi$, $\vartheta$, and $\beta$, is independently updated and the constraints on the Lagrangian gradient (\ref{eq:voi-jacobian}) and Hessian (\ref{eq:voi-hessian}) are satisfied, then solution approximations of the policy can be formed.  In the limit, the approximations will converge to the policy that solves (\ref{eq:voi1})--(\ref{eq:voi2}) for the current set of agent experiences.  This assumes certain constraints on the parameters, though.

Many path-following methods have been developed \cite{GovaertsWJF-book2000a,AllgowerEL-book2003a} that can be adapted to the value of information.  A popular method, parameter path-following, traces a solution trajectory by repeatedly perturbing a given parameter until a desired maximal or minimal value is reached.  After a parameter-value change, the final solution from the previous step is adjusted to represent what a potential solution could look like for this new value.  There is no guarantee, however, that this initial guess is a valid solution.  The iterate can be corrected so it approximately lies along a solution curve.  Branch-detection and switching processes are also carried out to handle intersecting solution branches.

This multi-stage process of guessing and correcting solutions is intuitively appealing.  It does, however, have drawbacks.  It fails whenever curvature of the solution surface is too high.  It also encounters issues whenever the system's Jacobian is singular, which is usually at a solution-branch bifurcation.  The correction step may either diverge at these points or not return to the same solution path.  Since singular points are frequently encountered for the value of information, a means of overcoming this latter issue is needed to preempt returning sub-par policies.

The shortcomings of parameter path-following at singularities can be remedied by re-parameterizing the entire problem by pseudo-arc-length.  That is, an approximate arc-length parameter is introduced so that the original solution vector is a function of it.  This yields a new equation system to be solved, which can be done via parameter path-following.  The path-following applied to this new system permits the iterates to jump over singular points, under some relatively mild conditions (see \hyperref[secA]{Appendix A}).  It thus permits continuing the optimization process for changing values of the exploration-rate hyperparameter and Lagrange multipliers.

We show that parameter path-following can be applied to the value of information (see \hyperref[sec4.1]{Section 4.1}).  We then propose a pseudo-arc-length re-parameterization of path-following for the value of information.  A byproduct of using pseudo-arc-length path-following is that the exploration-rate adjustment is specified automatically according to the agent's experiences.  No prior knowledge about either the environment or its dynamics is hence needed to tune this parameter.  Afterwards, we outline how to combine pseudo-arc-length path-following with $Q$-learning-based reinforcement learning (see \hyperref[sec4.2]{Section 4.2}).  Theoretical and practical aspects of path following, as it relates to reinforcement learning, are investigated in an associated online appendix (see \hyperref[secA]{Appendix A}).  We additionally prove, in the online appendix, when state-action-group bifurcations occur for changing exploration rates.  This specifies when the state abstraction changes.  We also outline how to handle switching to new solution branches in the appendix.

\phantomsection\label{sec4.1}
%%%%%%%%%%%%%%%%%%%%%%%%%%%%%%%%%%%%%%%%%%%%%%%%%%%%%%%%%%%%%%%%%%%%%%%%%%%
%%%%%%%%%%%%%%%%%%%%%%%%%%%%%%%%%%%%%%%%%%%%%%%%%%%%%%%%%%%%%%%%%%%%%%%%%%%
\subsection*{\small{\sf{\textbf{4.1.$\;\;\;$Finding Value-of-Information Solutions}}}}

%%%%%%%%%%%%%%%%%%%%%%%%%%%%%%%%%%%%%%%%%%%%%%%%%%%%%%%%%%%%%%%%%%%%%%%%%%%
%%%%%%%%%%%%%%%%%%%%%%%%%%%%%%%%%%%%%%%%%%%%%%%%%%%%%%%%%%%%%%%%%%%%%%%%%%%
\subsection*{\small{\sf{\textbf{4.1.1.$\;\;\;$Parameter Path-Following}}}}

An approach for optimally solving such systems as certain parameters are iteratively adjusted is to employ parameter path-following.  Parameter path-following operates by tracing a solution path $\nabla_{\!\pi,\beta}\mathcal{L}((\pi,\beta),\vartheta) \!=\! 0$ for perturbations in the hyperparameter $\vartheta$.  That is, it permits optimally updating the action-selection policy, using second-order information, for changes in the exploration amount $\vartheta$; it does not, however, yield a way to optimally update $\vartheta$ across either each episode or a set of episodes.

Geometrically, parameter path-following amounts to approximating the equilibrium $\nabla_{\!\pi,\beta}\mathcal{L}((\pi,\beta),\vartheta) \!=\! 0$, at a\\ \noindent point, by a tangent vector.  Following this vector updates the action-selection policy and associated multipliers for a change in $\vartheta$, but often causes the iterate to lie outside of the original trajectory.  A correction step must be applied to ensure that the iterate is projected back onto the solution path.  This two-step process of predicting and correcting is repeated until a desired maximum value of the exploration rate $\vartheta$ is reached.

More specifically, at time $k$, parameter path-following uses the tangent $\partial_\beta (\pi_{k},\beta_{k})$ at the point $((\pi_{k},\beta_{k}),\vartheta_{k})$ from the previous time to construct a preliminary guess $((\pi_{k+1}^{0},\beta_{k+1}^0),\vartheta_{k+1})$ for the equilibrium.  This is done by setting 
\begin{equation}
\Bigg(\begin{matrix} (\pi_{k+1}^{0},\beta_{k+1}^0)\vspace{0.075cm}\\ \vartheta_{k+1} \end{matrix}\Bigg) = \Bigg(\begin{matrix} (\pi_{k},\beta_{k}) \!+\! \delta\partial_\vartheta (\pi_{k},\beta_{k})\vspace{0.075cm}\\ \vartheta_{k} \!+\! \delta_\vartheta,\end{matrix}\Bigg)
\label{eq:param-cont-1}
\end{equation}

\noindent where $\delta, \delta_\vartheta \!\in\! \mathbb{R}_+$ are positive perturbation scalars.  This preliminary guess is used as a seed for Newton's method to\\ \noindent project onto the next equilibrium point $\nabla_{\!\pi,\beta}\mathcal{L}((\pi_{k+1}^*,\beta_{k+1}^*),\vartheta_{k+1}) \!=\! 0$ on the solution path; $\vartheta_k$ is kept fixed while Newton's method is being run to find the projection back onto the solution path.  Specifics of this approach are detailed below and summarized in \hyperref[alg:pathfollowing]{Algorithm 1}.  We provide a visual overview in \cref{fig:pathfollowing}.

\newcommand*\circled[2]{\tikz[baseline=(char.base)]{
\node[circle,white,draw,scale=#2,inner sep=1pt] (char) {#1};}}

\newcommand*\blackwhitecircled[2]{\tikz[baseline=(char.base)]{
\node[circle,black,fill=black,draw,scale=#2,inner sep=1pt] (char) {#1};}}

\begin{figure*}[t!]
\vspace{-0.5cm}
\hspace{-0.3cm}\begin{tabular}{c}
\imagetop{\parbox{1.0\linewidth}{
{\singlespacing\begin{algorithm}[H]
\label{alg:pathfollowing}
\RaggedRight%\fontdimen2\font=1.9pt\selectfont
\DontPrintSemicolon
\SetAlFnt{\small} \SetAlCapFnt{\small}
\caption{Value-of-Information-Based Parameter Path-Following}
\AlFnt{\small}\KwIn{An\! initial\! equilibrium\! point\! $(\pi_0,\beta_0)$\! of\! the\! system\! $(\dot{\pi},\dot{\beta}) \!=\! \nabla_{\!\pi,\beta}\mathcal{L}((\pi,\beta),\vartheta)$.\vspace{0.2cm}}
\AlFnt{\small} \For{\!{\bf each} $k \!=\! 1,2,\ldots$}{
\AlFnt{\small}    Find\! the\! tangent\! vector\! $\partial_\beta \pi_k$\! by\! solving\! $\partial_\pi \mathcal{L}((\pi_k,\beta_k),\vartheta_k)\partial_\beta \pi_k \!=\! -\partial_\beta \mathcal{L}((x_k,\beta_k),\vartheta_k)$.\vspace{0.05cm}\;
\AlFnt{\small}    Specify\! the\! preliminary\! iterate\! guess\! $\pi_k^{0} \!=\! \pi_{k-1} \!+\! \delta\partial_\beta \pi_{k-1}$.\vspace{0.05cm}\;
\AlFnt{\small}    Update\! $\beta_k \!=\! \beta_{k-1} \!+\! \delta$, $\delta \!>\! 0$, \!and\! $\vartheta_k \!=\! \vartheta_{k-1} \!+\! \delta_\vartheta$, $\delta_\vartheta \!>\! 0$.\;
\AlFnt{\small} \For{\!{\bf each} $i \!=\! 0,1,\ldots$ {\bf until} $\pi_k^i \to \pi_k$}{
\AlFnt{\small}    Update\! the\! iterate\! $\pi_k^{i+1}$\! by\! solving\! $\partial_\pi \mathcal{L}^i((\pi_k^i,\beta_k),\vartheta_k)(\pi_k^{i+1} \!-\! \pi_k^i) \!=\! -\mathcal{L}^i((\pi_k^i,\beta_k),\vartheta_k)$.
}
}
\end{algorithm}}}}
\end{tabular}\vspace{-0.35cm}
\end{figure*}
\begin{figure*}[t!]

\hspace{-5.75cm}\scalebox{0.775}{
\begin{tikzpicture}

 {\includegraphics[width=7cm]{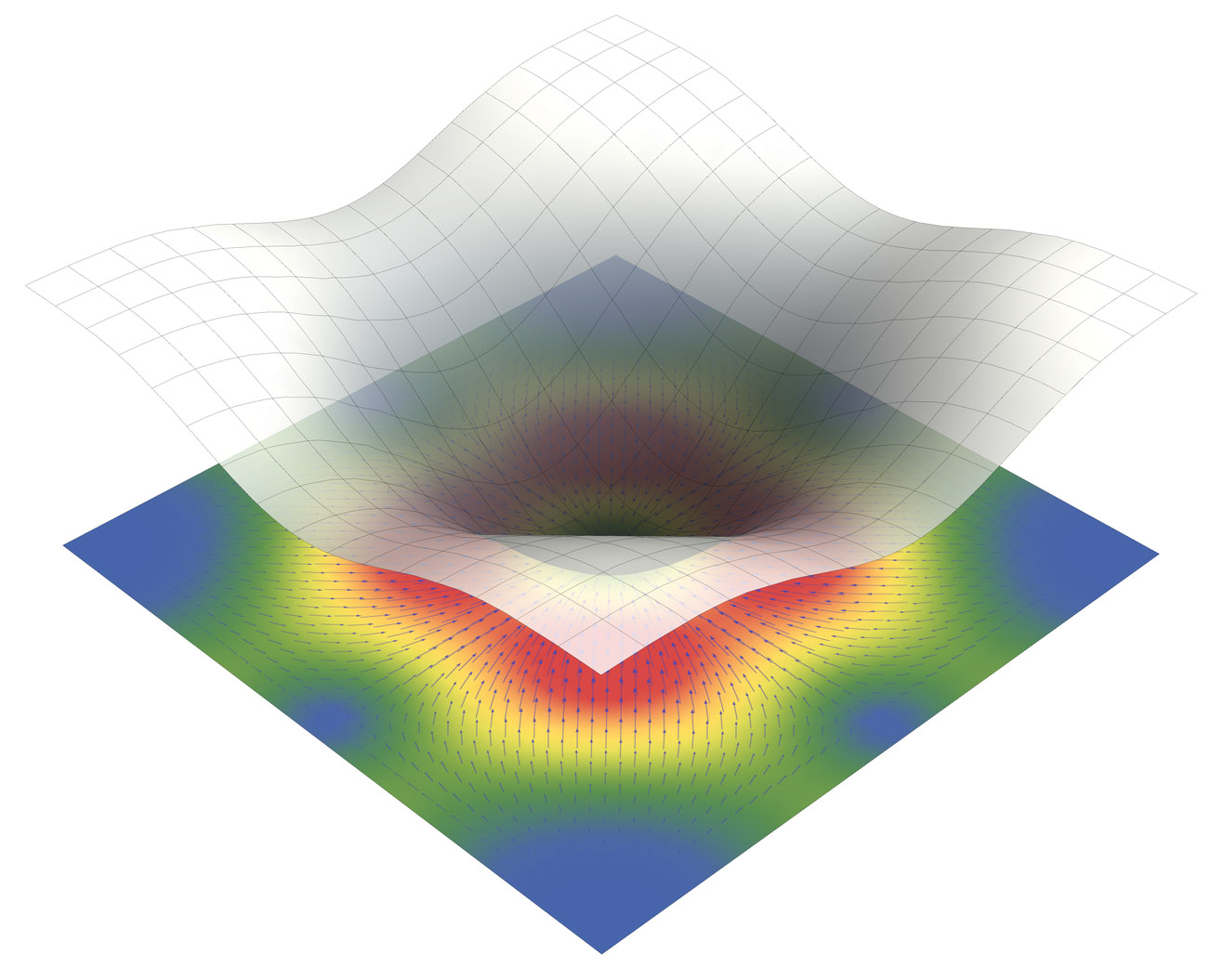}};

  \node[] at (0.0,0) {\includegraphics[width=5.5in]{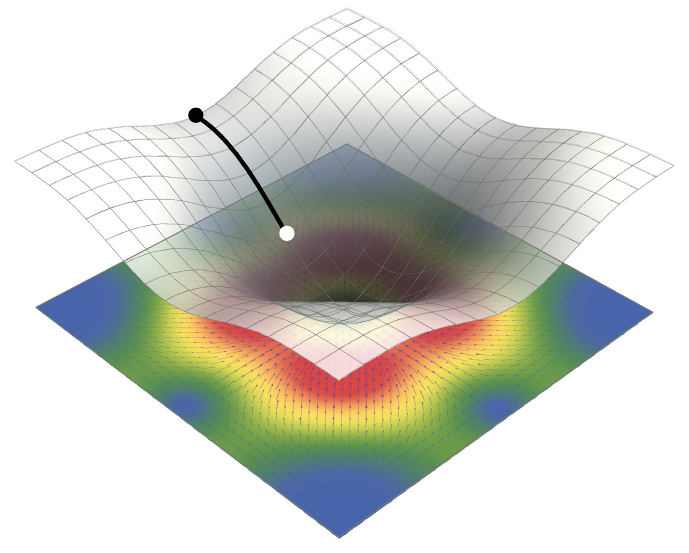}};

  \node at (-0.2,-3.7) {\large \textcolor{white}{$-\nabla_{\!\pi,\beta}\mathcal{L}((\pi,\beta),\vartheta)$}};
  \node at (4,3.6) {\large $\mathcal{L}((\pi,\beta),\vartheta)$};

  \node at (-4.2,3.6) {$((\pi_k,\beta_k),\vartheta_k)$};
  \node at (-4.4,-3.1) {\large $((\pi,\beta),\vartheta)_1$};
  \node at (4.4,-3.1) {\large $((\pi,\beta),\vartheta)_2$};

  \node at (2.2,0.85) {\textcolor{white}{\boldmath$\partial_\beta ((\pi_k,\beta_k),\vartheta_k)$}};
  \node at (-1.1,0.35) {\textcolor{white}{\boldmath$((\pi_{k+1},\beta_{k+1}),\vartheta_{k+1})$}};
  \node at (1.475,2.0) {\textcolor{white}{\boldmath$((\pi_{k}^0,\beta^0_{k+1}),\vartheta_{k+1})$}};

  \definecolor{tempcolor}{rgb}{0.995,0.995,0.995}
  \node at (-0.9,2.6) {\textcolor{tempcolor}{\boldmath$\delta$}};

  \node at (-3.7,1.85) {\textcolor{black}{$\nabla_{\!\pi,\beta}\mathcal{L}((\pi,\beta),\vartheta) \!=\! 0$}};

  \node at (-0.075,-1.3) {\textcolor{black}{$((\pi^*,\beta^*),\vartheta)$}};

  \draw[->,>=stealth,white,line width=0.85mm] (-3.03,3.285) -- (0.75,1.05);
  \draw[dashed,white,line width=0.85mm] (-0.0,1.6) -- (-1.25,0.775);

  \draw[fill, black] (-3.0,3.25) circle (4pt);
  \draw[fill, white] (-0.075,1.525) circle (4.25pt);
  \draw[fill, white] (-1.165,0.85) circle (4pt);

  \draw[fill, black] (-0.075,-0.875) circle (4.375pt);

  \node at (2.2,0.325) {\textcolor{black}{\bf \circled{\small \textcolor{white}{2}}{1} }};
  \node at (1.1,2.5) {\textcolor{black}{\bf \circled{\small \textcolor{white}{3}}{1} }};
  \node at (1.8,2.5) {\textcolor{black}{\bf \circled{\small \textcolor{white}{4}}{1} }};
  \node at (-1.1,-0.15) {\textcolor{black}{\bf \circled{\small \textcolor{white}{6}}{1} }};

  \node at (-9.0,0.1) {\large $Q_k(s,a)$};
  \node at (-7.275,-2.05) {\large $s$};
  \node at (-9,-3.78) {\large $a$};
  \node at (9,-3.78) {\large $a$};
  \node at (7.275,-2.05) {\large $s$};
  \node at (9.0,0.1) {\large $Q_{k+1}(s,a)$};

  \node at (-9,-7.6) {\large $(\pi_k,\beta_k)$};
  \node at (-5.8,-7.6) {\large $(\pi_{k}^0,\beta_{k}^0)$};
  \node at (5.8,-7.6) {\large $(\pi_{k}^i,\beta_{k}^i)$};
  \node at (9,-7.6) {\large $(\pi_{k+1},\beta_{k+1})$};

  \setlength{\fboxrule}{0.75pt}
  \setlength{\fboxsep}{0.025pt}
  \node at (-9.0,-0.25) {\includegraphics[width=1.39in]{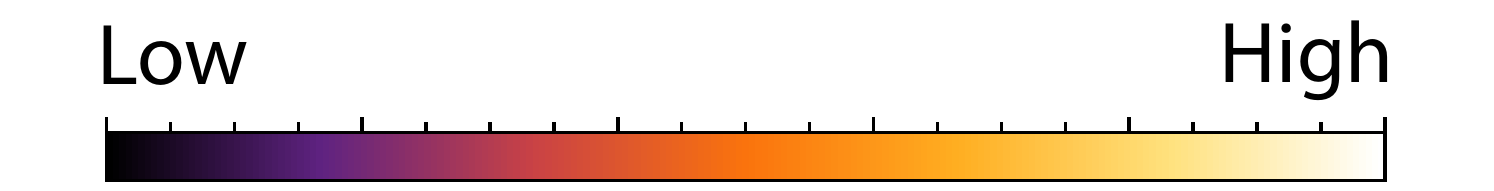}};
  \node at (-9,-2.05) {\framebox{\includegraphics[width=1.175in]{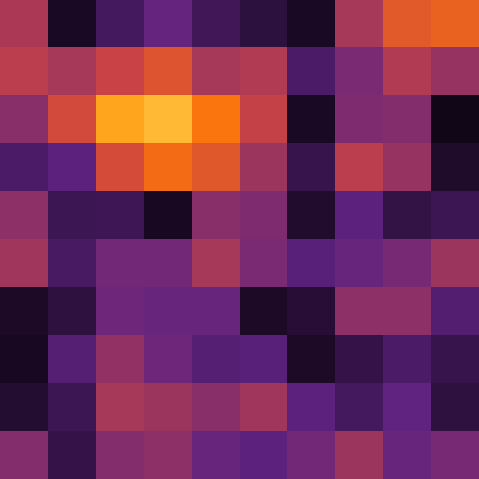}}};
  \node at (-9,-5.5) {\framebox{\embedvideo{\includegraphics[width=1.175in]{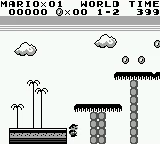}}{pathfollowing-update-0-1-noaudio.mp4}}};
  \node at (-5.8,-5.5) {\framebox{\embedvideo{\includegraphics[width=1.175in]{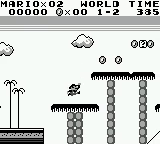}}{pathfollowing-update-1-1-noaudio.mp4}}};
  \node at (5.8,-5.5) {\framebox{\embedvideo{\includegraphics[width=1.175in]{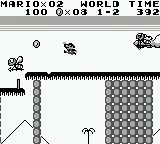}}{pathfollowing-update-2-1-noaudio.mp4}}};
  \node at (9,-5.5) {\framebox{\embedvideo{\includegraphics[width=1.175in]{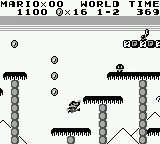}}{pathfollowing-update-3-1-noaudio.mp4}}};
  \node at (9,-2.05) {\framebox{\includegraphics[width=1.175in]{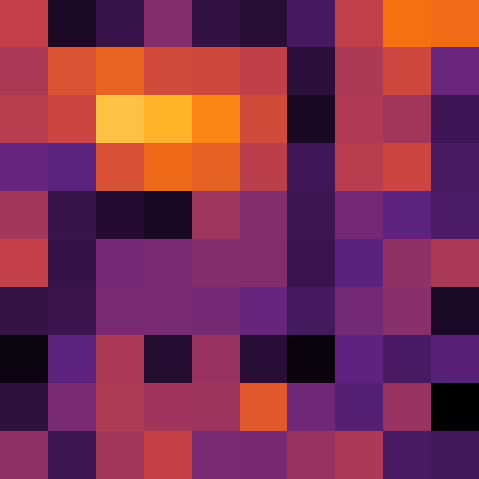}}};
  \node at (9.0,-0.25) {\includegraphics[width=1.39in]{sunsetbar.pdf}};

  \node at (-9,-7.15) {\includegraphics[width=1.39in]{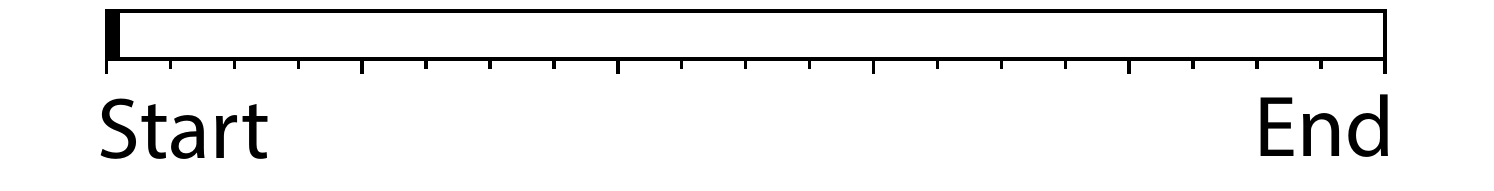}};
  \node at (-5.8,-7.15) {\includegraphics[width=1.39in]{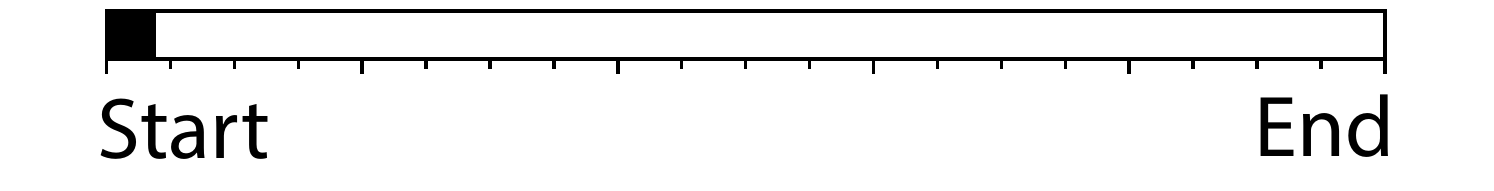}};
  \node at (5.8,-7.15) {\includegraphics[width=1.39in]{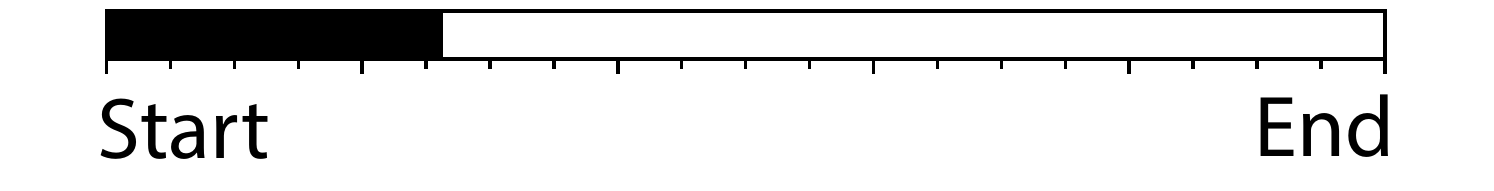}};
  \node at (9,-7.15) {\includegraphics[width=1.39in]{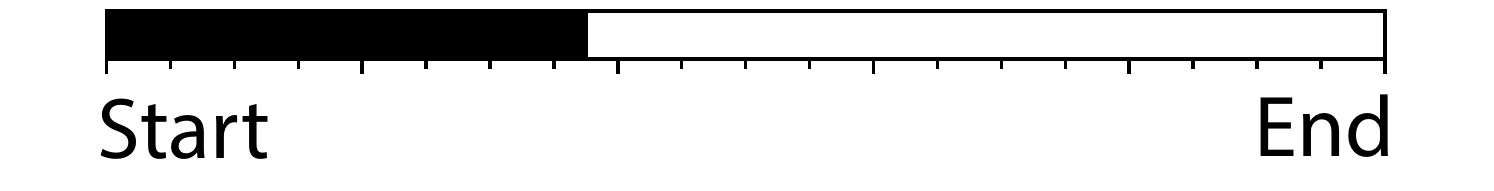}};

\end{tikzpicture}\vspace{-0.15cm}
}
   \caption[]{(middle) A visual overview of parameter path-following for the value of information.  For a given starting point, $((\pi_k,\beta_k),\vartheta_k)$, the tangent vector, $\partial_\varphi (\pi_k,\beta_k)$, (white arrow) is formed (step 2, \hyperref[alg:pathfollowing]{Algorithm 1}).  Given a step size tuple, $(\delta,\delta_\vartheta)$, $((\pi_k,\beta_k),\vartheta_k)$ is translated to a new point $((\pi_k^0,\beta_{k+1}),\vartheta_{k+1})$ along the tangent vector (steps 3--4, \hyperref[alg:pathfollowing]{Algorithm 1}).  This point is then iteratively retracted (white dashed line) back to the curve $\nabla_{\!\pi,\beta}\mathcal{L}((\pi,\beta),\vartheta) \!=\! 0$ (black line) along the Lagrangian surface (step 6, \hyperref[alg:pathfollowing]{Algorithm 1}).  Note that, depending on the magnitude of $\delta,\delta_\vartheta$, this update process may fail to converge to an equilibrium, $((\pi^*,\beta^*),\vartheta)$, of the gradient flow.  It may, instead, endlessly oscillate around this local optimum.  For each of the major updates shown in this overview, we provide corresponding embedded videos, for \emph{Super\! Mario\! Land}.  These videos illustrate the agent's improved understanding of the environment dynamics (left).  The level progress bars beneath them corroborate it.  We also provide quantized $Q$-value tables for the ten dominant state-action groups.  Once the iterates converge to the solution curve, the agent understands how to better react in certain situations (right).  However, it may explore either too much or too little, since the exploration rate is not automatically adjusted.  The $Q$-value table, and hence the policy, may not change greatly across successive episodes.  We recommend viewing this document within Adobe Acrobat DC; click on an image and enable content to start playback of the corresponding video.\vspace{-0.2cm}}
   \label{fig:pathfollowing}
\end{figure*}

The tangent vector $\partial_\beta (\pi_{k},\beta_{k})$ in (\ref{eq:param-cont-1}) can be constructed as follows whenever the derivative of the system\\ \noindent $\partial_\pi \nabla_{\!\pi,\beta}\mathcal{L}((\pi_k,\beta_k),\vartheta_k) \!=\! \nabla^2_{\!\pi,\beta}\mathcal{L}((\pi_k,\beta_k),\vartheta_k)$ is non-singular.  First, we note that, from the implicit function theo-\\ \noindent rem, we can take the total derivative of $\nabla_{\!\pi,\beta}\mathcal{L}((\pi_k,\beta_k),\vartheta_k) \!=\! 0$, which yields $\partial_\beta \nabla_{\!\pi,\beta}\mathcal{L}((\pi_k,\beta_k),\vartheta_k) \!=\! 0$, and hence 
\begin{align}
\partial_\beta (\pi_k,\beta_k) + \Bigg((\nabla^2_{\!\pi,\beta}\mathcal{L}((\pi_k,\beta_k),\vartheta_k))^{-1}\partial_\beta \nabla_{\!\pi,\beta}\mathcal{L}((\pi_k,\beta_k),\vartheta_k)\Bigg) &\!= 0\\
\Bigg(\nabla^2_{\!\pi,\beta}\mathcal{L}((\pi_k,\beta_k),\vartheta_k)\partial_\beta (\pi_k,\beta_k)\Bigg) + \Bigg(\partial_\beta \nabla_{\!\pi,\beta}\mathcal{L}((\pi_k,\beta_k),\vartheta_k)\Bigg) &\!= 0.\label{eq:param-cont-2}
\end{align}

\noindent Equation (\ref{eq:param-cont-2}) specifies a practical equation for finding the tangent vector $\partial_\beta (\pi_k,\beta_k)$ at the current equilibrium\\ \noindent point $\nabla_{\!\pi,\beta}\mathcal{L}((\pi_k^*,\beta_k^*),\vartheta_k) \!=\! 0$.

As we noted above, once the preliminary guess has been formed by way of the tangent vector, Newton's method can be applied to find $\nabla_{\!\pi,\beta}\mathcal{L}((\pi_k^*,\beta_k^*),\vartheta_k) \!=\! 0$.  Newton's method works by considering a sequence of linear approximations to the system and determining the solutions to those approximate systems $\nabla_{\!\pi,\beta}\mathcal{L}^i((\pi_k^i,\beta_k^i),\vartheta_k) \!=\! 0$ for a fixed $\vartheta_k$.  The linear approximation of the Lagrangian about an iterate can be found from Taylor's theorem.  This yields a series of equations that can be solved for projection steps $i \!=\! 1,2,\ldots$

\begin{equation}
\nabla_{\!\pi,\beta}\mathcal{L}^i((\pi_k,\beta_k),\vartheta_k) =\! \Bigg(\nabla_{\!\pi,\beta}\mathcal{L}^i((\pi_k^i,\beta_k^i),\vartheta_k)\Bigg) + \Bigg(\nabla^2_{\!\pi,\beta}\mathcal{L}^i((\pi_k^i,\beta_k^i),\vartheta_k)((\pi,\beta) \!-\! (\pi_k^i,\beta_k^i))\Bigg).
\label{eq:param-cont-3}
\end{equation}

\noindent The corresponding solution $(\pi_k^{i+1},\beta_k^{i+1})$ of (\ref{eq:param-cont-3}) can be constructed by solving the equation 

\begin{equation}
\Bigg(\nabla_{\!\pi,\beta}\mathcal{L}^i((\pi_k^i,\beta_k^i),\vartheta_k)\Bigg) + \Bigg(\nabla^2_{\!\pi,\beta}\mathcal{L}^i((\pi_k^i,\beta_k^i),\vartheta_k)((\pi_k^{i+1},\beta_k^{i+1}) \!-\! (\pi_k^i,\beta_k^i))\Bigg) \!= 0.
\label{eq:param-cont-4}
\end{equation}

\noindent For good initializations $(\pi_k^0,\beta_k^0)$, provided that the Hessian $\nabla^2_{\!\pi,\beta}\mathcal{L}((\pi_k,\beta_k),\vartheta_k)$ is non-singular, the iterates\\ \noindent $\{(\pi_k^i,\beta_k^i)\}_{i=1,2,\ldots} \!=\! \{(\pi_k^1,\beta_k^1),(\pi_k^2,\beta_k^2),\ldots\}$ provably converge to the true solution on the solution curve as the\\ \noindent number of iterations becomes infinite.  Practically, only a few steps $i$ are needed for (\ref{eq:param-cont-4}) to approach a solution.\vspace{0.15cm}

\begin{itemize}
\item[] \-\hspace{0.0cm}{\small{\sf{\textbf{Proposition 4.1.}}}} Assume that $\mathcal{L}^i((\pi_k^i,\beta_k^i),\vartheta_k^i)$ is Lipschitz differentiable, where $\mathcal{L}^i((\pi_k^0,\beta_k^0),\vartheta_k^0) \!=\! 0$ and\\ \noindent $\nabla_{\pi,\beta} \mathcal{L}^i((\pi_k^0,\beta_k^0),\vartheta_k^0)$ is non-singular.  There is an $\epsilon \!>\! 0$ that depends on the Lipschitz constants of\\ \noindent $\partial_\vartheta \mathcal{L}^i((\pi_k^0,\beta_k^0),\vartheta_k^0)$ and $\nabla_{\pi,\beta} \mathcal{L}^i((\pi_k^0,\beta_k^0),\vartheta_k^0)$ such that \hyperref[alg:pathfollowing]{Algorithm 1} converges $q$-quadratically to the solution $(\pi_{k+1},\beta_{k+1})$ of $\mathcal{L}((\pi_{k+1},\beta_{k+1}),\vartheta_{k+1}) \!=\! 0$ for $|\vartheta_{k+1} \!-\! \vartheta_k^0| \!<\! \epsilon$.\vspace{0.15cm}
\end{itemize}

\noindent The proof of this claim is given in \hyperref[secA.2]{Appendix A}.

For parameter path-following, the hyperparameter perturbation amount $\delta_\vartheta$ needs to be manually specified.  Choosing good values is troublesome, though.  Values of $\delta_\vartheta$ that are too high can cause the corrector step to sometimes converge to a point on a different branch or even completely diverge.  Small values of $\delta_\vartheta$ often avoid these issues.  However, they may not change the iterates much per step, which is computationally wasteful.

%%%%%%%%%%%%%%%%%%%%%%%%%%%%%%%%%%%%%%%%%%%%%%%%%%%%%%%%%%%%%%%%%%%%%%%%%%%
%%%%%%%%%%%%%%%%%%%%%%%%%%%%%%%%%%%%%%%%%%%%%%%%%%%%%%%%%%%%%%%%%%%%%%%%%%%
\subsection*{\small{\sf{\textbf{4.1.2.$\;\;\;$Pseudo-Arc-length Path-Following}}}}

Although parameter path-following is straightforward, it fails as a non-isolated solution is approached.  That is, it fails whenever the Hessian of the Lagrangian is singular.  While it may be possible to skip over some singular points, parameter path-following is unable to avoid saddle bifurcations.  Also, at other bifurcations, such as the pitchfork variety, some special procedures are required to jump from one branch to another.  Parameter path-following does not natively implement branch switching.

A way to remedy this defect of parameter path-following is to re-parameterize the problem by incorporating an approximate arc-length parameter so that both the policy and the Lagrange multipliers depend on it.  This idea, which is known as pseudo-arc-length path-following, introduces such a parameter and treats both the policy and its associated Lagrange multiplier as a function of it.  A new system of equations is hence produced, which can be solved by parameter path-following.  For pseudo-arc-length path-following to succeed, the corresponding Hessian for this new system must be non-singular.  It can be shown that this is the case for simple folds and hence where the original Lagrangian is non-singular, as a pseudo-arc-length constraint is appended to the original system's Jacobian to ensure it is full rank for sufficiently small parameter-value perturbations.

For pseudo-arc-length path-following, the vector $((\pi_{k},\beta_{k}),\vartheta_{k})$ of value-of-information variables is parameterized by a variable $\varphi_{k}$.  Here, $\varphi_k$ represents the arc-length along a solution curve $((\pi(\varphi_k),\beta(\varphi_k)),\vartheta(\varphi_k))$.  Under sufficient smoothness and regularity assumptions for the Lagrangian, we have that the following equality is satisfied
\begin{equation}
\Bigg(\nabla^2_{\!\pi,\beta}\mathcal{L}((\pi(\varphi_k),\beta(\varphi_k)),\vartheta(\varphi_k))(\dot{\pi}(\varphi_k),\dot{\beta}(\varphi_k))\Bigg) + \Bigg(\partial_\vartheta\nabla_{\!\pi,\beta}\mathcal{L}(\pi(\varphi_k),\beta(\varphi_k),\vartheta(\varphi_k))\dot{\vartheta}(\varphi_k)\Bigg) \!= 0
\label{eq:parc-1}
\end{equation}
at a solution $((\pi(\varphi_k),\beta(\varphi_k)),\vartheta(\varphi_k))$ of $\nabla_{\!\pi,\beta}\mathcal{L}((\pi_k,\beta_k),\vartheta_k) \!=\! 0$.  This solution is expected to jointly satisfy the\\ \noindent constraint $\theta\|(\dot{\pi}(\varphi_k),\dot{\beta}(\varphi_k))\|^2 \!+\! (1 \!-\! \theta)\dot{\vartheta}(\varphi_k)^2 \!=\! 1$, $\theta \!\in\! (0,1)$, which ensures that the orientation of the branch is\\ \noindent preserved if the steplength is sufficiently small.  Both conditions influence how the solution will be constructed.

\begin{figure*}[t!]

\vspace{-0.5cm}
\hspace{-0.3cm}\begin{tabular}{c}
\imagetop{\parbox{1.0\linewidth}{
{\singlespacing\begin{algorithm}[H]
\label{alg:pseudoarclengthpathfollowing}
\DontPrintSemicolon
\SetAlFnt{\small} \SetAlCapFnt{\small}
\caption{Value-of-Information-Based Pseudo-Arc-length Path-Following}
\AlFnt{\small}\KwIn{An\! initial\! equilibrium\! point\! $(\pi_0,\beta_0)$\! of\! the\! system\! $(\dot{\pi},\dot{\beta}) \!=\! \nabla_{\!\pi,\beta}\mathcal{L}((\pi,\beta),\vartheta)$.\vspace{0.2cm}}
\AlFnt{\small} \For{\!{\bf each} $k \!=\! 1,2,\ldots$}{
\AlFnt{\small}    Find\! the\! tangent\! vector\! $(\partial_\varphi\pi_k(\varphi_k)^\top,\partial_\varphi\beta_k(\varphi_k)^\top)^\top$ by solving $\phantom{\;\;\;\;}\nabla^2_{\!\pi,\beta}\mathcal{L}((\pi_k,\beta_k),\vartheta_k)(\partial_\varphi \pi_k(\varphi_k), \partial_\varphi \beta_k(\varphi_k))^\top \!=\! -(\nabla_{\!\pi} f(\pi_k), 0)^\top$.\vspace{0.05cm}\;
\AlFnt{\small}    Specify\! the\! preliminary\! iterate\! guess\! $\pi_k^{0}(\varphi_k) \!=\! \pi_{k-1}(\varphi_k) \!+\! \delta\partial_\varphi \pi_{k-1}(\varphi_k)$, $\;\delta \!>\! 0$.\vspace{0.05cm}\;
\AlFnt{\small}    Update\! $\vartheta_k(\varphi_k) \!=\! \vartheta_{k-1}(\varphi_k) \!+\! \delta\partial_\varphi \vartheta_{k-1}(\varphi_k)$, $\;\delta \!>\! 0$.\;
\AlFnt{\small} \For{\!{\bf each} $i \!=\! 0,1,\ldots$ {\bf until} $(\pi_k^i,\beta_k^i) \to (\pi_{k+1},\beta_{k+1})$, $\vartheta_k^i \to \vartheta_{k+1}$}{
\AlFnt{\small}    Update\! the\! iterates $(\pi_k^{i+1},\beta_k^{i+1},\vartheta_k^{i+1})$\! by\! solving\! $\phantom{\;\;\;\;}\Bigg(\begin{matrix} \nabla^2_{\!\pi,\beta}\mathcal{L}^i(\gamma_k^i(\varphi_k)) & \partial_\vartheta \nabla_{\!\pi,\beta}\mathcal{L}^i(\gamma_k^i(\varphi_k))\\ \partial_\varphi (\pi_k(\varphi_k)^\top,\beta_k(\varphi_k)^\top) & \partial_\varphi \vartheta_k(\varphi_k) \end{matrix}\Bigg)\Bigg(\begin{matrix}(\pi(\varphi_k) \!-\! \pi_k^i(\varphi_k),\beta(\varphi_k) \!-\! \beta_k^i(\varphi_k))\\ \vartheta(\varphi_k) \!-\! \vartheta_k^i(\varphi_k)\end{matrix}\Bigg) =$\vspace{0.075cm} $\phantom{\;\;\;\;\;\;\;\;\;\;\;\;\;\;\;\;\;\;\;\;\;\;\;\;\;\;\;\;\;\;\;\;\;\;\;\;\;\;\;\;\;\;\;\;\;\;\;\;\;\;\;\;\;\;\;\;\;\;\;\;\;\;\;\;\;\;\;\;\;\;\;\;\;\;\;\;\;\;\;\;\;\;\;\;\;\;\;\;\;\;\;\;\;\;\;\;\;\;\;\;\;\;\;\;\;\;\;\;\;\;\;\;\;\;\;\;\;\;\;\;\;\;\;\;\;}- \Bigg(\begin{matrix} \nabla_{\!\pi,\beta}\mathcal{L}^i(\gamma_k^i(\varphi_k))\\ \mathcal{K}^i(\gamma^i_k(\varphi_k)) \!-\! \delta \end{matrix}\Bigg).$\;
}
}
\end{algorithm}}}}
\end{tabular}\vspace{-0.6cm}
\end{figure*}

\begin{figure*}[t!]

\hspace{-0.295cm}\scalebox{0.775}{
\begin{tikzpicture}
  \node[] at (0.0,0) {\includegraphics[width=5.5in]{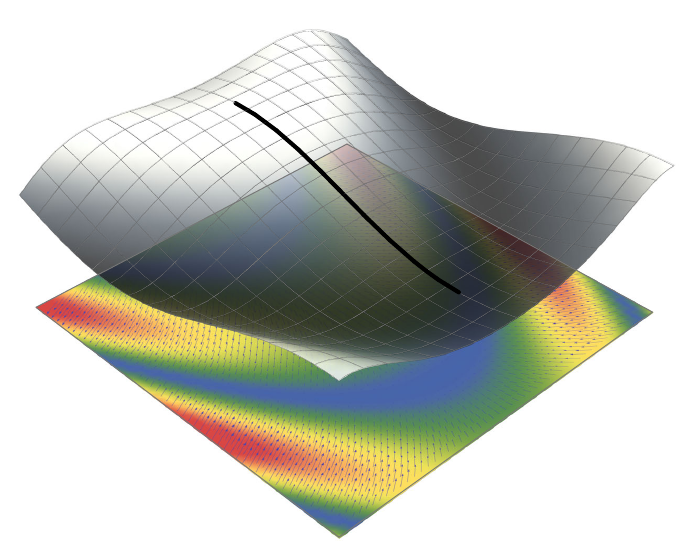}};

  \node at (-0.2,-3.7) {\large \textcolor{black}{$-\nabla_{\!\pi,\beta}\mathcal{L}((\pi,\beta),\vartheta)$}};
  \node at (3.7,3.45) {\large $\mathcal{L}((\pi,\beta),\vartheta)$};

  \node at (-3.3,3.85) {$((\pi_k,\beta_k),\vartheta_k)$};

  \node at (-4.65,-3.1) {\large $((\pi,\beta),\vartheta)(\varphi)_1$};
  \node at (4.65,-3.1) {\large $((\pi,\beta),\vartheta)(\varphi)_2$};

  \node at (4.1,1.85) {\textcolor{white}{\boldmath$\partial_\varphi (\pi_k(\varphi_k),\beta_k(\varphi_k))$}};
  \node at (2.6,-0.975) {\textcolor{white}{\boldmath$((\pi_{k+1},\beta_{k+1}),\vartheta_{k+1})$}};
  \node at (-0.0,1.025) {\textcolor{white}{\boldmath$((\pi_{k}^0(\varphi_k),\beta_{k+1}(\varphi_k)),\vartheta_{k+1})$}};
  \node at (2.6,-1.51) {\textcolor{white}{\boldmath$((\pi^*,\beta^*),\vartheta)$}};

  \definecolor{tempcolor}{rgb}{0.995,0.995,0.995}
  \node at (0.7,2.65) {\textcolor{tempcolor}{\boldmath$\delta$}};

  \node at (-2.95,2.45) {\textcolor{black}{$\nabla_{\!\pi,\beta}\mathcal{L}((\pi,\beta),\vartheta) \!=\! 0$}};

  \draw[->,>=stealth,white,line width=0.85mm] (-2.15,3.45) -- (3.9,0.8);
  \draw[dashed,white,line width=0.85mm] (2.551,1.2) -- (2.55,-0.5);

  \draw[fill, black] (-2.15,3.45) circle (4pt);
  \draw[fill, white] (2.551,1.35) circle (4.25pt);
  \draw[fill, white] (2.55,-0.5) circle (4pt);

  \node at (4.1,1.325) {\textcolor{black}{\bf \circled{\small \textcolor{white}{2}}{1} }};
  \node at (-0.325,0.49) {\textcolor{black}{\bf \circled{\small \textcolor{white}{3}}{1} }};
  \node at (0.325,0.49) {\textcolor{black}{\bf \circled{\small \textcolor{white}{4}}{1} }};
  \node at (2.615,-2.05) {\textcolor{black}{\bf \circled{\small \textcolor{white}{6}}{1} }};

  \node at (-9.0,0.1) {\large $Q_k(s,a)$};
  \node at (-7.275,-2.05) {\large $s$};
  \node at (-9,-3.78) {\large $a$};
  \node at (9,-3.78) {\large $a$};
  \node at (7.275,-2.05) {\large $s$};
  \node at (9.0,0.1) {\large $Q_{k+1}(s,a)$};

  \node at (-9,-7.6) {\large $(\pi_k,\beta_k)$};
  \node at (-5.8,-7.6) {\large $(\pi_{k}^0(\varphi_k),\beta_{k}(\varphi_k))$};
  \node at (5.8,-7.6) {\large $(\pi_{k}^i(\varphi_k),\beta_{k}(\varphi_k))$};
  \node at (9,-7.6) {\large $(\pi_{k+1},\beta_{k+1})$};

  \setlength{\fboxrule}{0.75pt}
  \setlength{\fboxsep}{0.025pt}
  \node at (-9.0,-0.25) {\includegraphics[width=1.39in]{sunsetbar.pdf}};
  \node at (-9,-2.05) {\framebox{\includegraphics[width=1.175in]{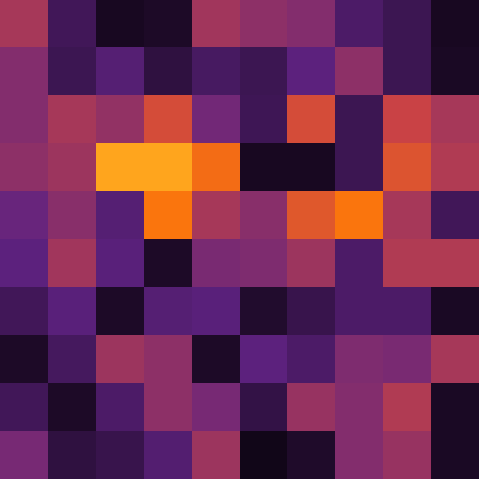}}};
  \node at (-9,-5.5) {\framebox{\embedvideo{\includegraphics[width=1.175in]{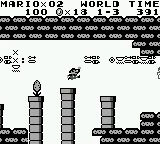}}{papathfollowing-update-0-1-noaudio.mp4}}};
  \node at (-5.8,-5.5) {\framebox{\embedvideo{\includegraphics[width=1.175in]{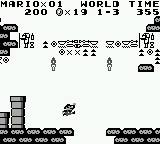}}{papathfollowing-update-1-1-noaudio.mp4}}};
  \node at (5.8,-5.5) {\framebox{\embedvideo{\includegraphics[width=1.175in]{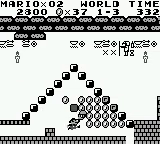}}{papathfollowing-update-2-1-noaudio.mp4}}};
  \node at (9,-5.5) {\framebox{\embedvideo{\includegraphics[width=1.175in]{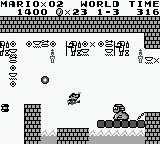}}{papathfollowing-update-3-1-noaudio.mp4}}};
  \node at (9,-2.05) {\framebox{\includegraphics[width=1.175in]{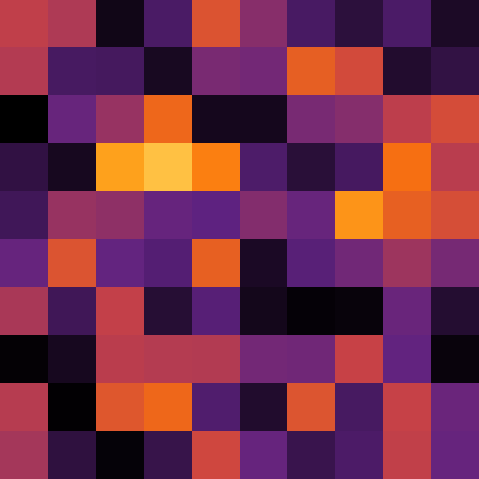}}};
  \node at (9.0,-0.25) {\includegraphics[width=1.39in]{sunsetbar.pdf}};

  \node at (-9,-7.15) {\includegraphics[width=1.39in]{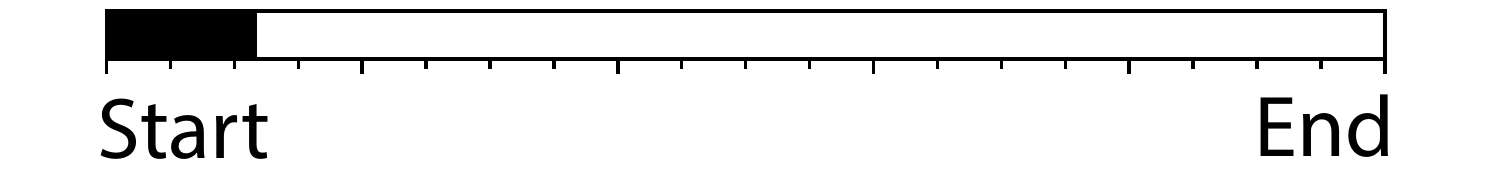}};
  \node at (-5.8,-7.15) {\includegraphics[width=1.39in]{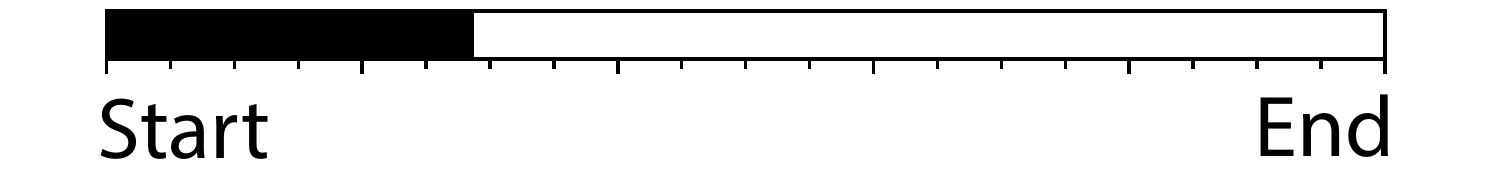}};
  \node at (5.8,-7.15) {\includegraphics[width=1.39in]{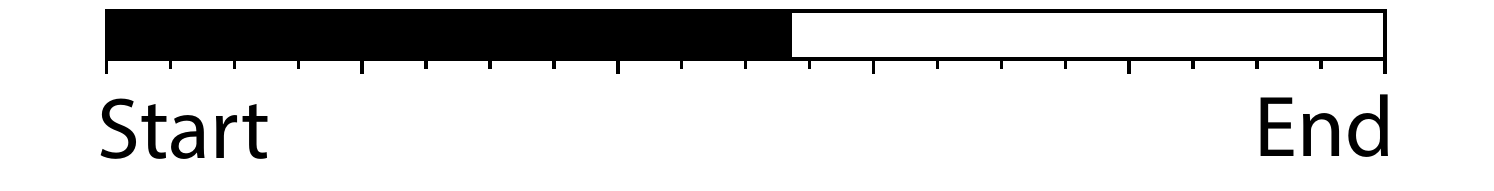}};
  \node at (9,-7.15) {\includegraphics[width=1.39in]{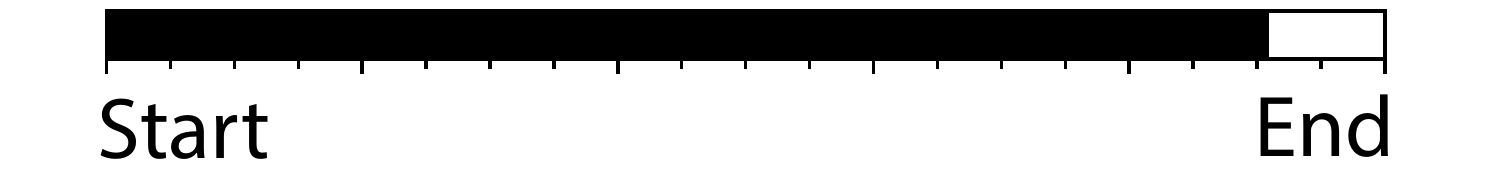}};

\end{tikzpicture}
}
   \caption[]{(middle) A visual overview of pseudo-arc-length path-following for the value of information.  For a given starting point, $((\pi_k,\beta_k),\vartheta_k)$, the parameterized tangent vector, $\partial_\beta (\pi_k(\varphi_k),\beta_k(\varphi_k))$, (white arrow) is formed (step 2, \hyperref[alg:pseudoarclengthpathfollowing]{Algorithm 2}).  Given an automatically determined step size, $\delta$, $((\pi_k,\beta_k),\vartheta_k)$ is translated to a new point $((\pi_k^0(\varphi_k),\beta_{k}(\varphi_k)),\vartheta_{k})$ along the tangent vector (steps 3--4, \hyperref[alg:pseudoarclengthpathfollowing]{Algorithm 2}).  This point is then iteratively retracted (white dashed line) back to the solution curve $\nabla_{\!\pi,\beta}\mathcal{L}((\pi,\beta),\vartheta) \!=\! 0$ (black line) along the Lagrangian surface (step 6, \hyperref[alg:pseudoarclengthpathfollowing]{Algorithm 2}).  Since $\delta$ is chosen automatically, this update process will usually converge to some point, $((\pi_{k+1}(\varphi_k),\beta_{k+1}(\varphi_{k+1})),\vartheta_{k})$, in an epsilon-ball around an equilibrium, $((\pi^*,\beta^*),\vartheta)$, of the gradient flow.  For each of the major updates shown in this overview, we provide corresponding embedded videos for \emph{Super\! Mario\! Land}.  They illustrate the agent's improved understanding of the environment dynamics (left).  The level progress bars beneath them corroborate it.  We also provide quantized $Q$-value tables for the ten dominant state-action groups.  Since the iterates converge to an equilibrium, which is a local solution for the value of information, the agent quickly adapts to the environment (right).  It determines how much it needs to explore based on its current experiences.  Rapid changes in the $Q$-value table is often seen early during learning, as shown here.  We recommend viewing this document within Adobe Acrobat DC; click on an image and enable content to start playback of the corresponding video.\vspace{-0.2cm}}
   \label{fig:papathfollowing}
\end{figure*}

The remaining mechanics are similar to that of parameter path-following.  In particular, the tangent vector\\ \noindent $\partial_\varphi ((\pi_{k}(\varphi_k),\beta_{k}(\varphi_k)),\vartheta_{k}(\varphi_k))$ to the curve $\nabla_{\!\pi,\beta}\mathcal{L}((\pi_{k}(\varphi_k),\beta_{k}(\varphi_k)),\vartheta_{k}(\varphi_k)) \!=\! 0$ at $(\pi_{k}(\varphi_k),\beta_{k}(\varphi_k))$ is\\ \noindent determined and then normalized.  This is used to supply an initial guess $((\pi_{k}^0(\varphi_k),\beta_{k}^0(\varphi_k)),\vartheta_{k}^0(\varphi_k))$ for the next equilibrium.  That is, we set 
\begin{equation}
\Bigg(\begin{matrix} (\pi_{k+1}^0(\varphi_{k+1}),\beta_{k+1}^0(\varphi_{k+1}))\vspace{0.075cm}\\ \vartheta_{k+1}^0(\varphi_{k+1}) \end{matrix}\Bigg) = \Bigg(\begin{matrix} (\pi_{k}^0(\varphi_k),\beta_{k}^0(\varphi_k)) \!+\! \delta \partial_\varphi (\pi_{k}(\varphi_k),\beta_{k}(\varphi_k))\vspace{0.075cm}\\ \vartheta_k \!+\! \delta\partial_\varphi\vartheta_k(\varphi_k)\end{matrix}\Bigg)
%(\pi_{k+1}^0,\beta_{k+1}^0) \!=\! (\pi_{k}^0,\beta_{k}^0) \!+\! \delta \partial_s (\pi_{k},\beta_{k}),\;\; \vartheta_k^0 \!=\! \vartheta_k \!+\! \delta\partial_s\vartheta_k,
\label{eq:arc-cont-1}
\end{equation}
\noindent where $\delta \!\in\! \mathbb{R}_+$ is a positive perturbation scalar that will be specified automatically using local properties of the sol-\\ \noindent ution path.  This initial guess is then modified by Newton's method so that it corresponds to an equilibrium on the solution path $\nabla_{\!\pi,\beta}\mathcal{L}((\pi_{k}(\varphi_k),\beta_{k})(\varphi_k),\vartheta_{k}(\varphi_k)) \!=\! 0$. Observe that the Newton-method correction step modifies the exploration rate, unlike in parameter path-following.  This two-step process is detailed below and outlined in \hyperref[alg:pseudoarclengthpathfollowing]{Algorithm 2}.  It is visually depicted in \cref{fig:papathfollowing}.

The tangent vector $(\partial_\varphi (\pi_k(\varphi_k),\beta_k(\varphi_k))^\top,\partial_\varphi \vartheta_k(\varphi_k))^\top$ in (\ref{eq:arc-cont-1}) for can be found as follows.  First the total derivative is taken, just as it was in parameter path-following.  Here, we use $\gamma_k(\varphi_k)$ to denote $((\pi_k(\varphi_k),\beta_k(\varphi_k)),\vartheta(\varphi_k))$ to make the notation more compact.
\begin{align}
\Bigg(\partial_\vartheta \nabla_{\!\pi,\beta}\mathcal{L}(\gamma_k(\varphi_k))\partial_\varphi \vartheta(\varphi_k)\Bigg) + \Bigg(\nabla^2_{\!\pi,\beta}\mathcal{L}(\gamma_k(\varphi_k))\partial_\varphi (\pi(\varphi_k),\beta(\varphi_k))\Bigg) &\!= 0 \label{eq:arc-cont-2}\\
\Bigg(\partial_\vartheta \nabla_{\!\pi,\beta}\mathcal{L}(\gamma_k(\varphi_k))\Bigg) + \Bigg(\nabla^2_{\!\pi,\beta}\mathcal{L}(\gamma_k(\varphi_k))\partial_\varphi (\pi_k(\varphi_k),\beta_k(\varphi_k))\Bigg) &\!= 0. \label{eq:arc-cont-3}
\end{align} 
$(\partial_\varphi \pi_k(\varphi_k)^\top,\partial_\varphi \beta_k(\varphi_k)^\top)^\top$ can be found by solving $\nabla_{\pi,\beta}^2 \mathcal{L}(\gamma_k(\varphi_k)) ((\partial_\varphi \pi_k(\varphi_k),\partial_\varphi \beta_k(\varphi_k)), \partial_\varphi \vartheta_k(\varphi_k))^\top \!=$\\ \noindent $- \partial_\vartheta \nabla_{\pi,\beta} \mathcal{L}(\gamma_k(\varphi_k))$, or, rather,
\begin{equation}
\Bigg(\begin{matrix} \nabla^2_{\!\pi,\beta}\mathcal{L}(\gamma_k(\varphi_k)) & \partial_\pi \nabla_{\!\beta}\mathcal{L}(\gamma_k(\varphi_k))^\top\vspace{0.05cm}\\ \partial_\pi \nabla_{\!\beta}\mathcal{L}(\gamma_k(\varphi_k)) & 0 \end{matrix}\Bigg)\Bigg(\begin{matrix}\partial_\varphi\pi_k(\varphi_k)\vspace{0.05cm}\\ \partial_\varphi\beta_k(\varphi_k)\end{matrix}\Bigg) = - \Bigg(\begin{matrix} \nabla_{\!\pi}f(\pi)\vspace{0.05cm}\\ 0 \end{matrix}\Bigg).
\label{eq:arc-cont-4}
\end{equation}
In (\ref{eq:arc-cont-3}) and (\ref{eq:arc-cont-4}), we show that (\ref{eq:arc-cont-2}) is solved for $\partial_\varphi (\pi(\varphi_k),\beta(\varphi_k))$ when $\partial_\varphi \vartheta_k(\varphi_k) \!=\! 1$.  It is permissible to set\\ \noindent $\partial_\varphi \vartheta_k(\varphi_k) \!=\! 1$ because $\partial_\varphi (\pi_k(\varphi_k),\beta_k(\varphi_k)) \!=\! -g (\nabla^2_{\!\pi,\beta}\mathcal{L}(\gamma_k(\varphi_k))^{-1}\partial_\beta \nabla_{\!\pi,\beta}\mathcal{L}((\pi_k,\beta_k),\vartheta_k)$ for $\partial_\varphi \vartheta(\varphi_k) \!=\! g$. This\\ \noindent implies that the two vectors, $\partial_\varphi (\pi_k(\varphi_k),\beta_k(\varphi_k)) \!=\! g \partial_\varphi (\pi_k(\varphi_k),\beta_k(\varphi_k))$, differ by only a scaling factor $g$.  Since the\\ \noindent tangent vector will be normalized, the effect of this scaling factor can be safely ignored.

When taking a step in the direction of the tangent vector $(\partial_\varphi (\pi_k(\varphi_k),\beta_k(\varphi_k)),\partial_\varphi \vartheta_k(\varphi_k))$, the preliminary guess may no longer be on the solution curve, just as in parameter path-following.  The corrector step finds a new equilibrium\\ \noindent $\gamma_{k+1}(\varphi_k)$ such that the norm of the projection of $((\pi_{k+1}^i(\varphi_k) \!-\! \pi_k(\varphi_k),\beta_{k+1}^i(\varphi_k) \!-\! \beta_k(\varphi_k)), \vartheta_{k+1}^i(\varphi_k) \!-\! \vartheta_k(\varphi_k))$\\ \noindent onto $((\partial_\varphi\pi_k(\varphi_k),\partial_\varphi\beta_k(\varphi_k)),\partial_\varphi\vartheta_k(\varphi_k))$ is bounded by the perturbation amount $\delta$
\begin{equation}
\textnormal{proj}(\gamma_k^i(\varphi_k)) = \Bigg\|\,\textnormal{proj}_{((\partial_\varphi(\varphi_k) \pi,\partial_\varphi \beta(\varphi_k)),\partial_\varphi \vartheta(\varphi_k))^\top}\Bigg(\begin{matrix}(\pi_{k+1}^i(\varphi_k) \!-\! \pi_k(\varphi_k),\, \beta_{k+1}^i(\varphi_k) \!-\! \beta_k(\varphi_k))\\ \vartheta_{k+1}^i(\varphi_k) \!-\! \vartheta_k(\varphi_k)\end{matrix}\Bigg)\Bigg\| \equiv \delta. %\Bigg(\begin{matrix}(\pi_{k+1}^i \!-\! \pi_k,\, \beta_{k+1}^i \!-\! \beta_k)\\ \vartheta_{k+1}^i \!-\! \vartheta_k\end{matrix}\Bigg)^\top\!\Bigg(\begin{matrix}\partial_s(\pi_k,\beta_k)\\ \partial_s\vartheta_k \end{matrix}\Bigg)
\label{eq:arc-cont-5}
\end{equation}
This is again facilitated via Newton's method.  The corresponding solution $\gamma_k^{i+1}(\varphi_k)$, for iteration $i \!+\! 1$, of the approximate system $\nabla_{\!\pi,\beta}\mathcal{L}^i(\gamma_k^i(\varphi_k)) \!=\! 0$ can be found by solving
\begin{equation}
\nabla_{\!\pi,\beta}\mathcal{K}^i(\gamma_k^i(\varphi_k))\Bigg(\begin{matrix}(\pi_k^{i+1}(\varphi_k),\beta_k^{i+1}(\varphi_k)) \!-\! (\pi_k^i(\varphi_k),\beta_k^i(\varphi_k))\vspace{0.05cm}\\ \vartheta_k^{i+1}(\varphi_k) \!-\! \vartheta_k^i(\varphi_k)\end{matrix}\Bigg) = -\mathcal{K}^i(\gamma_k^i(\varphi_k)).
\label{eq:arc-cont-6}
\end{equation}
In (\ref{eq:arc-cont-6}), $\mathcal{K}^i(\gamma_k^i(\varphi_k)) \!=\! (\nabla_{\!\pi,\beta}\mathcal{L}(\gamma_k^i(\varphi_k)),\textnormal{proj}(\gamma_{k+1}^i(\varphi_{k+1})) \!-\! \delta)^\top$ is a modified version of the Lagrangian, where\\ \noindent $\textnormal{proj}(\gamma_{k+1}^i(\varphi_{k+1})) \!-\! \delta \!=\! 0$.  This implies that the next iterate is specified by repeatedly solving
\begin{multline}
\Bigg(\begin{matrix} \nabla^2_{\!\pi,\beta}\mathcal{L}^i(\gamma_k^i(\varphi_k)) & \partial_\vartheta \nabla_{\!\pi,\beta}\mathcal{L}^i(\gamma_k^i(\varphi_k))\\ \partial_\varphi (\pi_k(\varphi_k)^\top,\beta_k(\varphi_k)^\top) & \partial_\varphi \vartheta_k(\varphi_k) \end{matrix}\Bigg)\Bigg(\begin{matrix}(\pi(\varphi_k) \!-\! \pi_k^i(\varphi_k),\beta(\varphi_k) \!-\! \beta_k^i(\varphi_k))\\ \vartheta(\varphi_k) \!-\! \vartheta_k^i(\varphi_k)\end{matrix}\Bigg) =\\ - \Bigg(\begin{matrix} \nabla_{\!\pi,\beta}\mathcal{L}^i(\gamma_k^i(\varphi_k))\\ \mathcal{K}^i(\gamma_k^i(\varphi_k)) \!-\! \delta \end{matrix}\Bigg),
\label{eq:arc-cont-7}
\end{multline}
which is guaranteed to converge to the next solution at the same rate as parameter path-following.\vspace{0.15cm}

\begin{itemize}
\item[] \-\hspace{0.0cm}{\small{\sf{\textbf{Proposition 4.2.}}}} Assume that $\mathcal{L}^i((\pi_k^i(\varphi_{k}),\beta_k^i(\varphi_{k})),\vartheta_k^i(\varphi_{k}))$ is Lipschitz differentiable, where\\ \noindent $\mathcal{L}^i((\pi_0^i(\varphi_0),\beta_0^i(\varphi_0)),\vartheta_0^i(\varphi_0)) \!=\! 0$ and $\nabla_{\pi,\beta} \mathcal{L}^i((\pi_0^i,(\varphi_0)\beta_0^i(\varphi_0)),\vartheta_0^i(\varphi_0))$ is non-singular.  There is an $\epsilon \!>\! 0$\\ \noindent that depends on $\langle \nabla_{\pi,\beta}\mathcal{L}^i((\pi_k^i(\varphi_k),\beta_k^i(\varphi_k)),\vartheta_k^i(\varphi_k)),\cdot\rangle$, the Lipschitz constant of \noindent $\partial_\vartheta \mathcal{L}^i((\pi_0^i(\varphi_0),\beta_0^i(\varphi_0)),\vartheta_0^i(\varphi_0))$, such that \hyperref[alg:pseudoarclengthpathfollowing]{Algorithm 2} converges $q$-quadratically to the solution $(\pi_{k+1}(\varphi_{k+1}),\beta_{k+1}(\varphi_{k+1}))$ of\\ \noindent $\mathcal{L}((\pi_{k+1}(\varphi_{k+1}),\beta_{k+1}(\varphi_{k+1})),\vartheta_{k+1}(\varphi_{k+1})) \!=\! 0$ for $|\varphi_{k+1} \!-\! \varphi_k^0| \!<\! \epsilon$.\vspace{0.15cm}
\end{itemize}

In \hyperref[secA.2]{Appendix A}, we prove that pseudo-arc-length path-following applied to the value of information can handle singular points, unlike parameter path-following.  It therefore will eventually converge to the optimal policy that solves (\ref{eq:voi1})--(\ref{eq:voi2}) and where the Hessian of the Lagrangian is negative semi-definite on the Jacobian nullspace.  In \cite{SledgeIJ-jour2018a} we showed that the value of information undergoes bifurcations whenever the exploration rate is increased past some critical value.  These bifurcations correspond to the formation a new state group.  Each state in a group is assigned a similar action-selection strategy as all other states in that group.  In the online appendix, we additionally specify how to decide which branch should be taken.

\phantomsection\label{sec4.2}
%%%%%%%%%%%%%%%%%%%%%%%%%%%%%%%%%%%%%%%%%%%%%%%%%%%%%%%%%%%%%%%%%%%%%%%%%%%
%%%%%%%%%%%%%%%%%%%%%%%%%%%%%%%%%%%%%%%%%%%%%%%%%%%%%%%%%%%%%%%%%%%%%%%%%%%
\subsection*{\small{\sf{\textbf{4.2.$\;\;\;$Value-of-Information-based Reinforcement Learning}}}}

Pseudo-arc-length path-following can be employed to optimally solve the value-of-information criterion.  It yields a systematic, second-order update for the action-selection policy whilst automatically tuning the uncertainty of the action-selection process. This is different than in our previous works \cite{SledgeIJ-jour2017a,SledgeIJ-jour2017b,SledgeIJ-jour2017c} where a first-order, soft-max-style of weighted-random exploration was obtained without a built-in mechanism for adjusting the exploration rate. % That is, it controls how much the agent should explore or exploit, by way of altering the conditional action-state probabilities, based upon the agent's experiences with the environment

%\begin{figure*}[t!]
\begin{tcolorbox}[blanker,float=tbp,grow to left by=0cm,grow to right by=0.05cm]
\vspace{-0.5cm}
{\singlespacing\begin{algorithm}[H]
\label{alg:voiqlearning}
\DontPrintSemicolon
\SetAlFnt{\small} \SetAlCapFnt{\small}
\caption{Coupled $Q$-Learning using Value-of-Information-Based Pseudo-Arc-length Path-Following}
\AlFnt{\small} Choose\! a\! non-negative\! values\! for\! the\! learning\! rates\! $\alpha$\! and\! $\omega$,\! discount\! factor\! $\gamma$,\! agent\! risk-taking\! parameter\! $\vartheta$,\! and\! steplength\! modulation\! factor\! $\delta'$.\!  Specify\! basis\! functions\! $\phi$.\;
\AlFnt{\small} Initialize\! the\! action-state\! value-function\! $Q(a,s)$.\!  Initialize\! the\! fast\! and\! slow\! time\! scales\! $u,v$.\;
\AlFnt{\small} \For{\!{\bf each\! episode\! until}\! $\vartheta_k$\! {\bf reaches\! some\! extremal\! value\!}}{
\AlFnt{\small} \For{\!{\bf each\! step}\! $k \!=\! 0,1,\ldots$ {\bf until\! an\! episode\! ends\!}}{
\AlFnt{\small}    Solve\! for\! the\! tangent\! vector\! $(\partial_\varphi \pi_k^\top,\partial_\varphi \beta_k^\top)^\top$\! using\! knowledge\! of\! the\! Hessian\vspace{0.05cm}\\ \nonl $\;\;\;\;\;\;\displaystyle \Bigg(\begin{matrix} \nabla^2_{\!\pi}\mathcal{L}((\pi_k,\beta_k),\vartheta_k) & \partial_\pi \nabla_\beta ((\pi_k,\beta_k),\vartheta_k)\vspace{0.05cm}\\ \partial_\pi \nabla_\beta ((\pi_k,\beta_k),\vartheta_k) & 0 \end{matrix}\Bigg)\Bigg(\begin{matrix} \partial_\varphi \pi_k(\varphi_k)\vspace{0.05cm}\\ \partial_\varphi \beta_k(\varphi_k) \end{matrix}\Bigg) = - \Bigg(\begin{matrix} \nabla_{\!\pi} f(\pi)\vspace{0.05cm}\\ 0\end{matrix}\Bigg)$.\vspace{0.05cm}\;
\AlFnt{\small}    Form\! an\! initial\! guess\! for\! the\! next\! iterate\! $(\pi_k^0,\beta_k^0,\vartheta_k)$\! using\! the\! tangent\! vector $(\partial_\varphi \pi_k^\top,\partial_\varphi \beta_k^\top)^\top$\vspace{0.05cm}\\ \nonl $\;\;\;\;\;\;\displaystyle \Bigg(\!\begin{matrix} (\pi_k^{0},\beta_k^0)\vspace{0.05cm}\\ \vartheta_k \end{matrix}\!\Bigg) = \Bigg(\!\begin{matrix} (\pi_{k},\beta_{k})\vspace{0.05cm}\\ \vartheta_{k} \end{matrix}\!\Bigg) + \frac{\delta'\textnormal{sign}(\textnormal{cos}(\theta))}{(1 \!+\! \|\partial_\varphi \pi_k(\varphi_k)\|^2 \!+\! \|\partial_\varphi \beta_k(\varphi_k)\|^2)^{1/2}} \Bigg(\begin{matrix} (\partial_\varphi\pi_{k}(\varphi_{k}),\; \partial_\varphi\beta_{k}(\varphi_{k}))\vspace{0.05cm}\\ 1 \end{matrix}\Bigg),$\vspace{0.05cm}\newline
\nonl where\! $\theta$\! is\! the\! angle\! between\! $(\partial_\varphi \pi_k(\varphi_k),\partial_\varphi \beta_k(\varphi_k),1)$\! and\! $(\partial_\varphi \pi_{k-1}(\varphi_{k-1}),\partial_\varphi \beta_{k-1}(\varphi_{k-1}),1)$.\vspace{0.05cm}\;
\AlFnt{\small} \For{\!{\bf each\! projection\! iteration}\! $i \!=\! 0,1,\ldots$ {\bf until}\! $(\pi_k^i,\beta_k^i,\vartheta_k^i)$\! {\bf has\! sufficiently\! converged\!}}{
\AlFnt{\small}    \AlFnt{\small}    Update\! the\! iterates $(\pi_k^{i+1},\beta_k^{i+1},\vartheta_k^{i+1})$\! by\! solving\! $\phantom{\;\;\;\;}\Bigg(\begin{matrix} \nabla^2_{\!\pi,\beta}\mathcal{L}^i(\gamma_k^i(\varphi_k)) & \partial_\vartheta \nabla_{\!\pi,\beta}\mathcal{L}^i(\gamma_k^i(\varphi_k))\\ \partial_\varphi (\pi_k(\varphi_k)^\top,\beta_k(\varphi_k)^\top) & \partial_\varphi \vartheta_k(\varphi_k) \end{matrix}\Bigg)\Bigg(\begin{matrix}(\pi(\varphi_k) \!-\! \pi_k^i(\varphi_k),\beta(\varphi_k) \!-\! \beta_k^i(\varphi_k))\\ \vartheta(\varphi_k) \!-\! \vartheta_k^i(\varphi_k)\end{matrix}\Bigg) =$\vspace{0.075cm} $\phantom{\;\;\;\;\;\;\;\;\;\;\;\;\;\;\;\;\;\;\;\;\;\;\;\;\;\;\;\;\;\;\;\;\;\;\;\;\;\;\;\;\;\;\;\;\;\;\;\;\;\;\;\;\;\;\;\;\;\;\;\;\;\;\;\;\;\;\;\;\;\;\;\;\;\;\;\;\;\;\;\;\;\;\;\;\;\;\;\;\;\;\;\;\;\;\;\;\;\;\;\;\;\;\;\;\;\;\;\;\;\;\;\;\;\;}- \Bigg(\begin{matrix} \nabla_{\!\pi,\beta}\mathcal{L}^i(\gamma_k^i(\varphi_k))\\ \mathcal{K}^i(\gamma^i_k(\varphi_k)) \!-\! \delta \end{matrix}\Bigg).$\;
}
               Set\! $(\pi_{k+1},\beta_{k+1},\vartheta_{k+1}) \leftarrow (\pi_k^{i+1},\beta_k^{i+1},\vartheta_k^{i+1})$\! after\! convergence.\;
\AlFnt{\small} \If{\!$(\pi_{k+1},\beta_{k+1},\vartheta_{k+1})$\! {\bf is\! a\! bifurcation\! point\!}}{
                  Enumerate\! bifurcating\! branches\! and\! perform\! a\! search\! over\! them.\;
}
               Choose\! an\! action\! $\pi_k(a_k|s_k) \!\rightarrow\! a_{k}$\! and\! perform\! a\! state\! transition\! $s_{k} \!\rightarrow\! s_{k+1} \!\in\!\mathcal{S}$.  Obtain\! a\! cost\! $r_{k+1} \!\in\! \mathbb{R}$.\;
\AlFnt{\small}    Update\! the\! fast\! time\! scale\! $u_{k+1} \leftarrow u_k + \alpha_k(\phi(s_k,a_k)Q_{v_k}(s_k,a_k) - u_k)$\! and\! the\! slow\! time\! scale\! $\phantom{\;\;\;}v_{k+1} \leftarrow v_k + \omega_k\phi(s_k,a_k)\Bigg(r_{k+1}(s_k,a_k) + \gamma_k\, \textnormal{inf}_{a \in \mathcal{A}}\; Q_{u_k}(s_{k+1},a) - Q_{v_k}(s_k,a_k)\Bigg)$.\;
\AlFnt{\small}    For\! $s_k \!\in\! \mathcal{S}$\! and\! $a_k \!\in\! \mathcal{A}$,\! update\! $\phantom{\;\;\;}Q_k(a_k,s_k) \leftarrow Q_{k-1}(s_k,a_k) + \alpha_k\Bigg(r_{k+1}(s_k,a_k) + \gamma_k\, \textnormal{inf}_{a \in \mathcal{A}}\; Q_{u_k}(s_{k+1},a) - Q_{v_k}(s_k,a_k)\Bigg)$.\;
\AlFnt{\small} }
\AlFnt{\small} Initialize\! the\! variables\! for\! the\! next\! episode\! using\! the\! ones\! from\! the\! current\! episode\! $(\pi_0,\beta_0,\vartheta_0,Q_0) \!\leftarrow\! (\pi_k,\beta_k,\vartheta_k,Q_k)$.\;
\AlFnt{\small} }
\end{algorithm}}\vspace{-0.4cm}
\end{tcolorbox}%
%\end{figure*}

This path-following-based action exploration can be combined with a Markov-decision-process abstraction to perform reinforcement learning.  It can hence be incorporated into learning methods like SARSA \cite{WieringM-jour1998a,SinghSP-jour2000a}, TD-learning with exploration \cite{MeynSP-conf2011a}, $Q$-learning \cite{WatkinsCJCH-jour1992a}, and various extensions of these algorithms \cite{AgarwalN-conf2022a}.  Here, we consider the value of information with coupled $Q$-learning \cite{CarvalhoDS-coll2020a}.  The discrete, tabular case for this methodology is given in \hyperref[alg:voiqlearning]{Algorithm 3}.  

Coupled $Q$-learning relies on a dual-time-scale inference.  For the faster time scale, an update similar to that of deep-$Q$ networks is used to reduce the effect of bootstrapping.  For the slower time scale, a modified version of the target network update is employed.  Experience replay is applied to break sample correlation and mitigate overfitting \cite{LinLJ-jour1992a,SchaulT-conf2016a,HorganD-conf2018a,FedusW-conf2020a,AndrychowiczM-coll2017a}.  In our simulations, we use prioritized experience replay \cite{SchaulT-conf2016a}.  This version of $Q$-learning utilizes linear experience-interpolation to improve the acquisition of agent behaviors for large environments.  It, however, guarantees convergence to an optimal policy, unlike other function approximators \cite{BoyanJA-coll1995a,TesauroG-coll1996a,SuttonRS-coll1999a,MahadevanS-coll2006a,BhatnagarS-coll2009a}.  For ease of presentation, these mechanisms are not included in \hyperref[alg:voiqlearning]{Algorithm 3}.

The value-of-information optimization steps are given in \hyperref[alg:voiqlearning]{Algorithm 3}, steps 4 through 8.  In \hyperref[alg:voiqlearning]{Algorithm 3}, step 5, we form the tangent vector $(\partial_\varphi \pi_k(\varphi_k)^\top,\partial_\varphi \beta_k(\varphi_k)^\top)^\top$ through knowledge of the Hessian.  This step comes about by re-writing $\nabla^2_{\!\pi,\beta}\mathcal{L}((\pi_k,\beta_k),\vartheta_k)(\partial_\varphi \pi_k(\varphi_k), \partial_\varphi \beta(\varphi_k))^\top$ from \hyperref[alg:pseudoarclengthpathfollowing]{Algorithm 2}, step 2, in terms of (\ref{eq:voi-hessian}).  The resulting tangent vector permits the calculation of a new candidate solution (\ref{eq:arc-cont-1}), which is done in \hyperref[alg:voiqlearning]{Algorithm 3}, step 6.  For this step, the exploration-rate update relies on knowledge of the tangent $\partial_\varphi \vartheta_{k-1}(\varphi_{k-1})$.  An expression for the tangent vector follows from the constraint that subsequent tangent vectors must have the same orientation.  We have used the fact that $\partial_\varphi \vartheta(\varphi_k)$ can equal one, as was assumed when going from (\ref{eq:arc-cont-2}) to (\ref{eq:arc-cont-3}); this is because the tangent vector will be normalized and the actual scaling factor can be ignored.

In step 6, we automatically calculate the steplength for the iterates.  This scalar has two components.  The numerator contains an orientation-preserving term, which ensures that the direction of the tangent vector does not change.  The denominator is a unit-length normalization term.  We can additionally augment the steplength by a small multiplicative term, $\delta'$.  This term specifies the size of a ball around a stationary point to which the iterates converge.  Smaller values of $\delta'$ are usually better for preventing iterate backtracking at the expense of more iterations.  However, it appears to be safe to consider a multiplicative term of one.  Such a value also does not impact our convergence theory.  

Lastly, in \hyperref[alg:voiqlearning]{Algorithm 3}, step 8, the candidate solutions are iteratively projected onto the solution curve to obtain an equilibrium that satisfies (\ref{eq:parc-1}).  This is done by repeatedly solving (\ref{eq:arc-cont-7}) for some small $\delta$.  Note that since the value of information is convex, every stationary point is a minimizer.  For non-convex criteria, (\ref{eq:arc-cont-7}) should be replaced with a minimization process that is subject to $\mathcal{K}^i((\pi_k^i,\beta_k^i),\vartheta_k^i) \!-\! \delta \!=\! 0$ so that optima are sought instead of stationary points.

As outlined in step 10, during the search process, there will be times where bifurcations are encountered.  These are singular points at which two conditions are met.  The first is $\textnormal{codim}(\textnormal{range}(\nabla^2_{\!\pi,\beta}\mathcal{L}(\pi_k(\varphi_k),\beta_k(\varphi_k),\vartheta_k(\varphi_k)))) \!=\! m$,\\ \noindent where $m$ is the dimensionality of the Hessian's nullspace at that singular point, $(\pi_k,\beta_k)$.  The second condition is that $\partial_\vartheta\nabla^2_{\!\pi,\beta}\mathcal{L}(\pi_k(\varphi_k),\beta_k(\varphi_k),\vartheta_k(\varphi_k)) \!\in\! \textnormal{range}(\nabla^2_{\!\pi,\beta}\mathcal{L}(\pi_k(\varphi_k),\beta_k(\varphi_k),\vartheta_k(\varphi_k)))$.  If both conditions are true, then each\\ \noindent of these solution branches needs to be investigated---a priori, we do not know which branch corresponds to the greatest reduction of total expected costs.  In \hyperref[secA.3]{Appendix A}, we specify possible approaches for enumerating these branches.  Each approach entails forming distinct tangent vectors and then running parallel searches using pseudo-arc-length path-following. %the number of branches that non-tangentially intersect at that singular point

Steps 13 and 14 in \hyperref[alg:voiqlearning]{Algorithm 3} correspond to updates of the action-state value-function.  Here, we assume that both the value function and the policy will be updated for every action choice in an episode.  In settings where the agent does not encounter novel situations frequently, this can be computationally wasteful.  There are possible ways to reduce the computational burden in such situations.  We can, for instance, abandon a Newton-type projection process in favor of a quasi-Newton one.  We can also forgo updating the policy with each taken action.  Instead, the update should be adjusted when a critical value of the learning-rate will be reached and for a brief period thereafter.  This is viable, since the variables often will not change much across consecutive action choices.  The action-selection probabilities will usually remain within a small band once they have stabilized between two critical values.  They will only begin to greatly change once an exploration-rate critical value is reached and a new state-group is formed \cite{SledgeIJ-jour2018a}.  

Policies produced by \hyperref[alg:voiqlearning]{Algorithm 3} will solve (\ref{eq:voi1})--(\ref{eq:voi2}) for a given information-bound amount.  If the maximal exploration rate has an associated information bound that is equivalent to the state-random-variable entropy, then the policies can be globally cost-optimal for the environment.  The state abstraction will be finely grained, as each state has the potential to be mapped to a unique action in the continuous case.  If the terminal exploration rate is set too low, then a coarse state abstraction will be obtained.  The policies may not be cost-optimal for the environment.  In either setting, the value of information should be trivially modified so that it is non-expansive everywhere \cite{AsadiK-conf2017a}.  By making this change, we can be guaranteed that globally cost-optimal policies will be consistently formed in the limit.

\phantomsection\label{sec5}
%%%%%%%%%%%%%%%%%%%%%%%%%%%%%%%%%%%%%%%%%%%%%%%%%%%%%%%%%%%%%%%%%%%%%%%%%%%
%%%%%%%%%%%%%%%%%%%%%%%%%%%%%%%%%%%%%%%%%%%%%%%%%%%%%%%%%%%%%%%%%%%%%%%%%%%
\subsection*{\small{\sf{\textbf{5.$\;\;\;$Simulations}}}}\addtocounter{section}{1}

In this section, we assess our exploration-rate-adaptation approaches on the classic arcade games \emph{Millipede} and \emph{Centipede} for the Nintendo GameBoy system.  Both games are challenging for reinforcement learning.

An aim of our simulations is to understand how path following influences the policy search process.  Toward this end, we compare parameter and pseudo-arc-length path-following, both with and without adaptive steplength sizes (see \hyperref[sec5.1.1]{Section 5.1.1}).  We also discuss the solution-surface bifurcation and tie observed performance improvements to implemented behaviors (see \hyperref[sec5.1.2]{Section 5.1.2}).  We additionally assess the performance gap between our path-following-based updates and both deterministic and adaptive-annealing-based updates when using the value of information (see \hyperref[sec5.2]{Section 5.2}).  We show that pseudo-arc-length path-following consistently outperforms the alternatives.

Gameplay mechanics and scoring details for our simulations are provided in an associated online appendix (see \hyperref[secB]{Appendix B}).  Other simulation aspects and additional results are presented in this appendix too.

\phantomsection\label{sec5.1}
%%%%%%%%%%%%%%%%%%%%%%%%%%%%%%%%%%%%%%%%%%%%%%%%%%%%%%%%%%%%%%%%%%%%%%%%%%%
%%%%%%%%%%%%%%%%%%%%%%%%%%%%%%%%%%%%%%%%%%%%%%%%%%%%%%%%%%%%%%%%%%%%%%%%%%%
\subsection*{\small{\sf{\textbf{5.1.$\;\;\;$Path Following Results and Discussions}}}}

\setcounter{figure}{0}
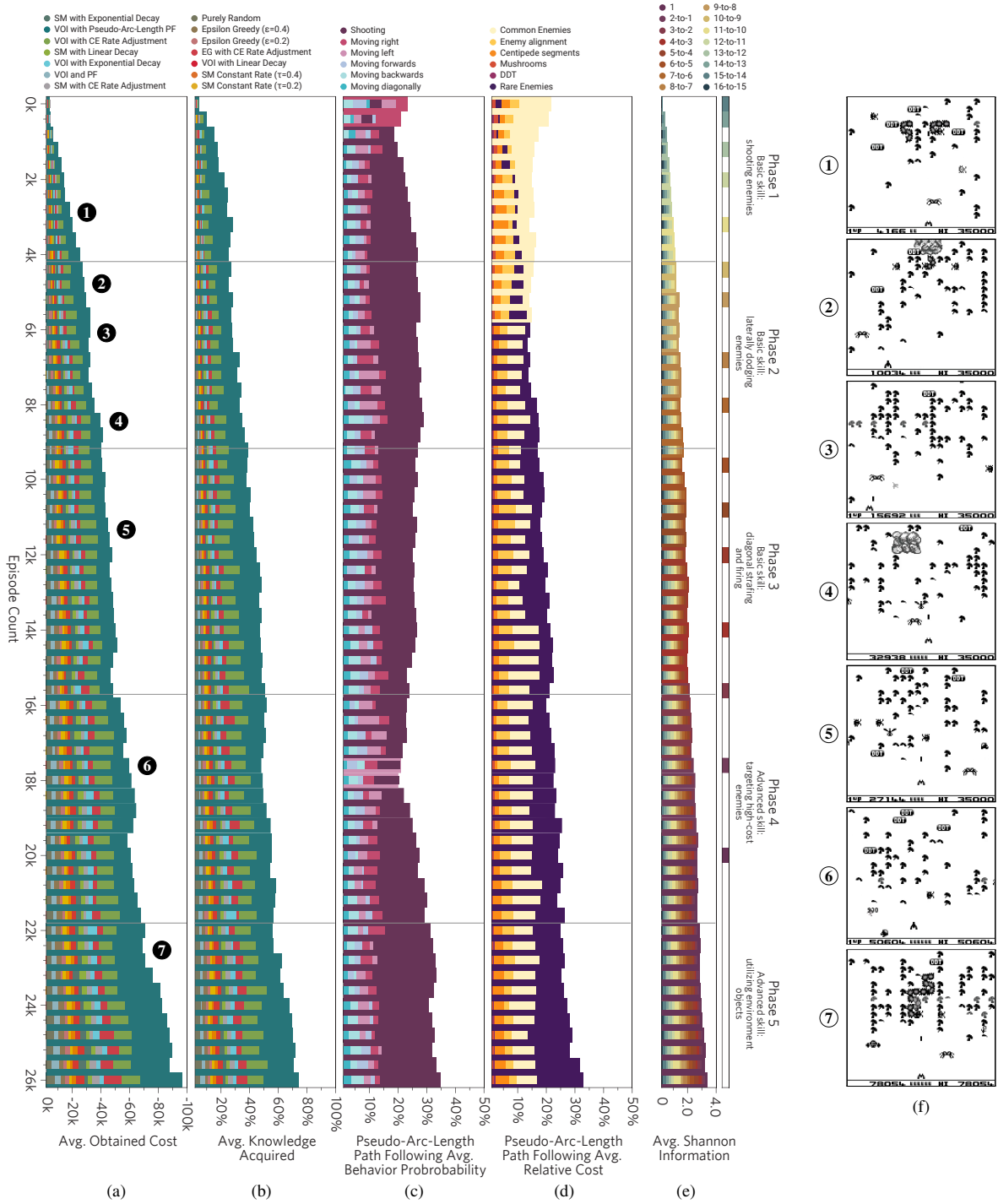
\begin{figure}[t!]
   \centering

   \scalebox{0.775}{
      \begin{tikzpicture} 
         \node at (-3.325,11.775) {\includegraphics[height=0.665in]{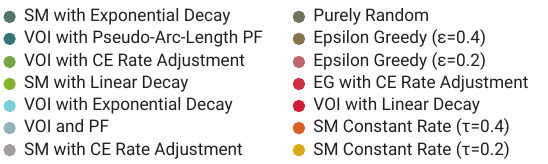}};
         \node at (0.81,11.645) {\includegraphics[height=0.57in]{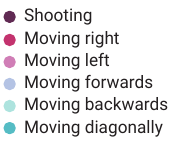}};
         \node at (3.915,11.645) {\includegraphics[height=0.57in]{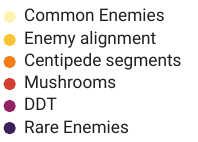}};
         \node at (7.325,11.875) {\includegraphics[height=0.76in]{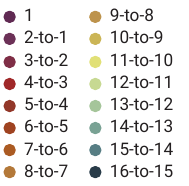}};

         \node[] at (1.0,-0.125) {\includegraphics[width=6.25in]{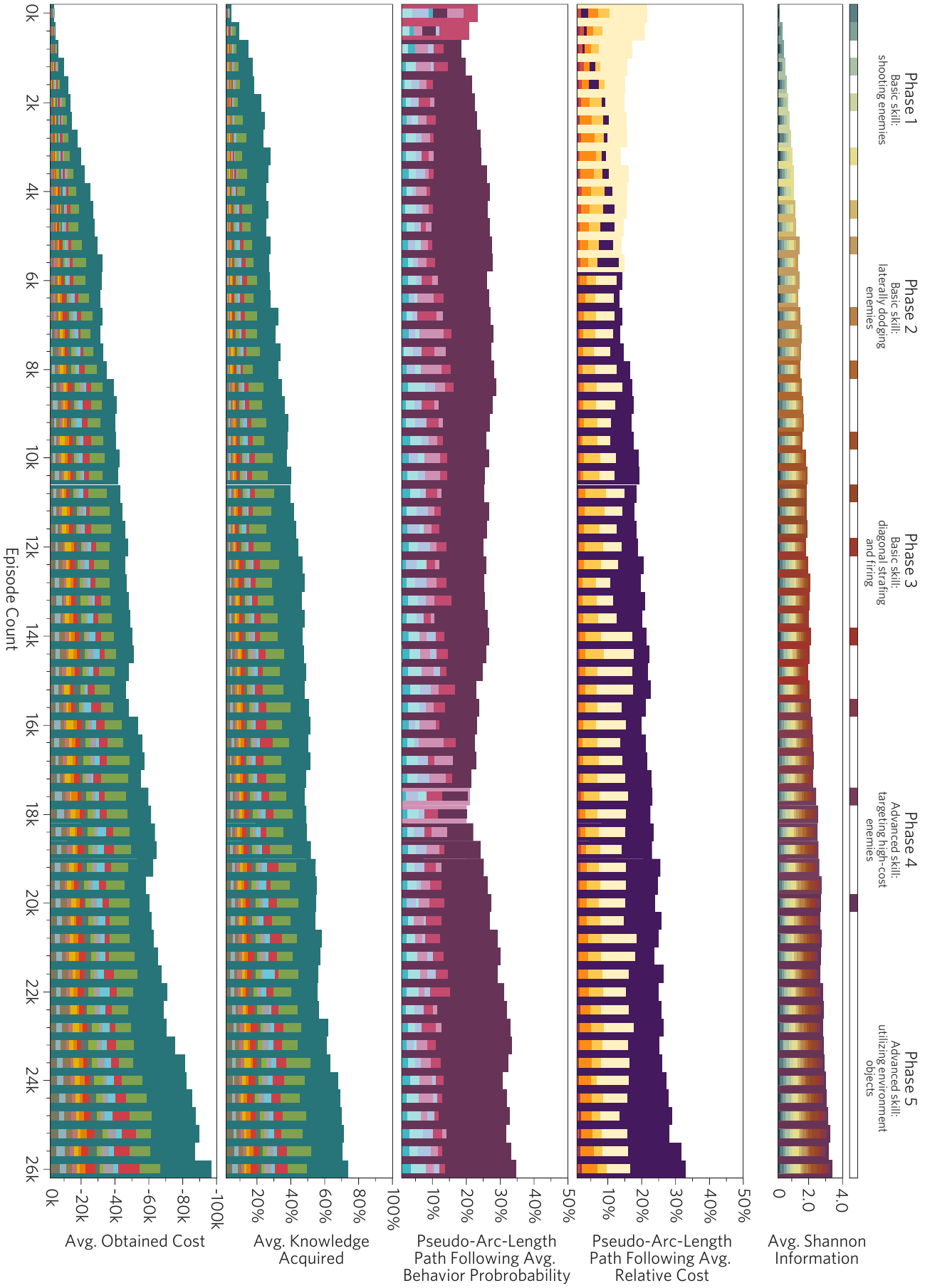}};

         \draw[line width=0.125mm, color={rgb,255:red,142;green,142;blue,142}] (-6.075,7.5) -- (-3.225,7.5);
         \draw[line width=0.125mm, color={rgb,255:red,142;green,142;blue,142}] (-3.06,7.5) -- (-0.21,7.5);
         \draw[line width=0.125mm, color={rgb,255:red,142;green,142;blue,142}] (-0.04,7.5) -- (2.815,7.5);
         \draw[line width=0.125mm, color={rgb,255:red,142;green,142;blue,142}] (2.975,7.5) -- (5.8,7.5);
         \draw[line width=0.125mm, color={rgb,255:red,142;green,142;blue,142}] (6.4175,7.5) -- (7.52,7.5);

         \draw[line width=0.125mm, color={rgb,255:red,142;green,142;blue,142}] (-6.075,3.7) -- (-3.225,3.7);
         \draw[line width=0.125mm, color={rgb,255:red,142;green,142;blue,142}] (-3.06,3.7) -- (-0.21,3.7);
         \draw[line width=0.125mm, color={rgb,255:red,142;green,142;blue,142}] (-0.04,3.7) -- (2.815,3.7);
         \draw[line width=0.125mm, color={rgb,255:red,142;green,142;blue,142}] (2.975,3.7) -- (5.8,3.7);
         \draw[line width=0.125mm, color={rgb,255:red,142;green,142;blue,142}] (6.4175,3.7) -- (7.52,3.7);

         \draw[line width=0.125mm, color={rgb,255:red,142;green,142;blue,142}] (-6.075,-1.3) -- (-3.225,-1.3);
         \draw[line width=0.125mm, color={rgb,255:red,142;green,142;blue,142}] (-3.06,-1.3) -- (-0.21,-1.3);
         \draw[line width=0.125mm, color={rgb,255:red,142;green,142;blue,142}] (-0.04,-1.3) -- (2.815,-1.3);
         \draw[line width=0.125mm, color={rgb,255:red,142;green,142;blue,142}] (2.975,-1.3) -- (5.8,-1.3);
         \draw[line width=0.125mm, color={rgb,255:red,142;green,142;blue,142}] (6.4175,-1.3) -- (7.52,-1.3);

         \draw[line width=0.125mm, color={rgb,255:red,142;green,142;blue,142}] (-6.075,-5.95) -- (-3.225,-5.95);
         \draw[line width=0.125mm, color={rgb,255:red,142;green,142;blue,142}] (-3.06,-5.95) -- (-0.21,-5.95);
         \draw[line width=0.125mm, color={rgb,255:red,142;green,142;blue,142}] (-0.04,-5.95) -- (2.815,-5.95);
         \draw[line width=0.125mm, color={rgb,255:red,142;green,142;blue,142}] (2.975,-5.95) -- (5.8,-5.95);
         \draw[line width=0.125mm, color={rgb,255:red,142;green,142;blue,142}] (6.4175,-5.95) -- (7.52,-5.95);

         \node at (-4.65,-11.425) {(a)};
         \node at (-1.7,-11.425) {(b)};
         \node at (1.4,-11.425) {(c)};
         \node at (4.45,-11.425) {(d)};
         \node at (6.925,-11.425) {(e)};

         \setlength{\fboxrule}{0.75pt}
         \setlength{\fboxsep}{0.025pt}
         \node at (11.7,9.46) {\framebox{\embedvideo{\includegraphics[width=1.175in]{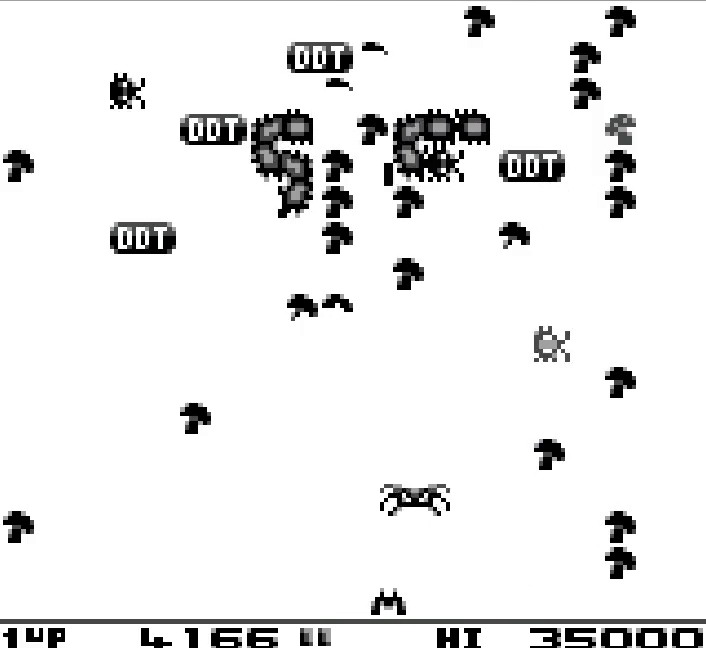}}{Millipede-1-1-small.mp4}}};
         \node at (11.7,6.57) {\framebox{\embedvideo{\includegraphics[width=1.175in]{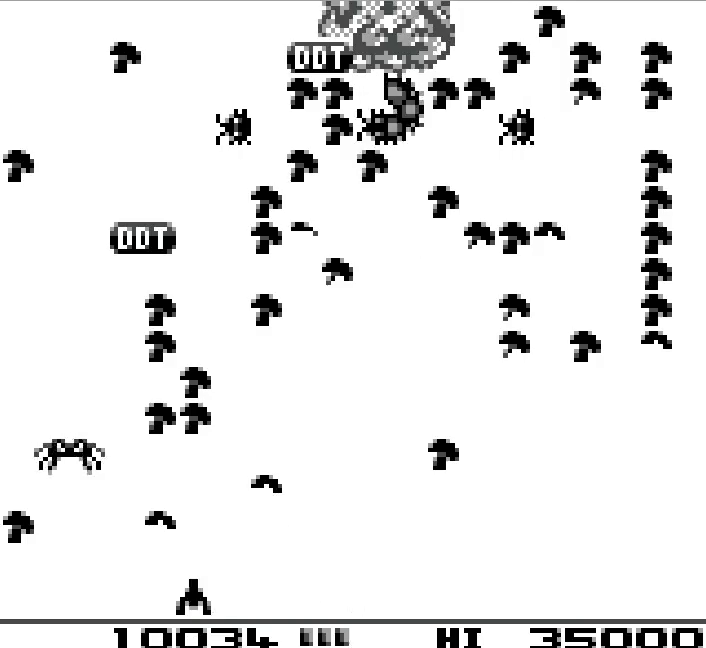}}{Millipede-1-2-small.mp4}}};
         \node at (11.7,3.68) {\framebox{\embedvideo{\includegraphics[width=1.175in]{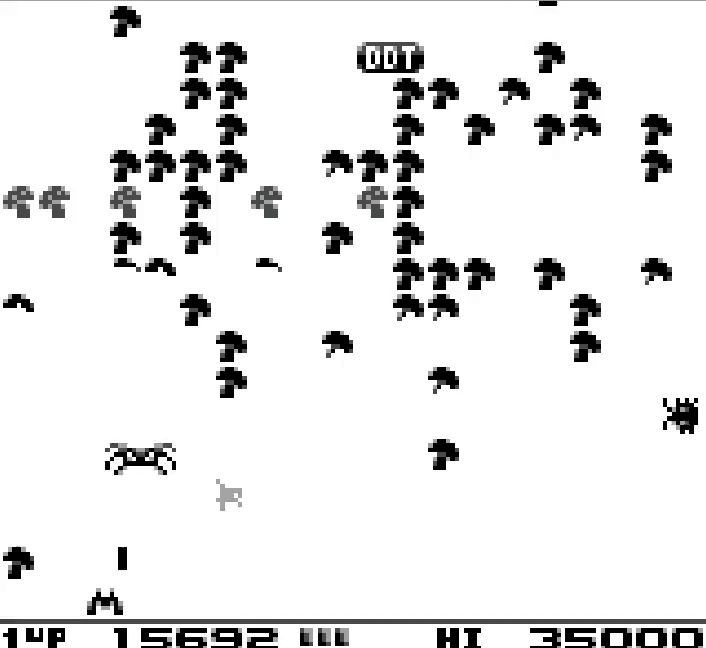}}{Millipede-1-3-small.mp4}}};
         \node at (11.7,0.79) {\framebox{\embedvideo{\includegraphics[width=1.175in]{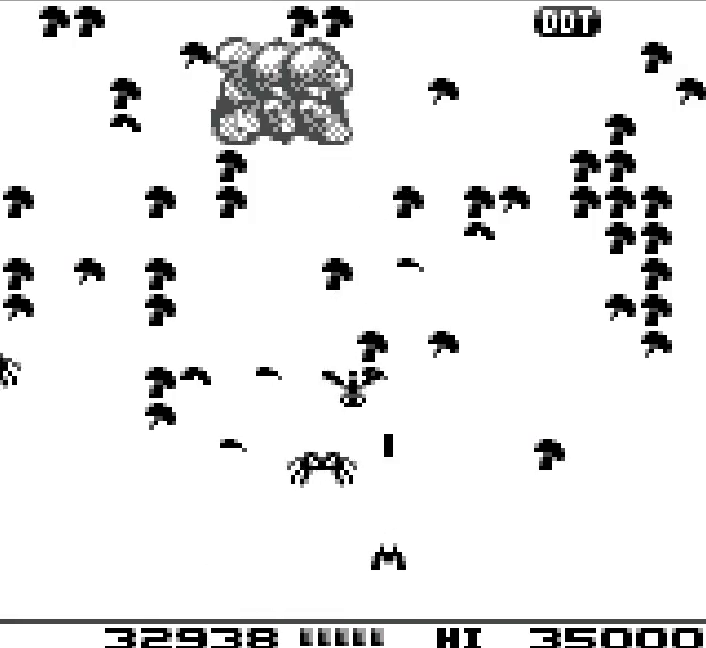}}{Millipede-1-4-small.mp4}}};
         \node at (11.7,-2.1) {\framebox{\embedvideo{\includegraphics[width=1.175in]{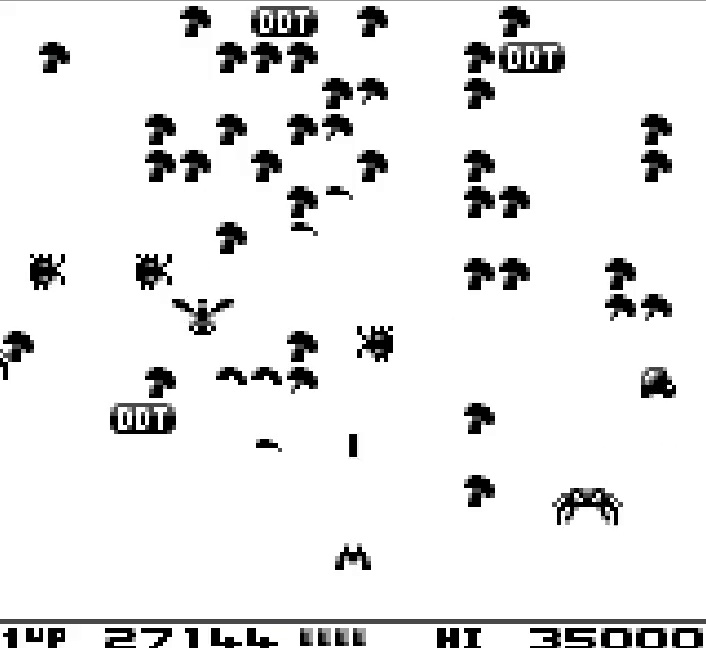}}{Millipede-1-5-small.mp4}}};
         \node at (11.7,-4.99) {\framebox{\embedvideo{\includegraphics[width=1.175in]{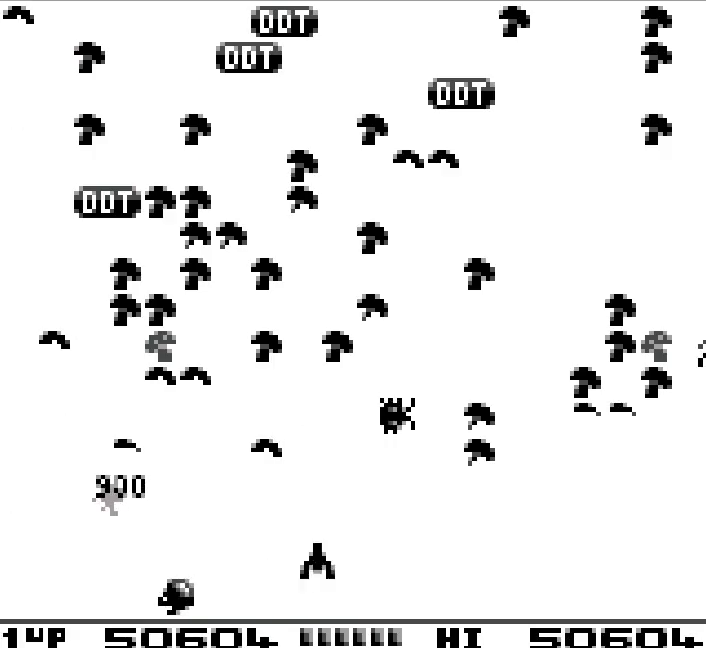}}{Millipede-1-6-small.mp4}}};
         \node at (11.7,-7.88) {\framebox{\embedvideo{\includegraphics[width=1.175in]{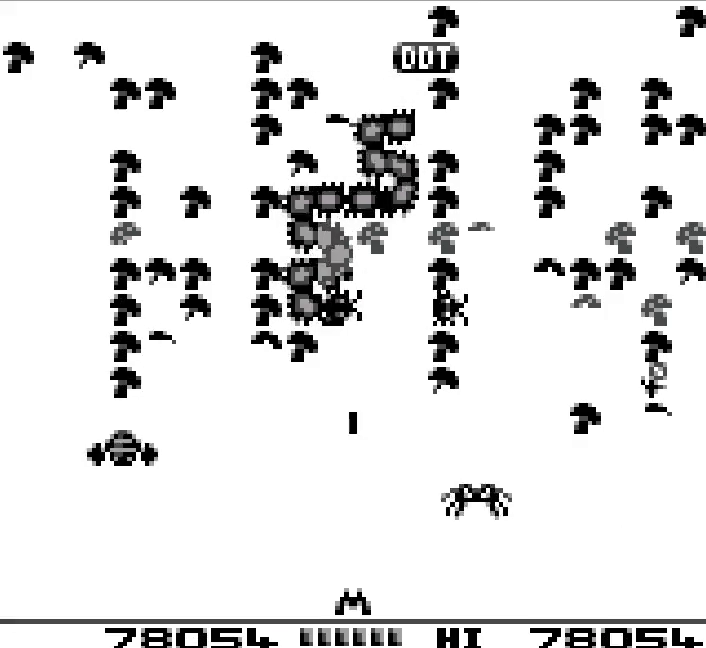}}{Millipede-1-7-small.mp4}}};
         \node at (11.7,-9.7) {(f)};

         \newcommand*\blackcircled[2]{\tikz[baseline=(char.base)]{
         \node[circle,black,draw,scale=#2,inner sep=1pt] (char) {#1};}}

         \node at (9.825,9.46) {\textcolor{white}{\bf \blackcircled{\small \textcolor{black}{1}}{1}}};
         \node at (9.825,6.57) {\textcolor{white}{\bf \blackcircled{\small \textcolor{black}{2}}{1}}};
         \node at (9.825,3.68) {\textcolor{white}{\bf \blackcircled{\small \textcolor{black}{3}}{1}}};
         \node at (9.825,0.79) {\textcolor{white}{\bf \blackcircled{\small \textcolor{black}{4}}{1}}};
         \node at (9.825,-2.1) {\textcolor{white}{\bf \blackcircled{\small \textcolor{black}{5}}{1}}};
         \node at (9.825,-4.99) {\textcolor{white}{\bf \blackcircled{\small \textcolor{black}{6}}{1}}};
         \node at (9.825,-7.88) {\textcolor{white}{\bf \blackcircled{\small \textcolor{black}{7}}{1}}};

         \node at (-5.25,8.5) {\textcolor{white}{\bf \blackwhitecircled{\small \textcolor{white}{1}}{1}}};
         \node at (-4.95,7.05) {\textcolor{white}{\bf \blackwhitecircled{\small \textcolor{white}{2}}{1}}};
         \node at (-4.85,6.05) {\textcolor{white}{\bf \blackwhitecircled{\small \textcolor{white}{3}}{1}}};
         \node at (-4.6,4.25) {\textcolor{white}{\bf \blackwhitecircled{\small \textcolor{white}{4}}{1}}};
         \node at (-4.45,2.05) {\textcolor{white}{\bf \blackwhitecircled{\small \textcolor{white}{5}}{1}}};
         \node at (-4.025,-2.75) {\textcolor{white}{\bf \blackwhitecircled{\small \textcolor{white}{6}}{1}}};
         \node at (-3.725,-6.5) {\textcolor{white}{\bf \blackwhitecircled{\small \textcolor{white}{7}}{1}}};

      \end{tikzpicture}}
   \caption[]{Depictions of the agent environment dynamics understanding, search performance, and implementations of agent gameplay behaviors for \emph{Millipede}.  (a) Average, smoothed costs for the different reinforcement learning approaches.  Lower values are better.  (b) Average, smoothed acquired knowledge of the environment transition dynamics, as a function of how much the policy differs from the highest-performing policy uncovered.  Here, we use policy-to-policy cross-entropy.  Higher values are better.  (c) Per-episode smoothed averages of the agent's relative costs for pseudo-arc-length path-following with an adaptive step size.  (d) Plot of the agent's smoothed average action-selection probability as a function of the number of learning episodes.  All averages are obtained over thirty Monte Carlo trials.  Note that the probabilities do not necessarily sum to one for each reported episode, since they are averages.  (e) A bifurcation diagram that shows, on average, when a new solution branch is encountered for the value of information.  Each added color denotes the emergence of a new branch.  For (a)--(e), we mark phase-transition boundaries where the agent skill set noticeably changes.  We refer to these as gameplay-skill boundaries in our discussions.  (f) Videos of the implemented game-play behaviors for each of the major gameplay phases.  Their corresponding episodes are highlighted in (a).  We recommend viewing this document within Adobe Acrobat DC; click on an image and enable content to start playback of the corresponding video.\vspace{-0.2cm}}
   \label{fig:exper-millipede}
\end{figure}

\begin{figure}[t!]
   \centering

   \scalebox{0.775}{
      \begin{tikzpicture} 
         \node at (-3.325,11.775) {\includegraphics[height=0.665in]{compmethods-millipede-1.pdf}};
         \node at (0.81,11.645) {\includegraphics[height=0.57in]{compmethods-2.pdf}};
         \node at (3.915,11.645) {\includegraphics[height=0.57in]{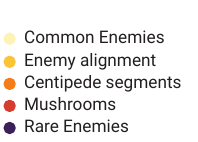}};
         \node at (7.325,11.875) {\includegraphics[height=0.76in]{branches-1-labels.pdf}};

         \node[] at (1.0,-0.125) {\includegraphics[width=6.25in]{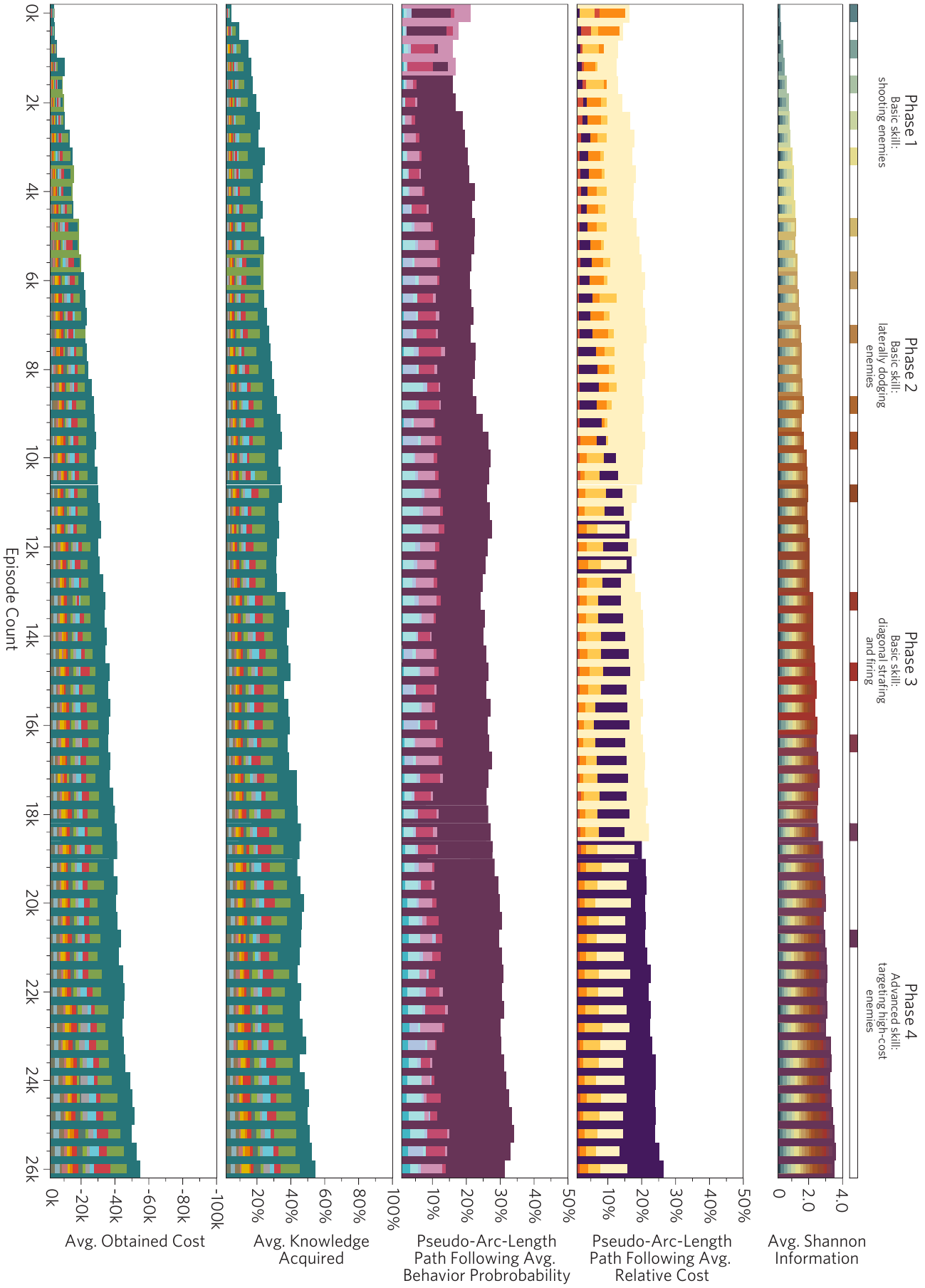}};

         \draw[line width=0.125mm, color={rgb,255:red,142;green,142;blue,142}] (-6.075,6.9) -- (-3.225,6.9);
         \draw[line width=0.125mm, color={rgb,255:red,142;green,142;blue,142}] (-3.06,6.9) -- (-0.21,6.9);
         \draw[line width=0.125mm, color={rgb,255:red,142;green,142;blue,142}] (-0.04,6.9) -- (2.815,6.9);
         \draw[line width=0.125mm, color={rgb,255:red,142;green,142;blue,142}] (2.975,6.9) -- (5.8,6.9);
         \draw[line width=0.125mm, color={rgb,255:red,142;green,142;blue,142}] (6.4175,6.9) -- (7.52,6.9);

         \draw[line width=0.125mm, color={rgb,255:red,142;green,142;blue,142}] (-6.075,2.4) -- (-3.225,2.4);
         \draw[line width=0.125mm, color={rgb,255:red,142;green,142;blue,142}] (-3.06,2.4) -- (-0.21,2.4);
         \draw[line width=0.125mm, color={rgb,255:red,142;green,142;blue,142}] (-0.04,2.4) -- (2.815,2.4);
         \draw[line width=0.125mm, color={rgb,255:red,142;green,142;blue,142}] (2.975,2.4) -- (5.8,2.4);
         \draw[line width=0.125mm, color={rgb,255:red,142;green,142;blue,142}] (6.4175,2.4) -- (7.52,2.4);

         \draw[line width=0.125mm, color={rgb,255:red,142;green,142;blue,142}] (-6.075,-3.5) -- (-3.225,-3.5);
         \draw[line width=0.125mm, color={rgb,255:red,142;green,142;blue,142}] (-3.06,-3.5) -- (-0.21,-3.5);
         \draw[line width=0.125mm, color={rgb,255:red,142;green,142;blue,142}] (-0.04,-3.5) -- (2.815,-3.5);
         \draw[line width=0.125mm, color={rgb,255:red,142;green,142;blue,142}] (2.975,-3.5) -- (5.8,-3.5);
         \draw[line width=0.125mm, color={rgb,255:red,142;green,142;blue,142}] (6.4175,-3.5) -- (7.52,-3.5);

         \node at (-4.65,-11.425) {(a)};
         \node at (-1.7,-11.425) {(b)};
         \node at (1.4,-11.425) {(c)};
         \node at (4.45,-11.425) {(d)};
         \node at (6.925,-11.425) {(e)};

         \setlength{\fboxrule}{0.75pt}
         \setlength{\fboxsep}{0.025pt}
         \node at (11.7,9.46) {\framebox{\embedvideo{\includegraphics[width=1.175in]{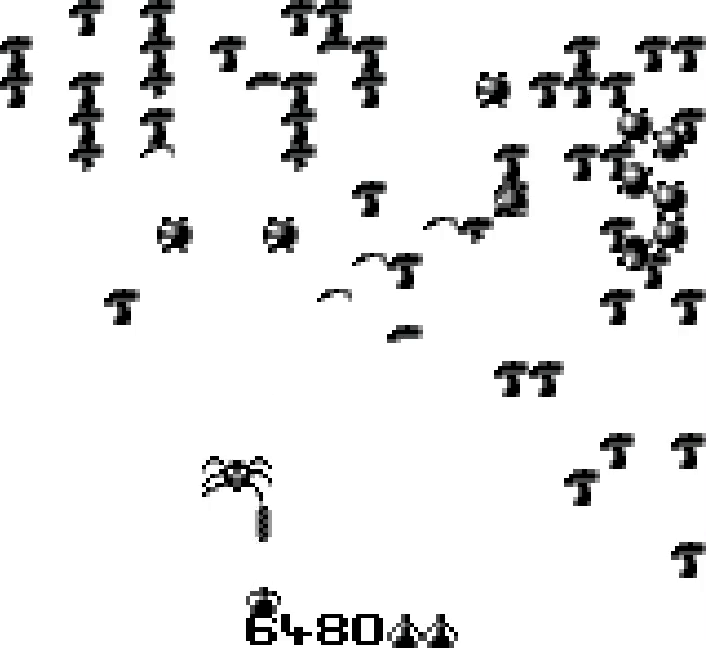}}{Centipede-1-1-small.mp4}}};
         \node at (11.7,6.57) {\framebox{\embedvideo{\includegraphics[width=1.175in]{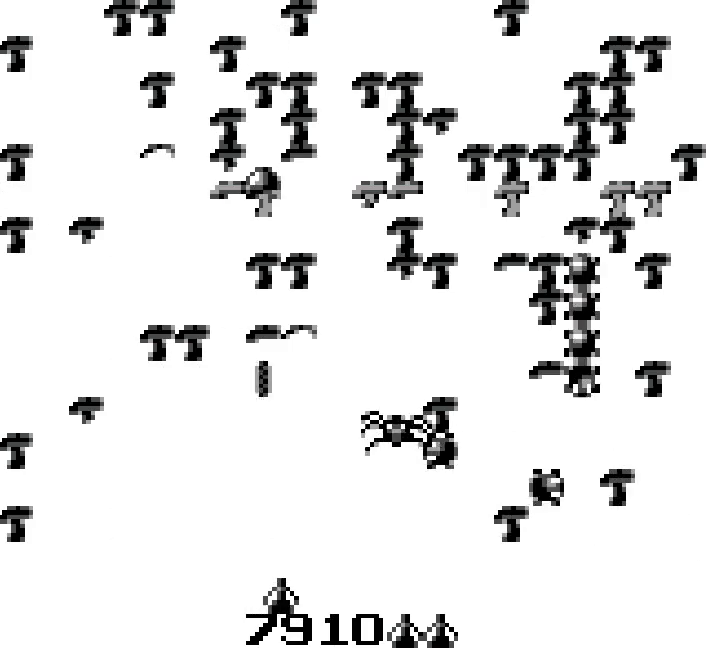}}{Centipede-1-2-small.mp4}}};
         \node at (11.7,3.68) {\framebox{\embedvideo{\includegraphics[width=1.175in]{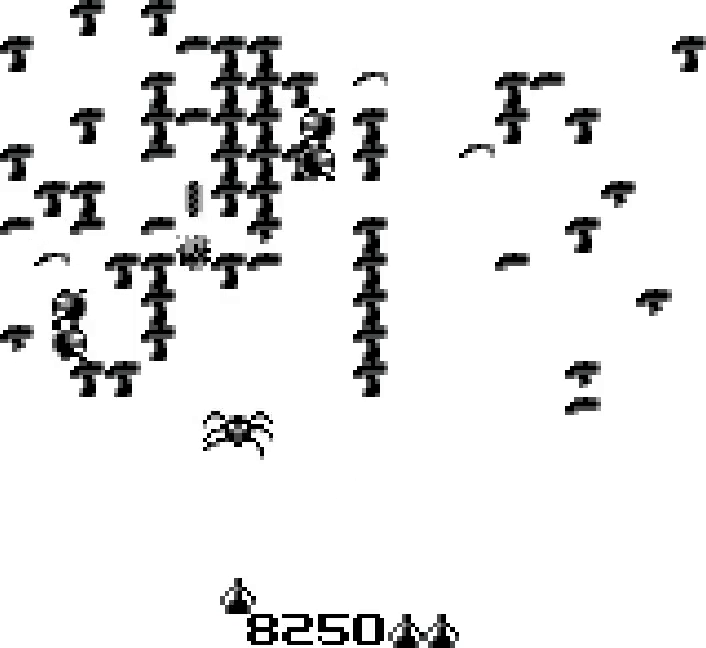}}{Centipede-1-3-small.mp4}}};
         \node at (11.7,0.79) {\framebox{\embedvideo{\includegraphics[width=1.175in]{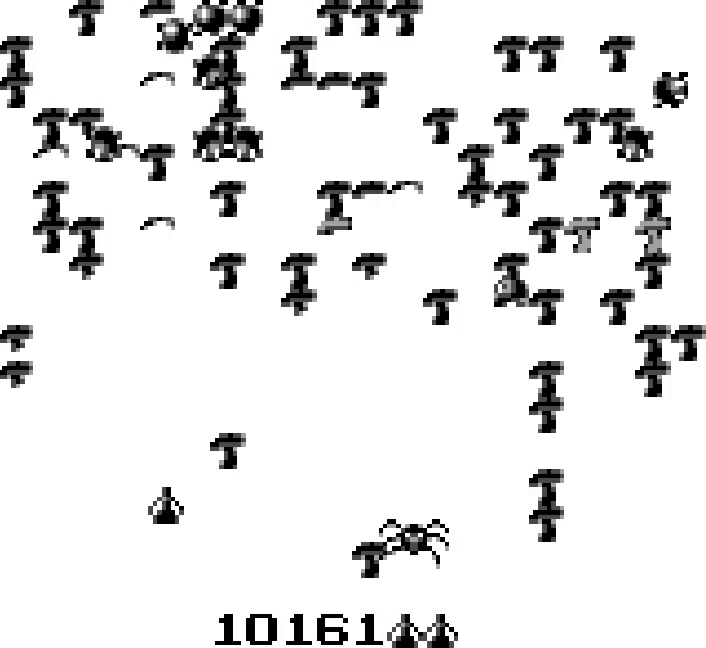}}{Centipede-1-4-small.mp4}}};
         \node at (11.7,-2.1) {\framebox{\embedvideo{\includegraphics[width=1.175in]{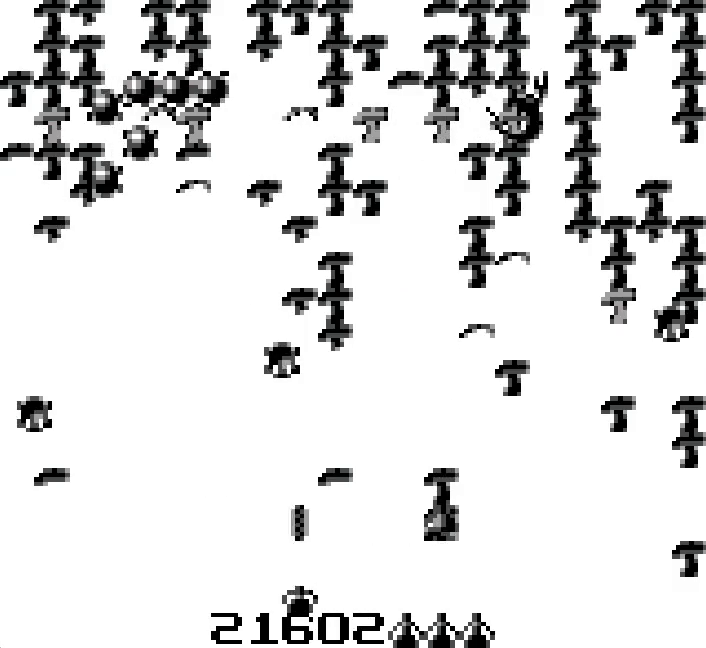}}{Centipede-1-5-small.mp4}}};
         \node at (11.7,-4.99) {\framebox{\embedvideo{\includegraphics[width=1.175in]{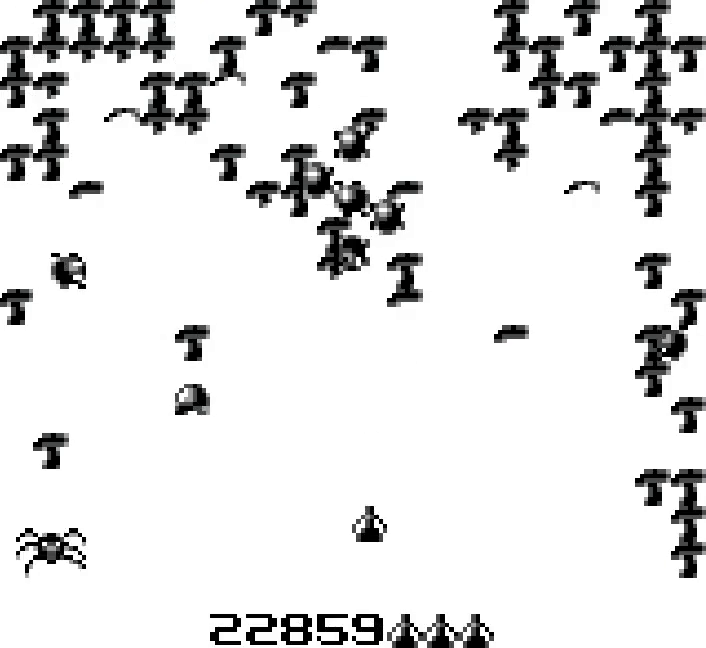}}{Centipede-1-6-small.mp4}}};
         \node at (11.7,-7.88) {\framebox{\embedvideo{\includegraphics[width=1.175in]{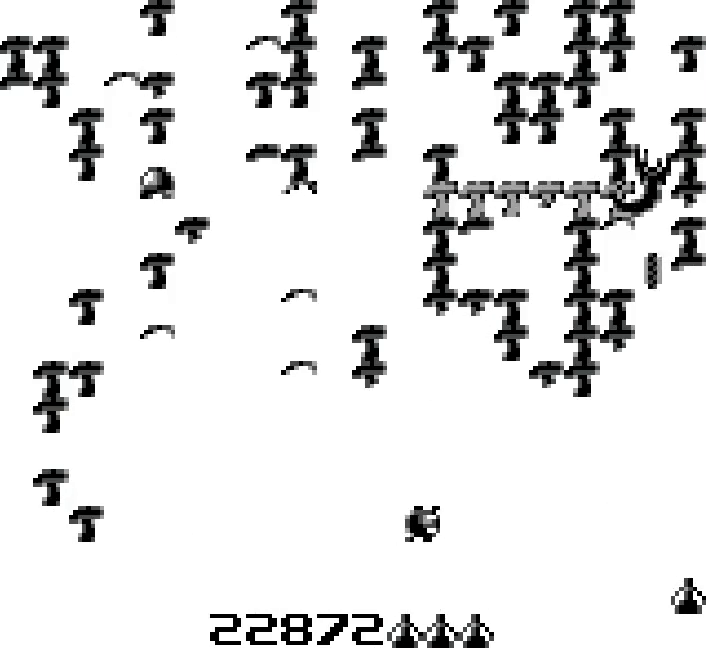}}{Centipede-1-7-small.mp4}}};
         \node at (11.7,-9.7) {(f)};

         \newcommand*\blackcircled[2]{\tikz[baseline=(char.base)]{
         \node[circle,black,draw,scale=#2,inner sep=1pt] (char) {#1};}}

         \node at (9.825,9.46) {\textcolor{white}{\bf \blackcircled{\small \textcolor{black}{1}}{1}}};
         \node at (9.825,6.57) {\textcolor{white}{\bf \blackcircled{\small \textcolor{black}{2}}{1}}};
         \node at (9.825,3.68) {\textcolor{white}{\bf \blackcircled{\small \textcolor{black}{3}}{1}}};
         \node at (9.825,0.79) {\textcolor{white}{\bf \blackcircled{\small \textcolor{black}{4}}{1}}};
         \node at (9.825,-2.1) {\textcolor{white}{\bf \blackcircled{\small \textcolor{black}{5}}{1}}};
         \node at (9.825,-4.99) {\textcolor{white}{\bf \blackcircled{\small \textcolor{black}{6}}{1}}};
         \node at (9.825,-7.88) {\textcolor{white}{\bf \blackcircled{\small \textcolor{black}{7}}{1}}};

         \node at (-5.315,7.75) {\textcolor{white}{\bf \blackwhitecircled{\small \textcolor{white}{1}}{1}}};
         \node at (-5.125,5.75) {\textcolor{white}{\bf \blackwhitecircled{\small \textcolor{white}{2}}{1}}};
         \node at (-4.95,3.15) {\textcolor{white}{\bf \blackwhitecircled{\small \textcolor{white}{3}}{1}}};
         \node at (-4.75,0.75) {\textcolor{white}{\bf \blackwhitecircled{\small \textcolor{white}{4}}{1}}};
         \node at (-4.685,-2.05) {\textcolor{white}{\bf \blackwhitecircled{\small \textcolor{white}{5}}{1}}};
         \node at (-4.535,-4.75) {\textcolor{white}{\bf \blackwhitecircled{\small \textcolor{white}{6}}{1}}};
         \node at (-4.275,-8.25) {\textcolor{white}{\bf \blackwhitecircled{\small \textcolor{white}{7}}{1}}};

      \end{tikzpicture}}
   \caption[]{Depictions of the agent environment dynamics understanding, search performance, and implementations of agent gameplay behaviors for \emph{Centipede}.  For plot descriptions, refer to \cref{fig:exper-millipede}.\vspace{-0.2cm}}
   \label{fig:exper-centipede}
\end{figure}

\phantomsection\label{sec5.1.1}
%%%%%%%%%%%%%%%%%%%%%%%%%%%%%%%%%%%%%%%%%%%%%%%%%%%%%%%%%%%%%%%%%%%%%%%%%%%
%%%%%%%%%%%%%%%%%%%%%%%%%%%%%%%%%%%%%%%%%%%%%%%%%%%%%%%%%%%%%%%%%%%%%%%%%%%
\subsection*{\small{\sf{\textbf{5.1.1.$\;\;\;$Path-Following Performance}}}}

We first illustrate empirical properties of pseudo-arc-length path-following, as they relate to agent performance.

As shown in \cref{fig:exper-millipede,fig:exper-centipede}, the exploration-rate adjustment strategy profoundly influences the agent's obtainable costs.  A variable, environment-sensitive steplength leads to the best results.  For \emph{Millipede}, there were marked decreases in the average costs at phase-transition boundaries when using pseudo-arc-length path-following with a variable steplength; this is presented in \cref{fig:exper-millipede}(a).  Reaching these episodes coincided with an improved coverage of the state space and hence a better understanding of the environment transition dynamics, which is codified by the policy's matrix-based cross-entropy \cite{SledgeIJ-jour2022a} in \cref{fig:exper-millipede}(b).  When using a fixed steplength of a small size, path following often cannot adequately adapt to the local geometry of the value-of-information Lagrangian solution curve.  While cost decreases can be observed near gameplay-skill boundaries, significantly more episodes are often needed to achieve comparable performance costs.  Consistently large changes in the exploration rate also led to poorly performing policies.  In either case, the overall accrued costs are typically worse compared the adaptive case in these simulations.  The agent also does not either as thoroughly or as broadly investigate the state space as when variable steplengths are used.  This is alluded to in \cref{fig:exper-millipede}(b).  The results in \cref{fig:exper-centipede} highlight that these trends hold for \emph{Centipede}.%, albeit for three exploration-rate values.

The type of path following used also has an impact on agent performance.  When relying on parameter path-following with an adaptive steplength regime, only initial cost decreases are typically encountered.  \hyperref[fig:exper-millipede]{Figures 5.1}(a) and \ref{fig:exper-centipede}(a) highlight that these occur the initial stages of learning.  Marginal cost decreases are often realized past this point.  Some cost increases are commonly witnessed instead.  \hyperref[fig:exper-millipede]{Figures 5.1}(b) and \ref{fig:exper-centipede}(b) indicate, however, that the agents continue to investigate parts of the state action-space despite the lack of cost improvement.  The agents are, after all, consistently exposed to novel states due to the random initializations of the environments.  The rate at which they search the space, particularly the available action choices, is subdued compared to earlier during training, though.  The agent consequently does not update its policy well and hence has a poor understanding of the transition dynamics.  Similar results are witnessed when fixed steplengths are used, regardless of their magnitude.  Comparatively, pseudo-arc-length path-following with an adaptive steplength tends to search much of the space.  The state-visitation plots in \cref{fig:millipedestatespace,fig:centipedestatespace} illustrate that the transition uncertainty is low, suggesting that the agent has extensively interacted with the environment and may have some insight of how to complete various objectives well (see \hyperref[secB]{Appendix B}).

\vspace{0.15cm}\noindent{\small{\sf{\textbf{Discussions.}}}} Finding value-of-information solutions is challenging, as there are many possible ways in which the solutions can evolve.  Here, we have demonstrated that our path-following procedure with an adaptive steplength does well for \emph{Millipede} and \emph{Centipede}.  It thoroughly investigates the state-action space, as we discuss in an online appendix, and hence obtains low costs.  Parameter continuation, in contrast, often does not.  It easily becomes stuck at certain stages of the learning process.  Upon subsequent investigations, we find that this occurs when encountering simple folds in the solution curves, which were where the Jacobian of the Lagrangian was singular.  Meaningful updates to the policy cease near these regions, and its performance is far worse as a consequence.

Pseudo-arc-length path-following works well for two reasons.  First, it can continue past simple folds, which we prove in the appendix.  This enables the search to proceed to supercritical bifurcated branches, which lead to finer partitionings of the policy and hence the formation of new agent behavior clusters.  We discuss this aspect more in the online appendix.  Secondly, pseudo-arc-length path-following relies on robust predictor-corrector continuation.  Virtually any point on the predicted curve segment will locally converge to some point back on the solution path.  Convergence is guaranteed provided the predicted starting point is sufficiently close to the solution path.  This indicates that the correction errors are independent of the corrector-iteration history and are solely determined by the iteration termination criterion, at the current step, provided there is convergence.  Any step along the predicted segment is, in principle, acceptable for which the resulting starting point is within the convergence domain of the corrector.  Solutions to the value of information will therefore almost always be uncovered for a Markov-decision-process abstraction.  One of the few exceptions are singular points where the error-surface curvature is too great to guarantee a retraction to the solution path (see \hyperref[secA]{Appendix A}).  Similar guarantees are difficult to furnish for parameter path-following.

Automatically adjusting the corrector step size also has an influence on policy performance.  The amount by which the step size, and hence the exploration rate, improves performance is dictated by the size of the local convergence region around stationary points.  Along certain sections of solution branches, the exploration rate increases slowly.  This typically occurs whenever the solution trajectories for the value-of-information Lagrangian gradient has steep curvatures.  Such areas coincide with segments of the trajectory just before and after symmetry-breaking bifurcations and hence either the emergence of new agent behaviors or the rapid refinement of existing ones.  Once these behaviors sufficiently stabilize, larger-magnitude updates can be made up to bifurcation points.  The solution curves are relatively flat in these areas and hence large steps can be taken without the risk of diverging.  The exploration-rate adjustment can also change dramatically to essentially bypass saddle-node bifurcations where symmetry is not broken.

When using small fixed steplengths, decreases in cost occurred more slowly compared to the adaptive case.  This is because path following cannot take advantage of flat-curvature regions of the solution trajectory.  If the constant steplength is too small, then a decreasing sequence of correction iterations may be encountered for which empirical policy convergence is slow.  For higher per-iteration adjustments, there is a chance that certain bifurcations will be missed.  The policies can therefore stagnate after some application-dependent number of episodes. 

The approach that we consider in this paper is but one possibility for investigating multiple solution branches.  Many alternatives often have either theoretical or empirical issues that impede learning, though.  For instance, explicit exploration rates for which state-group phase changes occur can be explicitly derived, assuming sufficient knowledge of the environment dynamics.  They are, however, difficult to explicitly compute a priori.  We are not aware of any way to do this well for the complicated environments that we consider in our experiments.  Therefore, in practice, we would need to repeatedly solve eigenvalue problems that rely on the second variation of the value-of-information Lagrangian.  A sufficient amount of agent-environment interactions is needed to ensure that spurious bifurcations are not returned.  If an erroneous branch switch occurs, then agent behaviors may unnecessarily require several episodes to materialize.  Pseudo-arc-length path-following, in comparison, uncovers phase transitions automatically during learning by evaluating determinants of the value-of-information Hessian (see \hyperref[secA]{Appendix A}).  While estimating this Hessian can be costly, it needs to be formed much more infrequently than would be required for eigendecompositions.

\phantomsection\label{sec5.1.2}
%%%%%%%%%%%%%%%%%%%%%%%%%%%%%%%%%%%%%%%%%%%%%%%%%%%%%%%%%%%%%%%%%%%%%%%%%%%
%%%%%%%%%%%%%%%%%%%%%%%%%%%%%%%%%%%%%%%%%%%%%%%%%%%%%%%%%%%%%%%%%%%%%%%%%%%
\subsection*{\small{\sf{\textbf{5.1.2.$\;\;\;$Path-Following Implemented Behaviors}}}}

We now illustrate that state-action groups are formed when using path following with value of information.  These are a byproduct of searching bifurcating branches of the first-order flow.  We also describe the effects of the groupings on the realized agent behaviors and how they influence the observed cost reductions. 

As indicated by \hyperref[fig:exper-millipede]{figures 5.1}(e) and \ref{fig:exper-centipede}(e), adapting the exploration rate induces bifurcations in the state-action assignment.  This yields phase transitions that increase the number of state-action groups, which is shown in \hyperref[fig:appendixa-bifurcation]{figure A.1} (see \hyperref[secA]{Appendix A}).  The increased state-action-space quantization granularity permits the acquisition of new behaviors once the agent had accrued enough experience through exploration.

We first focus on the 16-to-15 branch in \hyperref[fig:exper-millipede]{figures 5.1}(e) and \ref{fig:exper-centipede}(e).  For many of the simulations, the initial bifurcation along the 16-uniform solution branch occurs early during learning.  Before this bifurcation, the agent largely performs a single action, regardless of the game state.  The preferred action is to remain stationary and continuously fire bolts.  This was a way to reliably decrease costs.  A limited number of movement actions are also sometimes favored, which are, typically, just erratic movements. 

Beyond the first few game levels, having a fixed agent becomes a detriment.  Waves of bees, dragonflies, and other enemies appear and rapidly deplete the agent's lives in \emph{Millipede}.  For \emph{Centipede}, spiders are the biggest threat.  It therefore is advantageous for the agent to move laterally to avoid being hit.  It also enables the agent to better target certain enemies.  Moving either left or right often becomes the action associated with the new state group that coincided with the 15- and 14-solution branches in \hyperref[fig:exper-millipede]{figures 5.1}(e) and \ref{fig:exper-centipede}(e).  The choice of the lateral-movement direction for the initial movement group is dictated by the accumulated experience.  The remaining state group is mostly associated with firing bolts, as this is the only way for the agent to reduce costs.  The firing of bolts would occur almost independently of the agent's and enemies' positions.  It is often advantageous for the agent to do this, since stray bolts can weaken and remove mushrooms. 

Moving mainly in a single lateral direction is highly restrictive in both games.  The agent could become stuck either near or at the edges of the environment, leaving itself open to attack from spiders and earwigs that emerge and leave in those areas.  About a sixth of the way through the overall learning process, a new state group usually forms due to a symmetry-breaking bifurcation.  One of three remaining directions would initially be chosen as the preferred action for this group.  Eventually, this action would often correspond to moving in the opposite lateral direction.  This choice yields the greatest cost reduction due to the agent's ability to target and dodge certain enemies and therefore continue playing without losing a life.  The top-most video in \hyperref[fig:exper-millipede]{figure 5.1}(f) shows that the agent initially would remain relatively stationary in certain parts of the game environment.  Only after additional updates, would the agent later move more frequently to target and avoid enemies.  This behavior is depicted in the second and third videos of \cref{fig:exper-millipede}(f) and the first and second videos in \cref{fig:exper-centipede}(f).

Two additional movement state groups often arise for increasing exploration rates, again due to symmetry-breaking bifurcations that occur early during the learning process.  These groups, which are associated with the 13- and 12-solution branches, implement either vertical or diagonal movements.  Such movements allow the agent to avoid enemies that traverse the bottom row of the play area where the agent spawns.  They also enable the agent to get closer to enemies, thereby reducing the amount of time between bolt fires and increasing the number of targeted enemies.  Additionally, the agent could move either above or below the bouncing spiders and nearby millipede and centipede segments, as captured by the fourth through seventh videos in \hyperref[fig:exper-millipede]{figures 5.1}(f) and \ref{fig:exper-centipede}(f).  This latter behavior extends the agent's lifetime in later game stages when multiple enemies would normally surround it.

We found that increasing the exploration rate later during training would begin to fragment existing movement state clusters to execute additional diagonal movements.  Further bifurcations would allow the agent to target high-cost enemies more quickly and effectively.  These typically occurred for the 10- through 6-solution branches.  The costs in \hyperref[fig:exper-millipede]{figures 5.1}(a) and \ref{fig:exper-centipede}(a) and action-selection probabilities in \hyperref[fig:exper-millipede]{figures 5.1}(c) and \ref{fig:exper-centipede}(c) substantiate this claim.  \Cref{fig:exper-millipede}(c) shows that high-cost enemies that rarely spawn quickly became a routine point source.  In \emph{Millipede}, The agent would, for instance, target beetles, since they put the agent at significant risk by turning mushrooms into near-indestructible flowers.  The agent would also begin to reliably shoot DDT canisters whenever enemies were present, as indicated by \cref{fig:exper-millipede}(d).  Doing so would markedly decrease costs.  It would also clear nearby patches of mushrooms, reducing the number of environmental obstructions and allowing the agent to more quickly destroy enemies.  It would also free the agent to target commonly spawning enemies, like spiders, that could be a nuisance to the agent.  Such behaviors are captured in the sixth and seventh videos in \ref{fig:exper-millipede}(f).  In \emph{Centipede}, the agent would target scorpions for similar reasons to the beetles in \emph{Millipede}.  This is illustrated in the sixth and seventh videos in \cref{fig:exper-centipede}(f).  Since there are fewer high-cost enemies in \emph{Centipede} compared to \emph{Millipede}, the overall cost contribution is well below that of the low-cost enemies, as shown in \cref{fig:exper-centipede}(e).

Taken together, the behaviors that emerge from these bifurcations explain the average cost decreases observed in \hyperref[fig:exper-millipede]{figures 5.1}(a) and \ref{fig:exper-centipede}(a) after halfway through the training process. 

By the end of training, it is common for about sixteen state-action groups to form.  All of these groups are highly context specific, as depicted in \cref{fig:millipedestateactionspace,fig:centipedestateactionspace} (see \hyperref[secB]{Appendix B}).  For instance, there are compound-action groups that facilitate moving and shooting along with remaining stationary and shooting, which are usually associated with the 5- through 2-solution branches.  The former compound action allows the agent to quickly destroy one enemy and align with another.  The latter compound action is useful whenever centipede and millipede segments are funneled down a corridor of mushrooms.  It also aids in clearing vertical strands of mushrooms, which partly explains the higher contributions of mushroom-based points during later stages of training in \hyperref[fig:exper-millipede]{figures 5.1}(d) and \ref{fig:exper-centipede}(d).  Other action groups, like remaining stationary, typically form during the few remaining bifurcations.  Such an action is preferred when the agent is unable to shoot, due to recently firing a bolt, and is also unable to safely move, due to the presence of nearby enemies.  In all of these cases, the corresponding grouped states are strongly correlated with varying degrees of cost reductions, which can be seen when relating \cref{fig:millipedestateactionspace,fig:centipedestateactionspace}, respectively, to \cref{fig:millipedestatespace,fig:centipedestatespace} (see \hyperref[secB]{Appendix B}); we discuss these aspects, and others, in further detail in the associated online appendix.  

All of the above groups are formed by consistently switching to good solution branches after a bifurcation occurs.  However, as shown in \hyperref[fig:exper-millipede]{figures 5.1}(e) and \ref{fig:exper-centipede}(e), the agents can remain on earlier branches, due to our use of parallel search.  Few to no bifurcations are typically encountered on such branches, even as the exploration rate is adjusted.  This implies that the number of state groups remains mostly static despite the agent accruing more experience.  Advanced behaviors, such as evading enemies, are largely not realized as a consequence.  Agent performance often stagnates from a lack of meaningful policy updates.  Similar issues are encountered when poorly choosing an initial exploration rate.

\vspace{0.15cm}\noindent{\small{\sf{\textbf{Discussions.}}}} Here, we have established that bifurcations along the value-of-information solution trajectory are connected with the development and refinement of the agent's context-specific action responses.

Where bifurcations happen on the value-of-information solution trajectory is application dependent.  A search rate that is either too high or too low may cause the exploration process to move onto a sub-optimal solution branch and thus slow learning.  Having an approach that can detect these phase transitions and appropriately adjust the search amount to pursue good solution branches is crucial for quickly realizing good agent behaviors.  Our path-following methodology does just that.

Beyond detecting and switching between solution branches well, care must be taken in choosing a starting exploration rate when using path following.  Low rates seem to be better than high ones, in most situations, for helping to uncover good agent behaviors.  The preferred bifurcated trajectories that lead to cost-reduction acting choices will tend to be discovered for near-zero exploration rates.  We have empirically found that such trajectories emanate from the first encountered symmetry-breaking bifurcation.  Beginning the search process with too high an exploration rate leads to the possibility of missing this first bifurcation, especially if good estimates of the action value-function magnitudes have not been obtained by that stage in the learning process.  Alternate branches may therefore be encountered that do not split in the same way.  The solution iterates could hence become stuck on a branch where the underlying Shannon-information bound would not change enough to precipitate the creation of new state groups and hence the formation of potentially novel agent behaviors.  Backtracking might be necessary in an attempt to discover equilibria on different branches, which can impede the learning process.

The above results also illustrate a unique property of the value of information---it partitions the states according to the state-action value-function and assigns a, mostly distinct, action-selection probability vector to each state group.  New rows in this partition, representing the materialization of new state groups, are introduced whenever the Hessian of the Lagrangian is singular for a given exploration rate and once enough knowledge of the environment dynamics has been acquired by the agent.  These singular-solution points are accompanied by so-called symmetry-breaking bifurcations.  These are forks in the solution surface where the solutions are fixed by sub-groups of the algebraic permutation group with a certain number of symbols.  Following the bifurcation direction suggested by the Equivariant Branching Lemma leads the solution to a trajectory with a permutation group containing one less symbol.  Chains of such sub-groups with decreasing numbers of symbols are encountered as path-following continues along stable, supercritical solution branches.  Eventually, the solution iterates lie on a symmetry-less solution branch and no further bifurcations are generally possible.  Along this symmetry-less branch are clustered action-selection policies with as many state groups as unique states.  We have previously demonstrated that this symmetry-less branch is linked to a non-aggregated Markov decision process \cite{SledgeIJ-jour2018a}.  All of the previous branches have Markov decision processes with aggregated Markov chains.  They hence correspond to increasingly simple reinforcement learning problems as the number of permutation-group symbols increases. 

It is important to note that the state-action clustering offered by the path-following-based value of information is functionally similar to explicit state abstraction.  However, it is more practically appealing.  The value of information does not require knowledge of an environment transition function, unlike \cite{AbelD-conf2016a,DietterichTG-jour2000a,BoutilierC-jour2000a,GivenR-jour2003a}, when forming these groups.  The value of information is hence readily applicable to producing human-understandable policies for arbitrary problems.  We will emphasize this claim in a future paper.  Moreover, no empirical convergence issues are typically encountered when using the value of information.  This is in contrast to the irrelevant-state-variable method of \cite{JongNK-conf2005a}, which may not produce policies for an abstract Markov decision process that are optimal for the original process.

\phantomsection\label{sec5.2}
%%%%%%%%%%%%%%%%%%%%%%%%%%%%%%%%%%%%%%%%%%%%%%%%%%%%%%%%%%%%%%%%%%%%%%%%%%%
%%%%%%%%%%%%%%%%%%%%%%%%%%%%%%%%%%%%%%%%%%%%%%%%%%%%%%%%%%%%%%%%%%%%%%%%%%%
\subsection*{\small{\sf{\textbf{5.2.$\;\;\;$Comparative Performance}}}}

We now compare pseudo-arc-length path-following with three alternate action searches, which are epsilon-greedy, soft-max, and value-of-information exploration.  For each technique, we consider a variety of strategies for adjusting the exploration rate.  To provide a fair comparison, each approach relies on the same coupled $Q$-learning process with experience generalization.

As shown in \hyperref[fig:exper-millipede]{figures 5.1}(a) and \ref{fig:exper-centipede}(a), none of these other approaches perform as well as the value of information with pseudo-arc-length path-following.  Constant-exploration searches often do the worst toward the latter half of learning.  This occurs even when a reasonable action-selection rate is discerned after many simulations.

It is well established that epsilon-greedy exploration can converge to optimal policies, in certain situations, as the number of episodes grows.  Modifications of soft-max and value-of-information selection, which ensure that the action-probability update is a contraction operator, allow these techniques to have similar guarantees.  For both games, however, convergence to a low-cost policy does not occur within the number of episodes that we considered.  This can be seen in \hyperref[fig:exper-millipede]{figures 5.1}(a).  The results are worse than pseudo-arc-length path-following by anywhere forty to almost seventy percent, depending on the chosen methodology.  Using a linear exploration-rate decay schedule leads to poorer results, as does considering fixed action-exploration amounts.  \hyperref[fig:exper-millipede]{figures 5.1}(b) does show, however, that the value of information outperformed parameter path-following.  The remaining methods often did in the later stages of learning.

\vspace{0.15cm}\noindent{\small{\sf{\textbf{Discussions.}}}} Our results highlight the utility of pseudo-arc-length path-following for the value of information.  Regardless of how we tune the parameters for either epsilon-greedy, soft-max, or expectation-maximization-based value-of-information search, neither are able to reach similar costs in the same number of episodes.  There are two reasons for this.  Foremost, fixed-update search schedules cannot exploit well the local geometry of the solution curves.  They may change the exploration rate either too greatly or too little across an episode sequence, which impacts policy performance.  Heuristic schemes, such as ones relying on cross-entropy, may still suffer the same issues, despite being somewhat sensitive to the learning dynamics.  This is because they typically rely on pre-specified exploration-rate adjustments.  Secondly, with the exception of the value of information, these alternate exploration strategies must investigate the entire state-action space, not a quantized version of it where the Markov decision process has been aggregated.  They hence must contend with a much more difficult learning problem, as each state has the potential to be assigned a unique action.  Several more learning episodes are required, as a result, to achieve good performance.

These results also validate that pseudo-arc-length path-following scales well to high-dimensional state-action spaces.  Path following repeatedly discovers, switches to, and traverses solution branches that permit seemingly continuous improvements in agent behaviors.  For \emph{Centipede} and \emph{Millipede}, such behaviors entail initially shooting at and dodging enemies, as we explained in the previous section.  Later, the agents utilize aspects of the environment to quickly score points.  The alternate search mechanisms, in contrast, do not appear to scale as well.  They hence often fail to implement crucial gameplay behaviors before training concluded.  For instance, throughout many of the early episodes, the agents simply oscillate in a given area without shooting.  Such behavior sometimes persists later during training, increasing the chances that the agent would collide with an enemy.  Jerky movements are often witnessed, even though action smoothing is used.  This typically prevents reliably shooting highly mobile enemies like spiders.  It made it difficult to also track and destroy centipede and millipede segments.  The agents would frequently forgo targeting high-cost enemies.  They appeared to almost randomly shoot, even if no enemies were nearby.

Curiously, parameter path-following performed reasonably well to these alternate search mechanisms.  This was despite being trapped by simple folds along the solution curve.  Subsequent analyses revealed this was due to the agent's preference to remain nearly motionless and continuously fire bolts.  Doing so enabled parameter path-following to reliably accrue more points than haphazardly moving throughout the environment and shooting at non-periodic intervals, which was the standard game-play tactic for epsilon-greedy and soft-max agents.  Such behaviors for parameter path-following emerged due to the implicit action-state partitioning functionality offered by the value of information.

%%%%%%%%%%%%%%%%%%%%%%%%%%%%%%%%%%%%%%%%%%%%%%%%%%%%%%%%%%%%%%%%%%%%%%%%%%%
%%%%%%%%%%%%%%%%%%%%%%%%%%%%%%%%%%%%%%%%%%%%%%%%%%%%%%%%%%%%%%%%%%%%%%%%%%%
\subsection*{\small{\sf{\textbf{6.$\;\;\;$Conclusions}}}}\addtocounter{section}{1}

The value of information is a constrained, information-theoretic criterion.  It describes the maximum benefit that can be obtained from a piece of information for either increasing expected rewards or reducing average costs.  We have previously shown that this property facilitates optimal decision-making under uncertainty.  It is hence well suited for addressing the exploration-exploitation dilemma in reinforcement learning.

Converting the value of information into an unconstrained criterion gives rise to a free parameter that dictates the action exploration rate.  Here, we propose a principled way of adjusting this parameter during learning.  This approach involves first characterizing equilibria conditions of a dynamical system associated with the value-of-information Lagrangian for changing parameter values.  Knowledge of these conditions permits the formulation of a tangent vector to map the policy and Lagrange multipliers for the current equilibrium to a neighborhood around a new equilibrium.  There is no guarantee that this new initial set of variables actually lies on a solution path traced by the dynamical system, though.  A projection-based correction is used to force the intermediate variables back near a solution path and hence ensure that they are equilibria for an updated exploration rate.  Alternating between guessing and correcting continues until some terminal exploration rate is reached.  Theoretical convergence to the best value-of-information policy associated with that exploration rate is guaranteed.  Convergence to the global-best policy can also be achieved.

Our simulations highlight that this approach does well for discrete state-action spaces where tabular policies can be used.  For the Nintendo GameBoy environments \emph{Centipede} and \emph{Millipede}, we show that pseudo-arc-length path-following can outperform parameter path-following.  The latter often cannot progress past simple folds in the solution trajectories.  Hence its policies stagnate, despite continuing to search the state-action space.  We have additionally illustrated the bifurcation structure for this environment.  Improvements in the agent behaviors, and hence decreases in costs, are associated with switching to new branches after bifurcations.  Using deterministic steplength updates may sometimes miss these bifurcations; certain game-play strategies may not be realized too.

Using these games, we also highlight that path-following-based exploration-rate adjustments can outperform both deterministic annealing and adaptive, cross-entropy-based schedules for the value of information and other exploration mechanisms.  Constant exploration-rate updates may not balance the agent's need to sufficiently experience the environment dynamics with the desire to explore as little as possible.  Either too much or too little action search may hence be conducted over a finite number of episodes, leading to poor empirical policies.  Adaptive schedules can overcome this issue to a certain extent.  They can, however, possess difficult-to-set parameters that lead to non-adequate utilizations of the agent's experiences.  Path-following-based adjustments rely on local solution details to automatically change the exploration rate, in contrast.  This facilitates taking actions that better elucidate certain dynamics and implement cost-decreasing behaviors.  Moreover, path following relies only a single, easy-to-set parameter that controls the projection accuracy.  This parameter appears to have a minor impact on the policy quality.

We demonstrate, in an extended set of simulations, that path-following-based adjustments can scale well to continuous spaces where tabular policies are no longer viable.  There, we apply the value of information with pseudo-arc-length path-following to facilitate exploration when using a heavily modified double-deep $Q$-learning framework.  We evaluate this framework on over fifty Nintendo GameBoy environments, such as \emph{Dr.\! Mario}, \emph{Mega\! Man}, and \emph{Donkey\! Kong\! Land}, many of which are more complex than games from the Atari arcade learning environment.  We show that our deep-$Q$-learning network consistently outperforms other deep-reinforcement-learning strategies that rely on alternate exploration mechanisms and exploration-rate adjustments.

Although we used a path-following process for the value of information, the same ideas are applicable to any other search scheme that can be written as the optimization of either a constrained or an unconstrained criterion.  Soft-max exploration is a promising candidate, as it corresponds to a version of the value of information where the expectations with respect to the state-visitation probabilities are ignored in both the cost terms and the Shannon-information constraint term.  The information-bottleneck method is another possibility.  It too corresponds to a variant of the value of information, albeit where the penalty function is changed so that the cost term becomes proportional to Shannon information.  The theory that we have developed should readily apply, with few to no modifications, to these alternatives due to their connection to Stratonovich's criterion.

\setstretch{0.95}\fontsize{9.75}{10}\selectfont
\putbib
\end{bibunit}

\clearpage\newpage

\RaggedRight\parindent=1.5em
\fontdimen2\font=2.1pt\selectfont
\singlespacing
\allowdisplaybreaks

\phantomsection\label{tblB.1}
\begin{center}
\begin{tabular}{| p{1.2in} | p{3.2in} | p{1.45in} |}
\hline
{\sc Symbol} & {\sc Description} & {\sc Section(s)}\\
\hline
\hline

$\mathbb{E}$ & Expected value & 3.1, 3.2\\

$\mathbb{R}$ & Real numbers & 3.2, A.1--A.3\\

$\mathbb{R}_+$ & Positive real numbers & 4.1, A.1--A.3\\

$\mathbb{B}$ & Banach space & A.2\\

$C^g$ & Differentiability class of order $g$ & A.2\\

$O$ & Asymptotically bounded above & A.3, A.2\\

$k$ & Episode index & 4.1, A.2\\

$i$ & Projection iteration index & 4.1, 4.2, A.2\\

$t$ & Time index & 4.2\\

$a,b,j,p,q$ & Arbitrary indices & A.2\\

$\mathcal{A}$ & Agent action space & 3.1, 4.2, A.1\\

$\mathcal{S}$ & Environment state space & 3.1, 4.2, A.1\\

$a$ & Agent action & 3.1, 3.2, 4.2, A.1\\

$s$ & Environment state & 3.1, 3.2, 4.2, A.1\\

$r$ & Received cost & 4.2\\

$p(s)$, $p(a)$, $p(a|s)$ & Probabilities & 3.1, 3.2\\

$Q(s,a)$ & State-action value-function & 3.1, 4.2\\

$\pi$, $\pi(a|s)$ & Probabilistic action-selection policy & 3.1, 3.2, 4.1, 4.2, A.1--A.3\\

$\pi_*$ & Locally or globally optimal probabilistic policy & 3.2, 4.1, A.1\\

$\varphi_\textnormal{inf}$ & Positive information-bound amount & 3.1\\

$\varepsilon$ & Epsilon-greedy exploration rate & 5.1\\

$\tau$ & Soft-max exploration rate & 5.1\\

$\vartheta$ & Value-of-information exploration rate & 3.2, 4.1, 4.2, A.1, A.2\\

$\psi$ & Element of the Jacobian nullspace & 3.2, A.2\\

$h$ & Real-valued vector & 3.2\\

$\Gamma$ & Full-rank column matrix that spans the Jacobian nullspace & 3.2\\

$\varphi$ & Arc-length parameter & 4.1, 4.2, A.2, A.3\\

$\varphi_*$ & Arc-length parameter value for a solution & A.2, A.3\\

$\beta$ & Probability unit-summation Lagrange multipliers & 3.2, 4.1, 4.2, A.1--A.3\\

$\beta_*$ & Locally or globally optimal Lagrange multipliers & 3.2\\

$\phi$ & Probability non-negativity Lagrange multipliers & A.1\\

$c(\pi)$ & Value-of-information equality constraint & A.1\\

$f(\pi)$ & Value of information loss terms & 3.2, 4.1\\

$F(\pi)$ & Unconstrained value of information & 3.2, 4.1, A.1\\

$\mathcal{G}$ & Value of information equality constraints & A.1\\

$\mathcal{M}$ & Value of information inequality constraints & A.1\\

$\mathcal{A}(\pi)$ & Active set of constraints & A.1\\

$\mathcal{J}$ & Tangent cone to the feasible set & A.1\\

$\mathcal{L}((\pi,\beta),\vartheta)$ & Value of information Lagrangian & 3.1, 4.1, 4.2, A.1--A.3\\

$\mathcal{K}((\pi,\beta),\vartheta)$ & Modified value of information Lagrangian & 4.1, 4.2, A.2\\

$J$ & Jacobian of $\nabla_\beta \mathcal{L}((\pi,\beta),\vartheta)$ & 3.2, A.1\\

$\delta$, $\delta'$, $\delta_\vartheta$ & Solution perturbation amounts & 4.1, 4.2, A.2\\

$\epsilon$, $\epsilon_0$, $\rho$ & Scalars & 4.1, A.2, A.3\\

$\nabla_{\pi,\beta}$ & First-order gradient & 3.2, 4.1, 4.2, A.1--A.3\\

$\nabla_{\pi,\beta}^2$ & Second-order gradient & 3.2, 4.1, 4.2, A.1--A.3\\

$\nabla_{\pi,\beta}^3$ & Third-order gradient & A.3\\

$\partial_\pi$, $\partial_\beta$, $\partial_\varphi$ & Partial derivatives & 4.1, 4.2, A.1--A.3\\

$\theta$, $\omega$, $\omega'$ & Positive scalars on the unit interval & 4.1, A.2\\

$g$ & Scaling factor equal to $\partial_\varphi \vartheta(\varphi)$ & 4.1\\

$\alpha$ & Learning rate & 4.2\\

$\gamma$ & Discount factor & 4.2\\

$u$ & Fast time scale & 4.2\\

$v$ & Slow time scale & 4.2\\

$\Omega$ & Open set of the reals & A.2\vspace{0.045cm}\\

\hline

\end{tabular}\vspace{-0.05cm}
\end{center}

\begin{center}{\footnotesize Table 1: Paper notation by section}\end{center}

\clearpage\newpage

\begin{center}
\begin{tabular}{| p{1.2in} | p{3.2in} | p{1.45in} |}
\hline
{\sc Symbol} & {\sc Description} & {\sc Section(s)}\\
\hline
\hline

$w$ & Iterate difference inside $\epsilon$-ball & A.2\\

$\mu'$, $K$ & Lipschitz constants & A.2\\

$M$ & Maximum of $\|\partial_\vartheta\mathcal{L}((\pi,\beta),\vartheta)\|$ in $\Omega$ & A.2\\

$T((\pi,\beta),\vartheta)$ & Newton mapping & A.2\\

$\mathcal{Q}((\pi,\beta),\vartheta)$ & Integral terms of $\mathcal{L}((\pi,\beta),\vartheta)$ & A.2\\

$\gamma(\varphi)$ & Shortened expression for the iterate $((\pi(\varphi),\beta(\varphi)),\vartheta(\varphi))$ & 4.1, 4.2, A.2\\

range & Range & A.2\\

ker, null & Nullspace & A.2\\

$Q((\pi,\beta),\vartheta)$ & Block matrix & A.2\\

$\mathcal{M}_\varphi$, $\mathcal{M}_\varphi(\pi,\beta,\vartheta)$ & Joint solution constraint & A.2\\

$G((\pi,\beta),\vartheta)$ & Linear operator associated with $Q((\pi,\beta),\vartheta)$ & A.2\\

$\kappa(\varphi)$ & Non-negative bound factor & A.2\\

$\eta$ & Positive scalar depending on the operator eigenstructure & A.2\\

$\tau$ & Either a scalar or a scalar with geometric convergence & A.2\\

$C$ & Positive scalar & A.2\\

id & Banach-space identity operator & A.2\\

$\tau^*$ & Solution constructed from $\tau_0^*$ & A.2\\

$\tau_0^*$ & Solution of $\nabla_{\!\pi,\beta}^2\mathcal{L}(\gamma(\varphi))\tau_0^* \!+\! \nabla_{\!\pi,\beta}\mathcal{M}_\varphi(\gamma) \!=\! 0$ & A.4\\

$\tau_1$, $\tau_2$ & Eigenfunction-associated variables & A.2\\

$\sigma$ & Iterate-norm-bound constant factor & A.2\\

$\alpha(\varphi)$, $\lambda(\varphi)$ & Eigenvalues of the linear operator & A.2\\

$\phi(\varphi)$, $\mu(\varphi)$ & Eigenvectors of the linear operator & A.2, A.3\\

$\psi^*(\varphi)$, $\hat{\psi}^*(\varphi)$ & Adjoint eigenfunction & A.2, A.3\\

$\xi_j$ & Non-negative scalars & A.2, A.3\\

$\omega_i$, $\omega_{i,j}$, $\omega_{i,j,p}$ & Non-negative coefficients & A.2, A.3\\

$B$ & Block matrix & A.2\\

$\mathcal{Y}_1$, $\mathcal{Y}_2$ & Subspaces of the Banach space & A.2\\

$\mathcal{U}_1$, $\mathcal{U}_2$, $\mathcal{W}_1$, $\mathcal{W}_2$ & Subspaces of the Banach space & A.2\\

$q_1$, $q_2$, $p_1$, $p_2$ & Projections onto the subspaces & A.2\\

$U_1$, $U_2$, $W_1$, $W_2$ & Continuous, bounded functions & A.2\\

$U$, $V$, $W$ & Either bounded or uniformly-bounded functions & A.2\\

$\mathcal{H}$ & Continuous function & A.3\\

$V$ & Second- and third-order gradient sum & A.3\vspace{0.045cm}\\

\hline

\end{tabular}\vspace{-0.05cm}
\end{center}

\begin{center}{\footnotesize Table 1 (Continued)}\end{center}

\clearpage\newpage

\begin{bibunit}
\bstctlcite{IEEEexample:BSTcontrol}

\RaggedRight\parindent=1.5em
\fontdimen2\font=2.1pt\selectfont
\singlespacing
\allowdisplaybreaks

\setcounter{equation}{0}
\renewcommand{\thefigure}{A.\arabic{figure}}
\renewcommand\theequation{A.\arabic{equation}}
\singlespacing

\phantomsection\label{secA}
%%%%%%%%%%%%%%%%%%%%%%%%%%%%%%%%%%%%%%%%%%%%%%%%%%%%%%%%%%%%%%%%%%%%%%%%%%%
%%%%%%%%%%%%%%%%%%%%%%%%%%%%%%%%%%%%%%%%%%%%%%%%%%%%%%%%%%%%%%%%%%%%%%%%%%%
\subsection*{\small{\sf{\textbf{Appendix A}}}}

In this appendix, we provide proofs to the theoretical claims that we have made throughout.

We begin with the solution properties of the value of information (see \hyperref[secA.1]{Appendix A.1}).  We prove one of the major conditions used in our path-following approaches, which is that a solution to our information-theoretic criterion is obtained whenever the Hessian of the value-of-information Lagrangian is negative semi-definite on the nullspace of the Jacobian.  We then investigate behaviors of path following when applied to the value of information (see \hyperref[secA.2]{Appendix A.2}).  We show that unique solutions for the value of information exist, whenever the Hessian is non-singular, and that both parameter and pseudo-arc-length path-following will converge to them.  For the latter approach, any points on the solution surface in which the Hessian is singular will be safely ignored.  Lastly, we quantify when bifurcations will occur when adjusting the exploration rate (see \hyperref[secA.3]{Appendix A.3}).  We then describe a way to investigate these bifurcating solution branches.

\phantomsection\label{secA.1}
%%%%%%%%%%%%%%%%%%%%%%%%%%%%%%%%%%%%%%%%%%%%%%%%%%%%%%%%%%%%%%%%%%%%%%%%%%%
%%%%%%%%%%%%%%%%%%%%%%%%%%%%%%%%%%%%%%%%%%%%%%%%%%%%%%%%%%%%%%%%%%%%%%%%%%%
\subsection*{\small{\sf{\textbf{A.1$\;\;\;$ Solution Properties}}}}

For what follows, it is helpful to explicitly state the first-order necessary conditions.

\phantomsection\label{propA.1}
\begin{itemize}
\item[] \-\hspace{0.0cm}{\small{\sf{\textbf{Proposition A.1.}}}} Let $\pi^* \!\in\! \mathbb{R}_{+}^{m \times n}$ be a global solution of the value of information, where $m$ represents the\\ \noindent number of discrete action choices and $n$ the number of discrete states.  Suppose that the Jacobian of the equality constraint $c_i(\pi) \!=\! 0$ and the inequality constraint $c_i(\pi) \!\leq\! 0$, has full row rank for an arbitrary policy $\pi \!\in\! \mathbb{R}_{+}^{m \times n}$.\\ \noindent 

There exists a vector of positive Lagrange multipliers $\beta^* $ such that $\nabla_{\!\pi}F(\pi^*,\vartheta) \!=\! -\sum_{i} \beta_i^* \nabla c_i(\pi^*)$ and\\ \noindent $\beta^*c_i(\pi^*) \!=\! 0$, with $c_i(\pi^*) \!=\! 0$, $\forall i \!\in\! \mathcal{G}$, and $\beta^* c_i(\pi^*) \!=\! 0$, $\forall j \!\in\! \mathcal{G} \cup \mathcal{M}$, for the equality constraint, and $c_i(\pi^*) \!\geq\! 0$ and $\beta^* \!\geq\! 0$, $\forall j \!\in\! \mathcal{M}$, for the inequality constraint.  Here, $\mathcal{G}$ represents the equality constraints for the value of\\ \noindent information while $\mathcal{M}$ are the inequality constraints.\vspace{0.1cm}
\end{itemize}

\noindent The next proposition is a corollary of \hyperref[propA.1]{Proposition A.1}.

\vspace{0.15cm}\begin{itemize}
\item[] \-\hspace{0.0cm}{\small{\sf{\textbf{Proposition A.2.}}}} Let $\pi^* \!\in\! \mathbb{R}_{+}^{m \times n}$ be a global solution of the value of information for a fixed hyperparameter\\ \noindent value $\vartheta \!\in\! \mathbb{R}_+$.  There is a vector of Lagrange multipliers $\beta \!\in\! \mathbb{R}^n$ such that the gradient of the Lagrangian is equal to the zero vector, $\nabla_{\!\pi}\, \mathcal{L}((\pi^*,\beta^*),\vartheta) \!=\! 0$.  

As well, we have that the $s$th component of the absolute Lagrangian is zero, $|\nabla_{\!\beta}\,\mathcal{L}((\pi^*,\beta^*),\vartheta)|_{s} \!=\! 0$,\\ \noindent which implies that, for the equality constraint, $c_i \!=\! 0$.  Hence, the Karush-Kuhn-Tucker conditions are satisfied.\vspace{0.1cm}
\end{itemize}

In order to prove \hyperref[prop3.1]{Proposition 3.1}, we need the notion of a limiting direction of a feasible sequence.  We therefore define the set of all feasible directions.

\vspace{0.15cm}\begin{itemize}
\item[] \-\hspace{0.0cm}{\small{\sf{\textbf{Definition A.1.}}}} Let $\pi^* \!\in\! \mathbb{R}_{+}^{m \times n}$ be a local solution of the value of information for a fixed $\vartheta \!\in\! \mathbb{R}_+$.  Let $\mathcal{A}(\pi^*)$ be\\ \noindent the active set.  Let $\mathcal{J} \!=\! \{\alpha \varphi\, |\, \alpha \!>\! 0,\; \varphi^\top \nabla c_i(\pi^*) \!=\! 0, \forall i \!\in\! \mathcal{G},\; \varphi^\top \nabla c_i(\pi^*) \!\geq\! 0, \forall i \!\in\! \mathcal{A}(\pi^*) \cap \mathcal{M}\}$, where $\varphi$ is a\\ \noindent feasible direction.  $\mathcal{J}$ is a tangent cone to the feasible set $\pi^*$ whenever the constraint qualification is satisfied.  Let $\mathcal{K}(\beta^*) \!=\! \{\varphi \!\in\! \mathcal{J}\,|\, \nabla c_i(\pi^*)^\top \varphi \!=\! 0, \forall i \!\in\! \mathcal{A}(\pi^*) \cap \mathcal{M} \textnormal{ with } \beta^* \!>\! 0\}$ be a subset of this cone.\vspace{0.1cm}
\end{itemize}

\noindent With this, the second-order sufficient conditions can be verified.

\phantomsection\label{prop3.1}
\vspace{0.15cm}\begin{itemize}
\item[] \-\hspace{0.0cm}{\small{\sf{\textbf{Proposition 3.1.}}}} For a given optimal policy $\pi^* \!\in\! \mathbb{R}_{+}^{m \times n}$, we suppose that there is a vector of Lagrange\\ \noindent multipliers $\beta^* \!\in\! \mathbb{R}^n$ such that the Karush-Kuhn-Tucker conditions are satisfied.  If, for the Jacobian of the con-\\ \noindent straints $J$, we have that the Hessian $\varphi^\top \nabla^2_{\!\pi}\, \mathcal{L}((\pi^*,\beta^*),\vartheta)\varphi \!\leq\! 0$, then $\pi^*$ is a local solution of the value of\\ \noindent information.  Here, $\varphi$ is an element of the Jacobian nullspace, $\varphi \!\in\! \textnormal{ker}(J)$.  The converse is also true.\vspace{0.15cm}
\end{itemize}

\begin{itemize}
\item[] \-\hspace{0.5cm}{\small{\sf{\textbf{Proof:}}}} For the claim to be valid, we must have that, for any feasible sequence $\{\pi_k\}_{k=1}^\infty$ approaching $\pi^*$, $F(\pi_k,\vartheta) \!>\! F(\pi^*,\vartheta)$ for any fixed $\vartheta$ and all sufficiently large $k$.  

Given any feasible sequence, all of the limiting directions lie in the cone specified by $\mathcal{J}$.  Choosing an arbitrary subsequence $s_\pi$ of $\{\pi_k\}_{k=1}^\infty$ such that properties of the limiting direction are satisfied, we have that $\mathcal{L}((\pi_k,\beta^*),\vartheta) \!=\! F(\pi_k,\vartheta) \!-\! \sum_{i \in \mathcal{A}(\pi^*)} \beta_i^* c_i(\pi_k) \!\leq\! F(\pi_k,\vartheta)$.

Suppose that the limiting direction is in $\mathcal{J}$ but not in $\mathcal{K}(\beta^*)$.  In this case, an index $j \!\in\! \mathcal{A}(\pi^*) \cap \mathcal{M}$ can be\\ \noindent found such that $\beta^* \nabla c_j(\pi^*)^\top \varphi \!>\! 0$ is satisfied while the remaining indices $i \!\in\! \mathcal{A}(\pi^*)$ lead to $\beta^* \nabla c_i(\pi^*)^\top \varphi \!\geq\! 0$.  We therefore have that $\mathcal{L}((\pi_k,\beta^*),\vartheta) \!\leq\! F(\pi_k,\vartheta) \!-\! \beta^*\nabla c_j(\pi^*)^\top \varphi\|\pi_k \!-\! \pi^*\| \!+\! o(\|\pi_k \!-\! \pi^*\|)$. From the second-\\ \noindent order Taylor-series expansion of the value-of-information Lagrangian, we obtain the following expression $\mathcal{L}((\pi_k,\beta^*),\vartheta) \!=\! F(\pi^*,\vartheta) \!+\! O(\|\pi_k \!-\! \pi^*\|^2)$.  Combining the two together permits us to quantify the solution\\ \noindent quality of the subsequence with respect to that of the optimal solution and obtain that $F(\pi_k,\vartheta) \!>\! F(\pi^*,\vartheta)$.

Suppose now that the limiting direction is in $\mathcal{J}$ and in $\mathcal{K}(\beta^*)$.  Again, $F(\pi_k,\vartheta) \!>\! F(\pi^*,\vartheta)$ for all $k$ suf-\\ \noindent ficiently large.  Since either argument applies to all limiting directions of the arbitrary subsequence, each subsequence will converge.  A local solution is hence obtained. $\blacksquare$\vspace{0.15cm}
\end{itemize}

The equality constraints for the value of information can be written as $\{c_i(\pi)\}_{i \in \mathcal{G}} \!=\! \{\sum_{a \in \mathcal{A}} \pi(a|s) \!-\! 1\}_{s \in \mathcal{S}}$.\\ \noindent  If $\pi \!\in\! \{\pi' \!\in\! \mathbb{R}^{m \times n}| \sum_{a \in \mathcal{A}} \pi'(a|s) \!=\! 1, \forall s \!\in\! \mathcal{S}\}$, then $c_i(\pi) \!=\! 0$ for every $i \!\in\! \mathcal{G}$.  For the inequality constraints, we\\ \noindent have that $\{c_i(\pi)\}_{i \in \mathcal{M}} \!=\! \{\pi(a|s)\}_{a \in \mathcal{A}, s \in \mathcal{S}}$.  If $\pi \!\in\! \{\pi' \!\in\! \mathbb{R}^{m \times n}| \sum_{a \in \mathcal{A}} \pi'(a|s) \!=\! 1 \textnormal{ with } \pi'(a|s) \!\geq\! 0, \forall s \!\in\! \mathcal{S}\}$, then\\ \noindent $c_i(\pi) \!\geq\! 0$ for every $i \!\in\! \mathcal{M}$.

It can be seen that the constraints on the value of information are linear.  If we therefore track $\pi^*$ where the Karush-Kuhn-Tucker conditions are satisfied and where $\nabla \mathcal{L}(\pi^*,\beta)$ is negative definite on $\textnormal{ker}(J)$, then the assumptions of \hyperref[prop3.1]{Proposition 3.1} are satisfied.  This implies that $\pi^*$ is a local solution of the value of information and hence an equilibrium of the gradient flow $((\dot{\pi},\dot{\beta}),\dot{\vartheta}) \!=\! \nabla \mathcal{L}((\pi,\beta),\vartheta)$.  Since Shannon's mutual information is convex, local solutions are global solutions for the value of information.

It is important to note that we choose consider this flow and not $((\dot{\pi},\dot{\beta},\dot{\phi}),\dot{\vartheta}) \!=\! \nabla \mathcal{L}'((\pi,\beta,\phi),\vartheta)$, with
\begin{equation*}
\mathcal{L}'((\pi,\beta,\phi),\vartheta) = \textstyle F(\pi,\vartheta) + \sum_{s \in \mathcal{S}} \beta_{s}(\sum_{a \in \mathcal{A}} \pi(a|s) \!-\! 1) + \sum_{s \in \mathcal{S}}\sum_{a \in \mathcal{A}} \phi_{s,a} \pi(a|s),
\end{equation*}
where $\phi$ are Lagrange multipliers associated with the constraints that the policy probabilities must be non-negative.  This was not an arbitrary choice.  There are no equilibria for any $\vartheta$, since, if $\nabla_{\pi,\beta,\phi} \mathcal{L}'((\pi^*,\beta^*,\phi^*),\vartheta) \!=\! 0$, then\\ \noindent $\partial_\beta \mathcal{L}'((\pi^*,\beta^*,\phi^*),\vartheta) \!=\! 0$, and all of the equality constraints are active.  As well, $\nabla_\phi \mathcal{L}'((\pi^*,\beta^*,\phi^*),\vartheta) \!=\! 0$, indicating that all of the inequality constraints are active.  Both cannot be true simultaneously.

\phantomsection\label{secA.2}
%%%%%%%%%%%%%%%%%%%%%%%%%%%%%%%%%%%%%%%%%%%%%%%%%%%%%%%%%%%%%%%%%%%%%%%%%%%
%%%%%%%%%%%%%%%%%%%%%%%%%%%%%%%%%%%%%%%%%%%%%%%%%%%%%%%%%%%%%%%%%%%%%%%%%%%
\subsection*{\small{\sf{\textbf{A.2$\;\;\;$ Path-Following Convergence Behaviors}}}}

We first prove a variant of the Implicit Function Theorem, which will be useful throughout.

\phantomsection\label{propA.3}
\vspace{0.15cm}\begin{itemize}
\item[] \-\hspace{0.0cm}{\small{\sf{\textbf{Proposition A.3.}}}} Let $\Omega$ be an open subset of the reals.  Let $\nabla_{\pi,\beta}\mathcal{L}((\pi,\beta),\vartheta) \!\in\! C^g(\Omega)$ for some differentiability order $g \!>\! 0$.  Assume that $\partial_\vartheta \nabla_{\!\pi,\beta} \mathcal{L}((\pi,\beta),\vartheta)$ and $\nabla_{\!\pi,\beta}^2\mathcal{L}((\pi,\beta),\vartheta)$ are Lipschitz continuous on the closure of $\Omega$.\\ \noindent  If $((\pi^0,\beta^0),\vartheta^0) \!\in\! \Omega$, $\nabla_{\pi,\beta}\mathcal{L}((\pi^0,\beta^0),\vartheta^0) \!=\! 0$, and $\nabla_{\!\pi,\beta}^2\mathcal{L}((\pi^0,\beta^0),\vartheta^0)$ is non-singular, then there are some $\epsilon$\\ \noindent and $\rho$ such that
\begin{itemize}
\item[] \-\hspace{0.5cm}(i) $(\pi,\beta)(\vartheta) \!\in\! C^g(\vartheta^0 \!-\! \rho,\vartheta^0 \!+\! \rho)$.
\item[] \-\hspace{0.5cm}(ii) There is a unique solution of $\nabla_{\pi,\beta} \mathcal{L}((\pi,\beta),\vartheta) \!=\! 0$ that exists in, $\{(\pi,\beta) | ((\pi,\beta) \!-\! (\pi^0,\beta^0)) \!<\! \epsilon\}$,\\ \noindent for $\epsilon \!\geq\! 0$, which is an $\epsilon$-ball.  This solution exists for all $\vartheta \!\in\! (\vartheta^0 \!-\! \rho,\vartheta^0 \!+\! \rho)$.
\end{itemize}
\end{itemize}
\begin{itemize}
\item[] \-\hspace{0.5cm}{\small{\sf{\textbf{Proof:}}}} We first show that $(\pi,\beta)$ is $C^g$-smooth for all $\vartheta \!\in\! (\vartheta^0 \!-\! \rho,\vartheta^0 \!+\! \rho)$.  Let $\vartheta,\vartheta' \!\in\! (\vartheta_0 \!-\! \rho,\vartheta_0 \!+\! \rho)$.  As well, let\\ \noindent $w \!=\! (\pi,\beta) \!-\! (\pi',\beta')$, where $(\pi,\beta),(\pi',\beta')$ belong to the $\epsilon$-ball.  From the non-singularity of $\nabla_{\!\pi,\beta}^2\mathcal{L}((\pi,\beta),\vartheta)$ and the continuity of $\nabla_{\!\pi,\beta}\mathcal{L}((\pi,\beta),\vartheta)$ on the closure of $\Omega$, $\|w\| \!\leq\! (M|\xi|/2 \!+\! K\|w\|(\epsilon \!+\! \rho))\|\nabla_{\!\pi,\beta}^{-2}\mathcal{L}((\pi^0,\beta^0),\vartheta^0)\|$,\\ \noindent Here, $K$ is the Lipschitz constant of $\nabla_{\!\pi,\beta}^2\mathcal{L}((\pi,\beta),\vartheta)$ on the closure of $\Omega$.  $M \!=\! \textnormal{max}_{((\pi,\beta),\vartheta) \in \Omega}\|\partial_\vartheta \nabla_{\!\pi,\beta}\mathcal{L}((\pi,\beta),\vartheta)\|$, which is necessarily finite.  We thus have that
\begin{equation*}
\|w\| \leq \textstyle (\frac{M|\mu'|}{2}\|\nabla_{\!\pi,\beta}^{-2}\mathcal{L}((\pi^0,\beta^0),\vartheta^0)\|)/(1 \!-\! K(\epsilon \!+\! \rho)\|\nabla_{\!\pi,\beta}^{-2}\mathcal{L}((\pi^0,\beta^0),\vartheta^0)\|).
\end{equation*}
Hence, $\|w\| \!\leq\! O(|\mu'|)$.  This demonstrates continuity of a solution branch $(\pi,\beta)(\vartheta)$ as a function of $\vartheta$.  Differentiability is straightforward to demonstrate as long as $\nabla_{\!\pi,\beta} \mathcal{L}((\pi,\beta),\vartheta)$ supports differentiation.

We now show that we can define a mapping that is a contraction on the $\epsilon_0$-ball, $\{(\pi,\beta) | ((\pi^0,\beta^0) \!-\! (\pi,\beta)) \!\leq\! \epsilon_0\}$, $\epsilon_0 \!\geq\! 0$, whenever $\vartheta \!\in\! (\vartheta^0 \!-\! \rho,\vartheta^0 \!+\! \rho)$.  Notice that the $\epsilon$- and $\epsilon_0$-ball differ in terms of the inequality constraint.

Let $(\pi,\beta) \!=\! (\pi^0,\beta^0) \!+\! \omega$ and $\vartheta \!=\! \vartheta^0 \!+\! \xi$.  From the fundamental theorem of calculus, we have that\\ \noindent $\nabla_{\!\pi,\beta}\mathcal{L}((\pi,\beta),\vartheta) \!=\! \nabla_{\!\pi,\beta}^2\mathcal{L}((\pi^0,\beta^0),\vartheta^0)\omega \!+\! \partial_\vartheta\nabla_{\!\pi,\beta}\mathcal{L}((\pi^0,\beta^0),\vartheta^0)\xi \!+\! \mathcal{Q}((\pi,\beta),\vartheta)$, where $\mathcal{Q}((\pi,\beta),\vartheta)$ is\\ \noindent composed of dual integral difference equations.  

If $(\pi,\beta)$ belongs to the $\epsilon_0$-ball, then we can define the Newton map,
\begin{equation*}
T((\pi,\beta),\vartheta) = (\pi,\beta) - \nabla_{\!\pi,\beta}^{-2}\mathcal{L}((\pi^0,\beta^0),\vartheta^0)(\partial_\vartheta\nabla_{\!\pi,\beta}\mathcal{L}((\pi^0,\beta^0),\vartheta^0)(\vartheta \!-\! \vartheta^0) \!+\! \mathcal{Q}((\pi,\beta),\vartheta)).
\end{equation*}
Since $\|\mathcal{Q}((\pi,\beta),\vartheta)\| \!\leq\! \mu'(\epsilon \!+\! \epsilon\epsilon_0 \!+\! \epsilon \epsilon_0^2)$, for $\mu' \!>\! 0$, we have that,
\begin{equation*}
\|T((\pi,\beta),\vartheta) \!-\! (\pi^0,\beta^0)\| \leq \|\nabla_{\!\pi,\beta}^{-2}\mathcal{L}((\pi^0,\beta^0),\vartheta^0)\|(\|\partial_\vartheta\nabla_{\!\pi,\beta}\mathcal{L}((\pi^0,\beta^0),\vartheta^0)\|\epsilon + \mu'(\epsilon \!+\! \epsilon\epsilon_0 \!+\! \epsilon \epsilon_0^2)) \leq \epsilon.
\end{equation*}
This inequality is satisfied whenever $|\vartheta \!-\! \vartheta_0| \!\leq\! \rho$, for $\epsilon,\rho$ sufficiently small on $\Omega$.  To show that the Newton map yields a unique solution, we must have that it is a contraction on the $\epsilon_0$-ball.  This fact is a consequence of the\\ \noindent inequality $\|\mathcal{Q}((\pi,\beta),\vartheta)) \!-\! \mathcal{Q}((\pi',\beta'),\vartheta))\| \!\leq\! \mu'(\rho \!+\! \epsilon)\|(\pi,\beta) \!-\! (\pi',\beta')\|$, where $(\pi',\beta')$ belongs to the $\epsilon_0$-ball,
\begin{equation*}
\|T((\pi,\beta),\vartheta) \!-\! T((\pi',\beta'),\vartheta)\| \leq \mu'(\epsilon \!+\! \rho)\|\nabla_{\!\pi,\beta}^{-2}\mathcal{L}((\pi^0,\beta^0),\vartheta^0)\|\|(\pi,\beta) \!-\! (\pi',\beta')\|,
\end{equation*}
as it implies that points in the image are closer together than the source, except at a solution, for $\epsilon,\rho$ sufficiently small on $\Omega$.  Since this mapping is a contraction, then the Banach fixed-point theorem gives that there is a unique fixed point in the $\epsilon_0$-ball, which is a solution for the value of information.  

We can strengthen this claim so that it holds for the $\epsilon$-ball. $\blacksquare$
\end{itemize}

\vspace{0.15cm}The following proposition states that parameter path-following will converge to solutions when $\nabla_{\pi,\beta} \mathcal{L}^i((\pi_k,\beta_k),\vartheta_k)$ is non-singular.  It may fail, however, if the solution path contains simple folds, which is where the Lagrangian gradient $(\nabla_{\pi,\beta} \mathcal{L}^i((\pi_k^i,\beta_k^i),\vartheta_k^i)$ is singular for some corrector step $i$ or episode $k$.\vspace{0.15cm}

\phantomsection\label{prop4.1}
\begin{itemize}
\item[] \-\hspace{0.0cm}{\small{\sf{\textbf{Proposition 4.1.}}}} Assume that $\mathcal{L}^i((\pi_k^i,\beta_k^i),\vartheta_k^i)$ is Lipschitz differentiable, where $\mathcal{L}^i((\pi_k^0,\beta_k^0),\vartheta_k^0) \!=\! 0$ and\\ \noindent $\nabla_{\pi,\beta} \mathcal{L}^i((\pi_k^0,\beta_k^0),\vartheta_k^0)$ is non-singular.  There is an $\epsilon \!>\! 0$ that depends on the Lipschitz constants of\\ \noindent $\partial_\vartheta \mathcal{L}^i((\pi_k^0,\beta_k^0),\vartheta_k^0)$ and $\nabla_{\pi,\beta} \mathcal{L}^i((\pi_k^0,\beta_k^0),\vartheta_k^0)$ such that algorithm 1 converges $q$-quadratically to the solution $(\pi_{k+1},\beta_{k+1})$ of $\mathcal{L}((\pi_{k+1},\beta_{k+1}),\vartheta_{k+1}) \!=\! 0$ for $|\vartheta_{k+1} \!-\! \vartheta_k^0| \!<\! \epsilon$.\vspace{0.05cm}
\end{itemize}

\begin{itemize}
\item[] \-\hspace{0.5cm}{\small{\sf{\textbf{Proof:}}}} This proposition is a consequence of \hyperref[propA.3]{Proposition A.3}.  First, we define the Lipschitz constant\\ \noindent $\|\partial_\vartheta \mathcal{L}^i((\pi_k,\beta_k),\vartheta_k) \!-\! \partial_\vartheta \mathcal{L}^i((\pi_k',\beta_k'),\vartheta_k')\| \!\leq\! \mu\|(\pi_k,\beta_k) \!-\! (\pi_k',\beta_k')\| \!+\! \mu(\vartheta_k \!-\! \vartheta_k')$.  Differentiating the\\ \noindent value-of-information Lagrangian with respect to $\vartheta$ yields 
$$d\nabla_{\pi,\beta} \mathcal{L}^i((\pi_k,\beta_k),\vartheta_k)/d\vartheta \!=\! - (\nabla_{\pi,\beta} \mathcal{L}^i((\pi_k,\beta_k),\vartheta_k))^{-1} \partial_\vartheta \mathcal{L}^i((\pi_k,\beta_k),\vartheta_k).$$
\hyperref[propA.3]{Proposition A.3} gives that there is an $\epsilon$ such that if $|\vartheta_{k+1} \!-\! \vartheta_k^0| \!\leq\! \epsilon'$ then there is a solution path defined for\\ \noindent it.  Since $(\nabla_{\pi,\beta} \mathcal{L}^i((\pi_k,\beta_k),\vartheta_k))^{-1} \partial_\vartheta \mathcal{L}^i((\pi_k,\beta_k),\vartheta_k)$ is Lipschitz continuous, there is a $\mu'$ that depends only on $\|(\nabla_{\pi,\beta} \mathcal{L}^i((\pi_k,\beta_k),\vartheta_k))^{-1}\|$ and on the Lipschitz constants of $\nabla_{\pi,\beta}^2 \mathcal{L}^i((\pi_k,\beta_k),\vartheta_k)$ and $\partial_\vartheta \mathcal{L}^i((\pi_k,\beta_k),\vartheta_k)$.  Moreover, $\|d(\pi_k,\beta_k)(\vartheta)/d\vartheta\| \!\leq\! \mu'$.

From \cite{RyabenkiiVS-book2006a}, we know that a lower bound on a spherical convergence region for Newton's method is given by\\ \noindent $(2\mu\|(\nabla_{\pi,\beta} \mathcal{L}^i((\pi_k,\beta_k),\vartheta_k))^{-1}\|)^{-1}$.  Setting $\epsilon \!=\! \textnormal{min}(\epsilon', (2\mu\|(\nabla_{\pi,\beta} \mathcal{L}^i((\pi_k,\beta_k),\vartheta_k))^{-1}\|)^{-1})$, which is obvi-\\ \noindent ously greater than zero, yields the desired parameter.  The remainder of the proof follows from standard convergence arguments for Newton iterations. $\blacksquare$\vspace{0.15cm}
\end{itemize}

\noindent A consequence of this proposition is that the bound on $\|(\nabla_{\pi,\beta} \mathcal{L}^i((\pi_k,\beta_k),\vartheta_k))^{-1}\|$ also bounds the path-following steplength.  Similar results hold for psuedo-arc-length path-following.  In this latter case, the smallest allowable steplength relies on the smallest eigenvalue of $(\nabla_{\pi,\beta} \mathcal{L}^i((\pi^i_k(\varphi_k),\beta^i_k(\varphi_k)),\vartheta^i_k(\varphi_k)))(\nabla_{\pi,\beta} \mathcal{L}^i((\pi^i_k(\varphi_k),\beta^i_k(\varphi_k)),\vartheta^i_k(\varphi_k)))^\top$.

\hyperref[propA.3]{Proposition 4.1} can be used to additionally demonstrate convergence of pseudo-arc-length path-following.  This is because pseudo-arc-length path-following is nothing more than parameter path-following with the value-of-information Lagrangian parameterized by $\varphi_k$.\vspace{0.15cm}

\phantomsection\label{prop4.2}
\begin{itemize}
\item[] \-\hspace{0.0cm}{\small{\sf{\textbf{Proposition 4.2.}}}} Assume that $\mathcal{L}^i((\pi_k^i(\varphi_{k}),\beta_k^i(\varphi_{k})),\vartheta_k^i(\varphi_{k}))$ is Lipschitz differentiable, where\\ \noindent $\mathcal{L}^i((\pi_k^i(\varphi_k),\beta_k^i(\varphi_k)),\vartheta_k^i(\varphi_k)) \!=\! 0$ and $\nabla_{\pi,\beta} \mathcal{L}^i((\pi_0^i(\varphi_k),\beta_k^i(\varphi_k)),\vartheta_k^i(\varphi_k))$ is non-singular.  There is an $\epsilon \!>\! 0$\\ \noindent that depends on $\langle \nabla_{\pi,\beta}\mathcal{L}^i((\pi_k^0(\varphi_k),\beta_k^0(\varphi_k)),\vartheta_k^0(\varphi_k)),\cdot\rangle$ the Lipschitz constant of \noindent $\partial_\vartheta \mathcal{L}^i((\pi_k^0(\varphi_k),\beta_k^0(\varphi_k)),\vartheta_k^0(\varphi_k))$ such that \hyperref[alg:pseudoarclengthpathfollowing]{Algorithm 2} converges $q$-quadratically to the solution $(\pi_{k+1}(\varphi_{k+1}),\beta_{k+1}(\varphi_{k+1}))$ of\\ \noindent $\mathcal{L}((\pi_{k+1}(\varphi_{k+1}),\beta_{k+1}(\varphi_{k+1})),\vartheta_{k+1}(\varphi_{k+1})) \!=\! 0$ for $|\varphi_{k+1} \!-\! \varphi_k^0| \!<\! \epsilon$.\vspace{0.15cm}
\end{itemize}

\noindent An advantage of pseudo-arc-length path-following is that it can jump over singular points.  However, this claim is not present in \hyperref[prop4.2]{Proposition 4.2}.  We assume, in \hyperref[prop4.2]{Proposition 4.2}, that policy updates only occur in neighborhoods of non-singular points along the solution curve, which is not realistic.

We thus strengthen \hyperref[prop4.2]{Proposition 4.2} into \hyperref[propA.9]{Proposition A.9}.  We begin by noting that the joint solution constraint $\theta\|(\dot{\pi}(\varphi_k),\dot{\beta}(\varphi_k))\|^2 \!+\! (1 \!-\! \theta)\dot{\vartheta}(\varphi_k)^2 \!=\! 1$, $\theta \!\in\! (0,1)$, is often too restrictive, even for merely proving the existence\\ \noindent of solutions.  We thus, following the ideas of Mittelmann \cite{MittelmannHD-jour1986a} and Keller \cite{KellerHB-coll1977a}, instead constrain $\nabla_{\!\pi,\beta}\mathcal{L}((\pi_k,\beta_k),\vartheta_k) \!=\! 0$ by the expression
\begin{equation}
\Bigg(\omega (\dot{\pi}(\varphi_k),\dot{\beta}(\varphi_k))^*((\pi(\varphi),\beta(\varphi)) \!-\! (\pi(\varphi_k),\beta(\varphi_k)))\Bigg) + \Bigg((1-\omega)\dot{\vartheta}(\varphi_k)(\vartheta_k(\varphi) \!-\! \vartheta_k(\varphi_k))\Bigg) \!= \varphi_k - \varphi
\label{eq:parc-4}
\end{equation}
where $\omega \!\in\! (0,1)$ is a parameter within the unit interval.  The term $(\dot{\pi}_k(\varphi_k),\dot{\beta}_k(\varphi_k))^*$ is the dual element to $(\dot{\pi}_k(\varphi_k),\dot{\beta}_k(\varphi_k))$, which is guaranteed to exist by the Hahn-Banach Theorem.  We refer to the entirety of (\ref{eq:parc-4}) as $\mathcal{M}_\varphi((\pi_k,\beta_k),\vartheta_k)$.

We demonstrate that solution curves consisting exist for this version of the value-of-information Lagrangian.  These solution curves are composed of both so-called regular and normal-limit points.

\phantomsection\label{defA.2}
\begin{itemize}
\vspace{0.15cm}\item[] \-\hspace{0.0cm}{\small{\sf{\textbf{Definition A.2.}}}} Let $((\pi_k(\varphi_k),\beta_k(\varphi_k)),\vartheta_k(\varphi_k)) \!=\! ((\pi_*(\varphi),\beta_*(\varphi)),\vartheta_*(\varphi))$ be a solution that satisfies (\ref{eq:parc-1}).\\ \noindent  A regular solution along a solution path is one where (i) the Jacobian has full rank and (ii) the Hessian\\ \noindent $\nabla^2_{\!\pi,\beta}\mathcal{L}((\pi_*(\varphi),\beta_*(\varphi)),\vartheta_*(\varphi))$ is non-singular.
\end{itemize}

\phantomsection\label{defA.3}
\begin{itemize}
\vspace{0.15cm}\item[] \-\hspace{0.0cm}{\small{\sf{\textbf{Definition A.3.}}}} Assume that $((\pi_*(\varphi),\beta_*(\varphi)),\vartheta_*(\varphi))$ be a solution that satisfies (\ref{eq:parc-1}).  A normal-limit solution is one where (i) the dimensionality of the Hessian nullspace is one, $\textnormal{dim}\, \textnormal{null}(\nabla^2_{\!\pi,\beta}\mathcal{L}((\pi_*(\varphi),\beta_*(\varphi)),\vartheta_*(\varphi))) \!=\! 1$, and (ii) the derivative, with respect to the exploration rate, of the Lagrangian is not in the range of the Hessian, $\partial_\vartheta \nabla_{\!\pi,\beta}\mathcal{L}((\pi_*(\varphi),\beta_*(\varphi)),\vartheta_*(\varphi)) \!\notin\! \textnormal{range}(\nabla^2_{\!\pi,\beta}\mathcal{L}((\pi_*(\varphi),\beta_*(\varphi)),\vartheta_*(\varphi)))$.\vspace{0.15cm}
\end{itemize}

We show that a linear operator can be defined that is non-singular for these two solution types.  First, we outline the conditions in which this occurs.\vspace{0.15cm}

\phantomsection\label{propA.4}
\begin{itemize}
\item[] \-\hspace{0.0cm}{\small{\sf{\textbf{Proposition A.4.}}}} Let $Q((\pi_k(\varphi_k),\beta_k(\varphi_k)),\vartheta_k(\varphi_k))$ be equal to
\begin{equation*}
\Bigg(\begin{matrix}\nabla^2_{\!\pi,\beta}\mathcal{L}((\pi_k(\varphi_k),\beta_k(\varphi_k)),\vartheta_k(\varphi_k)) & \partial_\vartheta\nabla_{\!\pi,\beta}\mathcal{L}((\pi_k(\varphi_k),\beta_k(\varphi_k)),\vartheta_k(\varphi_k))\vspace{0.05cm}\\ \theta (\dot{\pi}_k(\varphi_k),\dot{\beta}_k(\varphi_k))^* & (1 \!-\! \theta)\dot{\vartheta}_k(\varphi_k)\end{matrix}\Bigg).
\end{equation*}
If the top-right sub-matrix, $\nabla^2_{\!\pi,\beta}\mathcal{L}((\pi_k(\varphi_k),\beta_k(\varphi_k)),\vartheta_k(\varphi_k))$, is singular and the dimensionality of its nullspace is one, then $Q((\pi_k(\varphi_k),\beta_k(\varphi_k)),\vartheta_k(\varphi_k))$ is non-singular if 
\begin{itemize}
\item[] \-\hspace{0.5cm}(i) $\textnormal{dim}\, \textnormal{range}(\partial_\vartheta\nabla_{\!\pi,\beta}\mathcal{L}(\pi_k(\varphi_k),\beta_k(\varphi_k),\vartheta_k(\varphi_k))) \!=\! 1$,\vspace{0.01cm}
\item[] \-\hspace{0.5cm}(ii) $\textnormal{dim}\, \textnormal{range}(\theta (\dot{\pi}_k(\varphi_k),\dot{\beta}_k(\varphi_k))^*) \!=\! 1$,
\item[] \-\hspace{0.5cm}(iii) $\textnormal{range}(\partial_\vartheta\nabla_{\!\pi,\beta}\mathcal{L}(\pi_k(\varphi_k),\beta_k(\varphi_k),\vartheta_k(\varphi_k))) \cap \textnormal{range}(\nabla^2_{\!\pi,\beta}\mathcal{L}(\pi_k(\varphi_k),\beta_k(\varphi_k),\vartheta_k(\varphi_k))) \!=\! 0$,\vspace{-0.02cm}
\item[] \-\hspace{0.5cm}(iv) $\textnormal{null}(\partial_\vartheta\nabla_{\!\pi,\beta}\mathcal{L}(\pi_k(\varphi_k),\beta_k(\varphi_k),\vartheta_k(\varphi_k))) \cap \textnormal{null}(\theta (\dot{\pi}_k(\varphi_k),\dot{\beta}_k(\varphi_k))^*) \!=\! 0$.\vspace{0.15cm}
\end{itemize}
\end{itemize}

\noindent Note that it is straightforward to verify that \hyperref[propA.4]{Proposition A.4} holds if both $\theta (\dot{\pi}_k(\varphi_k),\dot{\beta}_k(\varphi_k))^*$ and $(1 \!-\! \theta)\dot{\vartheta}_k(\varphi_k)$ are replaced with the approximations given in (\ref{eq:parc-4}).

Next, we show that these conditions are satisfied for the two solution types.\vspace{0.15cm}

\phantomsection\label{propA.5}
\begin{itemize}
\item[] \-\hspace{0.0cm}{\small{\sf{\textbf{Proposition A.5.}}}} Let $((\pi_*(\varphi),\beta_*(\varphi)),\vartheta_*(\varphi))$ be either a regular solution point or a normal limit solution.  Let $\nabla_{\!\pi,\beta}\mathcal{L}((\pi,\beta),\vartheta)$, $\forall \pi,\beta,\varphi$, have two continuous derivatives in a ball about $(\pi_*(\varphi),\beta_*(\varphi),\vartheta_*(\varphi))$.  Then, there exists a unique, smooth curve of solutions when using the normalization (\ref{eq:parc-4}).  On this curve, the directional derivative of the linear operator,
\begin{equation}
G((\pi_*(\varphi),\beta_*(\varphi)),\vartheta_*(\varphi)) =\! \Bigg(\begin{matrix}\nabla^2_{\!\pi,\beta}\mathcal{L}(\pi_*(\varphi),\beta_*(\varphi),\vartheta_*(\varphi)) & \partial_\vartheta\nabla_{\!\pi,\beta}\mathcal{L}(\pi_*(\varphi),\beta_*(\varphi),\vartheta_*(\varphi))\vspace{0.05cm}\\ \nabla_{\!\pi,\beta}\mathcal{M}_\varphi((\pi_*,\beta_*),\vartheta_*) & \partial_\vartheta\mathcal{M}_\varphi((\pi_*,\beta_*),\vartheta_*)\end{matrix}\Bigg)
\label{eq:parc-5}
\end{equation}
is non-singular.
\begin{itemize}
\item[] \-\hspace{0.5cm}{\small{\sf{\textbf{Proof:}}}} This is a consequence of \hyperref[propA.3]{Proposition A.3} applied to $\nabla_{\!\pi,\beta}\mathcal{L}((\pi_k(\varphi_k),\beta_k(\varphi_k)),\vartheta_k(\varphi_k)) \!=\! 0$, provided that $G((\pi(\varphi_k),\beta(\varphi_k)),\vartheta(\varphi_k))$ is non-singular.  We hence only need to verify non-singularity for the two solution types.  In both cases, we use \hyperref[propA.4]{Proposition A.4} to do this.

We first consider the case where $((\pi_k(\varphi_k),\beta_k(\varphi_k)),\vartheta_k(\varphi_k)) \!=\! ((\pi_*(\varphi),\beta_*(\varphi)),\vartheta_*(\varphi))$ is a regular\\ \noindent solution.  \hyperref[defA.2]{Definition A.2}(ii) implies that
\begin{equation*}
(\dot{\pi}_*(\varphi),\dot{\beta}_*(\varphi))/\dot{\vartheta}_*(\varphi) = -\nabla^{-2}_{\!\pi,\beta}\mathcal{L}((\pi_*(\varphi),\beta_*(\varphi)),\vartheta_*(\varphi)) \partial_\vartheta \nabla_{\!\pi,\beta}\mathcal{L}((\pi_*(\varphi),\beta_*(\varphi)),\vartheta_*(\varphi)).
\end{equation*}
It can be shown that $(1 \!-\! \theta)\dot{\vartheta}_*(\varphi) \!-\! \theta (\dot{\pi}_*,\dot{\beta}_*)^*(\dot{\pi}_*(\varphi),\dot{\beta}_*(\varphi))/\dot{\vartheta}_*(\varphi) \!\neq\! 0$ is non-singular whenever\\ \noindent $\dot{\vartheta}_*(\varphi) \!\neq\! 0$ and hence $(\dot{\pi}_*(\varphi),\dot{\beta}_*(\varphi)) \!\neq\! 0$.  A similar expression is obtainable for the approximate case.\\ \noindent  We thus need to show that this is not possible.  Assume the converse, that is, $\dot{\vartheta}_*(\varphi) \!=\! 0$.  If this is true, then by\\ \noindent \hyperref[defA.2]{Definition A.2}(ii) we have that $(\dot{\pi}_*(\varphi),\dot{\beta}_*(\varphi)) \!=\! 0$.  This contradicts the branch-orientation condition,\\ \noindent $\theta\|(\dot{\pi}_*(\varphi),\dot{\beta}_*(\varphi))\|^2 \!+\! (1 \!-\! \theta)\dot{\vartheta}_*(\varphi) \!>\! 0$, and its approximate version.  Therefore, $\dot{\vartheta}_*(\varphi) \!\neq\! 0$ and hence\\ \noindent $(\dot{\pi}_*(\varphi),\dot{\beta}_*(\varphi)) \!\neq\! 0$.  The directional derivative of the operator is thus non-singular for regular solutions.

We now consider $((\pi_*(\varphi),\beta_*(\varphi)),\vartheta_*(\varphi))$ to be a normal limit point.  As a consequence of\\ \noindent \hyperref[defA.3]{Definition A.3}(ii), we have that $\dot{\vartheta}_*(\varphi) \!=\! 0$.  Hence, $(\dot{\pi}_*(\varphi),\dot{\beta}_*(\varphi)) \!\in\! \textnormal{null}(\nabla^2_{\!\pi,\beta}\mathcal{L}(\pi_*(\varphi),\beta_*(\varphi),\vartheta_*(\varphi)))$.\vspace{-0.035cm}\\ \noindent  Additionally, from \hyperref[defA.3]{Definition A.3}(ii), we get that $(\dot{\pi}_*(\varphi),\dot{\beta}_*(\varphi))^*(\dot{\pi}_*(\varphi),\dot{\beta}_*(\varphi)) \!\neq\! 0$ and therefore that\\ \noindent $(\dot{\pi}_*(\varphi),\dot{\beta}_*(\varphi))^* \!\notin\! \textnormal{range}(\nabla^2_{\!\pi,\beta}\mathcal{L}^*((\pi_*(\varphi),\beta_*(\varphi)),\vartheta_*(\varphi)))$.  These results, coupled with \hyperref[defA.3]{Definition A.3}(i) and \hyperref[propA.4]{Proposition A.4}, demonstrate that the directional derivative of the operator is non-singular for normal limit solutions.  The results hold in the approximate case too. $\blacksquare$
\end{itemize}
\end{itemize}

\vspace{0.15cm}\noindent Any smooth branch of solutions composed of either regular points or normal limit points can be determined using, say, Euler-Newton path-following for the normalization in (\ref{eq:parc-4}).  Pseudo-arc-length path-following is one instance of such a scheme, as the preliminary guesses are first-order Euler predictors which are then corrected by a corresponding series of Newton steps \cite{BohlE-jour1980a,SchwetlickH-jour1987a}.  

Here, we consider a slightly different version of the process outlined in \hyperref[sec4]{Section 4}.  As before, we find the tangent vector, $\partial_\varphi ((\pi_k(\varphi_k), \beta_k(\varphi_k)),\vartheta_k(\varphi_k))$, and use it to construct an initial solution guess via (\ref{eq:arc-cont-1}), where $\delta \!=\! \varphi \!-\! \varphi_k$, for some $\varphi$ in an interval along a solution curve.  This initial guess is then corrected via Newton's method, which entails solving the following system for the approximate steplength constraint,
\begin{multline}
G((\pi^{i-1}(\varphi_k),\beta^{i-1}(\varphi_k),\vartheta^{i-1}(\varphi_k)),\varphi)\Bigg(\begin{matrix}\pi^{i}_k(\varphi_k) \!-\! \pi^{i-1}_k(\varphi_k),\, \beta^i_k(\varphi_k) \!-\! \beta^{i-1}_k(\varphi_k)\\ \vartheta^i_k(\varphi_k) \!-\! \vartheta^{i-1}_k(\varphi_k)\end{matrix}\Bigg) =\vspace{0.025cm}\\ -\Bigg(\begin{matrix}\nabla_{\!\pi,\beta}\mathcal{L}^{i-1}(\pi^{i-1}(\varphi_k),\beta^{i-1}(\varphi_k),\vartheta^{i-1}(\varphi_k))\vspace{0.05cm}\\ \mathcal{M}_\varphi(\pi^{i-1}_k,\beta^{i-1}_k,\vartheta^{i-1}_k)\end{matrix} \Bigg).
\label{eq:parc-6}
\end{multline}
To demonstrate convergence, we only need to show that $((\pi^0_k(\varphi_k), \beta^0_k(\varphi_k)),\vartheta^0_k(\varphi_k))$ is in the appropriate domain of attraction around a solution $((\pi_*(\varphi), \beta_*(\varphi)),\vartheta_*(\varphi))$.  We also need that $G((\pi^i_k(\varphi_k),\beta^i_k(\varphi_k),\vartheta^i_k(\varphi_k)))$ is non-singular for each iterate $i$.

With these concepts, we can formally show that pseudo-arc-length path-following can sometimes jump over certain singular points when transitioning from one solution to the next for the value of information.

\phantomsection\label{defA.5}
\begin{itemize}
\vspace{0.15cm}\item[] \-\hspace{0.0cm}{\small{\sf{\textbf{Definition A.5.}}}} Let $((\pi_k(\varphi_*),\beta_k(\varphi_*)),\vartheta_k(\varphi_*))$ be a solution such that $\nabla_{\!\pi,\beta}\mathcal{L}((\pi_k(\varphi_*),\beta_k(\varphi_*)),\vartheta_k(\varphi_*)) \!=\! 0$\\ \noindent and $\mathcal{M}_{\varphi_*}(\pi_k,\beta_k,\vartheta_k) \!=\! 0$.  A singular solution point, or singular point, is one such that (\ref{eq:parc-5}) is singular for $\varphi_*$.\vspace{0.075cm}
\end{itemize}

\phantomsection\label{propA.6}
\begin{itemize}
\item[] \-\hspace{0.0cm}{\small{\sf{\textbf{Proposition A.6.}}}} Let $((\pi_{k+1}(\varphi_{k+1}),\beta_{k+1}(\varphi_{k+1})),\vartheta_{k+1}(\varphi_{k+1}))$ be a twice-differentiable path of solutions, $\varphi_{k+1} \!\in\! [\varphi_{k}^a,\varphi_{k}^b] \!-\! \{\varphi_*\}$, exist for the system
\begin{equation*}
\Bigg(\begin{matrix}\nabla_{\!\pi,\beta}\mathcal{L}((\pi_{k+1}(\varphi_{k+1}),\beta_{k+1}(\varphi_{k+1})),\vartheta_{k+1}(\varphi_{k+1}))\\ \mathcal{M}_\varphi((\pi_{k+1},\beta_{k+1}),\vartheta_{k+1})\end{matrix}\Bigg) \!= 0,
\end{equation*}
where $|\varphi_{k+1} \!-\! \varphi_{k}^a| \!<\! \epsilon$, $\epsilon \!>\! 0$.  Here, $\varphi_*$ represents a value of $\varphi_{k+1}$ for which a solution is a singular point.  Assume that we have a solution, for some $k \!+\! 1$, \noindent $((\pi_{k+1}(\varphi_{k+1}),\beta_{k+1}(\varphi_{k+1})),\vartheta_{k+1}(\varphi_{k+1})) \!=\! ((\pi_*(\varphi),\beta_*(\varphi)),\vartheta_*(\varphi))$,\\ \noindent which satisfies
\begin{equation*}
\Bigg(\begin{matrix}\nabla_{\!\pi,\beta}\mathcal{L}((\pi_*(\varphi),\beta_*(\varphi)),\vartheta_*(\varphi))\\ \mathcal{M}_\varphi((\pi_*,\beta_*),\vartheta_*)\end{matrix}\Bigg)\Bigg(\begin{matrix}(\dot{\pi}_*(\varphi),\dot{\beta}_*(\varphi))\\ \dot{\vartheta}_*(\varphi)\end{matrix}\Bigg) = -G((\pi_*(\varphi),\beta_*(\varphi)),\vartheta_*(\varphi))\Bigg(\begin{matrix}(\dot{\pi}_*(\varphi),\dot{\beta}_*(\varphi))\\ \dot{\vartheta}_*(\varphi)\end{matrix}\Bigg)
\end{equation*}
along with the algebraic bifurcation equations.  As well, assume that, for some positive constant that depends on the reparameterization term, $\kappa(\varphi_{k+1})$, $\textnormal{max}_{\varphi \leq \varphi_{k+1}}\, \|((\ddot{\pi}_{k+1}(\varphi),\ddot{\beta}_{k+1}(\varphi)),\ddot{\vartheta}_{k+1}(\varphi))\| \!\leq\! \kappa(\varphi_{k+1})$.  Additionally,\\ \noindent assume that (\ref{eq:parc-5}), for $\gamma_{k+1}(\varphi_{k+1}) \!=\! ((\pi_{k+1}(\varphi_{k+1}),\beta_{k+1}(\varphi_{k+1})),\vartheta_{k+1}(\varphi_{k+1}))$, is Lipschitz continuous, with constant $K(\varphi_{k+1})$, wherever the inequality $\|\gamma_{k}^i(\varphi_{k}) \!-\! \gamma_{k+1}(\varphi_{k+1})\| \!\leq\! \textstyle\frac{1}{2}\kappa(\varphi_{k+1})(\varphi_{k+1} \!-\! \varphi_{k}^a)^2$ is satisfied.  If\\ \noindent $\|G^{-1}(\gamma_{k+1}(\varphi_{k+1}))\|\kappa(\varphi_{k+1})K(\varphi_{k+1})(\varphi_{k+1} \!-\! \varphi_{k}^a)^2 \!<\! \frac{1}{2}$, then the iterates of (\ref{eq:parc-6}) converge at a rate that is at\\ \noindent least geometric to a solution of the value of information.
\begin{itemize}
\item[] \-\hspace{0.5cm}{\small{\sf{\textbf{Proof:}}}} We follow along the lines of Doedel et al. \cite{DoedelE-jour1991a}, albeit using an induction argument versus a contraction argument.  That is, we show that there is a double cone about the next solution, with the vertex of the cone at a singular point, $\varphi_*$; a visualization is given in \cref{fig:appendixa-singularpoint}.  To skip over this singular point, the tangent vector to the next solution, at the current solution, needs to penetrate this cone for some $\varphi_{k+1} \!>\! \varphi_*$.  This occurs\\ \noindent provided that the normed-solution-difference inequality is satisfied.  If the curvature of the solution path is too great, and hence the inequality is violated for any of the Newton steps, then the tangent vector lies outside of the cone and divergence occurs.

Let $\gamma_{k}^i(\varphi_{k}) \!=\! ((\pi_{k}^i(\varphi_{k}),\beta_{k}^i(\varphi_{k})),\vartheta_{k}^i(\varphi_{k}))$.  We need to show that there exists a term, $|\tau| \!<\! 1$, $\tau^i \!\to\! 0$, such\\ \noindent that $\|\gamma_{k}^i(\varphi_{k}) \!-\! \gamma_{k+1}(\varphi_{k+1})\| \!\leq\! C\tau^i$, with $C \!>\! 0$.  Once we find this term, for a series of base cases, then we\vspace{-0.025cm}\\ \noindent can use induction to verify it holds for all other cases and hence that geometric convergence is attained.

Consider the first iteration of (\ref{eq:parc-6}).  Using the definition of the linear operator, we have
\begin{equation*}
\|\gamma_{k}^1(\varphi_{k}) \!-\! \gamma_{k+1}(\varphi_{k+1})\| = \|G^{-1}(\gamma_{k}^0(\varphi_{k}))(G(\gamma_{k}^0(\varphi_{k})) \!-\! G(\gamma_{k+1}'(\varphi_{k+1})))(\gamma_{k}^0(\varphi_{k}) \!-\! \gamma_{k+1}(\varphi_{k+1}))\|,
\end{equation*}
where $\gamma_{k+1}'(\varphi_{k+1}) \!=\! \omega' \gamma_{k+1}(\varphi_{k+1}) \!+\! (1 \!-\! \omega')\gamma_{k+1}(\varphi_{k+1})$, with $\omega' \!\in\! [0,1]$.  We can bound some of the terms\\ \noindent that appear here and hence the iterate norm.  That is, $\|\gamma_{k}^0(\varphi_{k}) \!-\! \gamma_{k+1}(\varphi_{k+1})\| \!\leq\! \textstyle\frac{1}{2}\kappa(\varphi_{k+1})(\varphi_{k+1} \!-\! \varphi_{k}^a)^2$ and\\ \noindent $\|\gamma_{k+1}'(\varphi_{k+1}) \!-\! \gamma_{k+1}(\varphi_{k+1})\| \!\leq\! \textstyle\frac{1}{2}\kappa(\varphi_{k+1})(\varphi_{k+1} \!-\! \varphi_{k}^a)^2$.  Therefore,
\begin{align*}
\|\gamma_{k}^1(\varphi_{k}) \!-\! \gamma_{k+1}(\varphi_{k+1})\| &\leq \|G^{-1}(\gamma_{k}^0(\varphi_{k}))\|\|\gamma_{k}^0(\varphi_{k}) \!-\! \gamma_{k+1}(\varphi_{k+1})\|\|G(\gamma_{k}^0(\varphi_{k})) \!-\! G(\gamma_{k+1}'(\varphi_{k+1}))\|\\
 &\leq \sigma_{k+1}\|\gamma_{k}^0(\varphi_{k}) \!-\! \gamma_{k+1}(\varphi_{k+1})\|.\vspace{-0.075cm}
\end{align*}
where $\sigma_{k+1} \!=\! \kappa(\varphi_{k+1})K(\varphi_{k+1})(\varphi_{k+1} \!-\! \varphi_{k}^a)^2/(1 \!-\! \kappa(\varphi_{k+1})K(\varphi_{k+1})(\varphi_{k+1} \!-\! \varphi_{k}^a)^2)$.  The tangent vector for the initial iterate thus intersects the double cone with radius $\frac{1}{2}\kappa(\varphi_{k+1})(\varphi_{k+1} \!-\! \varphi_{k}^a)^2$.  This establishes the base case, with geometric convergence factor $\tau \!=\! \sigma_{k+1}$ and positive constant $C \!=\! \|\gamma_{k}^{i-1}(\varphi_{k}) \!-\! \gamma_{k+1}(\varphi_{k+1})\|$.

We can now show that this holds.  Assume $\|\gamma_{k}^j(\varphi_{k}) \!-\! \gamma_{k+1}(\varphi_{k+1})\| \!\leq\! \sigma_{k+1}\|\gamma_{k}^{j-1}(\varphi_{k}) \!-\! \gamma_{k+1}(\varphi_{k+1})\|$ at\vspace{-0.02cm}\\ \noindent iteration $j$, with $G(\gamma_{k}^i(\varphi_{k}))$ invertible for $i \!=\! 0,\ldots,j \!-\! 2$.  We know that 
\begin{equation*}
\|G^{-1}(\gamma_{k}^0(\varphi_{k}))(G(\gamma_{k}^j(\varphi_{k})) \!-\! G(\gamma_{k}^j(\varphi_{k})))\| \leq \kappa(\varphi_{k+1})K(\varphi_{k+1})(\varphi_{k+1} \!-\! \varphi_{k}^a)^2.
\end{equation*}
As well,
\begin{equation*}
G(\gamma_{k}^{j-1}(\varphi_{k})) = G(\gamma_{k}^0(\varphi_{k}))(\textnormal{id} \!+\! G^{-1}(\gamma_{k}^0(\varphi_{k}))(G(\gamma_{k}^{j-1}(\varphi_{k})) \!-\! G(\gamma_{k}^0(\varphi_{k})))),
\end{equation*}
where $\textnormal{id}$ is the identity operator.  Banach's lemma can be applied to deduce that (\ref{eq:parc-5}) is invertible and thus\\ \noindent that $\|\gamma_{k}^j(\varphi_{k}) \!-\! \gamma_{k+1}(\varphi_{k+1})\| \!\leq\! \sigma_{k+1}\|\gamma_{k}^{j-1}(\varphi_{k}) \!-\! \gamma_{k+1}(\varphi_{k+1})\|$.  The trend continues to hold from the base\\ \noindent case.  Finally, we show that it can be extended for one more iteration.  Since $G(\gamma_{k}^q(\varphi_{k}))$, $q \!>\! j$, is still invertible, we again can bound the normed difference in solutions, 
\begin{align*}
\|\gamma_{k}^{q}(\varphi_{k}) \!-\! \gamma_{k+1}(\varphi_{k+1})\| &\leq \|G^{-1}(\gamma_{k}^{q-1}(\varphi_{k}))\|\|\gamma_{k}^{q-1}(\varphi_{k}) \!-\! \gamma_{k+1}(\varphi_{k+1})\|\|G(\gamma_{k}^{q-1}(\varphi_{k})) \!-\! G(\gamma_{k+1}'(\varphi_{k+1}))\|\\
 &\leq \sigma_{k+1}\|\gamma_{k}^{q-1}(\varphi_{k}) \!-\! \gamma_{k+1}(\varphi_{k+1})\|.\vspace{-0.075cm}
\end{align*}
Both the convergence factor and constant again remain the same as in the base case, since $\gamma_{k}^{q-1}(\varphi_{k}),\gamma_{k}^q(\varphi_{k})$\\ \noindent still lie within the double cone.  

Given that $\varphi_{k+1} \!\neq\! \varphi_*$, the linear operator (\ref{eq:parc-5}) in (\ref{eq:parc-6}) will be non-singular and hence \hyperref[propA.3]{Proposition A.3}\\ \noindent applies.  An induction argument can be used to show geometric convergence of $\gamma_{k}^i(\varphi_{k}) \!\to\! \gamma_{k+1}(\varphi_{k+1})$. $\blacksquare$\vspace{0.15cm}
\end{itemize}
\end{itemize}

\noindent \hyperref[propA.6]{Proposition A.6} amends \hyperref[prop4.2]{Proposition 4.2} to show that pseudo-arc-length path-following will not get stuck, unlike parameter path-following.

\setcounter{figure}{0}
\begin{figure*}[t!]

\hspace{-0.295cm}\scalebox{0.775}{
\begin{tikzpicture}
  \node[] at (0.0,-2) {\includegraphics[width=5.5in]{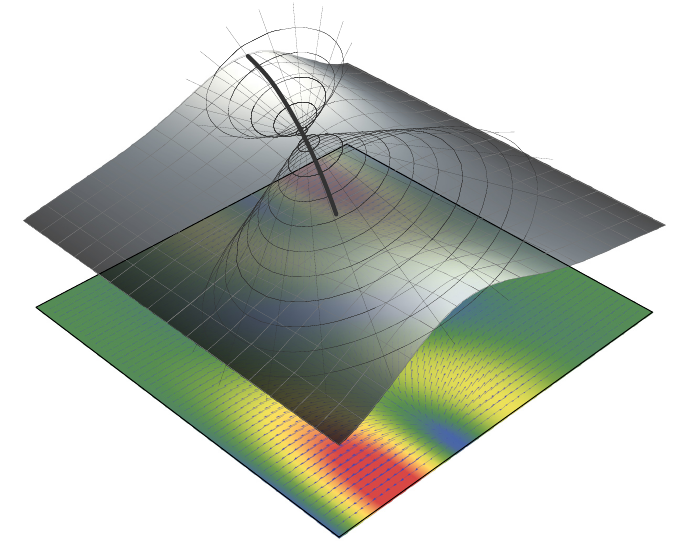}};

  \node at (-0.2,-5.7) {\large \textcolor{white}{$-\nabla_{\!\pi,\beta}\mathcal{L}((\pi,\beta),\vartheta)$}};
  \node at (3.7,1.45) {\large $\mathcal{L}((\pi,\beta),\vartheta)$};

  \node at (-4.1,1.95) {$((\pi_k(\varphi^a_k),\beta_k(\varphi^a_k)),\vartheta_k(\varphi^a_k))$};

  \node at (-4.65,-5.1) {\large $((\pi,\beta),\vartheta)(\varphi)_1$};
  \node at (4.65,-5.1) {\large $((\pi,\beta),\vartheta)(\varphi)_2$};

  \node at (-2.0,2.95) {\textcolor{black}{\boldmath$\varphi^a_k$}};
  \node at (-0.65,-0.65) {\textcolor{white}{\boldmath$\varphi_k^b$}};
  \node at (-0.5,0.35) {\textcolor{white}{\boldmath$\varphi_*$}};
  \node at (-1.0,-1.45) {\textcolor{white}{\boldmath$((\pi_{k}(\varphi_{k}^b),\beta_{k}(\varphi_{k}^b)),\vartheta_{k}(\varphi_{k}^b))$}};

  \draw[->,>=stealth,white,line width=0.85mm] (-1.95,2.4) -- (3.45,-0.675);

  \draw[dashed,white,line width=0.85mm] (-0.125,-0.925) -- (1.5,0.45);

  \node at (1.25,-0.475) {\textcolor{white}{\boldmath$r(\varphi)$}};

  \draw[fill, black] (-1.95,2.4) circle (4pt);
  \draw[fill, white] (1.5,0.45) circle (4.25pt);
  \draw[fill, black] (-0.125,-0.925) circle (4pt);
  \draw[fill, white] (-0.8,0.85) circle (4pt);

  \node at (3.7,-1.075) {\textcolor{white}{\boldmath$\partial_\varphi \gamma_k(\varphi_k))$}};

  \node at (-9.0,0.1) {\large $Q_k(s,a)$};
  \node at (-7.275,-2.05) {\large $s$};
  \node at (-9,-3.78) {\large $a$};
  \node at (9,-3.78) {\large $a$};
  \node at (7.275,-2.05) {\large $s$};
  \node at (9.0,0.1) {\large $Q_{k}(s,a)$};

  \node at (-9,-7.6) {\large $(\pi_k(\varphi_k^a),\beta_k(\varphi_k^a))$};
  \node at (9,-7.6) {\large $(\pi_{k}(\varphi_k^b),\beta_{k}(\varphi_k^b))$};

  \setlength{\fboxrule}{0.75pt}
  \setlength{\fboxsep}{0.025pt}
  \node at (-9.0,-0.25) {\includegraphics[width=1.39in]{sunsetbar.pdf}};
  \node at (-9,-2.05) {\framebox{\includegraphics[width=1.175in]{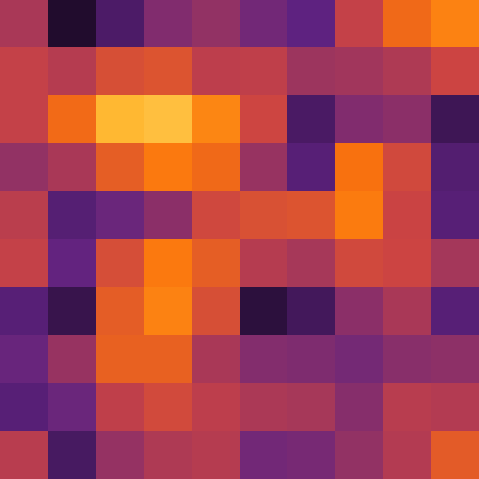}}};
  \node at (-9,-5.5) {\framebox{\embedvideo{\includegraphics[width=1.175in]{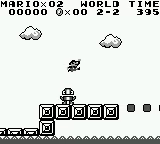}}{Level-2-2-1-noaudio.mp4}}};
  \node at (9,-5.5) {\framebox{\embedvideo{\includegraphics[width=1.175in]{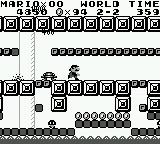}}{Level-2-2-2-noaudio.mp4}}};
  \node at (9,-2.05) {\framebox{\includegraphics[width=1.175in]{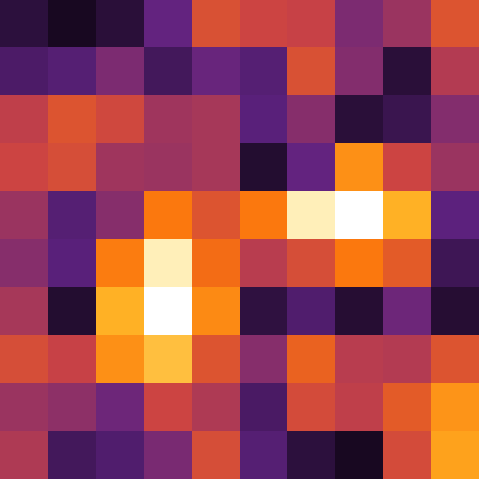}}};
  \node at (9.0,-0.25) {\includegraphics[width=1.39in]{sunsetbar.pdf}};

  \node at (-9,-7.15) {\includegraphics[width=1.39in]{papathfollowing-levelprogress-1.pdf}};
  \node at (9,-7.15) {\includegraphics[width=1.39in]{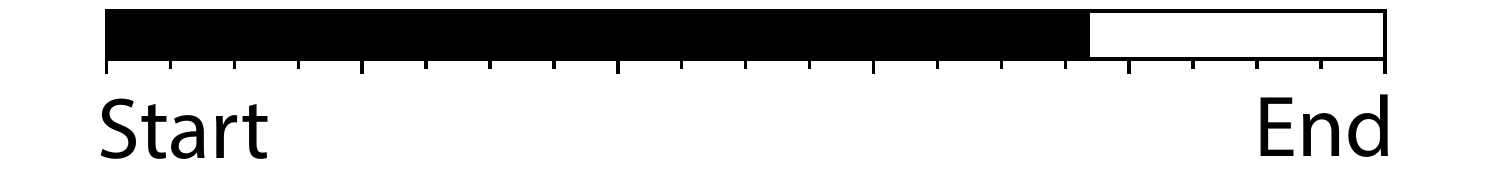}};

\end{tikzpicture}
}
   \caption[]{(middle) A visual overview of pseudo-arc-length path-following when encountering a singular point.  For a given starting point, $\gamma_k(\varphi_k) \!=\! ((\pi_k(\varphi_k),\beta_k(\varphi_k)),\vartheta_k(\varphi_k))$, we form the tangent vector $\partial_\varphi \gamma_k(\varphi_k))$ (white line).  We then have an interval of values for the arc-length parameter $\varphi$ to consider, which range from $\varphi_k^a$ (black circle) to $\varphi_k^b$ (black circle).  In this example, within this interval, a singular point, $\varphi_*$, exists (white circle).  A double cone can be fit about this singular point, the radius of which is bounded like $r(\varphi) \!=\! \frac{1}{2}\kappa(\varphi)(\varphi \!-\! \varphi_*)^2$ (dashed white line).  Here, $\varphi$ is a free parameter that changes along the solution curve $\nabla_{\pi,\beta}\mathcal{L}(\gamma(\varphi)) \!=\! 0$ (black line).  As long as the tangent vector intersects this double cone for some $\varphi \!>\! \varphi_*$ and $(\varphi \!-\! \varphi_k^a)^2 \kappa(\varphi) \!\leq\! 2r(\varphi)$, both of which occur for this example, then pseudo-arc-length path-following can jump over $\varphi_*$.  The update process can then proceed to a new iterate $\gamma_{k+1}(\varphi_{k+1})$, where, in this case, $\varphi_{k+1} \!=\! \varphi_k^b$.  For the end points of the arc-length spectrum, we provide corresponding embedded videos for \emph{Super\! Mario\! Land}.  (left)  At this point during learning, the agent has uncovered how to make it through several obstacles in this level.  However, if the update process became stuck at the singular point, then no new state groups would form.  It's likely that only small changes would be made to the policing and learning would effectively stop.  Based on a number of simulations, the gameplay behaviors are nearly equivalent to that depicted in this video.  (right) If path-following can hop over the singular point, either onto a new branch that intersects at that point or on the current branch, then learning can progress.  In this example, the agent learns to navigate deeper into the level and avoid troublesome enemies.  We also provide quantized $Q$-value tables for the ten dominant state-action groups.  We recommend viewing this document within Adobe Acrobat DC; click on an image and enable content to start playback of the corresponding video.\vspace{-0.2cm}}
   \label{fig:appendixa-singularpoint}
\end{figure*}

There is, however, an important issue which was not addressed in \cite{DoedelE-jour1991a}, which concerns the rate at which the\\ \noindent normed solution difference changes.  As $\varphi_{k+1} \!\to\! \varphi_*$, $G(\gamma_{k+1}(\varphi_{k+1}))$ becomes singular.  The closer the parameter\\ \noindent gets to the singular point, the smaller the double-cone radius also becomes.  This causes the convergence-rate factor\\ \noindent to become unbounded, implying that geometric convergence can no longer be obtained.

We therefore quantify, in \hyperref[propA.9]{Proposition A.9}, how quickly the convergence-rate factor becomes unbounded.  This permits us to suitably modify the conditions of \hyperref[propA.6]{Proposition A.6}, which we do in \hyperref[propA.11]{Proposition A.11}.

Toward this end, we first outline when the linear operator $G(\gamma_k(\varphi_k))$ possesses a similar structure to the system $(G(\gamma_k(\varphi_k)),\mathcal{M}_\varphi(\gamma_k))$ where the search-direction orientation is explicitly preserved. The following result will be crucial for this purpose.  We will explain, after introducing some additional notation in \hyperref[propA.9]{Proposition A.9}, why it is.

\phantomsection\label{propA.7}
\begin{itemize}
\vspace{0.15cm}\item[] \-\hspace{0.0cm}{\small{\sf{\textbf{Proposition A.7.}}}} Consider a linear operator of the form $G(\gamma_k(\varphi_k))$ in (\ref{eq:parc-5}).  Suppose that the top-left element of $G(\gamma_k(\varphi_k))$, $\nabla_{\!\pi,\beta}^2 \mathcal{L}(\gamma_k(\varphi_k))$, is a Fredholm operator of index zero that has zero as a simple eigenvalue.  Then $G(\gamma_k(\varphi_k))$ is a Fredholm operator of index zero.  We therefore have that $\textnormal{null}(\nabla_{\!\pi,\beta}^2 \mathcal{L}(\gamma_k(\varphi_k))) \!=\! \textnormal{span}(\phi_1)$ and\\ \noindent $\textnormal{null}(\nabla_{\!\pi,\beta}^2 \mathcal{L}(\gamma_k(\varphi_k))^*) \!=\! \textnormal{span}(\psi_1^*)$, where the eigenfunctions obey $\psi_1^*\phi_1 \!=\! 1$.  The nullspaces for the linear oper-\\ \noindent ator share a similar form, with $\textnormal{null}(G(\gamma_k(\varphi_k))) \!=\! \textnormal{span}(\phi)$ and $\textnormal{null}(G(\gamma_k(\varphi_k))^*) \!=\! \textnormal{span}(\psi^*)$, for eigenfunction $\phi$ and adjoint eigenfunction $\psi^*$.

This is true only in the following cases:
\begin{itemize}
\item[] \-\hspace{0.0cm}(i) The top-right element of $G(\gamma_k(\varphi_k))$, $\partial_\varphi\nabla_{\!\pi,\beta}\mathcal{L}(\gamma_k(\varphi_k))$, is not in the range of $\nabla_{\!\pi,\beta}^2 \mathcal{L}(\gamma_k(\varphi_k))$.  Moreover, $\phi_1 \!\in\! \textnormal{null}(\nabla_{\!\pi,\beta}\mathcal{M}_\varphi(\gamma_k))$.  In this case, we have that $\phi \!=\! (\phi_1,0)$ and $\psi^* \!=\! (\tau_0^* \!+\! \tau_1 \psi_1^*,1)$, where $\tau_0^*$ is the unique solution of $\nabla_{\!\pi,\beta}^2\mathcal{L}(\gamma_k(\varphi_k))\tau_0^* \!+\! \nabla_{\!\pi,\beta}\mathcal{M}_\varphi(\gamma_k) \!=\!0$ with the constraint that $\tau_0^*\phi_1 \!=\! 0$.  The term $\tau_1 \!=\! \psi^*\phi$, where $\psi^*\phi \!=\! -(\tau_0^*\partial_\vartheta \nabla_{\!\pi,\beta}\mathcal{L}(\gamma_k(\varphi_k)) \!+\! \partial_\vartheta \mathcal{M}_\varphi(\gamma_k))/(\psi_1^*\partial_\vartheta \nabla_{\!\pi,\beta}\mathcal{L}(\gamma_k(\varphi_k)))$.

\vspace{0.15cm}\item[] \-\hspace{0.0cm}(ii) $\partial_\varphi\nabla_{\!\pi,\beta}\mathcal{L}(\gamma_k(\varphi_k))$ is in the range of $\nabla_{\!\pi,\beta}^2 \mathcal{L}(\gamma_k(\varphi_k))$.  In this case, there is a unique solution $\tau_0^*$ such that $\nabla^2_{\!\pi,\beta}\mathcal{L}(\gamma_k(\varphi_k))\tau_0^* \!+\! \partial_\vartheta \nabla_{\!\pi,\beta}\mathcal{L}(\gamma_k(\varphi_i)) \!=\! 0$ where $\psi_1^*\tau_0^* \!=\! 0$.  Therefore, $(\tau_0^* \!+\! \tau_2\phi_1,1)\psi^* \!=\! (\psi_1^*,0)$ for\\ \noindent $\tau_2 \!=\! - (\nabla_{\!\pi,\beta}\mathcal{M}_\varphi(\gamma_k)\tau_0^* \!+\! \partial_\vartheta \mathcal{M}_\varphi(\gamma_k))/(\nabla_{\!\pi,\beta}\mathcal{M}_\varphi(\gamma_k)\phi_1)$ whenever $\nabla_{\!\pi,\beta}\mathcal{M}_\varphi(\gamma_k)\phi_1 \!\neq\! 0$.  If, however, the\\ \noindent denominator of $\tau_2$ is zero but the numerator is not, then $\psi^*\phi \!=\! 1$, where $\phi \!=\! (\phi_1,0)^\top$ and $\psi \!=\! (\psi_1^*,0)$.
\end{itemize}

\noindent In either case, the linear operator will possess a simple eigenvalue if $\tau_1 \!\neq\! 0$ and $\tau_2 \!\neq\! 0$.   
\end{itemize}
\begin{itemize}
\item[] \-\hspace{0.5cm}{\small{\sf{\textbf{Proof:}}}} In both instances, we use direct proofs to specify the forms of the eigenfunctions.  We then invoke the Fredholm Alternative Theorem to demonstrate the existence and uniqueness of solutions.

We consider the case $\partial_\varphi\nabla_{\!\pi,\beta}\mathcal{L}(\gamma_k(\varphi_k)) \!\notin\! \textnormal{range}(\nabla_{\!\pi,\beta}^2 \mathcal{L}(\gamma_k(\varphi_k)))$.  We can see that $G(\gamma_k(\varphi_k))(\rho\phi_1,0) \!=\! 0$.\\ \noindent  This implies that there is a unique eigenvector the linear operator, up to some multiplicative scalar $\rho \!\in\! \mathbb{R}$, provided\\ \noindent that $\nabla_{\!\pi,\beta}\mathcal{M}_\varphi(\gamma_k) \phi_1 \!=\! 0$.  We can thus take this eigenvector to be $\phi \!=\! (\phi_1,0)$.  Likewise, $(\tau^*,\upsilon)G(\gamma_k(\varphi_k)) \!=\! 0$ if\\ \noindent $\phi_1 \!\in\! \textnormal{null}(\nabla_{\!\pi,\beta}\mathcal{M}_\varphi(\gamma_k))$.  Multiplying the two terms, we get that $\nabla^2_{\!\pi,\beta}\mathcal{L}(\gamma_k(\varphi_k))^*\tau^* \!+\! \upsilon\nabla_{\!\pi,\beta}\mathcal{M}_\varphi(\gamma_k) \!=\! 0$ and\\ \noindent $\tau^*\partial_\vartheta\nabla_{\!\pi,\beta}\mathcal{L}(\gamma_k(\varphi_k)) \!+\! \upsilon \partial_\vartheta \mathcal{M}_\varphi(\gamma_k) \!=\! 0$.  Since $\nabla_{\!\pi,\beta}\mathcal{M}_\varphi(\gamma_k) \!\in\! \textnormal{range}(\nabla_{\!\pi,\beta}^2 \mathcal{L}(\gamma_k(\varphi_k)))^*$, there exists a unique\\ \noindent $\tau_0^*$, for $\tau_0^*\phi_1 \!=\! 0$, such that $\nabla^2_{\!\pi,\beta}\mathcal{L}(\gamma_k(\varphi_k))^*\tau^*_0 \!+\! \nabla_{\!\pi,\beta}\mathcal{M}_\varphi(\gamma_k) \!=\! 0$.  Therefore, $\tau^* \!=\! \upsilon \tau_0^* \!+\! \rho \psi_1^*$, which implies that 
$$\upsilon(\tau_0^* \!+\! \partial_\vartheta \nabla_{\!\pi,\beta}\mathcal{L}(\gamma_k(\varphi_k))) + \rho\psi_1^*\partial_\vartheta \nabla_{\!\pi,\beta}\mathcal{L}(\gamma_k(\varphi_k)) = 0.$$
After re-arranging terms, we can arrive at an expression for the multiplicative scalar and therefore $\tau_1$,
$$\rho = -\upsilon(\tau_0^* \partial_\vartheta \nabla_{\!\pi,\beta}\mathcal{L}(\gamma_k(\varphi_k)) \!+\! \partial_\vartheta\mathcal{M}_\varphi(\gamma_k))/(\psi_1^* \partial_\vartheta \nabla_{\!\pi,\beta}\mathcal{L}(\gamma_k(\varphi_k))),\;\; \tau_1 = \upsilon^{-1}\rho.$$
Hence, $\psi^* \!=\! (\tau_0^* \!+\! \tau_1\psi_1^*,1)$ is a unique adjoint eigenvector, since it corresponds to a distinct eigenvalue.\\ \noindent  Additionally, it follows that $\psi^*\phi \!=\! \tau_1$.

We now consider when $\partial_\varphi\nabla_{\!\pi,\beta}\mathcal{L}(\gamma_k(\varphi_k)) \!\in\! \textnormal{range}(\nabla_{\!\pi,\beta}^2 \mathcal{L}(\gamma_k(\varphi_k)))$.  We have $G(\gamma_k(\varphi_k))(\tau\tau_0^* \!+\! \rho \phi_1,\tau)$,\\ \noindent with $\rho,\tau \!\in\! \mathbb{R}$.  The term $\tau_0^*$ is the unique solution of $\nabla^2_{\!\pi,\beta}\mathcal{L}(\gamma_k(\varphi_k))^*\tau_0^* \!+\! \partial_\vartheta\nabla_{\!\pi,\beta}\mathcal{L}(\gamma_k(\varphi_k)) \!=\! 0$ with the con-\\ \noindent straint that $\psi_1^*\tau_0^* \!=\! 0$.  We have that 
$$\tau(\nabla_{\!\pi,\beta}\mathcal{L}(\gamma_k(\varphi_k))^*\tau_0^* \!+\! \partial_\vartheta \mathcal{M}_\varphi(\gamma_k)) + \rho\nabla_{\!\pi,\beta}\mathcal{L}(\gamma_k(\varphi_k))^*\phi_1 \!=\! 0.$$
We can solve for both $\tau$ and $\rho$ if both $\nabla_{\!\pi,\beta}\mathcal{L}(\gamma_k(\varphi_k))^*\phi_1$ and $\nabla_{\!\pi,\beta}\mathcal{L}(\gamma_k(\varphi_k))^*\tau_0^* \!+\! \partial_\vartheta \mathcal{M}_\varphi(\gamma_k)$ do not evaluate\\ \noindent to zero.  If this is true, then, as in the first case, we can specify a term 
$$\tau_2 \!=\! - (\nabla_{\!\pi,\beta}\mathcal{M}_\varphi(\gamma_k)\tau_0^* + \partial_\vartheta \mathcal{M}_\varphi(\gamma_k))/(\nabla_{\!\pi,\beta}\mathcal{M}_\varphi(\gamma_k)\phi_1),$$
with $\phi \!=\! (\tau_0^* \!+\! \tau_2\phi_1,1)\psi^*$.  If, however, $\nabla_{\!\pi,\beta}\mathcal{L}(\gamma_k(\varphi_k))^*\phi_1 \!=\! 0$, then $\phi \!=\! (\phi_1,0)$.  To find $\psi$, we proceed in a\\ \noindent manner similar to that of $\phi$ in the first case.  We rely on the fact that $\nabla^2_{\!\pi,\beta}\mathcal{L}(\gamma_k(\varphi_k))\tau^* \!+\! \upsilon \nabla_{\!\pi,\beta}\mathcal{M}_\varphi(\gamma_k)^* \!=\! 0$ and $\tau^* \partial_\vartheta \nabla_{\!\pi,\beta} \mathcal{L}(\gamma_k(\varphi_k)) \!+\! \upsilon \partial_\vartheta \mathcal{M}_\varphi(\gamma_k)^* \!=\! 0$ and solve to find $\rho$ and hence $\tau_2$.  If we assume $\nabla_{\!\pi,\beta}\mathcal{M}_\varphi(\gamma_k)^*\phi_1 \!\neq\! 0$,\\ \noindent then $\tau^* \!=\! \rho \psi_1^*$.  As well, we get that $\upsilon \!=\! 0$.  We therefore have the requirement that $\rho \psi_1^* \partial_\vartheta \nabla_{\!\pi,\beta} \mathcal{L}(\gamma_k(\varphi_k)) \!=\! 0$,\\ \noindent which occurs for any $\rho \!\in\! \mathbb{R}$.  Therefore, $\psi^* \!=\! (\psi_1^*,0)$.  Now, suppose that $\nabla_{\!\pi,\beta}\mathcal{M}_\varphi(\gamma_k)^*\phi_1 \!=\! 0$ but where we have\\ \noindent $\nabla_{\!\pi,\beta}\mathcal{M}_\varphi(\gamma_k)^*\phi_0 \!+\! \partial_\vartheta \mathcal{M}_\varphi(\gamma_k) \!\neq\! 0$.  In this instance, $\nabla_{\!\pi,\beta}\mathcal{M}_\varphi(\gamma_k)^* \!\in\! \textnormal{range}(\nabla_{\!\pi,\beta}^2 \mathcal{L}(\gamma_k(\varphi_k))^*)$.  We can see that $\tau^* \!=\! \upsilon \tau_0^* \!+\! \rho \psi_1^*$ under the condition 
$$\upsilon (\tau_0^*\partial_\vartheta \nabla_{\!\pi,\beta}\mathcal{L}(\gamma_k(\varphi_k))) \!+\! \partial_\vartheta\mathcal{M}_\varphi(\gamma_k)) + \rho\psi_1^*\partial_\vartheta \nabla_{\!\pi,\beta}\mathcal{L}(\gamma_k(\varphi_k)) = 0.$$
Since $\psi_1^* \partial_\vartheta \nabla_{\!\pi,\beta}\mathcal{L}(\gamma_k(\varphi_k)) \!=\! 0$, we get $\upsilon (\tau_0^*\partial_\vartheta \nabla_{\!\pi,\beta}\mathcal{L}(\gamma_k(\varphi_k))) \!=\! 0$ too.  It is straightforward to show that\\ \noindent $\nabla_{\!\pi,\beta}\mathcal{M}_\varphi(\gamma_k)\phi_0 \!+\! \partial_\vartheta\mathcal{M}_\varphi(\gamma_k) \!\neq\! 0$ and $\tau_0^* \partial_\vartheta\nabla_{\!\pi,\beta}\mathcal{L}(\gamma_k(\varphi_k)) \!+\! \partial_\vartheta\mathcal{M}_\varphi(\gamma_k) \!\neq\! 0$.  Both expressions follow from\\ \noindent $\partial_\vartheta\nabla_{\!\pi,\beta} \mathcal{L}(\gamma_k(\varphi_k)) \!=\! -\nabla^2_{\!\pi,\beta}\mathcal{L}(\gamma_k(\varphi_k))\phi_0$, which implies that $\tau_0^* \partial_\vartheta\nabla_{\!\pi,\beta} \mathcal{L}(\gamma_k(\varphi_k))\!=\! -\tau_0^* \nabla^2_{\!\pi,\beta}\mathcal{L}(\gamma_k(\varphi_k))\phi_0 \!=$\\ \noindent $\nabla_{\!\pi,\beta}\mathcal{M}_\varphi(\gamma_k)^*\phi_0$.  Therefore, $\upsilon \!=\! 0$, just like it did before, and the adjoint eigenfunction is $\psi^* \!=\! (\psi_1^*,0)$.

Above, we have made the assumption that $\tau_0^*$ is a unique solution for various equations.  This, however, needs to be verified.

Since $G(\gamma_k(\varphi_k))$ is a Fredholm operator of index zero, we can use a weark form of the Fredholm Alternative Theorem \cite{RammAG-jour2001a} to demonstrate the existence and uniqueness of $\tau_0^*$.  For the theorem to apply, we need only show that the range space of the operator is closed.  The remaining conditions of the theorem are trivially satisfied. $\blacksquare$\vspace{0.15cm}

\end{itemize}

\noindent \hyperref[propA.7]{Proposition A.7} enables us to work with just $G(\gamma_k(\varphi_k))$ in subsequent proofs.  The linear operator is much less cumbersome to analyze than $(G(\gamma_k(\varphi_k)),\mathcal{M}_\varphi(\gamma_k))$, since we do not have to handle an added constraint.

We will now demonstrate that, for $\varphi_k$ near a singular point $\varphi_*$, the linear operator $G(\gamma_k(\varphi_k))$ has a small simple eigenvalue that is real.  Moreover, if the operator is differentiable, then so are $\alpha(\varphi_k)$ and $\phi(\varphi_k)$.  Differentiability is lost without a simple eigenvalue, but we show that the eigenvalue will usually be simple.

\phantomsection\label{propA.8}
\begin{itemize}
\vspace{0.15cm}\item[] \-\hspace{0.0cm}{\small{\sf{\textbf{Proposition A.8.}}}} Suppose that $G(\gamma_k(\varphi_k))$ in (\ref{eq:parc-5}) satisfies the conditions in \hyperref[propA.7]{Proposition A.7} at a solution point for the parameter $\varphi_*$, where $|\varphi_k \!-\! \varphi_*| \!<\! \epsilon_0$, $\epsilon_0 \!>\! 0$.  As well, assume that $\psi^*(\varphi_*) \!=\! \psi^*$, where $G(\gamma_k(\varphi_*))^*\psi^* \!=\! 0$.  There exists some $\epsilon_1 \!>\! 0$ such that, for $|\varphi_k \!-\! \varphi_*| \!<\! \epsilon_1$, $G(\gamma_k(\varphi_k))\phi(\varphi_k) \!=\! \alpha(\phi_k)\phi(\varphi_k)$, with $\psi^*(\varphi_k)\phi(\varphi_k) \!=\! 1$,\\ \noindent where $\alpha(\varphi_k)$ is an eigenvalue and $\phi(\varphi_k)$ is an eigenfunction.  At $\varphi_k \!=\! \varphi_*$, $\alpha(\varphi_*) \!=\! 0$ and $G(\gamma_k(\varphi_*))\phi(\varphi_*) \!=\! 0$.\\ \noindent  The eigenvalue remains simple in this case.\vspace{0.15cm}
\end{itemize}
\begin{itemize}
\item[] \-\hspace{0.5cm}{\small{\sf{\textbf{Proof:}}}} We define two operators, $T_1(\gamma_k(\varphi_k);u,v) \!=\! G(\gamma_k(\varphi_k))u \!-\! vu$ and $T_2(\gamma_k(\varphi_k);u,v) \!=\! \psi^*(\varphi_k)u \!-\! 1$,\\ \noindent both of which stem from the two claims that we wish to prove.  If we assume that $\varphi_k \!=\! \varphi_*$, then it is easy to see\\ \noindent that $T_1(\gamma_k(\varphi_*);\phi,0) \!=\! T_2(\gamma_k(\varphi_*);\phi,0) \!=\! 0$.  Taking the directional derivative $\varphi_k \!=\! \varphi_*$, we have that\\ \noindent $\partial T_1/\partial (u,v)|_{\varphi_k = \varphi_*}$ and $\partial T_2/\partial (u,v)|_{\varphi_k = \varphi_*}$ are non-singular, since we assume that $\psi^*\phi \!=\! 1$.  Given that the\\ \noindent operators are continuously differentiable, \hyperref[propA.3]{Proposition A.3} can be applied to show solution existence.

The proof for eigenvalue simplicity is nearly the same.  We define a linear operator
\begin{equation*}
T_3(\gamma_k(\varphi_k);u,v) = \Bigg(\begin{matrix}G(\gamma_k(\varphi_k)) \!-\! \alpha(\varphi_k)\textnormal{id} & \phi(\varphi_k)\vspace{0.025cm}\\ \psi^*(\varphi_k) & 0\end{matrix}\Bigg)\Bigg(\begin{matrix}u\vspace{0.025cm}\\ v\end{matrix}\Bigg),
\end{equation*}
and evaluate its directional derivatives at $\varphi_k \!=\! \varphi_*$, $\partial T_3/\partial (u,v)|_{\varphi_k = \varphi_*}$.  The directional derivative yields a matrix which is non-singular and does not depend on either $u$ or $v$.  Therefore, the trivial solution of $T_3(\gamma_k(\varphi_k);u,v) \!=\! 0$ is the only solution, for any $(u,v)$, which is sufficient to demonstrate eigenvalue simplicity. $\blacksquare$
\end{itemize}

\vspace{0.15cm}\noindent \hyperref[propA.3]{Proposition A.3} only guarantees the uniqueness of a formed solution around a known solution point.  Our arguments in \hyperref[propA.8]{Proposition A.8} do not suffer from this issue.  This is because, in the linear case, local uniqueness of a solution is equivalent to global uniqueness.

Since the eigenfunctions are provably differentiable, we can study their rate of change as $\varphi_k$ approaches $\varphi_*$.  This will be needed for our convergence result in \hyperref[propA.12]{Proposition A.12}.

\phantomsection\label{propA.9}
\begin{itemize}
\vspace{0.15cm}\item[] \-\hspace{0.0cm}{\small{\sf{\textbf{Proposition A.9.}}}} Let there be a set of smooth functions, which are at least twice differentiable, that contain $\gamma_k(\varphi_*)$, a solution point of the value of information.  For the linear operator $G(\gamma_k(\varphi_k))$ in (\ref{eq:parc-5}), we have that,
\begin{itemize}
\item[] \-\hspace{0.5cm}(i) If $G(\gamma_{k}(\varphi_{k}))$ has an single eigenvalue of zero, then it goes to zero like $\alpha(\varphi_k) \!=\! O(|\varphi_k \!-\! \varphi_*|)$. 
\item[] \-\hspace{0.5cm}(ii) If $G(\gamma_{k}(\varphi_{k}))$ has two zero eigenvalues, then both go to zero like $\alpha(\varphi_k) \!=\! O(|\varphi_k \!-\! \varphi_*|^{1/2})$.
\end{itemize}
\end{itemize}
\begin{itemize}
\item[] \-\hspace{0.5cm}{\small{\sf{\textbf{Proof:}}}} We assume that the linear operator can be decomposed as $G(\gamma_{k}(\varphi_{k})) \phi(\varphi_k) \!=\! \alpha(\varphi_k)\phi(\varphi_k)$, where\\ \noindent $\alpha(\varphi_k)$ are eigenvalues and $\phi(\varphi_k)$ are eigenvectors, both of which naturally depend on arc-length.  Since we are interested in the rate at which one or both eigenvalues approach zero, for a changing arc-length, we differentiate the eigenfunction expression,
\begin{equation*}
\textstyle \psi^*(\varphi_{k}) G(\gamma_{k}(\varphi_{k}))\partial_\varphi \phi(\varphi_k) + \psi^*(\varphi_{k})\partial_\varphi G(\gamma_{k}(\varphi_{k})) \phi(\varphi_k) = \partial_\varphi\alpha(\varphi_k) \psi^*(\varphi_{k}) \phi(\varphi_k) + \alpha(\varphi_k) \psi^*(\varphi_{k}) \partial_\varphi \phi(\varphi_k).
\end{equation*}
Here, we have applied the adjoint eigenfunction, $\psi(\varphi_k)$, which satisfies $(\nabla_{\pi,\beta}\mathcal{L}^*(\gamma_k) \!-\! \alpha(\varphi_k)\textnormal{id})\psi^*(\varphi_k) \!=\! 0$.\\ \noindent We use \hyperref[propA.7]{Proposition A.7} to normalize the adjoint eigenfunction as $\psi^*(\varphi_k)\phi(\varphi_k) \!=\! 1$.

We now evaluate the eigenfunction derivative at $\varphi_k \!=\! \varphi_*$, which corresponds to a solution point.  Since\\ \noindent $\alpha(\varphi_*) \!=\! 0$ and $\psi^*(\varphi_*)G(\gamma_{k}(\varphi_*)) \!=\! 0$, we can reduce the expression to a more manageable one, that facilitates\\ \noindent finding $\partial_\varphi\alpha(\varphi_k)$, $\psi^*(\varphi_*)\partial_\varphi G(\gamma_{k}(\varphi_*)) \phi(\varphi_*) \!=\! \partial_\varphi\alpha(\varphi_*) \psi^*(\varphi_*) \phi(\varphi_*)$.  We can do this whenever the linear\\ \noindent operator has a zero eigenvalue with algebraic multiplicity one.  This condition implies that $\psi^*(\varphi_*) \phi(\varphi_*) \!\neq\! 0$,\\ \noindent since it is actually equal to one according to our normalization condition.  Hence, a trivial solution of $\partial_\varphi\alpha(\varphi_*) \!=\! 0$ is not realized and we can bound the rate of change for the eigenvalue. %\vspace{-0.025cm}\\ \noindent

We first consider when the linear operator has a single eigenvalue of zero.  In this case, we can systematically reduce the eigenfunction expression, $\partial_\varphi\alpha(\varphi_*) \psi^*(\varphi_*) \phi(\varphi_*) \!=\! \partial_\varphi\alpha(\varphi_*)$, which we will do in two stages.\\ \noindent  First, we will show that, in some instances, we can heavily simplify the infinite-dimensional problem of finding a value-of-information policy to that of solving a finite set of constrained polynomial equations \cite{StakgoldI-jour1971a,SattingerDH-book1979a}.  The solution to these equations permit quantifying $\partial_\varphi (\pi_k(\varphi_*),\beta_k(\varphi_*))$ and $\partial_\varphi \vartheta_k(\varphi_*)$ and hence $\partial_\varphi G(\gamma_{k}(\varphi_*))$.  Second, we will show that if $\partial_\varphi G(\gamma_{k}(\varphi_*)) \!\neq\! 0$, then it becomes possible to specify how $\alpha(\varphi_*)$ tends to zero.

Since $\partial_\vartheta \nabla_{\pi,\beta}\mathcal{L}(\gamma_{k}(\varphi_*)) \!\in\! \textnormal{range}(\nabla_{\pi,\beta}^2\mathcal{L}(\gamma_{k}(\varphi_*)))$, we have that $\partial_\varphi({\pi}_k(\varphi_*)),{\beta}_k(\varphi_*)) \!=\! \sum_{j=0}^m \xi_j\phi_j$.  Here,\vspace{-0.025cm}\\ \noindent $\xi_j$ are scalars, with the first element being $\xi_* \!=\! \partial_\varphi\vartheta_k(\varphi_*)$, while $\phi_*$ is the unique solution of
\begin{align*}
\nabla^2_{\pi,\beta}\mathcal{L}(\gamma_{k}(\varphi_*))\phi_* + \partial_\vartheta \nabla_{\pi,\beta}\mathcal{L}(\gamma_{k}(\varphi_*)) &= 0\\
\psi_j^* \phi_* &= 0\vspace{-0.02cm}\\
\textstyle\sum_{j=1}^m \sum_{p=1}^m \omega_{i,j,p}\xi_j\xi_p + 2 \sum_{j=1}^m \omega_{i,j} \xi_j\xi_0 + \omega_i \xi_0^2 &= 0
\end{align*}
We refer to the left portion of the last line as $\rho_{i}(\xi_0,\ldots,\xi_i)$, which, in \cite{KeenerJP-jour1979a,JepsonAD-jour1986a}, is called the algebraic bifurcation equation.  For $i,j,p \!\in\! 1,\ldots,m$, the coefficients of this equation are given by\vspace{-0.05cm}
\begin{align*}
\psi_i^*(\varphi_k)\nabla_{\pi,\beta}^3\mathcal{L}(\gamma_{k}(\varphi_*)) \phi_j\phi_p &= \omega_{i,j,p}\\
\psi_i^*(\varphi_k) (\nabla_{\pi,\beta}^3\mathcal{L}(\gamma_{k}(\varphi_*))\phi_0 + \partial_\vartheta\nabla^2_{\pi,\beta}\mathcal{L}(\gamma_{k}(\varphi_*))) \phi_j &= \omega_{i,j}\\
\psi_i^*(\varphi_k) (\nabla_{\pi,\beta}^3\mathcal{L}(\gamma_{k}(\varphi_*))\phi_0\phi_0 + 2\partial_\vartheta\nabla_{\pi,\beta}^2\mathcal{L}(\gamma_{k}(\varphi_*))\phi_0 + \partial_\vartheta^2\nabla_{\pi,\beta}\mathcal{L}(\gamma_{k}(\varphi_*))) &= \omega_i
\end{align*}
with $m$ being the zero-eigenvalue multiplicity for $\nabla^2_{\pi,\beta}\mathcal{L}(\gamma_{k}(\varphi_*))$.  For the remaining unknowns, we have used the assumption that $\nabla^2_{\pi,\beta}\mathcal{L}(\gamma_{k}(\varphi_*)\partial_\varphi^2(\pi_k(\varphi_*),\beta_k(\varphi_*)) \!+\! \partial_\vartheta\nabla_{\pi,\beta}\mathcal{L}(\gamma_{k}(\varphi_*) \partial_\varphi^2 \vartheta(\varphi_k) \!\in\! \textnormal{range}(\nabla_{\pi,\beta}^2\mathcal{L}(\gamma_{k}(\varphi_*)))$.

We now suppose that $\gamma_k(\varphi_*)$ are such that $\partial_\varphi (\pi_k(\varphi_*),\beta_k(\varphi_*)) \!=\! \xi_0\phi_0 \!+\! \xi_1 \phi_1$ and $\partial_\varphi \vartheta_k(\varphi_*) \!=\! \xi_0$, where\\ \noindent $\xi_0,\xi_1$ are the solutions to $\rho_i(\xi_0,\xi_1) \!=\! 0$.  Since $\psi^*(\varphi_*)\phi(\varphi_*) \!=\! (\psi_1^*,0)(\phi_1,0)^\top \!=\! 1$ whenever $G(\gamma_{k}(\varphi_*))$ has a\\ \noindent zero eigenvalue with multiplicity one, we get that
\begin{equation*}
\partial_\varphi \alpha(\varphi_*) = (\psi_1^*,0)\partial_\varphi G(\gamma_{k}(\varphi_*)) (\phi_1,0)^\top = (\xi_0 \omega_{1,1} \!+\! \xi_1 \omega_{1,1,1}) - (\xi_1/2\xi_0)(\xi_0\omega_1 \!+\! \xi_1 \omega_{1,1}),
\end{equation*}
which is strictly positive.  Here, the $\xi_1/2\xi_0$ term emerges from \hyperref[propA.8]{Proposition A.8} along with the relationships\\ \noindent $\nabla_{\pi,\beta}\mathcal{M}_\varphi(\gamma_k(\varphi_*))\phi_0 \!+\! \partial_\vartheta\mathcal{M}_\varphi(\gamma_k(\varphi_*)) \!=\! 2 \xi_0$ and $\nabla_{\pi,\beta}\mathcal{M}_\varphi(\gamma_k(\varphi_*))\phi_1 \!=\! \xi_1$.  If $\xi_0 \!\neq\! 0$, then we can use the\\ \noindent definition of $\mathcal{M}_\varphi(\gamma_k(\varphi_*))$, in (\ref{eq:parc-4}) to algebraically simplify the eigenvalue derivative further.  We then find that $\alpha(\varphi_k)$ goes to zero like $O(|\varphi_k \!-\! \varphi_*|)$ as $\varphi_k \!\to\! \varphi_*$.

We can now consider the situation where the linear operator is singular.  As before, we systematically reduce the eigenfunction expression, $\partial_\varphi\alpha(\varphi_*) \psi^*(\varphi_*) \phi(\varphi_*) \!=\! \partial_\varphi\alpha(\varphi_*)$.  The process is a bit more complicated, though, than the above case.  We will first show that we can consider a separate linear operator, $B(\varphi_k)$, in one of two subspaces of the iterate Banach space.  This operator has the same eigenvalues as $G(\gamma_{k}(\varphi_{k}))$ when restricted to the other subspace.  We will then use the solutions for the constrained polynomial equations to simplfy $\partial_\varphi B(\varphi_{k}))$\\ \noindent and hence $\partial_\varphi G(\gamma_{k}(\varphi_{k}))$ to assess the change in $\alpha(\varphi_k)$.

In this case, the normalization constraints are $\nabla_{\pi,\beta}\mathcal{M}_\varphi(\gamma_k(\varphi_*)) \!=\! \phi^*_1$ and $\partial_\varphi\mathcal{M}_\varphi(\gamma_k(\varphi_*)) \!=\! 0$.  After taking\\ \noindent into account \hyperref[propA.8]{Proposition A.8}, we can conclude that $\textnormal{range}(\nabla^2_{\pi,\beta}\mathcal{L}(\gamma_{k}(\varphi_*)))$ has a co-dimension of one.  This fact allows us to decompose the Banach space in a way that makes the operator $G(\gamma_{k}(\varphi_*))$ non-singular on one of the subspaces.  Such a property is crucial, since we will be attempting to use the inferred eigenvalue rate to bound the inverse operator.

Let $H(\gamma_{k}(\varphi_{k}))$ be a block matrix with $G(\gamma_{k}(\varphi_{k}))$ on the diagonal and zeros on the off-diagonals.  Let $B(\varphi_k)$ be a block matrix, where, at a solution, it becomes a block lower-triangular matrix of ones.  From \cite{McLeodJB-jour1973a}, we know\vspace{-0.015cm}\\ \noindent that $H(\gamma_{k}(\varphi_{k}))(\phi(\varphi_k),\phi'(\varphi_k))^\top \!=\! B(\varphi_k)(\phi(\varphi_k),\phi'(\varphi_k))^\top$, $\phi(\varphi_k) \!=\! (\phi_0,1)^\top$ and $\phi'(\varphi_k) \!=\! (\phi_1 \!+\! \phi_2,0)^\top$,\vspace{-0.015cm}\\ \noindent where the eigenvalues of $B(\varphi_k)$ are those of $G(\gamma_{k}(\varphi_{k}))$ restricted to an invariant subspace of the iterate Banach space.  Differentiating this equality, with respect to arc-length $\varphi$, and evaluating it at $\varphi_k \!=\! 0$, with $\xi_0 \!=\! 0$ and\\ \noindent $\xi_1 \!=\! 1$, we have that
\begin{equation*}
\partial_\varphi \alpha(\varphi_*) = \psi^*(\varphi_*) \partial_\varphi G(\gamma_{k}(\varphi_*)) \phi(\varphi_*) = \partial_\varphi \omega_{1,1} \psi^*(\varphi_*) \phi(\varphi_*) + \partial_\varphi \omega_{1,2} \phi'(\varphi_*)
\end{equation*}
where $\psi^*(\varphi_*) \!=\! (\psi_1^*,0)$.  Here, we have again made the assumption that the iterates, $\gamma_k(\varphi_*)$, are such that\vspace{-0.015cm}\\ \noindent $\partial_\varphi (\pi_k(\varphi_*),\beta_k(\varphi_*)) \!=\! \xi_0\phi_0 \!+\! \xi_1 \phi_1$ and $\partial_\varphi \vartheta_k(\varphi_*) \!=\! \xi_0$.  Since $\psi^*(\varphi_*)\phi(\varphi_*) \!=\! 1$ and $\psi^*(\varphi_*)\phi'(\varphi_*) \!=\! 1$, we find that $\partial_\varphi \alpha(\varphi_*) \!=\! \omega_{1,1}$. Therefore, $\alpha(\varphi_k)$ goes to zero like $O(|\varphi_k \!-\! \varphi_*|^{1/2})$ as $\varphi_k \!\to\! \varphi_*$. $\blacksquare$\vspace{0.15cm}
\end{itemize}

\noindent It is possible to give an explicit form of the linear operator in terms solutions to the algebraic bifurcation equation.

\phantomsection\label{propA.10}
\begin{itemize}
\vspace{0.15cm}\item[] \-\hspace{0.0cm}{\small{\sf{\textbf{Proposition A.10.}}}} Assume that $\gamma_k(\varphi_k) \!=\! \gamma_k(\varphi_*)$ is a singular point of $\nabla_{\!\pi,\beta}\mathcal{L}(\gamma_k(\varphi_k))$.  Let $\xi_0,\xi_1$ be roots of\\ \noindent the algebraic bifurcation equation in \hyperref[propA.9]{Proposition A.9}.  Let $\phi_0^*,\phi_1^*$ be chosen to satisfy $\phi_0^*\phi_1 \!=\! \phi_1^*\phi_0 \!=\! 0$ and\\ \noindent $\phi_0^*\phi_0 \!=\! \phi_1^*\phi_1 \!=\! 1$.  The operator $G(\gamma_k(\varphi_*))$, 
\begin{equation*}
G(\gamma_k(\varphi_*)) = \Bigg(\begin{matrix} \nabla_{\!\pi,\beta}^2\mathcal{L}(\gamma_k(\varphi_*)) & \partial_\vartheta\nabla_{\!\pi,\beta}\mathcal{L}(\gamma_k(\varphi_*))\vspace{0.025cm}\\ \xi_0\phi_0^* \!+\! \xi_1\phi_1^* & \xi_0\end{matrix}\Bigg),
\end{equation*}
is of the Fredholm class, with index zero.  The nullspace of $G(\gamma_k(\varphi_*))$ is given by $\textnormal{span}((\xi_1\phi_0 \!-\! 2\xi_0\phi_1,\xi_1)^\top)$.  The range of $G(\gamma_k(\varphi_*))$ is the set of all $y \!\in\! \mathbb{B} \!\times\! \mathbb{R}$ where $\phi^*(\varphi_*)y \!=\! 0$, $\phi(\varphi_*) \!=\! (\phi_1^*,0)$.
\end{itemize}

\vspace{0.15cm}We are now able to tie \hyperref[propA.7]{Propositions A.7}--\hyperref[propA.9]{A.9} to our goal of quantifying the convergence rate of pseudo-arc-length path-following around singular points.

Suppose that we have a path of solutions that satisfy $(G(\gamma_k(\varphi_k)),\mathcal{M}_\varphi(\gamma_k)) \!=\! 0$.  Assume that $\varphi_k \!=\! \varphi_*$.  In this\\ \noindent case, $\nabla_{\!\pi,\beta}^2 \mathcal{L}(\gamma_k(\varphi_*))$ is a Fredholm operator with index zero that has a simple eigenvalue of zero.  We also assume that this solution corresponds to a path bifurcation, which is satisified whenever $\partial_\vartheta \nabla_{\!\pi,\beta}\mathcal{L}(\gamma_k(\varphi_*)) \!\in\! \textnormal{range}(\nabla_{\!\pi,\beta}^2 \mathcal{L}(\gamma_k(\varphi_*)))$ (see \hyperref[defA.3]{Definition A.3}).

Now, if $\phi_0$ is the unique solution of $\nabla_{\!\pi,\beta}^2 \mathcal{L}(\gamma_k(\varphi_*))\phi_0(\varphi_*) \!+\! \partial_\vartheta \nabla_{\!\pi,\beta}\mathcal{L}(\gamma_k(\varphi_*)) \!=\! 0$, with $\psi_1^*\phi_0 \!=\! 0$, then from\\ \noindent (\ref{eq:parc-4}), we can deduce that
\begin{align*}
\nabla_{\!\pi,\beta}^2 \mathcal{L}(\gamma_k(\varphi_*))(\dot{\pi}_k(\varphi_*),\dot{\beta}_k(\varphi_*)) + \partial_\vartheta \nabla_{\!\pi,\beta}\mathcal{L}(\gamma_k(\varphi_*))\dot{\vartheta}_k(\varphi_*) &= 0\\
(\dot{\pi_k}(\varphi_*),\dot{\beta_k}(\varphi_*))^*(\dot{\pi_k}(\varphi_*),\dot{\beta_k}(\varphi_*)) + \dot{\vartheta}_k(\varphi_*)^2 &= 1.
\end{align*}
A solution to these expressions is $(\dot{\pi}_k,\dot{\beta}_k) \!=\! \xi_0\phi_0(\varphi_*) \!+\! \xi_1\phi_1(\varphi_*)$ and $(\dot{\pi}_k,\dot{\beta}_k)^* \!=\! \xi_0\phi_0(\varphi_*)^* \!+\! \xi_1\phi_1(\varphi_*)^*$, with\\ \noindent $\dot{\vartheta}_k(\varphi_*) \!=\! \xi_0$.  Therefore, $(\dot{\pi_k}(\varphi_*),\dot{\beta_k}(\varphi_*))^*(\dot{\pi_k}(\varphi_*),\dot{\beta_k}(\varphi_*)) \!=\! \xi_0^2 \!+\! \xi_1^2$, which is always non-zero unless $(\dot{\pi}_k,\dot{\beta}_k) \!=\! 0$.\\ \noindent Moreover, the constraint $2\xi_0^2 \!+\! \xi_1^2 \!-\! 1 \!=\! 0$, which is obtained from the algebraic bifurcation equation in \hyperref[propA.9]{Proposition A.9},\\ \noindent ensures that the normalization in (\ref{eq:parc-4}) is obeyed.

Since $\partial_\vartheta \nabla_{\!\pi,\beta}\mathcal{L}(\gamma_k(\varphi_*)) \!\in\! \textnormal{range}(\nabla_{\!\pi,\beta}^2 \mathcal{L}(\gamma_k(\varphi_*)))$, we get from \hyperref[propA.7]{Proposition A.7} that $\nabla_{\!\pi,\beta}\mathcal{M}(\gamma_k)\phi_1(\varphi_*) \!=\! \xi_1$ and\\ \noindent $\nabla_{\!\pi,\beta}\mathcal{M}(\gamma_k)\phi_0(\varphi_*) \!+\! \partial_\vartheta\mathcal{M}_\varphi(\gamma_k) \!=\! 2\xi_0$.  As well, $|\nabla_{\!\pi,\beta}\mathcal{M}(\gamma_k)\phi_0(\varphi_*) \!+\! \partial_\vartheta\mathcal{M}_\varphi(\gamma_k)| \!+\! |\nabla_{\!\pi,\beta}\mathcal{M}(\gamma_k)\phi_1(\varphi_*)| \!\neq\! 0$\\ \noindent from \hyperref[propA.9]{Proposition A.9}.  Hence, $G(\gamma_k(\varphi_k))$ has a unique eigenvector that is inherited from $\nabla_{\!\pi,\beta}^2\mathcal{L}(\gamma_k(\varphi_*))$.  It also has a zero eigenvalue.  According to \hyperref[propA.7]{Propositions A.7} and \hyperref[propA.8]{A.8}, the eigenvalue is simple if $\dot{\vartheta}_k(\varphi_*) \!=\! 0$ and $\psi^*(\varphi_*)\phi(\varphi_*) \!\neq\! 0$,\\ \noindent where $\psi^*(\varphi_*)\phi(\varphi_*) \!=\! \tau_2 \!=\! -2\xi_0/\xi_1$.  If, however, $\dot{\vartheta}_k(\varphi_*) \!\neq\! 0$, and thus $\xi_0 \!\neq\! 0$, then we can redefine the eigenvectors\\ \noindent so that we still get $\psi^*(\varphi_*)\phi(\varphi_*) \!=\! 1$.

\hyperref[propA.7]{Propositions A.7}, in conjunction with the above arguments, can be used to show that $(G(\gamma_k(\varphi_k)),\mathcal{M}_\varphi(\gamma_k))$ is a Fredholm operator of index zero, just like $G(\gamma_k(\varphi_k))$, except when $\dot{\vartheta}_k(\varphi_*) \!=\! 0$.  The eigendecomposition is the same in both cases.  \hyperref[propA.7]{Propositions A.7} therefore enables us to work with $G(\gamma_k(\varphi_k))$ versus $(G(\gamma_k(\varphi_k)),\mathcal{M}_\varphi(\gamma_k))$ even when bifurcations occur and hence the solution is singular.

In some instances, the linear operator will have non-simple eigenvalues that are zero.  We quantify when this occurs, which will be necessary for our convergence result in \hyperref[propA.12]{Propositions A.12}.

\phantomsection\label{propA.11}
\begin{itemize}
\vspace{0.075cm}\item[] \-\hspace{0.0cm}{\small{\sf{\textbf{Proposition A.11.}}}} Assume the same conditions as in \hyperref[propA.6]{Proposition A.6}, except that the linear-operator inequality,\vspace{-0.02cm}\\ \noindent is modified to be $\|G^{-1}(\gamma_{k}(\varphi_{k}))\|\kappa(\varphi_{k})K'(\varphi_*)K(\varphi_{k})|\varphi_{k} \!-\! \varphi_*|^\eta \!\nless\! \frac{1}{2}$, with $K'(\varphi_*) \!\in\! \mathbb{R}_+$.  Here, the variable\\ \noindent $\eta \!\in\! \mathbb{R}_+$ depends on the eigenstructure of $G(\gamma_{k}(\varphi_{k}))$ for the Newton iterates as it approaches a solution $\gamma_k(\varphi_*)$,
\begin{itemize}
\item[] \-\hspace{0.5cm}(i) If $G(\gamma_{k}(\varphi_*))$ has an eigenvalue of zero, with algebraic multiplicity one, then $\eta \!=\! 1$. 
\item[] \-\hspace{0.5cm}(ii) If $G(\gamma_{k}(\varphi_*)$ has an eigenvalue of zero, with algebraic multiplicity two, then $\eta \!=\! \frac{1}{2}$.
\end{itemize}
In both cases, the iterates of (\ref{eq:parc-6}) converge at a rate that is at least geometric.
\end{itemize}
\begin{itemize}
\item[] \-\hspace{0.5cm}{\small{\sf{\textbf{Proof:}}}} We have shown, in \hyperref[propA.7]{Proposition A.7}, when $G(\gamma_{k}^i(\varphi_{k}))$ inherits the structure of $\nabla_{\pi,\beta}\mathcal{L}(\gamma_{k}^i(\varphi_{k}))$.  In particular, it does when $\dot{\vartheta}_{k}(\varphi_k) \!\neq\! 0$.  In these cases, it becomes a Fredholm operator of index zero and has zero as\\ \noindent a simple eigenvalue.  If, however, $\dot{\vartheta}_{k}(\varphi_k) \!=\! 0$, then the linear operator has zero has non-simple eigenvalues.\\ \noindent  Regardless of which occurs, we can bound how much $\|G^{-1}(\gamma_{k}^i(\varphi_{k}))\|$ is changing, or, rather, how quickly its corresponding eigenfunctions change, as $\varphi_k \!\to\! \varphi_*$, and appropriately modify the associated conditions in \hyperref[propA.6]{Proposition A.6} to reflect this.

We will show that, regardless of the algebraic multiplicity, the norm of the linear operator can be bounded in terms of its eigenvalues.  We can then analyze the rate of change for the eigenvalues.  We will do this only for the first Newton step, since, for subsequent ones, analogous expressions can be derived.

Let an initial approximation to a solution be $\gamma_k^0(\varphi_k) \!=\! \gamma_k(\varphi_*) \!+\! (\varphi_k \!-\! \varphi_*)\partial_\varphi \gamma_k^0(\varphi_*)$.

We first consider when the linear operator has a single eigenvalue of zero.  In this case, there exist a pair $(\alpha(\varphi_k),\phi(\varphi_k))$, continuously differentiable to $\varphi$, for which $G(\gamma^0_{k}(\varphi_{k}))\phi(\varphi_k) \!=\! \alpha(\varphi_k)\phi(\varphi_k)$.  This existence\\ \noindent is guaranteed by \hyperref[propA.8]{Proposition A.8}.  We also can define two subspaces that decompose the underlying Banach space\\ \noindent $\mathcal{U}_1 \!=\! \textnormal{null}(G(\gamma^0_{k}(\varphi_{k})) \!-\! \alpha(\varphi_k)\textnormal{id})$ and $\,\mathcal{U}_2 \!=\! \textnormal{range}(G(\gamma^0_{k}(\varphi_{k})) \!-\! \alpha(\varphi_k)\textnormal{id})$.  In both cases, $\textnormal{id}$ is the identity operator.\\ \noindent  There exist projections onto these subspaces, $q_1,q_2$, with $q_1(\varphi_k) \!+\! q_2(\varphi_k) \!=\! \textnormal{id}$.  We therefore can re-write the\\ \noindent linear-operator norm, for the first Newton step, as
\begin{align*}
\|G^{-1}(\gamma_{k}^0(\varphi_{k}))\| &= \|G^{-1}(\gamma_{k}^0(\varphi_{k}))(q_1(\varphi_{k}) \!+\! q_2(\varphi_{k}))\|\\
 &\leq \|G^{-1}(\gamma_{k}^0(\varphi_{k}))q_1(\varphi_{k})\| + \|G^{-1}(\gamma_{k}^0(\varphi_{k}))q_2(\varphi_{k})\|.
\end{align*}
Since $\,\mathcal{U}_1 \!=\! \textnormal{span}(\phi(\varphi_k))$, we get $\|G^{-1}(\gamma_{k}^0(\varphi_{k}))q_1(\varphi_{k})\| \!\leq\! \alpha^{-1}(\varphi_k)U_1(\varphi_k)$, where $U_1$ is a continuous, bounded\\ \noindent function.  Additionally, $(G(\gamma_{k}^0(\varphi_{k})) \!-\! \alpha(\varphi_k)\textnormal{id})\mathcal{U}_2 \!=\! \mathcal{U}_2$ and hence $G(\gamma_{k}^0(\varphi_{k}))\mathcal{U}_2 \!=\! \mathcal{U}_2$.  From \cite{CrandallMG-jour1971a}, we know that the linear operator is a bijection onto $\,\mathcal{U}_2$.  Therefore $\|G^{-1}(\gamma_{k}^0(\varphi_{k}))q_2(\varphi_{k})\| \!\leq\! U_2(\varphi_k)$, where $U_2$ is a continuous,\\ \noindent bounded function.  Taken together, both inequalities imply the existence of a continuous, uniformly bounded function, $U$, where $\|G^{-1}(\gamma_{k}^0(\varphi_{k}))\| \!\leq\! |\alpha^{-1}(\varphi_k)|U(\varphi_{k})$.

We now consider when the linear operator has dual eigenvalues that are zero.  As in the above case, from\vspace{-0.01cm}\\ \noindent \hyperref[propA.8]{Proposition A.8}, we know that there exist a pair $(\mu(\varphi_k),\lambda(\varphi_k))$, that are continuously differentiable to $\varphi$, for which $G(\gamma^0_{k}(\varphi_{k}))\lambda(\varphi_k) \!=\! \mu(\varphi_k)\lambda(\varphi_k)$.  The iterate Banach space can be decomposed into two subspaces\\ \noindent $\mathcal{W}_1 \!=\! \textnormal{null}(G(\gamma^0_{k}(\varphi_{k})) \!-\! \mu(\varphi_k)\textnormal{id})$ and $\,\mathcal{W}_2 \!=\! \textnormal{range}(G(\gamma^0_{k}(\varphi_{k})) \!-\! \mu(\varphi_k)\textnormal{id})$.  There exist projections onto these\\ \noindent subspaces, $p_1,p_2$, with $p_1(\varphi_k) \!+\! p_2(\varphi_k) \!=\! \textnormal{id}$.  We therefore can re-write the linear-operator norm
\begin{align*}
\|G^{-1}(\gamma_{k}^0(\varphi_{k}))\| &= \|G^{-1}(\gamma_{k}^0(\varphi_{k}))(p_1(\varphi_{k}) \!+\! p_2(\varphi_{k}))\|\\
 &\leq \|G^{-1}(\gamma_{k}^0(\varphi_{k}))p_1(\varphi_{k})\| + \|G^{-1}(\gamma_{k}^0(\varphi_{k}))p_2(\varphi_{k})\|.
\end{align*}
As $G^{-1}(\gamma_{k}^0(\varphi_{k}))p_1$ restricts the linear operator's inverse to $\mathcal{W}_1$, $\|G^{-1}(\gamma_{k}^0(\varphi_{k}))p_1(\varphi_{k})\| \!\leq\! \mu^{-1}(\varphi_k)W_1(\varphi_k)$,\\ \noindent where $W_1$ is a continuous, bounded function.  Here, $\mu(\varphi_k)$ is either of the eigenvalues for the linear operator, as they approach zero at the same rate.  Moreover, we have that the linear operator is a bijection onto $\,\mathcal{W}_2$, so, for a continuous, bounded function, $W_2$, $\|G^{-1}(\gamma_{k}^0(\varphi_{k}))p_2(\varphi_{k})\| \!\leq\! W_2(\varphi_k)$.  These inequalities both imply that\\ \noindent $\|G^{-1}(\gamma_{k}^0(\varphi_{k}))\| \!\leq\! |\mu^{-1}(\varphi_k)|W(\varphi_{k})$ for continuous, uniformly bounded function $W$. % as $\varphi_k \!\to\! \varphi_0$

\hyperref[propA.9]{Proposition A.9} can be invoked to show that $\alpha(\varphi_k) \!=\! O(|\varphi_k \!-\! \varphi_*|)$.  Similarly, for the other eigenvalues,\\ \noindent $\mu(\varphi_k) \!=\! O(|\varphi_k \!-\! \varphi_*|^{1/2})$.  These bounds hold not only for the first Newton step, but also for subsequent ones,\\ \noindent and thus can be inserted into \hyperref[propA.6]{Proposition A.6} to obtain geometric convergence. $\blacksquare$\vspace{0.15cm}
\end{itemize}

\noindent \hyperref[propA.11]{Proposition A.11} relies on a decomposition of the Banach space.  This is a consequence of \hyperref[propA.8]{Proposition A.8}.

More specifically, we assume that $G(\gamma_k(\varphi_*))$ is a Fredholm operator of index zero.  $G(\gamma_k(\varphi_k)) \!-\! \alpha(\varphi_k)\textnormal{id}$\\ \noindent shares this property too, since $\|G(\gamma_k(\varphi_*)) \!-\! (G(\gamma_k(\varphi_k)) \!-\! \alpha(\varphi_k)\textnormal{id})\| \!\leq\! \epsilon$, for some small $\epsilon \!\geq\! 0$.  According to\\ \noindent \hyperref[propA.7]{Propositions A.7}--\hyperref[propA.8]{A.8}, it is natural to conclude that both have a simple eigenvalue of zero.  We can therefore re-write the Banach space as $\mathcal{Y}_1 \!\oplus\! \mathcal{Y}_2$ with subspaces $\mathcal{Y}_1 \!=\! \textnormal{null}(G(\gamma_k(\varphi_i)) \!-\! \alpha(\varphi_k)\textnormal{id})$ and $\mathcal{Y}_2 \!=\! \textnormal{range}(G(\gamma_k(\varphi_i)) \!-\! \alpha(\varphi_k)\textnormal{id})$.

In \hyperref[propA.11]{Proposition A.11}, we characterize the subspace $\mathcal{Y}_1$, or, rather, $\mathcal{U}_1$ and $\mathcal{W}_1$, via the adjoint eigenfunctions, which are either $\psi^*(\varphi_k)$ or $\mu^*(\varphi_k)$.  In practice, though, we only know $\psi^*(\varphi_*)$ and $\mu^*(\varphi_*)$.  To obtain this eigenfunction, we could take $\psi^*(\varphi_k) \!=\! \psi^*$ to be a smooth mapping.  However, since the existence of $\psi(\varphi_k)$ and $\alpha(\varphi_k)$ are guar-\\ \noindent anteed, and since $G(\gamma_k(\varphi_k)) \!-\! \alpha(\varphi_k)\textnormal{id}$ is also Fredholm, we get the existence of some $\hat{\psi}^*(\varphi_k) \!\equiv\! \psi^*(\varphi_k)$, where\\ \noindent $G(\gamma_k(\varphi_k))^*\hat{\psi}^*(\varphi_k) \!=\! \alpha(\varphi_k)\hat{\psi}^*(\varphi_k)$.  The existence of $\hat{\psi}^*(\varphi_k)$ permits defining the subspace projectors that we used in \hyperref[propA.11]{Proposition A.11}.

We are now in a position to strengthen \hyperref[propA.6]{Proposition A.6}.  This proposition guarantees geometric convergence of pseudo-arc-length path-following, which occurs a $q$-linear rate.  We would like to obtain the full $q$-quadratic convergence rate offered by Newton's method, though.  We show that this is possible in many cases.

\phantomsection\label{propA.12}
\begin{itemize}
\vspace{0.15cm}\item[] \-\hspace{0.0cm}{\small{\sf{\textbf{Proposition A.12.}}}} Assume that the linear operator $G(\gamma_k(\varphi_k))$ in (\ref{eq:parc-5}) is thrice differentiable.  As well, assume that $G(\gamma_k^0(\varphi_k))$, for $\gamma_k^0(\varphi_k) \!=\! \gamma_{k}(\varphi_*) \!+\! (\varphi_k \!-\! \varphi_*)\partial_\varphi \gamma_{k}(\varphi_*)$, has either a single eigenvalue of zero or dual eigen-\\ \noindent values that are zero.  If the Kantorovich conditions \cite{KantorovichLV-book1964a} are satisfied, then the iterates (\ref{eq:parc-6}) converge $q$-quadratically to a solution $\gamma_*(\varphi_{k}) \!=\! \gamma_{k+1}(\varphi_k)$, where $G(\gamma_*(\varphi_k)) \!=\! 0$ and hence $\nabla_{\pi,\beta}\mathcal{L}(\gamma_*(\varphi_k)) \!=\! 0$.
\vspace{0.075cm}
\end{itemize}
\begin{itemize}
\item[] \-\hspace{0.5cm}{\small{\sf{\textbf{Proof:}}}} We first consider when the linear operator has two eigenvalues that are zero, as this is the simpler case.  From \hyperref[propA.6]{Proposition A.6}, we can show that the linear operator can be bounded like 
\begin{equation*}
\|G(\gamma_k(\varphi_k))\| \leq K(\varphi_*)(1 \!-\! 2\kappa(\varphi_k)K(\varphi_k)\|G^{-1}(\gamma_k(\varphi_k))\|)^{-1}|\varphi_k \!-\! \varphi_*|^{-1/2}.
\end{equation*}
Additionally,\vspace{-0.075cm}
\begin{multline*}
\|G^{-1}(\gamma_k^i(\varphi_k))(\nabla_{\pi,\beta}\mathcal{L}(\gamma_k^i(\varphi_k)),\mathcal{M}_\varphi(\gamma_k^i(\varphi_k)))^\top\| \leq\\ \|G^{-1}(\gamma_k^i(\varphi_k))\|\|(\nabla_{\pi,\beta}\mathcal{L}(\gamma_k^i(\varphi_k)),\mathcal{M}_\varphi(\gamma_k^i(\varphi_k)))^\top \!-\! (\nabla_{\pi,\beta}\mathcal{L}(\gamma_*(\varphi_k)),\mathcal{M}_\varphi(\gamma_*(\varphi_k)))^\top\|.
\end{multline*}
For the latter term, if we assume a bound on the linear operator, $\|G(\gamma_k^i(\varphi_k))\| \!\leq\! V(\varphi_k)$, near a solution arc for\\ \noindent episode $k$, $\gamma_*(\varphi_k)$, then we can non-strictly constrain it above by $V(\varphi_k)\|\gamma_k^i(\varphi_k) \!-\! \gamma_*(\varphi)\|$.  Since the Kantorovich\\ \noindent conditions are assumed to be satisfied, we get that its three constants $v_1,v_2,v_3$ can be bounded as
\begin{equation*}
v_1v_2v_3 \leq (K(\varphi_*)(1 \!-\! 2\kappa(\varphi_k)K(\varphi_k)\|G^{-1}(\gamma_k(\varphi_k))\|)^{-1}|\varphi_k \!-\! \varphi_*|^{-1})^2 K(\varphi_k)V(\varphi_k)\|\gamma_k^i(\varphi_k) \!-\! \gamma_*(\varphi_k)\|.
\end{equation*}
Note, however, that $\|\gamma_k^i(\varphi_k) \!-\! \gamma_k(\varphi_k)\| \!\leq\! \sigma_{k}^i\kappa(\varphi_k)(\varphi_k \!-\! \varphi_*)^2/2$, where the definition of $\sigma_k$ is given in \hyperref[propA.6]{Proposition A.6}.  Here, $\sigma_k^i$ denotes $\sigma_k$ raised to the $i$th power.  This condition permits further reducing the inequality for the Kantorovich constants to $v_1v_2v_3 \!\leq\! W(\varphi_k)\sigma_k^i(\varphi_k \!-\! \varphi_*)^2|\varphi_k \!-\! \varphi_*|^{-1}$.  Here, $W(\varphi_k)$ is a well-behaved function,\\ \noindent even as $\varphi_k$ approaches $\varphi_*$; this function was introduced in \hyperref[propA.11]{Proposition A.11}.  Due to the eigenvalue assumptions of the linear operator, $v_1v_2v_3 \!<\! \frac{1}{2}$ and therefore $\gamma_k^i(\varphi_k) \!\to\! \gamma_*(\varphi_k)$ $q$-quadratically.

If the linear operator has a single eigenvalue that is zero, then, through a similar process, we can revise the solution inequality to $v_1v_2v_3 \!\leq\! U(\varphi_k)\sigma_k^i(\varphi_k \!-\! \varphi_*)^2|\varphi_k \!-\! \varphi_*|^{-2}$.  Here, $U(\varphi_k)$ is a well-behaved function, even as $\varphi_k$ approaches $\varphi_*$; this function was introduced in \hyperref[propA.11]{Proposition A.11}. However, $v_1v_2v_3 \!<\! \frac{1}{2}$ is not necessarily\\ \noindent guaranteed for the first iteration, since we have been using rather loose bounds on various terms. 

We therefore tighten the bounds on $\|G^{-1}(\gamma_k^i(\varphi_k))(\nabla_{\pi,\beta}\mathcal{L}(\gamma_k^i(\varphi_k)),\mathcal{M}_\varphi(\gamma_k^i(\varphi_k)))^\top\|$ to see if the Kantorovich conditions can be obeyed.  We will do this by evaluating how the iterates change across a single Newton step; in fact, it will be the first step after forming an initial guess.

As in \hyperref[propA.11]{Proposition A.11}, we can do an eigendecomposition of the linear operator and define dual subspaces and projections onto them.  This permits us to state that\vspace{-0.075cm}
\begin{multline*}
\|G^{-1}(\gamma_k^0(\varphi_k))(\nabla_{\pi,\beta}\mathcal{L}(\gamma_k^0(\varphi_k)),\mathcal{M}_\varphi(\gamma_k^0(\varphi_k)))^\top\| \leq\vspace{0.025cm}\\
\begin{array}{c}
\|G^{-1}(\gamma_k^0(\varphi_k))q_1(\varphi_k)\|\|q_1(\varphi_k)(\nabla_{\pi,\beta}\mathcal{L}(\gamma_k^0(\varphi_k)),\mathcal{M}_\varphi(\gamma_k^0(\varphi_k)))^\top\| \,+ \vspace{0.125cm}\\
\|G^{-1}(\gamma_k^0(\varphi_k))q_2(\varphi_k)\|\|q_2(\varphi_k)(\nabla_{\pi,\beta}\mathcal{L}(\gamma_k^0(\varphi_k)),\mathcal{M}_\varphi(\gamma_k^0(\varphi_k)))^\top\|
\end{array}
\end{multline*}
Let $P(\gamma_k(\varphi_k)) \!=\! (\nabla_{\pi,\beta}\mathcal{L}(\gamma_k(\varphi_k)),\mathcal{M}_\varphi(\gamma_k(\varphi_k)))^\top$.  We can consider a Taylor expansion about $\varphi_k \!=\! \varphi_*$, which yields that $P(\gamma_k^0(\varphi_k)) \!=\! \frac{1}{2}(\varphi_{k} \!-\! \varphi_*)^2 \nabla_{\pi,\beta}G(\gamma_k^0(\varphi_k)) \partial_\varphi \gamma_k^0(\varphi_*) \partial_\varphi\gamma_k^0(\varphi_*) \!+\! O(|\varphi_k \!-\! \varphi_*|^3)$.  Since we know from\\ \noindent \hyperref[propA.9]{Proposition A.9} that the adjoint and non-adjoint eigenvectors are normalized so that $\psi^*(\varphi_k)\phi(\varphi_k) \!=\! 1$, we can\\ \noindent write $q_1(\varphi_k) P(\gamma_k^0(\varphi_k)) \!=\! (\psi^*(\varphi_k)P(\gamma_k^0(\varphi_k)))\phi(\varphi_k)$.  We evaluate $\psi^*(\varphi_k)P(\gamma_k^0(\varphi_k))$ for $\gamma_k(\varphi_*)$, where we\\ \noindent assume that $\partial_\varphi (\pi_k(\varphi_*),\beta_k(\varphi_*)) \!=\! \xi_0\phi_0 \!+\! \xi_1 \phi_1$ and $\partial_\varphi \vartheta_k(\varphi_*) \!=\! \xi_0$, with $\xi_0,\xi_1$ being the solutions to the algebraic\\ \noindent bifurcation equation.  This yields, for a continuous, bounded function $A_1$,
\begin{equation*}
\psi^*(\varphi_k)P(\gamma_k^0(\varphi_*)) = \textstyle\frac{1}{2}(\omega_{1,1,1}\xi_1^2 \!+\! 2\omega_{1,1}\xi_0\xi_1 \!+\! \omega_1 \xi_0^2)(\varphi_k \!-\! \varphi_*)^2 + A_1(\varphi_k)|\varphi_k \!-\! \varphi_*|^3;
\end{equation*}
we have used the relationship $\psi^*(\varphi_*) \!=\! (\varphi_1^*,0)$ to simplify the above expression.  For the first term, we have that\\ \noindent $\omega_{1,1,1}\xi_1^2 \!+\! 2\omega_{1,1}\xi_0\xi_1 \!+\! \omega_1 \xi_0^2 \!=\! 0$, which is due to \hyperref[propA.9]{Proposition A.9}.  Thus, $q_1(\varphi_k) P(\gamma_k^0(\varphi_k)) \!=\! A_1(\varphi_k)|\varphi_k \!-\! \varphi_*|^3$.\\ \noindent Using a similar process, we find $q_2(\varphi_k) P(\gamma_k^0(\varphi_k)) \!=\! A_2(\varphi_k)|\varphi_k \!-\! \varphi_*|^2$, for a continuous, bounded function $A_2$.\\ \noindent  With these equalities, we get that
\begin{equation*}
\|G^{-1}(\gamma_k^0(\varphi_k))P(\gamma_k^0(\varphi_k))\| \leq K'(\varphi_*)A_1(\varphi_k)|\varphi_k \!-\! \varphi_*|^2 + A_2(\varphi_k)U_2(\varphi_k)|\varphi_k \!-\! \varphi_*|^2
\end{equation*}
for a continuous, bounded function $U_2$ taken from \hyperref[propA.9]{Proposition A.9}.  Hence,
\begin{equation*}
v_1v_2v_3 \leq K'(\varphi_*)K(\varphi_k)|\varphi_k \!-\! \varphi_*|(K'(\varphi_*)A_1(\varphi_k)|\varphi_k \!-\! \varphi_*|^2 + A_2(\varphi_k)U_2(\varphi_k)|\varphi_k \!-\! \varphi_*|^2)
\end{equation*} 
for $|\varphi_k \!-\! \varphi_*| \!<\! \epsilon$.  If $\epsilon$ is sufficiently small, then $v_1v_2v_3 \!<\! \frac{1}{2}$ and $\gamma_k^i(\varphi_k) \!\to\! \gamma_*(\varphi_k)$ $q$-quadratically. $\blacksquare$

\end{itemize}

\vspace{0.15cm}It is important to provide some context for this theory.

Several numerical approaches for the solution of bifurcation problems have been developed over the last five decades.  For the numerical treatment of infinite-dimensional problems, many researchers, including us, have opted to discretize the original equation, with respect to the continuation parameter, to obtain a finite-dimensional solution space \cite{WeberH-jour1984a,WeberH-jour1985a,KloedenPE-jour1986a}.  This yields a finite-dimensional bifurcation problem, which facilitate the derivation of error estimates.  Other classes of approaches are available too.  As an example, some researchers opt to transform the original problem into a new one that is well-conditioned but no longer exhibits any branching phenomena \cite{LangfordWF-jour1977a,LangfordWF-jour1977b}.

Most of the theory on discretization-based techniques has been for analyzing bifurcations from the trivial solution in the case where the linear operator has only simple eigenvalues.  Weiss \cite{WeissR-jour1975a}, for instance, investigated bifurcations that occur in difference approximations to the two-point boundary value problem.  He used the iteration method of Keller and Langford \cite{KellerHB-jour1972a} to prove the existence of a non-trivial solution branch as well as a branch of the difference equations.  Under reasonable stability assumptions, he obtained a geometric rate of convergence.  Later, Atkinson \cite{AtkinsonKE-jour1977a} showed how to discretize certain types of problems via collectively-compact-operator approximation.  Using the Lyapunov-Schmidt method \cite{FarrWW-jour1989a}, he proved the existence of bifurcating branches for the continuous and the discrete problem and obtained linear convergence.  Westreich and Vaaroll \cite{WestreichD-jour1979a} showed that the ideas of Atkinson could be used for non-linear integral equations and proposed an appropriate iteration scheme.

There are significant limitations of these discretization approaches.  The most prominent is that, without suitable modifications, they cannot treat bifurcations that arise for non-simple eigenvalues.  The reason for this is that the original continuous problem and the linearization of it generally possesses a different solution-set structure.  Similar issues are also encountered if secondary bifurcations are treated.  In general, the solution curves of the discrete, linearized equations no longer intersect and effects known from perturbed bifurcations will appear \cite{KeenerJP-jour1973a}.  Here, however, we have demonstrated that it is possible to guarantee convergence when the linear operator possseses both simple and non-simple eigenvalues.  This addresses some of these concerns, though not completely.  Moreover, we have been able to retain the full quadratic rate of convergence offered by Newton's method, not just the geometric rate that is offered by existing contributions.  This bodes well for ensuring that solutions can be quickly uncovered, at each episode, for the value of information.

\phantomsection\label{secA.3}
%%%%%%%%%%%%%%%%%%%%%%%%%%%%%%%%%%%%%%%%%%%%%%%%%%%%%%%%%%%%%%%%%%%%%%%%%%%
%%%%%%%%%%%%%%%%%%%%%%%%%%%%%%%%%%%%%%%%%%%%%%%%%%%%%%%%%%%%%%%%%%%%%%%%%%%
\subsection*{\small{\sf{\textbf{A.3$\;\;\;$ Branch Switching at Bifurcation Points}}}}

Bifurcations of the solution path may be encountered several times when performing pseudo-arc-length path-following.  That is, the solution curve may split into multiple paths at equilibria points, with each path containing viable solutions.  It is important to detect when bifurcations may occur.  As well, it is important to determine which branch should be taken so as to best optimize the value of information.

\phantomsection\label{defA.3}
\begin{itemize}
\vspace{0.15cm}\item[] \-\hspace{0.0cm}{\small{\sf{\textbf{Definition A.3.}}}} Let $\gamma_k(\varphi_*)$ be a solution that satisfies (\ref{eq:parc-1}) but where (\ref{eq:parc-5}) is singular.  Such a solution is a bifurcation point: two or more branches of solutions have non-tangential intersections at this point.  Moreover, we\\ \noindent have that $\textnormal{dim}\, \textnormal{null}(\nabla^2_{\!\pi,\beta}\mathcal{L}(\gamma_k(\varphi_*))) \!=\! \textnormal{codim}\, \textnormal{range}(\nabla^2_{\!\pi,\beta}\mathcal{L}(\gamma_k(\varphi_*))),$ which is equal to some scalar $m$.  As well, $\partial_\vartheta \nabla_{\!\pi,\beta}\mathcal{L}(\gamma_k(\varphi_*)) \!\in\! \textnormal{range}(\nabla^2_{\!\pi,\beta}\mathcal{L}(\gamma_k(\varphi_*))).$
\vspace{0.15cm}
\end{itemize}

We can now show how to deduce the number of bifurcation branches.  

The first part of \hyperref[defA.3]{Definition A.3} implies that $\nabla^2_{\!\pi,\beta}\mathcal{L}(\gamma_k(\varphi_*))$ is a Freholm operator of index zero.  We therefore can state that $\textnormal{null}(\nabla^2_{\!\pi,\beta}\mathcal{L}(\gamma_k(\varphi_*))) \!=\! \textnormal{span}(\phi_1(\varphi_*),\ldots,\phi_m(\varphi_*))$, where the $\phi_j(\varphi_*)$s are eigenvectors.  Similarly, the\\ \noindent adjoint shares this trait, so $\textnormal{null}(\nabla^2_{\!\pi,\beta}\mathcal{L}(\gamma_k(\varphi_*)))^* \!=\! \textnormal{span}(\psi_1(\varphi_*),\ldots,\psi_m(\varphi_*))$, where the $\psi_p(\varphi_*)$s are adjoint eigen-\\ \noindent functions.  We also get that $\psi_p^*\phi_j \!=\! \delta_{p,j}$.  The second part of \hyperref[defA.3]{Definition A.3} indicates that there exists a unique $\phi_0$ such\\ \noindent that $\nabla^2_{\!\pi,\beta}\mathcal{L}(\gamma_k(\varphi_*))\phi_0 \!+\! \partial_\vartheta \nabla_{\!\pi,\beta}\mathcal{L}(\gamma_k(\varphi_*)) \!=\! 0$ and hence $\psi_j^*\phi_0 \!=\! 0$.

As in \hyperref[propA.9]{Proposition A.9}, since $\nabla^2_{\!\pi,\beta}\mathcal{L}(\gamma_k(\varphi_*))\partial_\varphi {\pi}_k(\varphi_*) \!+\! \partial_\vartheta \nabla_{\!\pi,\beta}\mathcal{L}(\gamma_k(\varphi_*)) \partial_\varphi {\beta}_k(\varphi_*) \!=\! 0$, it follows that there\\ \noindent exist scalars $\xi_j$ such that $\partial_\varphi({\pi}_k(\varphi_*)),{\beta}_k(\varphi_*)) \!=\! \sum_{j=0}^m \xi_j\phi_j$.  Moreover, $\xi_0 \!=\! \partial_\varphi {\beta}_k(\varphi_*)$ and $\xi_j \!=\! \psi_j^* \partial_\varphi {\pi}_k(\varphi_*)$.  The\\ \noindent $\xi_j$s necessarily satisfy
\begin{align*}
\textstyle\sum_{j=1}^m \sum_{p=1}^m \omega_{i,j,p}\xi_j\xi_p + 2 \sum_{j=1}^m \omega_{i,j} \xi_j\xi_0 + \omega_i \xi_0^2 &= 0
\end{align*}
For $i,j,p \!\in\! 1,\ldots,m$, the coefficients $\omega$ are given by\vspace{-0.05cm}
\begin{align*}
\psi_i^*(\varphi_k)\nabla_{\pi,\beta}^3\mathcal{L}(\gamma_{k}(\varphi_*)) \phi_j\phi_p &= \omega_{i,j,p}\\
\psi_i^*(\varphi_k) (\nabla_{\pi,\beta}^3\mathcal{L}(\gamma_{k}(\varphi_*))\phi_0 + \partial_\vartheta\nabla^2_{\pi,\beta}\mathcal{L}(\gamma_{k}(\varphi_*))) \phi_j &= \omega_{i,j}\\
\psi_i^*(\varphi_k) (\nabla_{\pi,\beta}^3\mathcal{L}(\gamma_{k}(\varphi_*))\phi_0\phi_0 + 2\partial_\vartheta\nabla_{\pi,\beta}^2\mathcal{L}(\gamma_{k}(\varphi_*))\phi_0 + \partial_\vartheta^2\nabla_{\pi,\beta}\mathcal{L}(\gamma_{k}(\varphi_*))) &= \omega_i.
\end{align*}
We therefore have that the tangent vector $\partial_\varphi\gamma_k(\varphi_*)^\top$ to every smooth branch through a bifurcation point $\gamma_k(\varphi_*)$ must conform to the algebraic bifurcation equation.  If the algebraic bifurcation equation has $r \!\geq\! 2$ distinct, non-trivial roots, then there exist at least $r$ smooth solution branches that non-tangentially intersect.  This result was proved by Keller and Langford \cite{KellerHB-jour1972a} for $m \!>\! 1$, where $m$ defines the dimensionality of the nullspace for the Hessian.  For the special case\\ \noindent where $m \!=\! 1$, the algebraic bifurcation equation reduces to a single quadratic with two non-trivial roots.  This was\\ \noindent proved by Crandall and Rabinowitz \cite{CrandallMG-jour1971a}.

There are different ways that we can specify the bifurcating solution branches and hence explore them \cite{DeuflhardP-jour1987a}.

The most straightforward is to find the non-trivial roots of the algebraic bifurcation equation and then use them in the expression $\partial_\varphi({\pi}_k(\varphi_*)),{\beta}_k(\varphi_*)) \!=\! \sum_{j=0}^m \xi_j\phi_j$ to construct the various tangent vectors.  We can then insert each of these vectors in (\ref{eq:parc-4}) and apply pseudo-arc-length path-following.  \hyperref[propA.12]{Proposition A.12} guarantees that, for a sufficiently small ball around the singular point, the path-following iterates will converge to a point on the new solution arc.  To reduce the computational burden of this process, we can use a root-approximation scheme.  If $\phi_j$ and $\psi_j^*$ are known, then we can re-define the coefficients $\omega_i$, $\omega_{i,j}$, and $\omega_{i,j,p}$, for $i,j,p \!\in\! 1,\ldots,m$, as\vspace{-0.05cm}
\begin{align*}
\epsilon^{-1}\psi_i^*(\varphi_k)(\nabla_{\pi,\beta}^2\mathcal{L}(\gamma_k'(\varphi_*,\epsilon\phi_j)) - \nabla_{\pi,\beta}\mathcal{L}(\gamma_k(\varphi_*)))\phi_p &= \omega_{i,j,p}(\epsilon)\\
\epsilon^{-1}\psi_i^*(\varphi_k) ((\nabla_{\pi,\beta}^2\mathcal{L}(\gamma_k'(\varphi_*,\epsilon\phi_j)) - \nabla_{\pi,\beta}\mathcal{L}(\gamma_k(\varphi_*)))\phi_0 & \\
 + (\partial_\vartheta\nabla_{\pi,\beta}\mathcal{L}(\gamma_k'(\varphi_*,\epsilon\phi_j)) - \partial_\vartheta\nabla_{\pi,\beta}\mathcal{L}(\gamma_{k}(\varphi_*)))) &= \omega_{i,j}(\epsilon)\\
\epsilon^{-1}\psi_i^*(\varphi_k) ((\nabla_{\pi,\beta}^2\mathcal{L}(\gamma_k'(\varphi_*,\epsilon\phi_0)) - \nabla_{\pi,\beta}\mathcal{L}(\gamma_k'(\varphi_*)))\phi_0 + 2(\partial_\vartheta\nabla_{\pi,\beta}\mathcal{L}(\gamma_k'(\varphi_*,\epsilon\phi_0)) & \\
 - \partial_\vartheta\nabla_{\pi,\beta}\mathcal{L}(\gamma_{k}(\varphi_*))) + (\partial_\vartheta\nabla_{\pi,\beta}\mathcal{L}((\pi_k(\varphi_*),\beta_k(\varphi_*)),\vartheta_k(\varphi_*) \!+\! \epsilon) - \partial_\vartheta\nabla_{\pi,\beta}\mathcal{L}(\gamma_{k}(\varphi_*)))) &= \omega_i(\epsilon).
\end{align*}
Here, $\gamma_k'(\varphi_*,\epsilon\phi_j) \!=\! ((\pi_k(\varphi_*),\beta_k(\varphi_*)) \!+\! \epsilon \phi_j,\vartheta_k(\varphi_*))$.  As $\epsilon \!\to\! 0$, $\omega_i(\epsilon)$, $\omega_{i,j}(\epsilon)$, and $\omega_{i,j,p}(\epsilon)$ converge to the true\\ \noindent coefficients but without the need for evaluating third-order Fr\'{e}chet derivatives. 

\begin{figure*}
   \begin{adjustbox}{addcode={\begin{minipage}{\width}}{\caption{%
     A visual overview of bifurcations for pseudo-arc-length path following.  (left) For a given starting point, $((\pi_{k-1},\beta_{k-1}),\vartheta_{k-1})$, path following is performed as in \hyperref[alg:pseudoarclengthpathfollowing]{Algorithm 2}.  Once a solution, $(\pi_{k},\beta_{k}),\vartheta_{k})$, is found, then a check is performed to determine if the Jacobian is singular.  If it is, then a bifurcation is present.  Multiple tangent vectors are then formed.  The updates continue to trace the solution curve (black line) for the current branch, albeit with diminishing returns for this example.  (right) Switching to another branch and tracing the solution curve (black line) permits a further reduction in costs, eventually yielding a globally optimal solution $((\pi^*,\beta^*),\vartheta^*)$.  For each of the major updates shown in this overview, we provide corresponding embedded videos, for \emph{Super\! Mario\! Land}.  These videos illustrate the agent's improved understanding of the environment dynamics (left).  However, switching to a new branch facilitates learning better behaviors, due to the fragmentation of the state-action space (right).  This is corroborated by the state-space similarity plots $\Pi_k(\mathcal{S})$ and $\Pi_{k+1}(\mathcal{S})$, which show that a new state-action group has formed as a consequence of switching to a bifurcating branch.  We recommend viewing this document within Adobe Acrobat DC; click on an image and enable content to start playback of the corresponding video.
      }\end{minipage}},rotate=90,center}
      \scalebox{0.775}{
         \begin{tikzpicture} 
            \node[] at (0,0) {\includegraphics[width=5.5in]{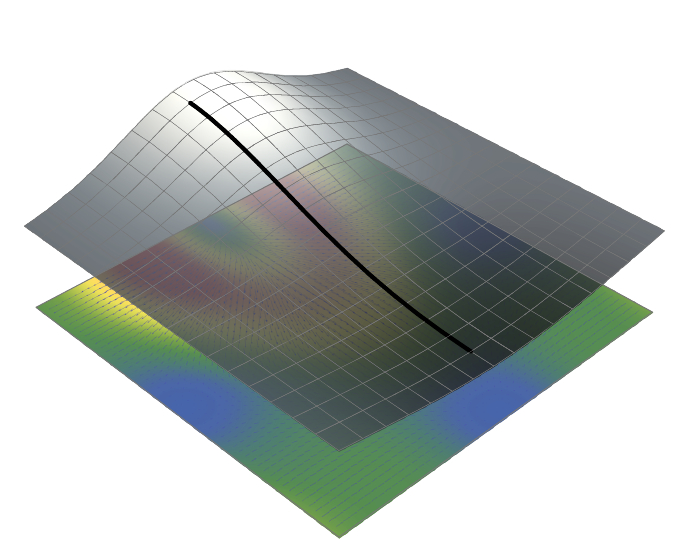}};
            \node at (-0.05,-3.7) {\large \textcolor{white}{$-\nabla_{\!\pi,\beta}\mathcal{L}((\pi,\beta),\vartheta)$}};

            \node at (-3.95,-3.1) {\large $(\pi,\beta)_1$};
            \node at (4.0,-3.1) {\large $(\pi,\beta)_2$};

            \node at (4,3.1) {\large $\mathcal{L}((\pi,\beta),\vartheta)$};

            \node at (1.05,1.45) {\textcolor{black}{\boldmath $\nabla_{\!\pi,\beta}\mathcal{L}((\pi,\beta),\vartheta) \!=\! 0$}};

            \node at (2.75,-0.015) {\textcolor{white}{\boldmath $|J((\pi_{k},\beta_{k}),\vartheta_{k})| \!=\! 0$}};

            \node at (-4.5,3.9) {$((\pi_{k-1},\beta_{k-1}),\vartheta_{k-1})$};
            \node at (-0.45,-0.55) {\textcolor{white}{\boldmath $((\pi_k,\beta_k),\vartheta_k)$}};
            \node at (0.25,-1.6) {\textcolor{white}{\boldmath $((\pi_{k'+1},\beta_{k'+1}),\vartheta_{k'+1})$}};

            \draw[fill, black] (-3.1,3.5) circle (4pt);
            \draw[fill, white] (1.2,-0.55) circle (4pt);
            \draw[fill, white] (2.65,-1.55) circle (4pt);

            \setlength{\fboxrule}{0.75pt}
            \setlength{\fboxsep}{0.025pt}

            \node at (-6.8,-5.5) {\framebox{\embedvideo{\includegraphics[width=1.175in,height=1.05in]{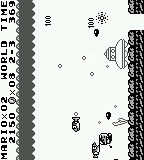}}{level2-3-1-rot-noaudio.mp4}}};
            \node at (-3.6,-5.5) {\framebox{\embedvideo{\includegraphics[width=1.175in,height=1.05in]{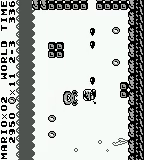}}{level2-3-2-rot-noaudio.mp4}}};
            \node at (-6.8,-7.15) {\includegraphics[width=1.39in]{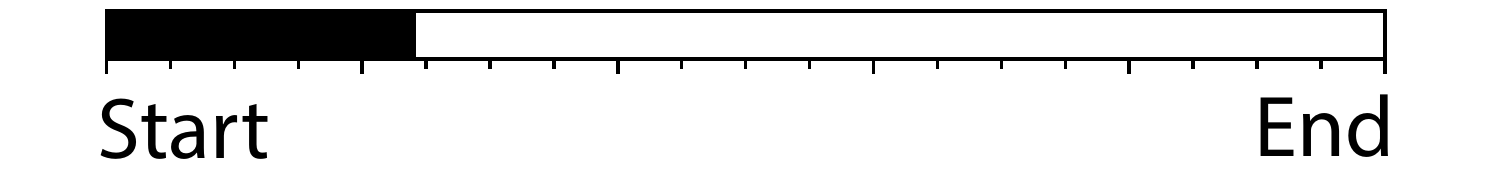}};
            \node at (-3.6,-7.15) {\includegraphics[width=1.39in]{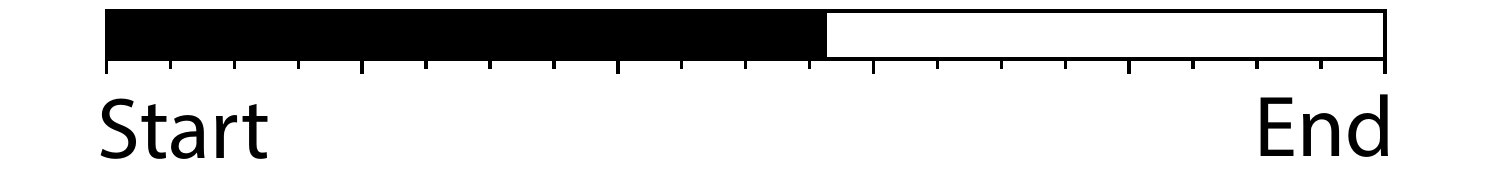}};
            \node at (-6.8,-7.6) {\large $(\pi_{k-1},\beta_{k-1})$};
            \node at (-3.6,-7.6) {\large $(\pi_{k'+1},\beta_{k'+1})$};

            \node at (3.4,-5.5) {\framebox{\includegraphics[height=1.05in]{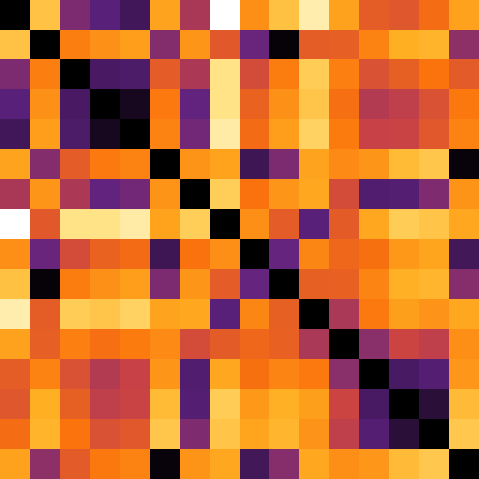}}};
            \node at (3.4,-7.15) {\includegraphics[width=1.25in]{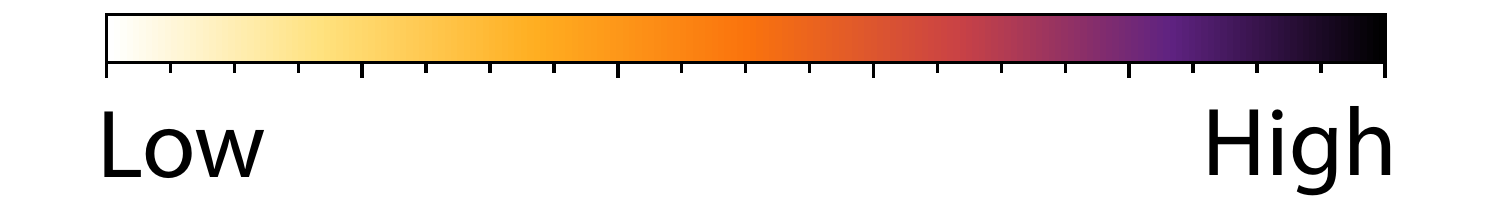}};

            \node at (3.4,-3.85) {\large $s'$};
            \node at (5.05,-5.5) {\large $s'$};
            \node at (3.4,-7.6) {\large $\Pi_{k'+1}\,(\mathcal{S})$};

            \node at (10.05,-5.5) {\framebox{\includegraphics[height=1.05in]{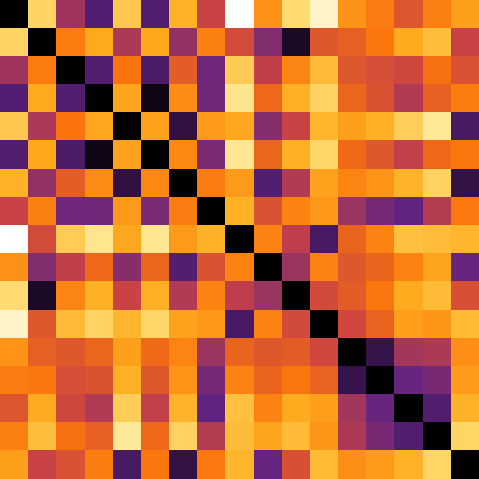}}};
            \node at (10.05,-7.15) {\includegraphics[width=1.25in]{sunsetbar-rot.pdf}};

            \node at (10.05,-3.85) {\large $s'$};
            \node at (8.4,-5.5) {\large $s'$};
            \node at (10.05,-7.6) {\large $\Pi_{k+1}\,(\mathcal{S})$};

            \node[] at (13.5,0) {\includegraphics[width=5.5in]{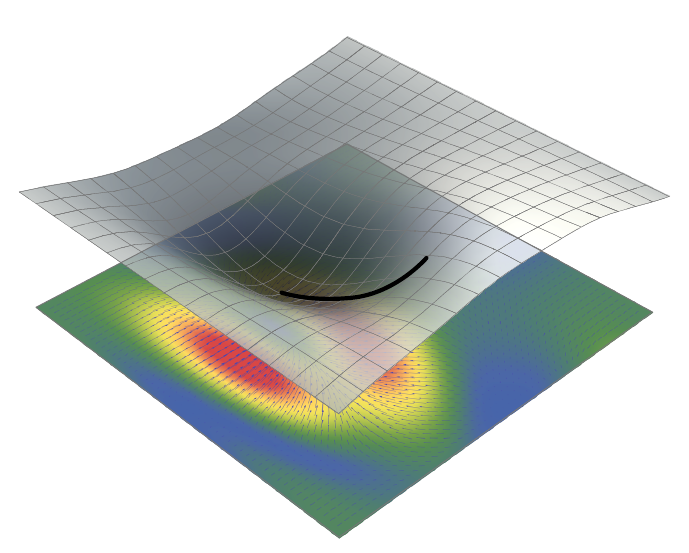}};
            \node at (13.4,-3.7) {\large \textcolor{white}{$-\nabla_{\!\pi,\beta}\mathcal{L}((\pi,\beta),\vartheta)$}};

            \node at (9.55,-3.1) {\large $(\pi,\beta)_1$};
            \node at (17.5,-3.1) {\large $(\pi,\beta)_2$};

            \node at (9.35,3.1) {\large $\mathcal{L}((\pi,\beta),\vartheta)$};

            \node at (16.15,-0.15) {\textcolor{white}{\boldmath $((\pi_k,\beta_k),\vartheta_k)$}};

            \draw[fill, white] (15.2,0.4) circle (4pt);
            \draw[fill, white] (12.2,-0.3) circle (4pt);

            \node at (12.2,0.175) {\textcolor{white}{\boldmath $((\pi_{k+1},\beta_{k+1}),\vartheta_{k+1})$}};
            \node at (12.2,-0.825) {\textcolor{white}{\boldmath $((\pi^*,\beta^*),\vartheta^*)$}};

            \node at (17,-5.5) {\framebox{\embedvideo{\includegraphics[width=1.175in,height=1.05in]{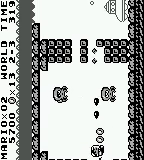}}{level2-3-3-rot-noaudio.mp4}}};
            \node at (20.2,-5.5) {\framebox{\embedvideo{\includegraphics[width=1.175in,height=1.05in]{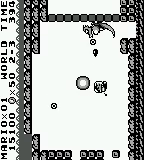}}{level2-3-4-rot-noaudio.mp4}}};

            \node at (17,-7.15) {\includegraphics[width=1.39in]{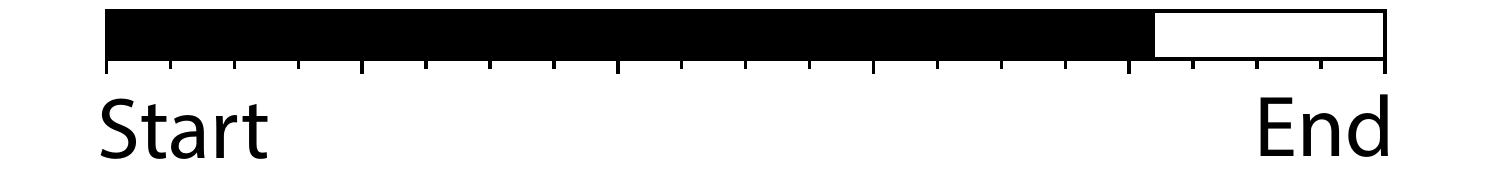}};
            \node at (20.2,-7.15) {\includegraphics[width=1.39in]{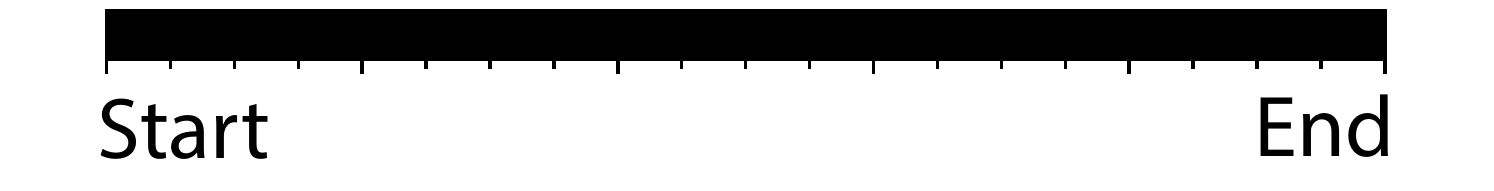}};

            \node at (17,-7.6) {\large $(\pi_{k},\beta_{k})$};
            \node at (20.2,-7.6) {\large $(\pi^*,\beta^*)$};
         \end{tikzpicture}}
   \end{adjustbox}
   \label{fig:appendixa-bifurcation}
\end{figure*}

We use this approach for our simulations in conjunction with parallel searches.  The idea, and how the policy state abstraction changes after taking the bifurcating branch, is depicted in \hyperref[fig:appendixa-bifurcation]{Figure A.4}.  Here, we have two intersecting solution surfaces, which, for visualization purposes, we have split.  For the first solution branch, pseudo-arc-length path-following continues until a singular point is reached.  The search can then continue on this branch and on the new one.  When the bifurcating path is taken, the state aggregation may become more finely grained, allowing for the agent to better specialize to the environment.  When the original path is pursued, then the state abstraction can the same and the policy may only marginally improved.  Sometimes, however, newly bifurcating path may lead to sub-optimal policies, which necessitates backtracking in the case of a single search process.  To avoid the need to re-explore the solution curve, which can be computationally wasteful, we spawn a new search process and explore the total space of policies in parallel.

In certain circumstances, we may wish to avoid deriving the coefficients of the algebraic bifurcation equations.  Even with the approximations that we consider, the computation time can still be high.  We thus can consider an alternative whenever one of the branches is known.  In this case, we can seek solutions on a subset of of a branch that is parallel to the tangent but displaced from the bifurcation in some direction that is normal to the tangent.  If we assume that the Hessian nullspace has unit dimensionality, then $[\partial_\varphi({\pi}_k(\varphi_*)),{\beta}_k(\varphi_*))]_0 \!=\! \partial_\varphi {\beta}_k(\varphi_*)\phi_0 \!+\! \psi_1^*\partial_\varphi \pi_k(\varphi_*)\phi_1$.  A\\ \noindent vector orthogonal to $\partial_\varphi({\pi}_k(\varphi_*)),{\beta}_k(\varphi_*))$, in the hyperplane spanned by $(\phi_1,0)$ and $(\phi_0,1)$, is $\xi_0'\phi_0 \!+\! \xi_1'\phi_1$, where the\\ \noindent coefficients are $\xi_0' \!=\! -\psi_1 \partial_\varphi {\pi}_k(\varphi_*)\|\phi_1\|^2$ and $\xi_1' \!=\! \partial_\varphi {\beta}_k(\varphi_*)(1 \!+\! \|\phi_0\|^2)$.  We thus want to find a second tangent,
\begin{equation*} 
\begin{array}{c}
[({\pi}_k(\varphi_*)),{\beta}_k(\varphi_*))]_1 = [({\pi}_k(\varphi_*)),{\beta}_k(\varphi_*))]_0 + \epsilon(\xi_0'\phi_0 \!+\! \xi_1'\phi_1) + \omega_1,\;\; [\varphi_k(\varphi_*)]_1 = [\varphi_k(\varphi_*)]_0 + \epsilon \xi_0' + \omega_2\vspace{0.05cm}\\
\textnormal{such that } \nabla_{\pi,\beta}\mathcal{L}([({\pi}_k(\varphi_*)),{\beta}_k(\varphi_*))]_1, [\varphi_k(\varphi_*)]_1) = 0 \textnormal{ and } (\xi_0'\phi_0^* \!+\! \xi_1'\phi_1^*)\omega_1 + \xi_0'\omega_2 = 0,
\end{array}
\end{equation*}
where $\omega_1,\omega_2 \!\in\! \mathbb{R}$ and with $\epsilon \!\in\! \mathbb{R}_+$ being sufficiently large.  This system of equations can be solved using either\\ \noindent Newton's method or, more efficiently, quasi-Newton methods.  For branches with more than one bifurcation, though, this approach cannot be used.  Instead, either of the root-finding methods should be applied so that each tangent vector can be specified.

Another option is to use the ideas outlined in \cite{KellerHB-jour1970a} whenever one of the branches is already known.  In this case, we seek a bifurcated branch of the form $([({\pi}_k(\sigma),{\beta}_k(\sigma))]_0 \!+\! \epsilon(v \!+\! \phi_0), [\vartheta_k(\sigma)]_0)$ such that $\psi_0^* v \!=\! 0$.  We set, for $\epsilon \!\neq\! 0$,
\begin{equation*}
\mathcal{H}(\sigma;\epsilon,v) = \psi_0^*(\nabla^2_{\!\pi,\beta}\mathcal{L}(\gamma_k(\varphi_*))v - \epsilon^{-1}\nabla_{\!\pi,\beta}\mathcal{L}([({\pi}_k(\sigma),{\beta}_k(\sigma))]_0 + \epsilon(v \!+\! \phi_0), [\vartheta_k(\sigma)]_0)).
\end{equation*}
To ensure that the right side of this expression is in $\textnormal{range}(\nabla^2_{\!\pi,\beta}\mathcal{L}(\gamma_k(\varphi_*)))$, we would like to pick $\sigma \!=\! \varphi$ such that\\ \noindent $\mathcal{H}(\sigma;\epsilon,v) \!=\! 0$.  For $\epsilon \!=\! 0$, $\mathcal{H}(\sigma;0,v) \!=\! \psi_0^*(\nabla^2_{\!\pi,\beta}\mathcal{L}(\gamma_k(\varphi_*))v \!-\! \epsilon^{-1}\nabla_{\!\pi,\beta}\mathcal{L}([({\pi}_k(\sigma),{\beta}_k(\sigma))]_0)(\phi_0 \!+\! v)$.  It can thus be\\ \noindent seen that regardless of the value of $\epsilon$, $v \!=\! 0$ guarantees $\mathcal{H}(\varphi_*;\epsilon,0) \!=\! 0$.  Moreover, $\partial_\varphi \mathcal{H}(\varphi_*;0,0) \!\neq\! 0$, where
\begin{equation*}
\partial_\varphi \mathcal{H}(\varphi_*;0,0) = - \psi_0^*(\nabla^3_{\!\pi,\beta}\mathcal{L}(\gamma_k(\varphi_*)) \partial_\varphi(\pi_k(\varphi_*),\beta_k(\varphi_*)) + \partial_\vartheta \nabla^2_{\!\pi,\beta}\mathcal{L}(\gamma_k(\varphi_*)) \partial_\varphi\vartheta(\varphi_*))\phi_0.
\end{equation*}
We can therefore use \hyperref[propA.3]{Proposition A.3} to guarantee that $\varphi \!=\! \sigma(\epsilon \!+\! v)$ is a root of $\mathcal{H}(\sigma;\epsilon,v) \!=\! 0$.  Since the Newton maps\\ \noindent are contractions, we have a unique solution $v \!=\! v(\epsilon)$ for sufficiently small $\epsilon$.  There is, however, one issue with this\\ \noindent approach that limits its appeal---we have to solve $\mathcal{H}(\varphi;\epsilon,v) \!=\! 0$ for $\varphi$.  There is no closed-form solution and an iterative process is needed \cite{RheinboldtWC-jour1976a}.  For example, we could employ a chord method to specify iterative solutions, which for $\sigma$ would be $\psi_0^* V \sigma^{p+1} \!=\! \psi_0^* V \sigma^p \!-\! \mathcal{H}(\sigma^p;\epsilon,v^p)$.  Similarly, $\nabla^2_{\!\pi,\beta}\mathcal{L}(\gamma_k(\varphi_*))v^{p+1} \!=\! \mathcal{H}(\varphi_*;\sigma^p,v^p)/\psi_0^* \!-\! V(\sigma^{p+1} \!-\! \sigma^p)$ for $v$.  In\\ \noindent both cases, $V \!=\! (\nabla^3_{\!\pi,\beta}\mathcal{L}(\gamma_k(\varphi_*))[\partial_\varphi({\pi}_k(\varphi_*),{\beta}_k(\varphi_*))]_0 \!+\! \partial_\varphi\nabla^2_{\!\pi,\beta}\mathcal{L}(\gamma_k(\varphi_*)) [\partial_\varphi\vartheta_k(\varphi_*)]_0)\phi_0$.  It is straightforward\\ \noindent to show convergence.  To avoid evaluating third-order Fr\'{e}chet derivatives, we can use the same approximation scheme as in the first approach.

Regardless of which approach is used, we are guaranteed, by the Equivariant Branching Lemma \cite{VanderbauwhedeA-book1982a}, that these bifurcations exist.  As long as each branch can be enumerated and explored, then, by \hyperref[propA.12]{Proposition A.12}, we are guaranteed that globally optimal policies will be uncovered.

\setstretch{0.95}\fontsize{9.75}{10}\selectfont
\putbib
\end{bibunit}

\clearpage\newpage

\begin{bibunit}
\bstctlcite{IEEEexample:BSTcontrol}

\RaggedRight\parindent=1.5em
\fontdimen2\font=2.1pt\selectfont
\singlespacing
\allowdisplaybreaks

\setcounter{figure}{0}
\setcounter{equation}{0}
\renewcommand{\thefigure}{B.\arabic{figure}}
\renewcommand\theequation{B.\arabic{equation}}
\phantomsection\label{secB}

%%%%%%%%%%%%%%%%%%%%%%%%%%%%%%%%%%%%%%%%%%%%%%%%%%%%%%%%%%%%%%%%%%%%%%%%%%%
%%%%%%%%%%%%%%%%%%%%%%%%%%%%%%%%%%%%%%%%%%%%%%%%%%%%%%%%%%%%%%%%%%%%%%%%%%%
\subsection*{\small{\sf{\textbf{Appendix B}}}}

In this appendix, we provide details about the \emph{Millipede} and \emph{Centipede} simulations presented in \hyperref[sec5]{Section 5}.

We first describe the training protocols and parameter values used for the various comparative methods (see \hyperref[secB.1]{Appendix B.1}).  We then discuss the gameplay and reward structure for these two games (see \hyperref[secB.2.1]{Appendix B.2.1}).  We also specify a state-action-space representation that is used to characterize both environments (see \hyperref[secB.2.2]{Appendix B.2.2}).

We additionally provide supplemental results to augment the discussions in \hyperref[sec5]{Section 5} (see \hyperref[secB.3]{Appendix B.3}).  We highlight the state abstractions that emerge for both games and relate them to observed gameplay behaviors and reductions in agent costs.

\phantomsection\label{secB.1}
%%%%%%%%%%%%%%%%%%%%%%%%%%%%%%%%%%%%%%%%%%%%%%%%%%%%%%%%%%%%%%%%%%%%%%%%%%%
%%%%%%%%%%%%%%%%%%%%%%%%%%%%%%%%%%%%%%%%%%%%%%%%%%%%%%%%%%%%%%%%%%%%%%%%%%%
\subsection*{\small{\sf{\textbf{B.1.$\;\;\;$Simulation Preliminaries}}}}

For each exploration strategy, we rely on coupled-$Q$-learning process.  The learning rate for the fast-time update follows an inverse polynomial decay schedule, from 0.6 to 0.0001.  In standard $Q$-learning, such an annealing helps facilitate polynomial-rate policy convergence \cite{EvenDarE-jour2003a}.  We find it works well for coupled $Q$-learning too.  The learning rate for the slow-time update uses the same type of schedule, albeit from 0.25 to 0.0001.  Both the fast- and slow-time decay schedules ensure that the corresponding state-action value-functions stabilize over time.  We set the discount factor to 0.85 so as to preempt slow convergence \cite{SzepesvariC-coll1997a}. 

Our version of coupled $Q$-learning relies on prioritized experience replay.  In all of our simulations, we use a prioritization constant of 0.6, an importance-sampling exponential factor of 0.4, and an proportional prioritization offset of 0.01.  A replay capacity of 100000 state transitions is used to provide large state-action coverage \cite{FedusW-conf2020a}.

When using epsilon-greedy and soft-max exploration, only a single parameter needs to be set, the exploration rate.  For both search methods, we consider a fixed exploration rate of 0.55 for some of our comparisons.  Such a value strikes a reasonable balance between trying new actions and favoring optimal ones.  We also consider a fixed, inverse-polynomial annealing schedule, which is from 0.75 to 0.01.

For value-of-information exploration using either path-following or pseudo-arc-length path-following, we consider a policy accuracy of 0.01.  Decreasing the value beyond this threshold did little to improve policy performance and simply increases the optimization time.  The performance changes were not statistically significant according to Friedman's tests and subsequent Nemenyi's tests.  For the exploration rate, we consider an initial value of 0.85.  Lower values increase that chance that solution-surface backtracking will be needed to find the optimal bifurcation.  Learning can stagnate during this period. 

For the non-path-following-based value-of-information, we evaluate both fixed and adaptive exploration rates.  The fixed case relies on the same parameter values as soft-max exploration.  In the adaptive case, we use a cross-entropy-based adjustment combined with an initial annealing schedule.  Whenever the cross-entropy between two probabilistic policies is at or above 0.35, then the exploration rate is decreased by a multiplicative factor of 0.925.  If the policy cross-entropy is above that threshold, then the exploration factor is multiplicatively increased by 1.025.  We perform this test for every pair of policies separated by twenty episodes to discern if many updates are being made to the policy entries.  This cross-entropy test has the effect of reducing exploration if too many policy changes are being made and increasing it if learning has stagnated.

All variants of the value of information assume access to the state prior probability.  A priori, this probability is not known.  Assuming that it is uniform discards much of the information about the environment dynamics.  Attempting to estimate it also proves difficult via conventional density-approximation methods, as the state features exist in a high-dimensional space.

Here, we derive this prior probability by solving a distributional pre-image problem \cite{SongL-conf2008a} in a dimensionally agonistic manner.  For a given set of initial state transitions, we compute their kernel mean embedding \cite{SriperumbudurBK-conf2008a,FukumizuK-coll2011a}.  We update this mean embedding for each additional state that is visited, including those in parallel solution-branch searches.  To actually form the mean embedding, we use a Gaussian kernel with a bandwidth of 0.25.  Such a kernel has many appealing traits.  Foremost, it is a universal kernel, which implies that the mean-element can distinguish between unique distributions \cite{SriperumbudurBK-conf2010a,SriperumbudurBK-jour2011a}.  Moreover, such a kernel simplifies the pre-image problem.  When using a mixture of Gaussians, which we do, each of the integral terms in the optimization process possesses a closed-form solution.

These parameter values were discerned from a finely-grained grid search conducted on a computing cluster with 128 NVIDIA Quadro RTX A6000s.  Each simulation was seeded with a random probabilistic policy.  For a given set of values, we ran five simulations to assess average performance.  The best-performing parameters were then used.

The results we present in \hyperref[sec5]{Section 5} were obtained from thirty Monte Carlo simulations performed for each method.  Learning was terminated after 24000 episodes.  We then averaged the results and smoothed them, via a fourth-order Savitzky-Golay process.  This was done to capture the dominant trends in the results.  Due to the large number of methods and quantities being compared, we plot only averages in \hyperref[sec5]{Section 5}.

%%%%%%%%%%%%%%%%%%%%%%%%%%%%%%%%%%%%%%%%%%%%%%%%%%%%%%%%%%%%%%%%%%%%%%%%%%%
%%%%%%%%%%%%%%%%%%%%%%%%%%%%%%%%%%%%%%%%%%%%%%%%%%%%%%%%%%%%%%%%%%%%%%%%%%%
\subsection*{\small{\sf{\textbf{B.2.$\;\;\;$Gameplay Environments}}}}

\phantomsection\label{secB.2.1}
%%%%%%%%%%%%%%%%%%%%%%%%%%%%%%%%%%%%%%%%%%%%%%%%%%%%%%%%%%%%%%%%%%%%%%%%%%%
%%%%%%%%%%%%%%%%%%%%%%%%%%%%%%%%%%%%%%%%%%%%%%%%%%%%%%%%%%%%%%%%%%%%%%%%%%%
\subsection*{\small{\sf{\textbf{B.2.1.$\;\;\;$Gameplay Overview}}}}

{\small{\sf{\textbf{Millipede Gameplay.}}}} In the game \emph{Millipede}, for the Nintendo GameBoy, the agent dictates the two-dimensional movement of a mobile platform.  The objective of every stage is to eliminate all of the millipede body segments, before they hit the agent, by firing bolts at them.  Destroying a body segment results in a small cost ($-$10), while destroying the head yields a large cost ($-$50).  The agent receives a small cost ($-$5) for being aligned with a millipede segment\\  

\setlength{\fboxrule}{0.8pt}
\setlength{\fboxsep}{0pt}
\begin{wrapfigure}{r}{0.375\textwidth}
   \vspace{-0.2cm}{\framebox{\includegraphics[width=0.175\textwidth]{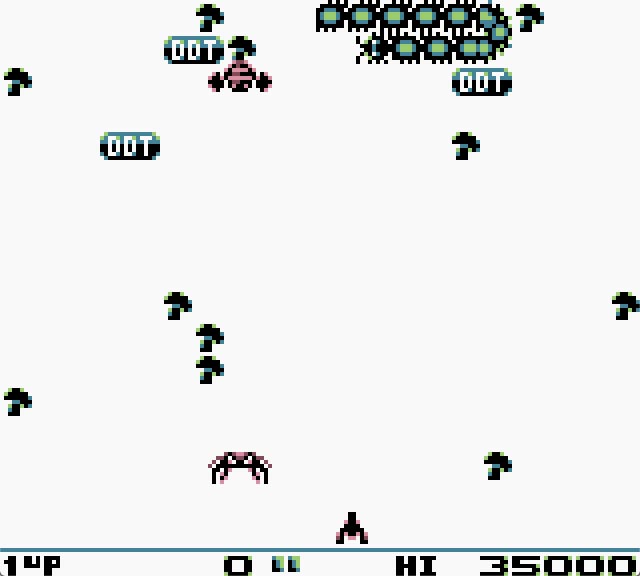}}} {\framebox{\includegraphics[width=0.175\textwidth]{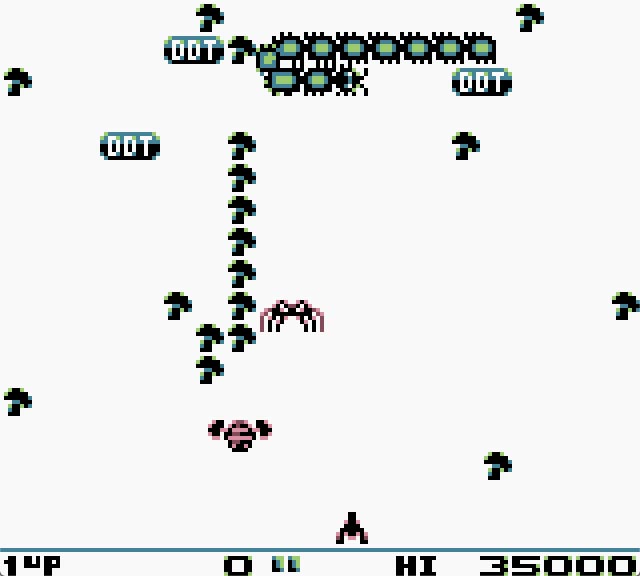}}}
   \caption[]{A bee leaving a trail of mushrooms as it moves from the top of the screen to the bottom in the game \emph{Millipede}.  The left image is earlier in time than the right image.\vspace{-0.2cm}}
   \label{fig:millipede-bee}
\end{wrapfigure}

\noindent and another ($-$10) for shooting while aligned, regardless of if the bolt connects.  An alignment cool-down period of approximately one second is used to prevent the agent from trivially accruing costs by continuously breaking and regaining alignment with the millipede.

The agent's objective is impeded in several ways.  Foremost, there are mushrooms present in the environment, which act as barriers to the bolts and make the millipede body segments harder to hit.  When a millipede encounters a mushroom in its path, it drops down a row and reverses direction.  Mushrooms can absorb multiple bolts before disappearing.  A minuscule cost is accrued as a mushroom is hit ($-$1) and when it is destroyed ($-$5).  Shooting any section of the millipede creates a new mushroom.  Mushrooms also randomly grow and are culled at various intervals.  Randomly spawned enemies, known as bees and dragonflies, have the ability to leave mushrooms as they travel from top to bottom in the environment (see \cref{fig:millipede-bee}).  Mushrooms can turn into impenetrable flowers when touched by an enemy known as beetles.  Flowers return to normal either when the agent dies or if a nearby DDT canister is hit.  Earwigs can poison the mushrooms so that a millipede segment\\ 

\setlength{\fboxrule}{0.8pt}
\setlength{\fboxsep}{0pt}
\begin{wrapfigure}{l}{0.375\textwidth}
   \vspace{-0.275cm}{\framebox{\includegraphics[width=0.175\textwidth]{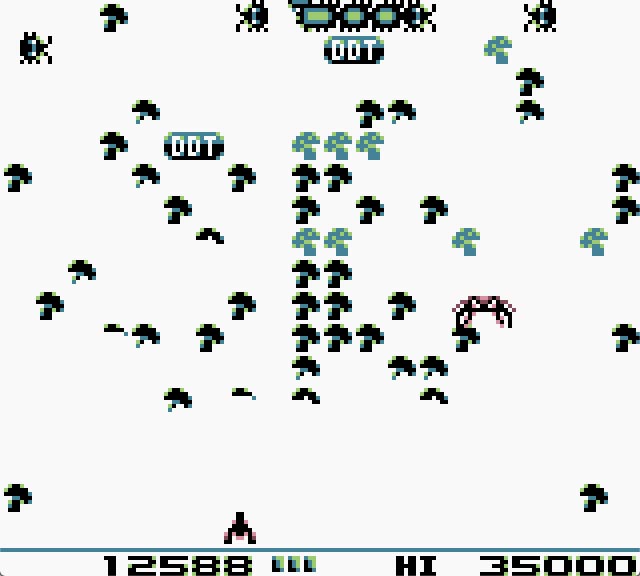}}} {\framebox{\includegraphics[width=0.175\textwidth]{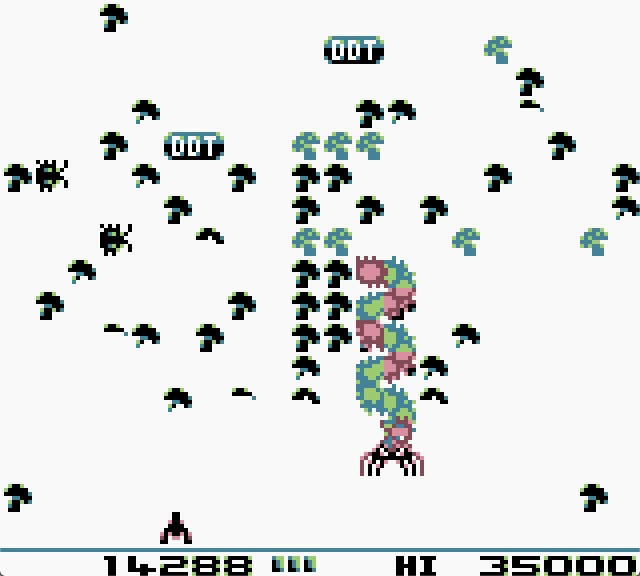}}}
   \caption[]{An example of a \emph{Millipede} game state where multiple poison mushrooms are visible in the environment.  These are denoted using light green mushroom sprites instead of the more common green-black mushrooms.  Once a millipede hits a poison mushroom, it immediately ignores boundary constraints and heads toward the bottom of the screen.  The left image is earlier in time than the right image.\vspace{-0.4cm}}
   \label{fig:millipede-poisonmushroom}
\end{wrapfigure}

\noindent hurtles towards the agent when touching one (see \cref{fig:millipede-poisonmushroom}).  This can create conditions where the agent becomes trapped in a small part of the screen.  Destroying poisoned mushrooms is encouraged using a large-magnitude cost ($-$500).

The agent also faces several enemies, each with different behaviors.  For instance, spiders bounce irregularly across the player area and consume mushrooms.  Multiple spiders can appear on the screen simultaneously later in the game.  Mosquitoes and beetles move in various parts of the environment.  Destroying mosquitoes scrolls the position of everything in the environment up one row.  Destroying beetles scrolls the position of everything down one row.  Hitting inchworms slows all enemies for a brief time.  The agent incurs negative costs for destroying such enemies.  Easily hit enemies like inchworms ($-$100), bees ($-$200), spiders ($-$300 to $-$1200) have low costs compared to ones that are either harder to hit or spawn less frequently like beetles ($-$300), mosquitoes ($-$400), dragonflies ($-$500), and earwigs ($-$1000).  Bees have a moderate cost ($-$500), as they can clutter the environment with mushrooms and make targeting\\ 

\setlength{\fboxrule}{0.8pt}
\setlength{\fboxsep}{0pt}
\begin{wrapfigure}{r}{0.375\textwidth}
   \vspace{-0.275cm}{\framebox{\includegraphics[width=0.175\textwidth]{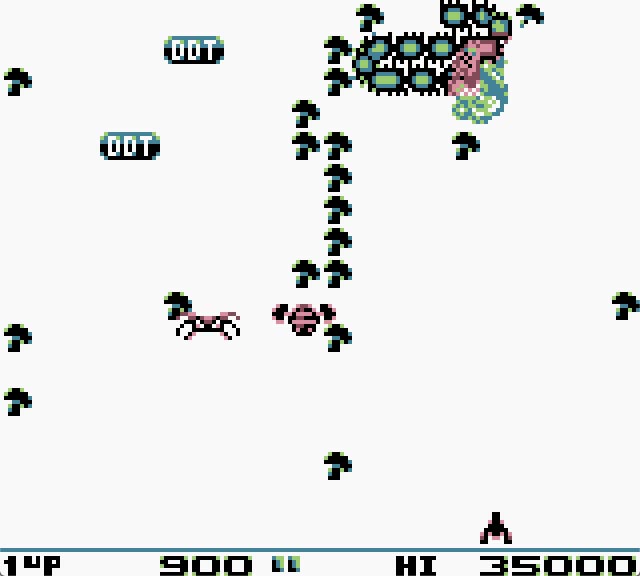}}} {\framebox{\includegraphics[width=0.175\textwidth]{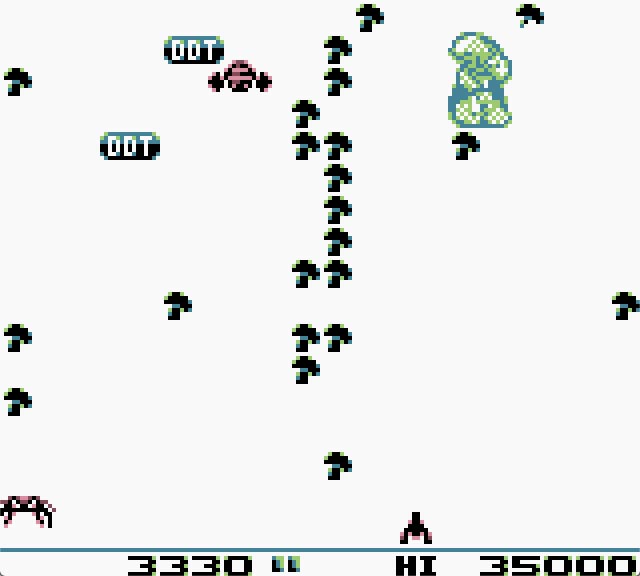}}}
   \caption[]{Using DDT canisters to take out enemies in the game \emph{Millipede}.  Here, the DDT cloud destroys an entire millipede.  The left image is earlier in time than the right image.\vspace{-0.225cm}}
   \label{fig:millipede-ddt}
\end{wrapfigure}

\noindent certain enemies difficult.  The agent receives a small cost ($-$30) for being aligned with any enemy and another ($-$20) for shooting while aligned, regardless of if the bolt connects.  We use the same alignment cool-down strategy outlined above.  Activating DDT canisters causes a cloud of poison gas to spawn, which destroys nearby enemies (see \cref{fig:millipede-ddt}).  Any enemies that die within the cloud increase the accrued costs by one and one half times.  Triggering a DDT canister when an enemy is adjacent to it is encouraged using a moderate cost ($-$300).

The agent loses a life ($+$1000) when it is hit by any enemy.  A game ends when all of the agent's lives are gone.  Good policies should hence choose context-specific actions that minimize the total cost.

An in-game score is supplied and can be used to track agent performance.  However, based on initial experiments, we opted to fashion the above scoring system.  Such a system provides denser rewards, compared to the in-game score, which promotes better self-supervision.  For instance, the agent learns that it can destroy mushrooms to clear out sections of the environment.  Doing so enables the agent to hit enemies.  The in-game reward for destroying mushrooms is too low for this to readily occur early during learning.  Likewise, the agent learns to track enemies more effectively early during the learning process.  

\vspace{0.15cm}{\small{\sf{\textbf{Centipede Gameplay.}}}} The gameplay for \emph{Centipede} is highly related to that of \emph{Millipede}.  The agent controls the two-dimensional movement of a mobile platform and fires bolts at enemies that appear on the screen.  The most common enemies are centipedes.  They have a similar behavioral pattern to their counterparts in \emph{Millipede} but leave mushrooms when they are shot.  Since other enemies are rare in \emph{Centipede}, destroying a centipede segment results in a moderate cost ($-$25).  Destroying the head yields a larger cost ($-$100).  Finishing all segments, and hence transitioning between levels, is encouraged ($-$250).  We use the same alignment cost and cool-down strategy as in \emph{Millipede} to help the agent learn to track the centipede segments.

\setlength{\fboxrule}{0.8pt}
\setlength{\fboxsep}{0pt}
\begin{wrapfigure}{r}{0.375\textwidth}
   \vspace{-0.4cm}{\framebox{\includegraphics[width=0.175\textwidth]{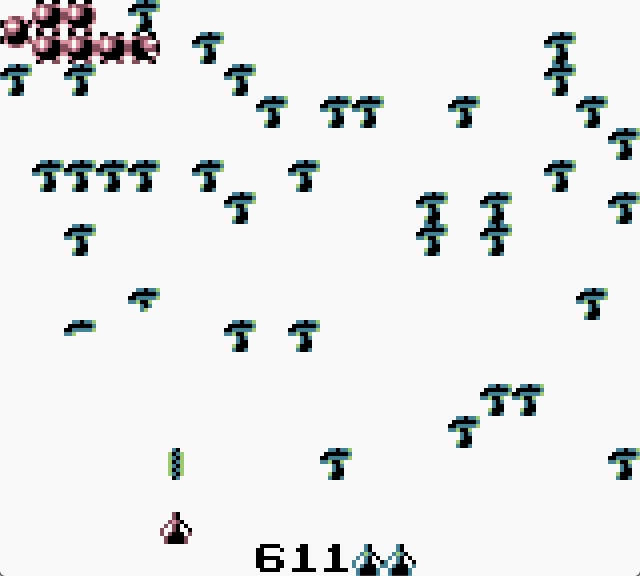}}} {\framebox{\includegraphics[width=0.175\textwidth]{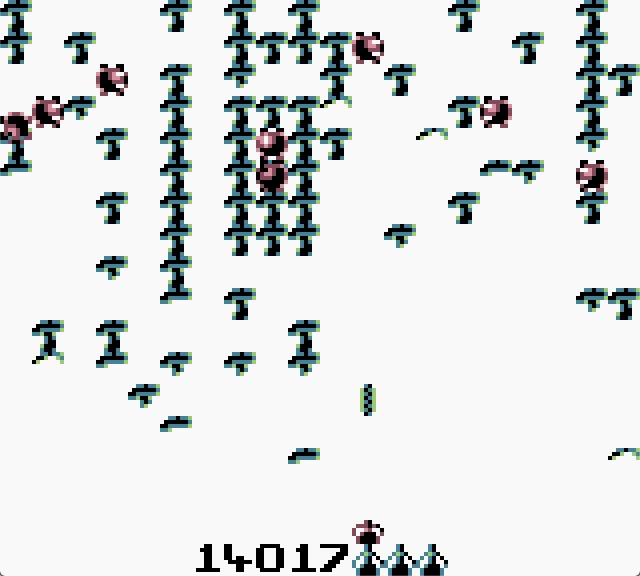}}}
   \caption[]{A large number of mushrooms can quickly emerge in the game \emph{Centipede}, as centipede segments leave them when destroyed.  In later levels of the game, only centipede heads spawn.  They hence distribute mushrooms near-uniformly on the screen.  The left image is much earlier in time than the right image.\vspace{-0.2cm}}
   \label{fig:centipede-flea}
\end{wrapfigure}

As in \emph{Millipede}, mushrooms are present in \emph{Centipede}.  However, they are far more prevalent and more difficult to destroy in \emph{Centipede}, as they require four shots.  There are no DDT canisters to remove large fields of mushrooms.  The agent dying also restores partially destroyed mushrooms.  It is therefore common for many mushrooms to be present in later levels, complicating the agent's progress.  Increasing amounts of mushrooms also cause the centipede segments to reach the agent quickly, limiting the agent's action choices in certain situations.  We encourage the agent to destroy mushrooms whenever possible.  The agent receives a small cost ($-$2) for shooting at a mushroom and weakening it.  This cost increases ($-$5, $-$7, $-$10) with each additional shot that weakens and eventually removes a mushroom from the environment.  However, the agent is also encouraged to leave vertical tunnels of six or more contiguous mushrooms ($-$500) so that centipedes are funneled into them and their segments can be easily destroyed.

\setlength{\fboxrule}{0.8pt}
\setlength{\fboxsep}{0pt}
\begin{wrapfigure}{l}{0.375\textwidth}
   \vspace{-0.2cm}{\framebox{\includegraphics[width=0.175\textwidth]{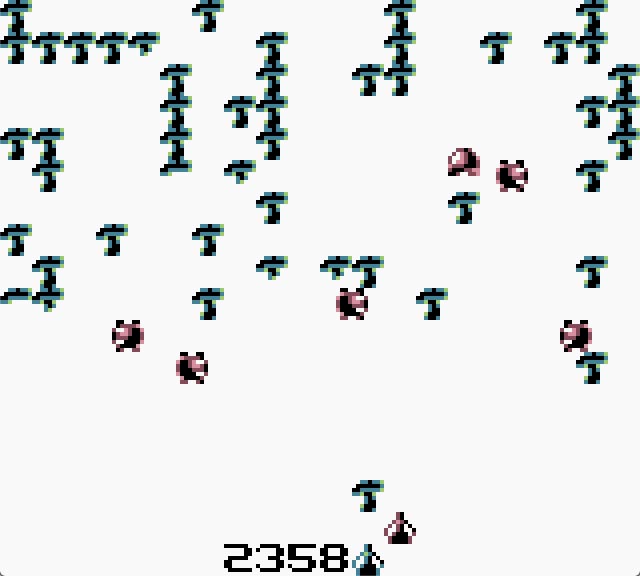}}} {\framebox{\includegraphics[width=0.175\textwidth]{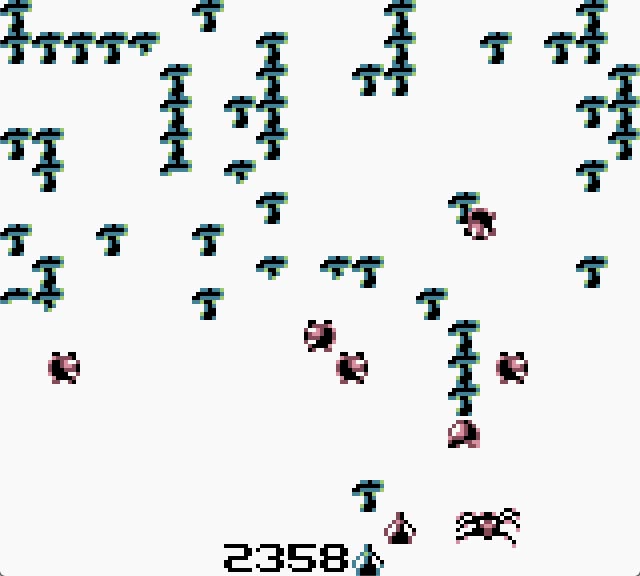}}}
   \caption[]{A flea leaving a trail of mushrooms as it moves from the top of the screen to the bottom in the game \emph{Centipede}.  The left image is earlier in time than the right image.\vspace{-0.2cm}}
   \label{fig:centipede-flea}
\end{wrapfigure}

There are fewer enemies in \emph{Centipede} than \emph{Millipede}.  Fleas take the place of bees and yield the same cost ($-$300).  Both require two shots to destroy.  However, unlike bees, fleas move more quickly after the first shot.  Another enemy, scorpions, possess a high destruction cost ($-$500).  Scorpions cause mushrooms they touch to turn poisonous; these mushrooms behave just as in \emph{Millipede}.  Lastly, spiders are present and yield a cost that is proportional to how close they are to the agent when destroyed ($-$300 to $-$1200).  Spiders zig-zag through the environment, sometimes blocking the agent.  They do, however, randomly clear mushrooms in their path.

The agent loses a life ($+$1000) when it is hit by any enemy.  The agent gains a life, up to a maximum of six, for every 12000 in-game points earned ($-$1000).  A game ends when all of the agent's lives are gone.  Good policies should hence choose context-specific actions that minimize the total cost.

\phantomsection\label{secB.2.2}
%%%%%%%%%%%%%%%%%%%%%%%%%%%%%%%%%%%%%%%%%%%%%%%%%%%%%%%%%%%%%%%%%%%%%%%%%%%
%%%%%%%%%%%%%%%%%%%%%%%%%%%%%%%%%%%%%%%%%%%%%%%%%%%%%%%%%%%%%%%%%%%%%%%%%%%
\subsection*{\small{\sf{\textbf{B.2.2.$\;\;\;$State-Action Space}}}}

{\small{\sf{\textbf{Action Space.}}}} The action spaces for both \emph{Centipede} and \emph{Millipede} are limited to six discrete actions.  For every twentieth step, the agent has the option of moving in one of four directions, up, down, left, and right, by simulating directional-pad button presses.  It can also remain stationary.  At any time, the agent can simulate a press of the action button in an attempt to fire a bolt, provided that one is loaded.  The chosen action is then repeated over the next nineteen steps.  Repeating the motions in this way prevents significant jitter, which generally helps improve gameplay performance.  All other GameBoy buttons are disabled.

We permit the agent to string up to three arbitrary button presses together to form a compound action that is executed over up to a set number of game frames.  While performing a compound action, any additional button presses made by the agent are ignored.

\vspace{0.15cm}{\small{\sf{\textbf{State Space.}}}} We evaluated a variety of state spaces for both games.  We initially considered convolutional autoencoders that were pre-trained on a half-hour of human gameplay videos and fixed during reinforcement learning.  We then considered pre-trained convolutional and recurrent-convolutional autoencoders that could be updated during reinforcement learning.

\setlength{\fboxrule}{0.8pt}
\setlength{\fboxsep}{0pt}
\begin{wrapfigure}{r}{0.375\textwidth}
   \vspace{-0.6cm}{\framebox{\includegraphics[width=0.175\textwidth]{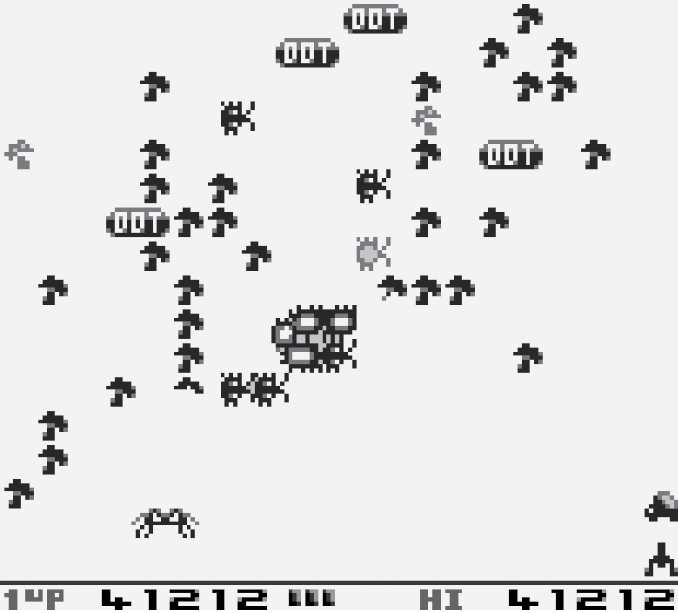}}} {\framebox{\includegraphics[width=0.175\textwidth]{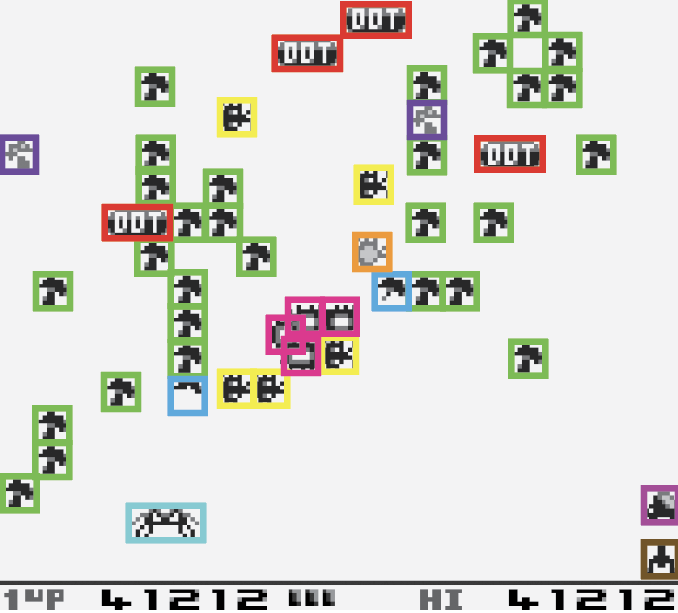}}}
   \caption[]{A visualization of template-correlation sprite recognition for the game \emph{Millipede}.  The right image shows the grid-occupancy labels for the game frame on the left.\vspace{-0.2cm}}
   \label{fig:millipede-features}
\end{wrapfigure}

Neither of these options fare well.  The former yields poor features for gameplay, as they are uncoupled from the extrinsic reward structure and hence the inferred policy.  The latter approach produces representations with the same issues.  They are also often altered too greatly over time to facilitate good learning convergence.  Moreover, online adaptation of the features can cause significant learning stagnations for strategies that used pre-defined annealing schedules for the exploration rate.  This makes a fair assessment of the exploration strategies difficult.

To facilitate fair comparison of the different strategies, we rely on a fixed state space composed of static and dynamic features.  For each game frame, we determine which grid cells are occupied and use knowledge of the game sprites to recognize the agent, enemies, and any environment objects.  An example of a labeled game frame is given in \cref{fig:millipede-features}.  The entire labeled occupancy grid is then taken as the static feature representation of a game state.  For the dynamic features, we characterize both the changes and the directional movement of any objects over the previous twenty game frames.  This is done using a primitive optical flow process at the sprite level, not the pixel level.  Objects which have not been altered in some way are ignored in the dynamic-feature representation.

\newcommand*\circled[2]{\tikz[baseline=(char.base)]{
\node[circle,white,draw,scale=#2,inner sep=1pt] (char) {#1};}}

\newcommand*\blackwhitecircled[2]{\tikz[baseline=(char.base)]{
\node[circle,black,fill=black,draw,scale=#2,inner sep=1pt] (char) {#1};}}

\begin{figure*}

\hspace{-0.125cm}\scalebox{0.775}{
\begin{tikzpicture}
  \node[] at (0.1,0.75) {\includegraphics[width=5.35in]{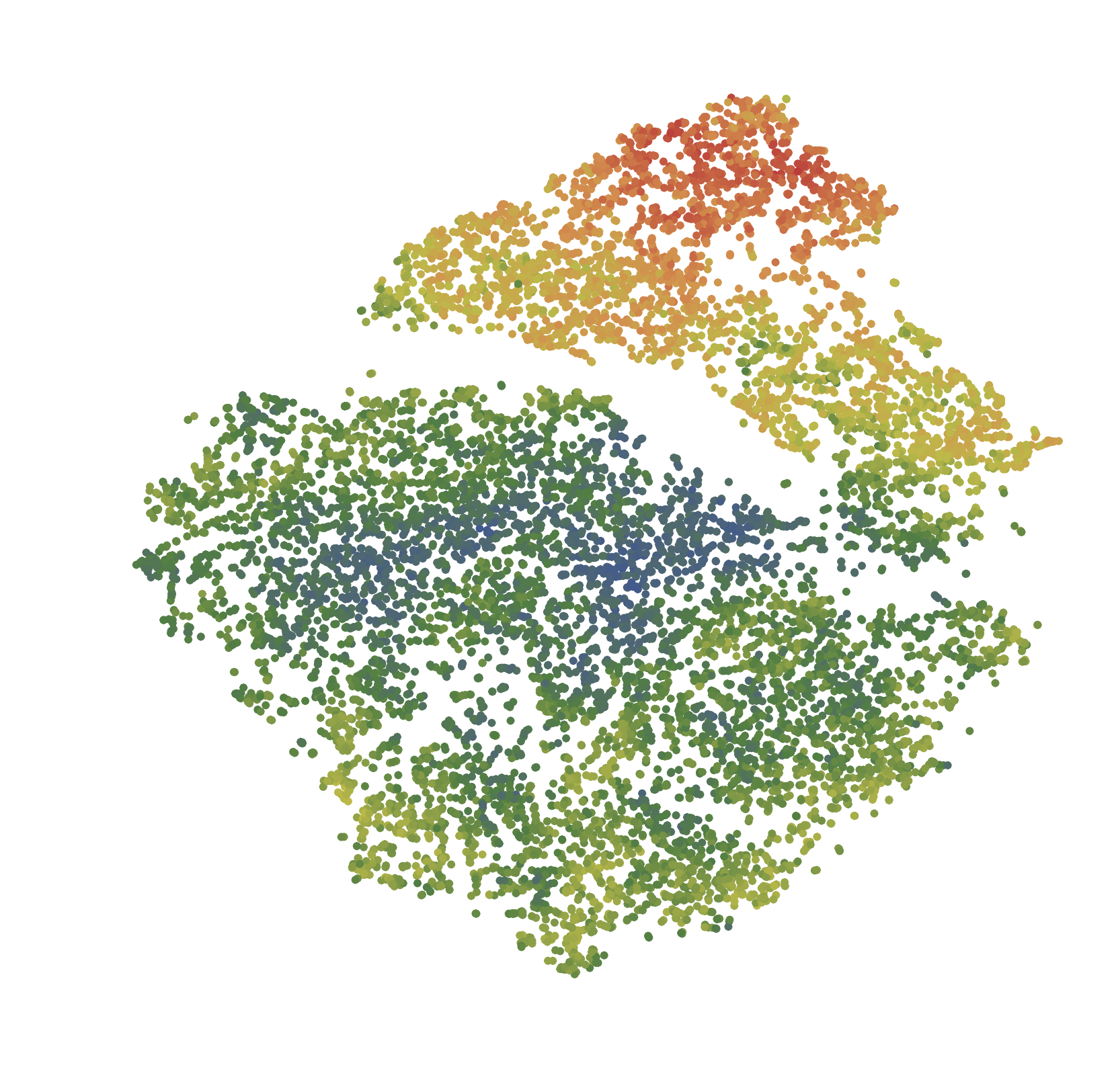}};
  \node[] at (-8.2,5.25) {\includegraphics[width=1.85in]{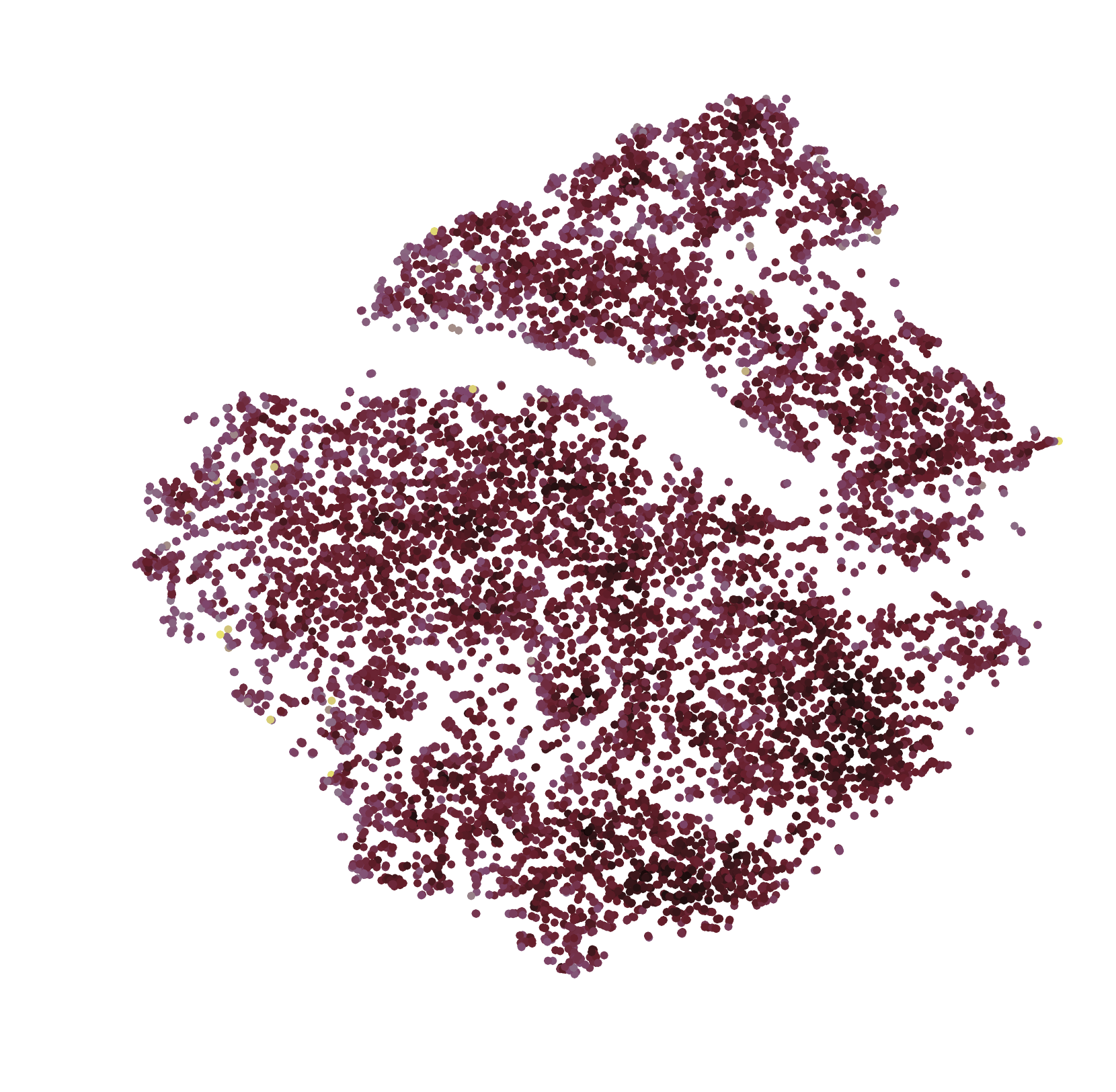}};

  \node at (-8.2,2.65) {\includegraphics[width=1.39in]{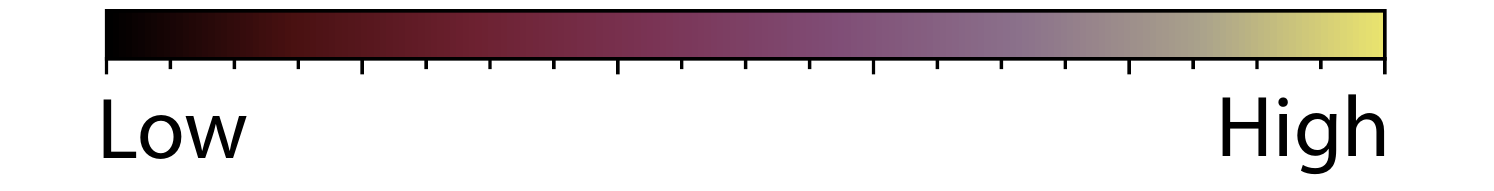}};
  \node at (-8.2,2.215) {$H\!(p(s_{t+1}|s_t,a_t))$};

  \setlength{\fboxrule}{0.5pt}
  \setlength{\fboxsep}{0.025pt}
  \node at (-9,-5.5) {\framebox{\includegraphics[width=1.175in]{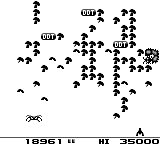}}};
  \node at (-5.825,-5.5) {\framebox{\includegraphics[width=1.175in]{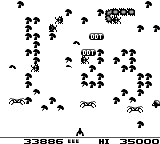}}};
  \node at (-9,-2.65) {\framebox{\includegraphics[width=1.175in]{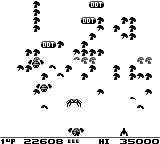}}};
  \node at (-9,0.2) {\framebox{\includegraphics[width=1.175in]{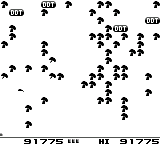}}};

  \node at (8.75,-5.5) {\framebox{\includegraphics[width=1.175in]{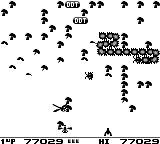}}};
  \node at (5.575,-5.5) {\framebox{\includegraphics[width=1.175in]{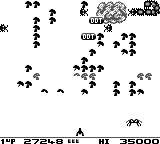}}};
  \node at (8.75,-2.65) {\framebox{\includegraphics[width=1.175in]{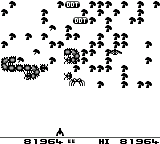}}};
  \node at (8.75,0.2) {\framebox{\includegraphics[width=1.175in]{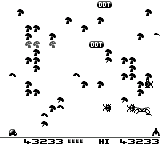}}};

  \node at (5.05,6.25) {\textcolor{black}{\bf\Large $\mathcal{S}$}};

  \newcommand*\blackcircled[2]{\tikz[baseline=(char.base)]{
  \node[circle,black,draw,scale=#2,inner sep=1pt] (char) {#1};}}

  \node at (-7.05,0.2) {\textcolor{black}{\bf \blackcircled{\small \textcolor{black}{1}}{1} }};
  \node at (-7.05,-2.65) {\textcolor{black}{\bf \blackcircled{\small \textcolor{black}{2}}{1} }};
  \node at (-8.95,-7.215) {\textcolor{black}{\bf \blackcircled{\small \textcolor{black}{3}}{1} }};
  \node at (-5.775,-7.215) {\textcolor{black}{\bf \blackcircled{\small \textcolor{black}{4}}{1} }};

  \node at (6.925,0.2) {\textcolor{black}{\bf \blackcircled{\small \textcolor{black}{5}}{1} }};
  \node at (6.925,-2.65) {\textcolor{black}{\bf \blackcircled{\small \textcolor{black}{6}}{1} }};
  \node at (8.8,-7.215) {\textcolor{black}{\bf \blackcircled{\small \textcolor{black}{7}}{1} }};
  \node at (5.65,-7.215) {\textcolor{black}{\bf \blackcircled{\small \textcolor{black}{8}}{1} }};

  \node at (2.25,5.2) {\textcolor{black}{\bf \blackwhitecircled{\small \textcolor{white}{8}}{1} }};
  \node at (0.0,5.35) {\textcolor{black}{\bf \blackwhitecircled{\small \textcolor{white}{7}}{1} }};
  \node at (3.7,1.9) {\textcolor{black}{\bf \blackwhitecircled{\small \textcolor{white}{6}}{1} }};
  \node at (-2.5,2.95) {\textcolor{black}{\bf \blackwhitecircled{\small \textcolor{white}{5}}{1} }};
  \node at (-4.25,2.8) {\textcolor{black}{\bf \blackwhitecircled{\small \textcolor{white}{4}}{1} }};
  \node at (-1.75,-4.5) {\textcolor{black}{\bf \blackwhitecircled{\small \textcolor{white}{3}}{1} }};
  \node at (-2.05,-1.6) {\textcolor{white}{\bf \blackwhitecircled{\small \textcolor{white}{2}}{1} }};
  \node at (1.75,1.7) {\textcolor{white}{\bf \blackwhitecircled{\small \textcolor{white}{1}}{1} }};

  \node at (0,-6.1) {\includegraphics[width=1.39in]{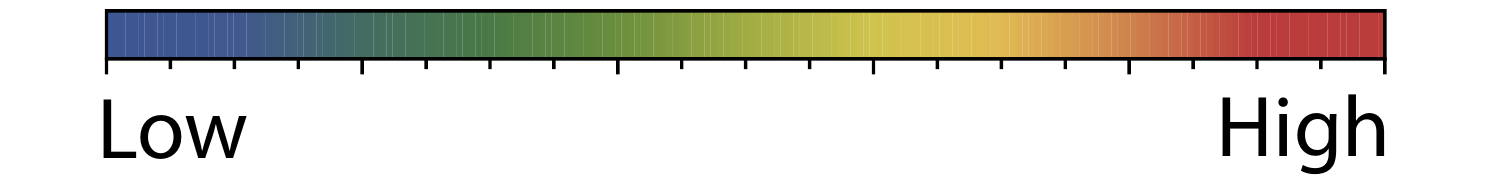}};
  \node at (0,-6.475) {$Q(s,a)$};

\end{tikzpicture}
}
   \caption[]{(middle) Depiction of the state-space organization for \emph{Millipede}.  Here, we randomly selected ten thousand visited states, across ten random runs, and projected them via UMAP.  Each state is color coded according to it's maximum learned $Q$-value across all available actions.  The plot shows that the states are roughly divided into two groups, those that have moderate to high expected costs and those with low expected costs.  (bottom left) Call-outs one through four correspond to the former group.  The first call-out, for instance, corresponds to the agent just having been hit by an enemy and losing a life.  The remaining call-outs correspond to states where the agent cannot readily lower costs greatly.  (bottom right) Call-outs five through eight correspond to the latter group.  For example, the eighth call-out shows that the agent is about to receive a large reduction in costs due to the DDT canister being active and taking out several millipede segments.  Each call-out is spatially referenced to the state space plot in the middle.  (top left) The average next-state transition surprise after training has concluded.  The plot shows that the agent has sufficiently explored much of the space and understands the transition dynamics well.\vspace{-0.2cm}}
   \label{fig:millipedestatespace}
\end{figure*}

\begin{figure*}

\hspace{-0.295cm}\scalebox{0.775}{
\begin{tikzpicture}
  \node[] at (-8.2,5.25) {\phantom{\includegraphics[width=1.85in]{millipede-statevisitation-1.pdf}}};
  \node[] at (0.1,0.75) {\includegraphics[width=5.35in]{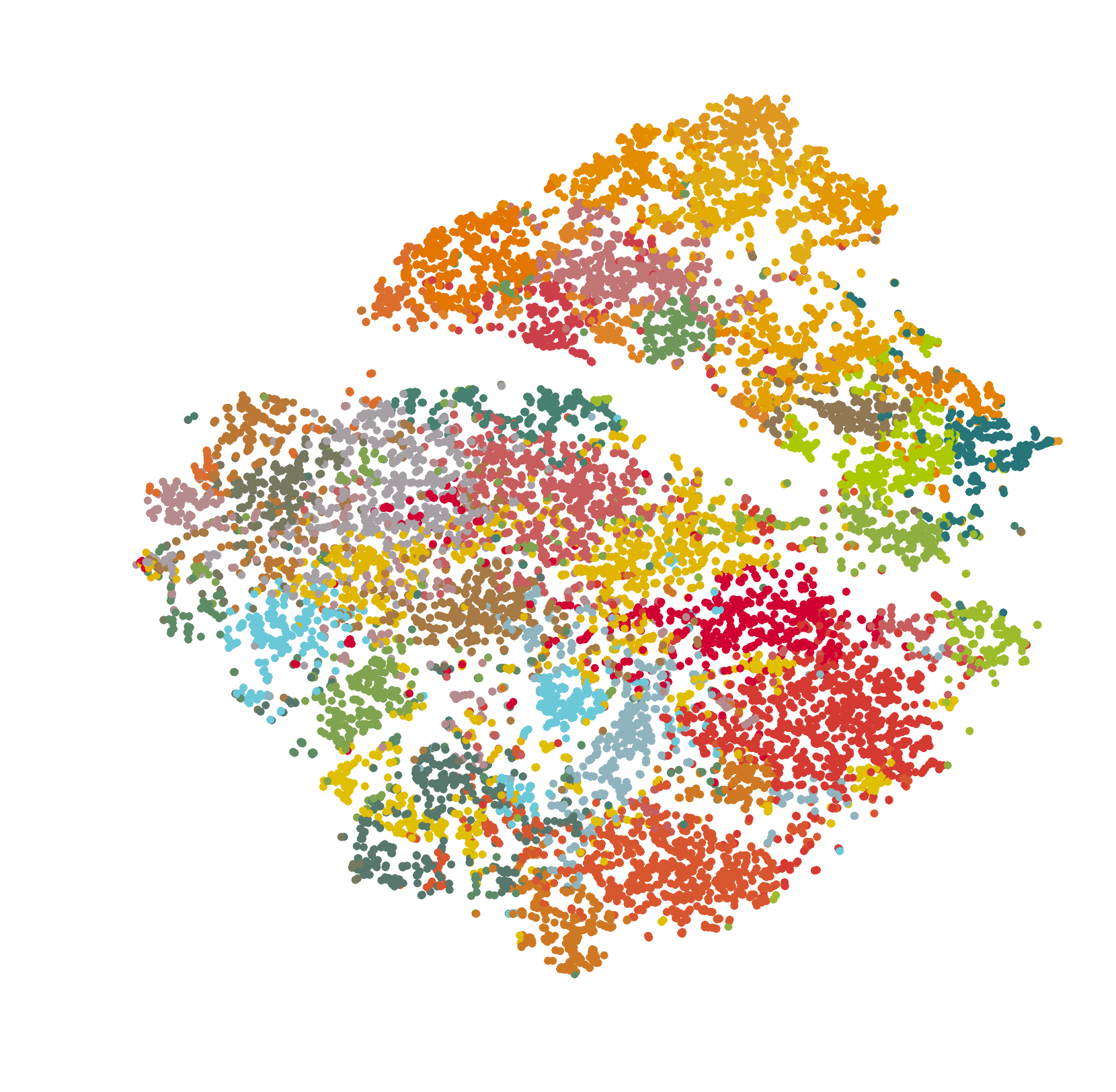}};

  \node at (-9,-2.65) {\framebox{\includegraphics[width=1.175in]{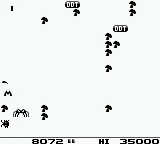}}};
  \node at (-5.825,-2.65) {\framebox{\includegraphics[width=1.175in]{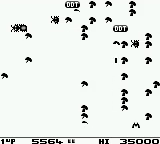}}};

  \node at (-9,-5.5) {\framebox{\includegraphics[width=1.175in]{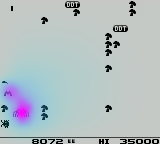}}};
  \node at (-5.825,-5.5) {\framebox{\includegraphics[width=1.175in]{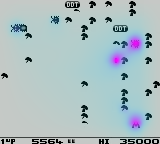}}};

  \node at (8.75,-2.65) {\framebox{\includegraphics[width=1.175in]{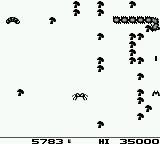}}};
  \node at (5.575,-2.65) {\framebox{\includegraphics[width=1.175in]{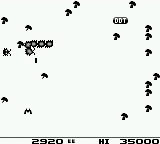}}};

  \node at (8.75,-5.5) {\framebox{\includegraphics[width=1.175in]{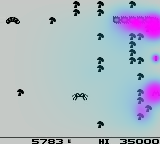}}};
  \node at (5.575,-5.5) {\framebox{\includegraphics[width=1.175in]{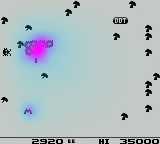}}};

  \node at (-9.0,-0.95) {\includegraphics[width=1.39in]{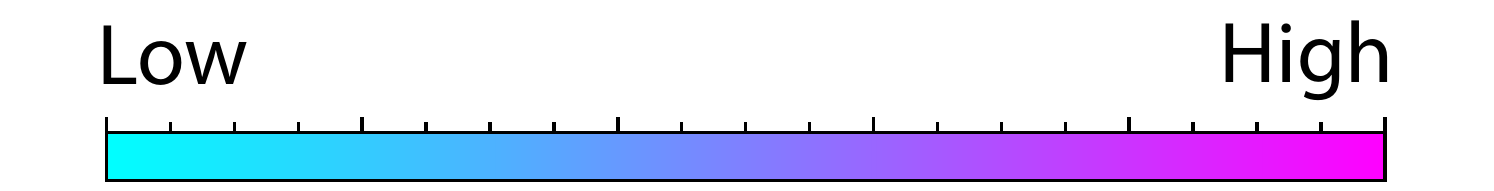}};

  \node at (-9.0,-0.5) {$\|\partial Q(s_t,a_t)/\partial s_{t,ij}\|_1$};

  \node at (-0.05,-6.375) {\includegraphics[height=0.475in]{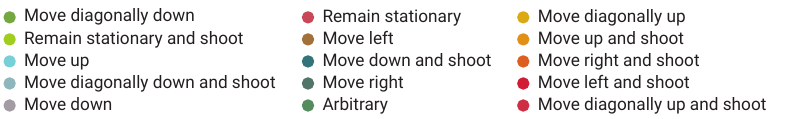}};

  \newcommand*\blackcircled[2]{\tikz[baseline=(char.base)]{
  \node[circle,black,draw,scale=#2,inner sep=1pt] (char) {#1};}}
  \newcommand*\redcircled[2]{\tikz[baseline=(char.base)]{
  \node[circle,red,draw,scale=#2,inner sep=1pt] (char) {#1};}}

  \node at (5.05,6.25) {\textcolor{black}{\bf\Large $\mathcal{S}$}};

  \node at (2.25,0.5) {\textcolor{black}{\bf \blackwhitecircled{\small \textcolor{white}{1}}{1} }};
  \node at (3.25,0.25) {\textcolor{black}{\bf \blackwhitecircled{\small \textcolor{white}{2}}{1} }};
  \node at (4.25,1.65) {\textcolor{black}{\bf \blackwhitecircled{\small \textcolor{white}{3}}{1} }};
  \node at (4.85,2.05) {\textcolor{black}{\bf \blackwhitecircled{\small \textcolor{white}{4}}{1} }};

  \node at (-9,-7.215) {{\textcolor{black}{\bf \blackcircled{\small \textcolor{black}{1}}{1}}}};
  \node at (-5.825,-7.215) {{\textcolor{black}{\bf \blackcircled{\small \textcolor{black}{2}}{1}}}};
  \node at (5.575,-7.215) {{\textcolor{black}{\bf \blackcircled{\small \textcolor{black}{3}}{1}}}};
  \node at (8.75,-7.215) {{\textcolor{black}{\bf \blackcircled{\small \textcolor{black}{4}}{1}}}};

\end{tikzpicture}
}
   \caption[]{(middle) Depiction of the state-action-space organization for \emph{Millipede}.  Here, we use the same states as in \cref{fig:millipedestatespace}.  Each state is color coded according to the dominant action chosen after learning concluded.  The plot shows that contiguous groups of action clusters emerge for scenarios with related $Q$-values.  (bottom left) Call-outs one and two correspond to cases where the agent cannot readily lower its costs.  In the first call-out, for example, the enemies are located behind the agent.  The agent's best course of action is to move to the right and down so that it can begin to target some of the enemies.  (bottom right) Call-outs three and four correspond to cases where the agent can achieve moderate cost reductions.  In both situations, the agent's best option is to remain stationary and shoot, as it will eventually destroy all of the millipede segments.  Due to the similarity of states depicted in the call-outs, they naturally cluster together in the UMAP embedding.  For each call-out, we provide feature gradient maps that illustrate what features the agent uses to make its decision.  The maps show that the agent fixates on local features that are relevant over the next time step and subsequent ones for a short-term horizon.\vspace{-0.2cm}}
   \label{fig:millipedestateactionspace}
\end{figure*}

\begin{figure*}

\hspace{-0.125cm}\scalebox{0.775}{
\begin{tikzpicture}
  \node[] at (0.0,1.05) {\includegraphics[width=5.15in]{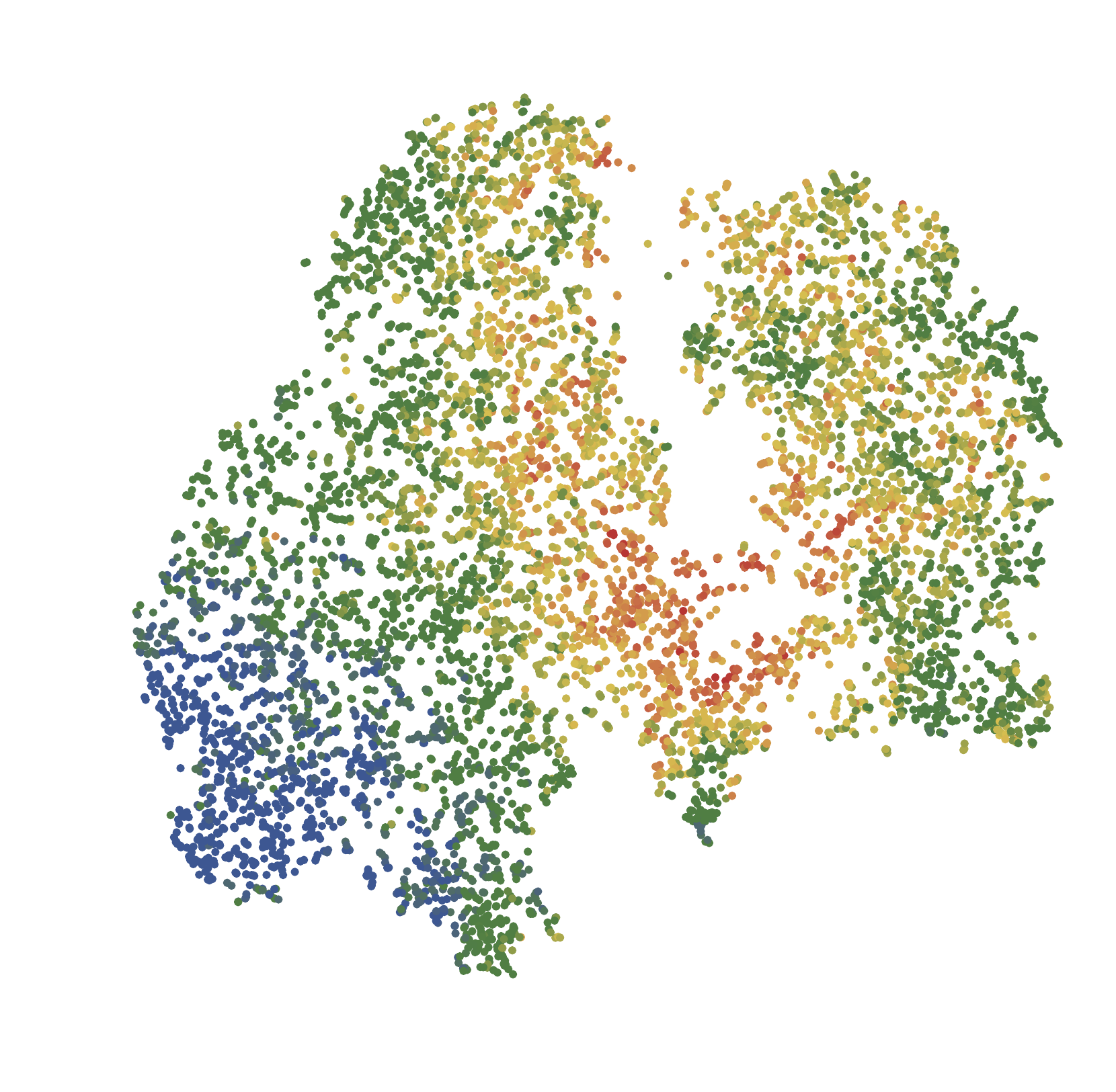}};
  \node[] at (-8.2,5.25) {\includegraphics[width=1.75in]{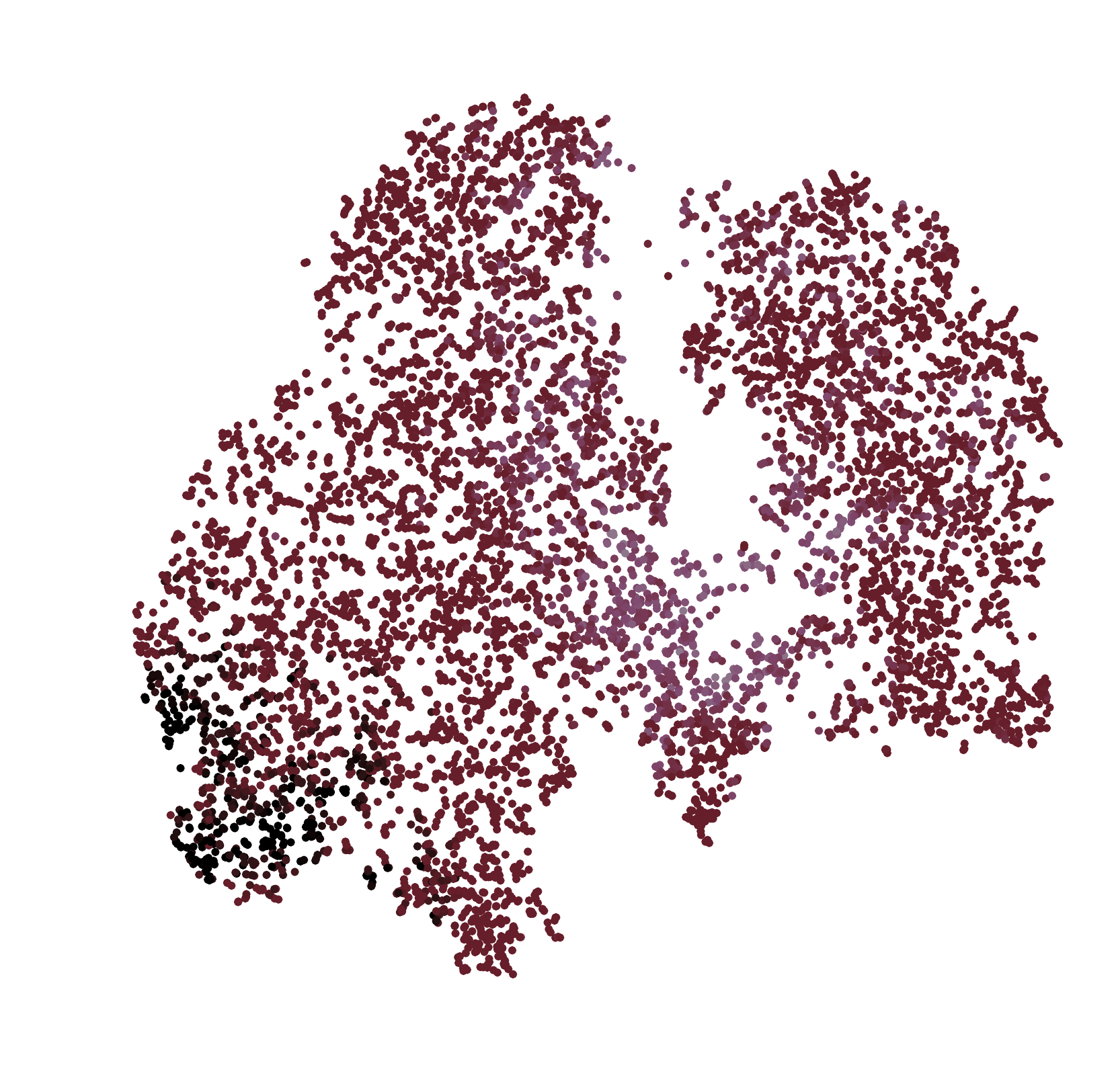}};

  \node at (-8.2,2.65) {\includegraphics[width=1.39in]{plumbar.pdf}};
  \node at (-8.2,2.215) {$H\!(p(s_{t+1}|s_t,a_t))$};

  \setlength{\fboxrule}{0.5pt}
  \setlength{\fboxsep}{0.025pt}
  \node at (-9,-5.5) {\framebox{\includegraphics[width=1.175in]{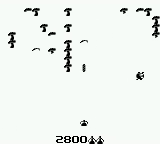}}};
  \node at (-5.825,-5.5) {\framebox{\includegraphics[width=1.175in]{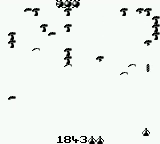}}};
  \node at (-9,-2.65) {\framebox{\includegraphics[width=1.175in]{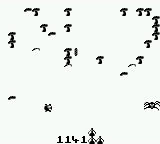}}};
  \node at (-9,0.2) {\framebox{\includegraphics[width=1.175in]{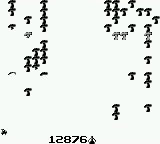}}};

  \node at (8.75,-5.5) {\framebox{\includegraphics[width=1.175in]{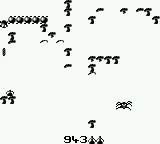}}};
  \node at (5.575,-5.5) {\framebox{\includegraphics[width=1.175in]{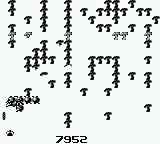}}};
  \node at (8.75,-2.65) {\framebox{\includegraphics[width=1.175in]{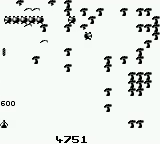}}};
  \node at (8.75,0.2) {\framebox{\includegraphics[width=1.175in]{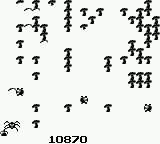}}};

  \newcommand*\blackcircled[2]{\tikz[baseline=(char.base)]{
  \node[circle,black,draw,scale=#2,inner sep=1pt] (char) {#1};}}
  \newcommand*\redcircled[2]{\tikz[baseline=(char.base)]{
  \node[circle,red,draw,scale=#2,inner sep=1pt] (char) {#1};}}

  \node at (5.05,6.25) {\textcolor{black}{\bf\Large $\mathcal{S}$}};

  \node at (-7.05,0.2) {\textcolor{black}{\bf \blackcircled{\small \textcolor{black}{1}}{1} }};
  \node at (-7.05,-2.65) {\textcolor{black}{\bf \blackcircled{\small \textcolor{black}{2}}{1} }};
  \node at (-8.95,-7.215) {\textcolor{black}{\bf \blackcircled{\small \textcolor{black}{3}}{1} }};
  \node at (-5.775,-7.215) {\textcolor{black}{\bf \blackcircled{\small \textcolor{black}{4}}{1} }};

  \node at (6.925,0.2) {\textcolor{black}{\bf \blackcircled{\small \textcolor{black}{5}}{1} }};
  \node at (6.925,-2.65) {\textcolor{black}{\bf \blackcircled{\small \textcolor{black}{6}}{1} }};
  \node at (8.8,-7.215) {\textcolor{black}{\bf \blackcircled{\small \textcolor{black}{7}}{1} }};
  \node at (5.65,-7.215) {\textcolor{black}{\bf \blackcircled{\small \textcolor{black}{8}}{1} }};

  \node at (0.45,-0.95) {\textcolor{black}{\bf \blackwhitecircled{\small \textcolor{white}{8}}{1} }};
  \node at (-1.95,1.25) {\textcolor{black}{\bf \blackwhitecircled{\small \textcolor{white}{7}}{1} }};
  \node at (1.7,-2.25) {\textcolor{black}{\bf \blackwhitecircled{\small \textcolor{white}{6}}{1} }};
  \node at (1.3,-2.7) {\textcolor{black}{\bf \blackwhitecircled{\small \textcolor{white}{5}}{1} }};
  \node at (-3.05,0.75) {\textcolor{black}{\bf \blackwhitecircled{\small \textcolor{white}{4}}{1} }};
  \node at (-2.75,0.25) {\textcolor{black}{\bf \blackwhitecircled{\small \textcolor{white}{3}}{1} }};
  \node at (-3.25,-0.5) {\textcolor{white}{\bf \blackwhitecircled{\small \textcolor{white}{2}}{1} }};
  \node at (-4,-3.0) {\textcolor{white}{\bf \blackwhitecircled{\small \textcolor{white}{1}}{1} }};

  \node at (0,-5.6) {\includegraphics[width=1.39in]{darkrainbowbar.pdf}};
  \node at (0,-5.975) {$Q(s,a)$};
\end{tikzpicture}
}
   \caption[]{(middle) Depiction of the state-space organization for \emph{Centipede}.  Here, we randomly selected ten thousand visited states, across ten random runs, and projected them via UMAP.  Each state is color coded according to it's maximum learned $Q$-value across all available actions.  The plot shows that the states are roughly divided into two groups, those that have moderate to high expected costs and those with low expected costs.  (bottom left) Call-outs one through four correspond to the former group.  The first call-out, for instance, corresponds to the agent just having been hit by an enemy and losing a life.  The remaining call-outs correspond to states where the agent cannot readily lower costs greatly.  (bottom right) Call-outs five through eight correspond to the latter group.  For example, the eighth call-out shows that the agent is about to receive a large reduction in costs due to the DDT canister being active and taking out several millipede segments.  Each call-out is spatially referenced to the state space plot in the middle.  (top left) The average next-state transition surprise after training has concluded.  The plot shows that the agent has sufficiently explored much of the space and understands the transition dynamics well.\vspace{-0.2cm}}
   \label{fig:centipedestatespace}
\end{figure*}

\begin{figure*}
\hspace{-0.325cm}\scalebox{0.775}{
\begin{tikzpicture}
  \node[] at (-8.2,5.25) {\phantom{\includegraphics[width=1.75in]{centipede-statevisitation-1.pdf}}};
  \node[] at (0.0,1.05) {\includegraphics[width=5.15in]{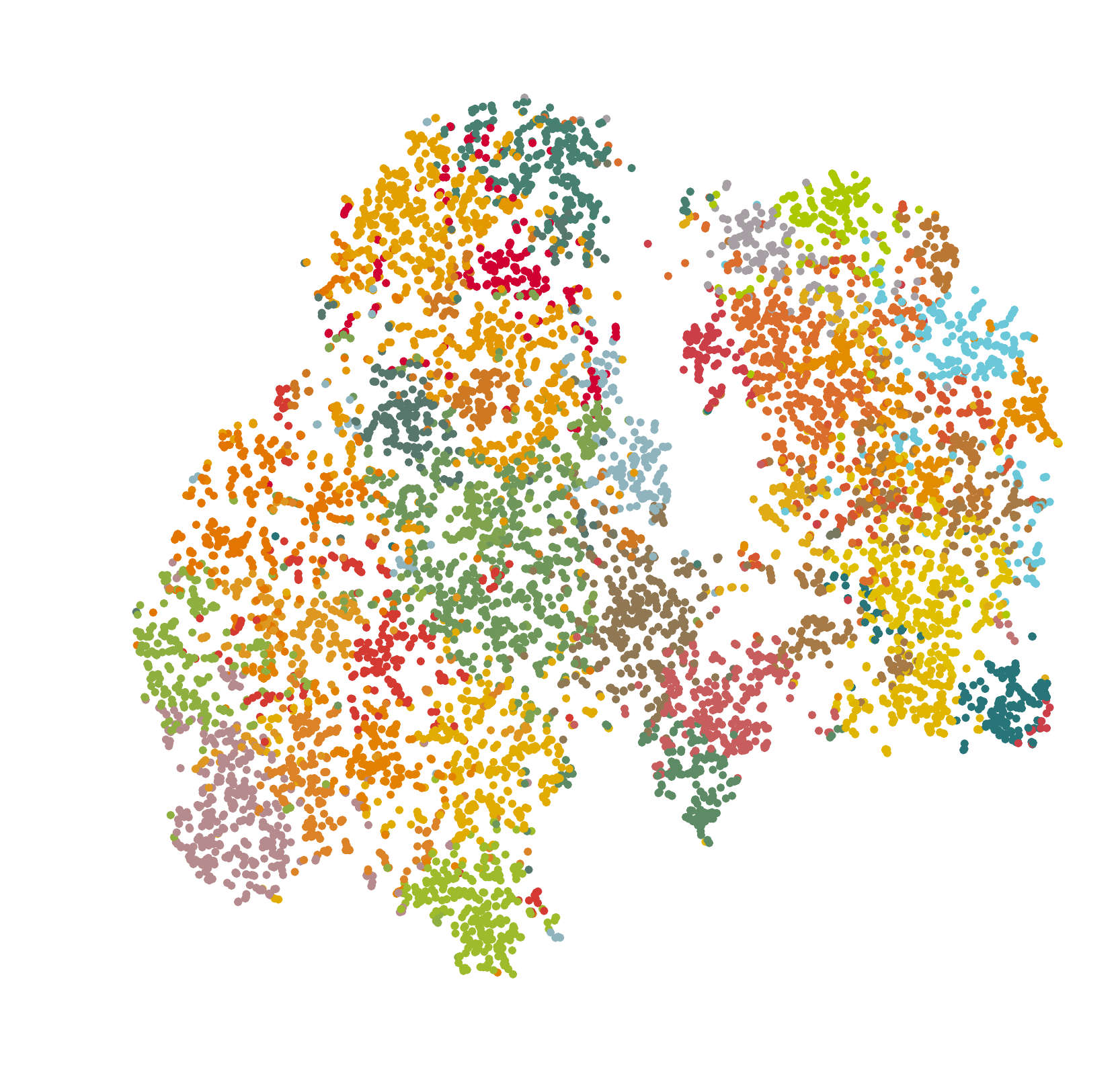}};

  \node at (-9,-2.65) {\framebox{\includegraphics[width=1.175in]{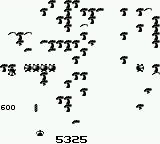}}};
  \node at (-5.825,-2.65) {\framebox{\includegraphics[width=1.175in]{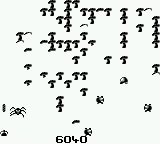}}};

  \node at (-9,-5.5) {\framebox{\includegraphics[width=1.175in]{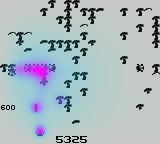}}};
  \node at (-5.825,-5.5) {\framebox{\includegraphics[width=1.175in]{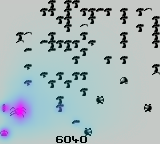}}};

  \node at (5.575,-2.65) {\framebox{\includegraphics[width=1.175in]{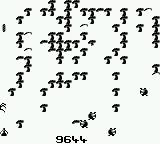}}};
  \node at (8.75,-2.65) {\framebox{\includegraphics[width=1.175in]{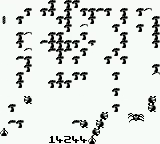}}};

  \node at (5.575,-5.5) {\framebox{\includegraphics[width=1.175in]{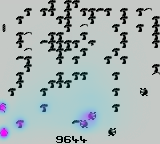}}};
  \node at (8.75,-5.5) {\framebox{\includegraphics[width=1.175in]{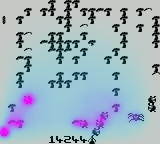}}};

  \node at (-9.0,-0.95) {\includegraphics[width=1.39in]{coolbar.pdf}};

  \node at (-9.0,-0.5) {$\|\partial Q(s_t,a_t)/\partial s_{t,ij}\|_1$};

  \node at (-0.05,-6.375) {\includegraphics[height=0.475in]{stateactionclusters-2.pdf}};

  \newcommand*\blackcircled[2]{\tikz[baseline=(char.base)]{
  \node[circle,black,draw,scale=#2,inner sep=1pt] (char) {#1};}}
  \newcommand*\redcircled[2]{\tikz[baseline=(char.base)]{
  \node[circle,red,draw,scale=#2,inner sep=1pt] (char) {#1};}}

  \node at (5.05,6.25) {\textcolor{black}{\bf\Large $\mathcal{S}$}};

  \node at (-0.8,-1.4) {\textcolor{black}{\bf \blackwhitecircled{\small \textcolor{white}{1}}{1} }};
  \node at (-0.85,-0.65) {\textcolor{black}{\bf \blackwhitecircled{\small \textcolor{white}{2}}{1} }};
  \node at (-0.5,-1.5) {\textcolor{black}{\bf \blackwhitecircled{\small \textcolor{white}{3}}{1} }};
  \node at (-1.25,-1.05) {\textcolor{black}{\bf \blackwhitecircled{\small \textcolor{white}{4}}{1} }};

  \node at (-9,-7.215) {{\textcolor{black}{\bf \blackcircled{\small \textcolor{black}{1}}{1}}}};
  \node at (-5.825,-7.215) {{\textcolor{black}{\bf \blackcircled{\small \textcolor{black}{2}}{1}}}};
  \node at (5.575,-7.215) {{\textcolor{black}{\bf \blackcircled{\small \textcolor{black}{3}}{1}}}};
  \node at (8.75,-7.215) {{\textcolor{black}{\bf \blackcircled{\small \textcolor{black}{4}}{1}}}};

\end{tikzpicture}
}
   \caption[]{(middle) Depiction of the state-action-space organization for \emph{Centipede}.  Here, we use the same states as in \cref{fig:millipedestatespace}.  Each state is color coded according to the dominant action chosen after learning concluded.  The plot shows that contiguous groups of action clusters emerge for scenarios with related $Q$-values.  (bottom left) Call-outs one and two correspond to cases where the agent cannot readily lower its costs.  In the first call-out, for example, the enemies are located behind the agent.  The agent's best course of action is to move to the right and down so that it can begin to target some of the enemies.  (bottom right) Call-outs three and four correspond to cases where the agent can achieve moderate cost reductions.  In both situations, the agent's best option is to remain stationary and shoot, as it will eventually destroy all of the millipede segments.  Due to the similarity of states depicted in the call-outs, they naturally cluster together in the UMAP embedding.  For each call-out, we provide feature gradient maps that illustrate what features the agent uses to make its decision.  The maps show that the agent fixates on local features that are relevant over the next time step and subsequent ones for a short-term horizon.\vspace{-0.2cm}}
   \label{fig:centipedestateactionspace}
\end{figure*}

This latter characterization of states is appropriate for both \emph{Centipede} and \emph{Millipede}.  The features are tied to the games' objectives and hence the reward structures.  Due to how we compute the grid-occupancy features, their interpretability remains the same throughout the entire learning process.  All of these traits aid in efficiently discerning good agent behaviors.

There are additional practical appeals to using such a feature representation.  Foremost, it is straightforward to specify.  This is because, in both games, the grid size and shape that defines the environment remains constant.  Only the objects and their locations within the grid change over time.  Tabular policies with a finite state count can thus be considered.  Secondly, the features can be reliably discerned in real time using simple correlation-based template recognition.  This stems from the fact that the appearances of the agent, enemies, and objects also do not change greatly.  There are also few sprite animations in both games.

\phantomsection\label{secB.3}
%%%%%%%%%%%%%%%%%%%%%%%%%%%%%%%%%%%%%%%%%%%%%%%%%%%%%%%%%%%%%%%%%%%%%%%%%%%
%%%%%%%%%%%%%%%%%%%%%%%%%%%%%%%%%%%%%%%%%%%%%%%%%%%%%%%%%%%%%%%%%%%%%%%%%%%
\subsection*{\small{\sf{\textbf{B.3.$\;\;\;$Simulation Supplement}}}}

We illustrate that context-specific groupings arise when learning using the value of information.

Similar to the results presented in \cite{ZahavyT-conf2016a}, there appears to be a hierarchical, spatio-temporal aggregation of the state space when using value-of-information-based exploration.  In both games, the agent begins in a low-cost state.  It is rarely in a position to immediately score points and thus must navigate in the environment to align with an enemy and fire bolts.  This is shown, for instance, in the second and third call-outs in \cref{fig:millipedestatespace} and \cref{fig:centipedestatespace}.  As the agent progresses through the early parts of the games, it predominantly shifts between low- and moderate-cost states.  The former are visited whenever the agent has no ability to score, such as when it must move from one side of the environment to the other to target an enemy or when enemies are blocked by mushrooms.  The latter case is illustrated in the third call-out in \cref{fig:millipedestatespace}.  These states also correspond to whenever the agent is clearing blocks of mushrooms, as in the third and forth call-outs in \cref{fig:centipedestatespace}.  Moderate-cost states are visited when the agent can target one or more common enemies, like spiders, in quick succession.  Once the agent has cleared a few levels, its opportunities for scoring greatly improve.  Rare enemies begin to appear in these levels.  Common enemies also spawn more rapidly.  The agent thus spends more time in moderate- to high-cost states.  Examples of these states are depicted in the fifth through eighth call-outs of \cref{fig:millipedestatespace} and the seventh and eighth call-outs in \cref{fig:centipedestatespace}.  Eventually, though, the agent is overwhelmed.  Sometimes, it cannot clear mushrooms quickly enough, leaving it vulnerable to waves of quickly-moving enemies.  Other times, enemies spawn at the fringes of the environment and the agent has little time to dodge them.  It thus always moves to a low-cost death state.  Examples are given in the first call-out of \cref{fig:millipedestatespace} and \cref{fig:centipedestatespace}.

Alongside the state-space aggregation is one of the action space.  By the end of training, there are fifteen action groups that emerge, as shown in \cref{fig:millipedestateactionspace} and \cref{fig:centipedestateactionspace}, which are spread across some thirty well-defined clusters for each game.  All of these groups is usually well correlated with cost.  Not surprisingly, clusters associated with firing bolts correspond either to states or near states with large costs.  Sometimes, however, bolt firing groups correspond with moderate-cost states, since there is a delay for a bolt to strike an enemy or environmental object, like a mushroom.   Those action groups related to movement have varying degrees of association with low- and moderate-cost states.  For example, certain types of movement, like left or right, may coincide with moderate-cost states, since the agent achieves alignment with an enemy.  Alternatively, the agent may move away from an enemy, thus allowing it to avoid being hit and playing the game.  Other movement directions, like up and down, are typically associated with low-cost states.  Unless the agent has happened to score, due to a bolt hit, then there is typically no cost reduction for such actions.  There does not appear to be a strong correlation between states associated with agent death and actions, however.  Any type of action can feasibly be executed as the agent is struck and dies.  Based on the transition-uncertainty plots in \cref{fig:millipedestatespace} and \cref{fig:centipedestatespace}, we can be relatively assured that the chosen action groups are stable for these simulations.  The highest transition entropy appears to be moderately low, suggesting that the agent understands well the environment dynamics and hence what it should do to consistently attain low costs. 

The action groups in \cref{fig:millipedestateactionspace} and \cref{fig:centipedestateactionspace} depict decision making at the macro scale.  At the local scale, there are multiple factors that influence the agent's action choices and hence the observed action aggregation.  Examples of the factors, which are illustrated by inferred saliency maps, are provided at the bottom of the figures.  These saliency maps highlight that, at least for value-of-information-based searches, the positions of the agents and nearest enemies are paramount for decision making.  The type of enemy also dictates how the agent will respond.  Rare enemies hence precedence over common ones, unless the agent is either threatened or currently engaged with an enemy.  If the agent is threatened by nearby enemies, then spiders are often targeted more readily than centipede segments.  The former move more quickly and less predictably and hence have a greater chance of colliding with the agent.  Earwigs are also prioritized over many other enemy types, since hitting them can cause them to accelerate quickly toward the agent.  The location of fired bolts is of additional importance.  It, alongside other features, determines whether the agent can move on to another objective or must continue pursuing its current one.  Surprisingly, nearby objects, like mushrooms, often do not influence the action choice, and hence macro-level behaviors.  It would appear that the agent mainly favors targeting enemies and that clearing mushrooms is a byproduct of that.  It is only when the environment is littered with mushrooms that the agent begins to destroy them frequently after dispatching enemies.  Alternatively, stray bolts naturally remove them from the environment.

The other search strategies that we consider \hyperref[sec5]{Section 5} do not consistently yield an easily interpretable aggregation.  They often explore too ineffectively to uncover a near-optimal estimate of the value function.  The action clusters are hence more diffuse and mixed.  The local features used for decision making are also much less coherent.

\setstretch{0.95}\fontsize{9.75}{10}\selectfont
\putbib
\end{bibunit}

\begin{bibunit}
\bstctlcite{IEEEexample:BSTcontrol}

\RaggedRight\parindent=1.5em
\fontdimen2\font=2.1pt\selectfont
\singlespacing
\allowdisplaybreaks

\clearpage\newpage
\setcounter{figure}{0}
\setcounter{equation}{0}
\renewcommand{\thefigure}{C.\arabic{figure}}
\renewcommand\theequation{C.\arabic{equation}}
\phantomsection\label{secC}
%%%%%%%%%%%%%%%%%%%%%%%%%%%%%%%%%%%%%%%%%%%%%%%%%%%%%%%%%%%%%%%%%%%%%%%%%%%
%%%%%%%%%%%%%%%%%%%%%%%%%%%%%%%%%%%%%%%%%%%%%%%%%%%%%%%%%%%%%%%%%%%%%%%%%%%
\subsection*{\small{\sf{\textbf{Appendix C}}}}

In this appendix, we provide additional experimental results to motivate our approach for adapting the exploration rate of the value of information. 

We begin by specifying a double-deep-$Q$ network for mapping game frames into action-value magnitudes and expected environment costs (see \hyperref[secC.1]{Appendix C.1}).  We then significantly augment the capabilities of this network, since the games that we consider are rather complex and agent skill acquisition can be slow.  We integrate tree-based searches, with fast roll-outs, to evaluate potential action strings and how they enable the agent to complete various objectives.  We also use uncertainty-based searches to force the agent into under-investigated regions of the state-action space.  The overall exploration process is guided by the value of information with pseudo-arc-length path-following.

We then present experimental results on over forty game environments (see \hyperref[secC.2]{Appendix C.2}).  We show that our network and exploration mechanism outperforms state-of-the-art alternatives on each game.  It also completes games more effectively than human players in many instances.

\phantomsection\label{secC.1}
%%%%%%%%%%%%%%%%%%%%%%%%%%%%%%%%%%%%%%%%%%%%%%%%%%%%%%%%%%%%%%%%%%%%%%%%%%%
%%%%%%%%%%%%%%%%%%%%%%%%%%%%%%%%%%%%%%%%%%%%%%%%%%%%%%%%%%%%%%%%%%%%%%%%%%%
\subsection*{\small{\sf{\textbf{C.1.$\;\;\;$Simulation Preliminaries}}}}

%%%%%%%%%%%%%%%%%%%%%%%%%%%%%%%%%%%%%%%%%%%%%%%%%%%%%%%%%%%%%%%%%%%%%%%%%%%
%%%%%%%%%%%%%%%%%%%%%%%%%%%%%%%%%%%%%%%%%%%%%%%%%%%%%%%%%%%%%%%%%%%%%%%%%%%
\subsection*{\small{\sf{\textbf{C.1.1.$\;\;\;$Deep Value-of-Information Search}}}}

For the games that we used in \hyperref[sec5]{Section 5}, the environments were simplistic enough to permit manually defining the state-action space.  This is not the case for the games that we consider in this appendix.  The environments here are typically much more visually rich, which prohibits easily specifying and extracting gameplay features.

Here, we use double-deep-$Q$ networks \cite{vanHasseltH-conf2016a} with prioritized experience replay \cite{SchaulT-conf2016a,HorganD-conf2018a} to implicitly uncover game-specific state features from images.  Such networks utilize continuous-valued features to regress action-value magnitudes and infer discrete action choices for each game frame.

Due to the large number of the environments that we consider, we are unable to manually provide dense reward signals.  We often rely on game-supplied scores, which can be sparse and delayed.  These scores can also be deceptive, in the sense that they do not necessarily reflect the agent's true progress.  They therefore do not always provide enough supervision that enable the agents to complete multifaceted objectives.  Even our game-tailored metrics can have these flaws.  Deep-$Q$-based approaches can thus stall early during learning, including the version that we use.

\begin{figure*}
   \begin{adjustbox}{addcode={\begin{minipage}{\width}}{\caption{%
      A visual overview of deep, curiosity-based reinforcement learning with the value of information.  At each time step, the current game state is fed into a value-of-information-trained deep network that assesses a potential best action and estimates its action-state-value-function magnitude.  This game state, $s_t$, is also used as a local root node that represents the starting point for a Monte Carlo tree search.  Each simulation for this search traverses the edge with the best action value, $V(s_t,a_{t+m})$, plus some augmentation term, $u(p(s_t,a_{t+m}))$, $m \!\geq\! 0$, that depends on some prior probability for that edge.  A corresponding action, in this case, button presses, is used experience a transition to a new state, $s_{t+q}$, $q \!>\! 0$, which becomes a leaf node of the tree.  If this leaf node is not a terminal state, then it may be expanded.  The new node is processed once by the deep network and the output probabilities are stored as priors for each action.  At the end of the simulation, the leaf node is evaluated in one of two ways.  The first is by using the deep network, which supplies an action-state magnitude, $Q_\theta(s_{t+q},a_{t+m})$.  The second is via a fast roll-out policy network, where a winning action sequence is chosen using a function $r_{\theta'}(s_{t+q},a_{t+m})$.  The action values are then updated to track the average value of all evaluations in the explored sub-trees.  After this back-propagation occurs, the best-performing action for the current game state is additionally evaluated from the context of how well it improves the agent's understanding of the state transition dynamics.  This yields a final action response.
      }\end{minipage}},rotate=90,center}
      \includegraphics[width=9.65in]{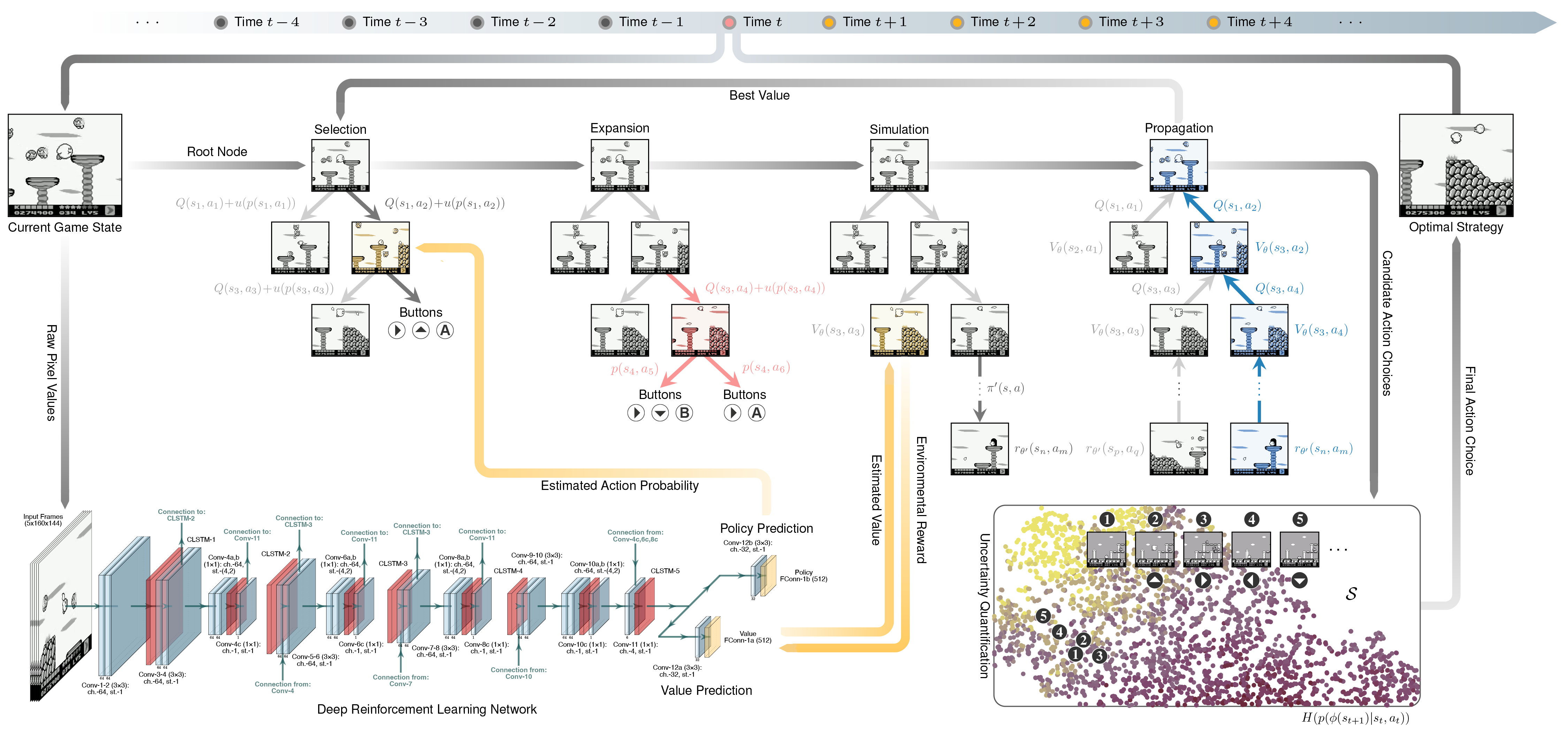}
   \end{adjustbox}
   \label{fig:drlmcts}
\end{figure*}

Somewhat analogous to expert iteration \cite{AnthonyT-coll2017a}, we consider a simultaneous, on-policy investigation of the state space to overcome stalling.  This process is illustrated in \hyperref[fig:drlmcts]{figure C.1}.  Much like Monte Carlo tree search \cite{GuoX-coll2014a,SilverD-jour2016a,SilverD-jour2017a}, each game frame becomes a base node of an ever-expanding $k$-ary tree.  Possible actions are chosen for this base node, yielding leaf nodes.  The simulation then moves to the branch with the best action value plus a bonus that depends on a stored probability for that edge.  Each new node on the branch is then processed by the double-deep-$Q$ network.  At the end of each simulation, the leaf node is evaluated in one two ways.  The first is by the deep network.  The second is via a fast, roll-out policy network, which chooses actions until some termination condition is met.  The tentative winning action for the base node is then selected using a game-specific cost function.  Finally, the state-action values are back-propagated to track the mean value of all evaluations in the sub-tree below that action.

Even with tree search, large parts of the state-action space may go uninvestigated.  Poor agent behaviors may be encountered in rare, but important, situations, stymieing progress.  We force the agent to explore such regions via an uncertainty-based constraint \cite{ChentanezN-coll2005a,BellemareM-coll2016a}.  That is, we implement a lightweight convolutional autoencoder, trained on game frames, which is fed into another deep network that approximates the game's transition function by predicting the next frame for the current action.  Whenever the next state is not properly predicted by this network, around some region about the true state, we add that transition to the experience replay buffer and impose that the associated action be taken.  We also weight that action's importance more heavily during value-of-information search to ensure that it will likely be chosen.  Here, we measure prediction accuracy using our matrix-based \cite{GiraldoLGS-jour2014a,SledgeIJ-jour2022a} cross-entropy-to-go criterion \cite{SledgeIJ-jour2022b}.  This criterion promotes minimax-optimal convergence, in a dimensionally agnostic way, so it is well suited for comparing empirical state transitions for high-dimensional observations.

Both the tree search and uncertainty-based searches supply principled guesses as to the action that should be taken.  We aggregate the scores and treat them as a modified action-state value-function for value-of-information exploration.  Pseudo-arc-length path-following is employed to automatically tune the exploration rate.

%%%%%%%%%%%%%%%%%%%%%%%%%%%%%%%%%%%%%%%%%%%%%%%%%%%%%%%%%%%%%%%%%%%%%%%%%%%
%%%%%%%%%%%%%%%%%%%%%%%%%%%%%%%%%%%%%%%%%%%%%%%%%%%%%%%%%%%%%%%%%%%%%%%%%%%
\subsection*{\small{\sf{\textbf{C.1.2.$\;\;\;$Network Architectures}}}}

The above training process leverages dual networks.  The first is a fast roll-out architecture for action selection during the tree searches.  The second is a feature backbone for deep-$Q$-based action selection.  

For the former, we use three convolutional layers.  The input to the first layer is a 160$\times$144-pixel grayscale image from the GameBoy emulator.  It is acted on by 64 filters that have a stride of 4.  The next two layers have strides of 2 and 1, respectively, with the same number of filters.  The receptive fields are of sizes 8$\times$8, 4$\times$4, and 3$\times$3.  Feature maps are appropriately mirror-padded where necessary.  Rectified-linear activation functions are applied throughout.  After the third layer, we cascade a convolutional-LSTM cell that has 64 filters each with a receptive field of 3$\times$3.  The recurrent length is 30 game frames, which corresponds to about half a second of real game-time for a GameBoy running at the default clock rate.  Gradient clipping is used for the LSTM cells to ensure learning stability and accelerate training \cite{ZhangJ-conf2020a}.

The feature backbone that we use for the double-deep-$Q$ network differs from convention.  We consider five blocks of two convolutional layers each with varying stride amounts.  The first layer uses 5$\times$5 filters with strides of 2, while the second through fifth layers rely on 3$\times$3 convolutions with unit strides.  The number of filters for each layer is fixed to 128 along the main feature path.  Rectified-linear activation functions are applied throughout.  Bi-directional convolutional-LSTM cells are added at the beginning of the second through fifth blocks to mix feature content across time.  These have 64 filters.  The receptive field size is consistent with the other convolutional layers in each block.  The LSTM cells have a frame length of 30.  After the second block, the backbone extracts multi-scale features using a combination of dilated convolutions \cite{ChenLC-jour2018a} and bi-directional convolutional-LSTM cells.  We use 3$\times$3 kernels with dilation rates of 2 and unit strides.  The filter sizes are the same for the convolutional-LSTM layers.  Both layer types use 64 filters each.  The outputs of the various multi-scale blocks are aggregated and flattened in dual fully-connected layers with 256 processing elements each.

%%%%%%%%%%%%%%%%%%%%%%%%%%%%%%%%%%%%%%%%%%%%%%%%%%%%%%%%%%%%%%%%%%%%%%%%%%%
%%%%%%%%%%%%%%%%%%%%%%%%%%%%%%%%%%%%%%%%%%%%%%%%%%%%%%%%%%%%%%%%%%%%%%%%%%%
\subsection*{\small{\sf{\textbf{C.1.3.$\;\;\;$Learning Protocols}}}}

Both of our networks have several parameters.  In each case, the discount factor is set to 0.99.  The learning rate is 0.001 and decreases exponentially to 0.00003 across 50000 episodes.  Each episode is anywhere from 50 to 2500 steps.  The number of steps between the target network updates is 5000.  The network relies on mini-batch sizes of 32, which helps preempt terminating at local optima \cite{HardtM-conf2016a}.  ADAM, with the default parameters, is used for training \cite{KingmaDP-conf2015a}.  Nearly identical parameter values are employed for the alternate approaches that we evaluate.  For some approaches, though, we use RMSProp \cite{MukkamalaMC-conf2017a} to be consistent with the recommendations of the authors.

Our double-deep-$Q$ network relies on prioritized experience replay.  In all of our simulations, we use a prioritization constant of 0.6, an importance-sampling exponential factor of 0.4, and an proportional prioritization offset of 0.01.  A replay capacity of 750000 state transitions is used to provide large state-action coverage \cite{FedusW-conf2020a}.  As noted above, we add state transitions suggested by the uncertainty-based search to this buffer.  We augment the buffer by 500000 entries to handle for these transitions.  The replay memory is sampled to update the network every four steps.  Mini-batches of size 32 are again used.  The same protocols are considered for the alternate learning approaches except where different experience replay mechanisms are explicitly considered.

Action exploration is conducted via epsilon-greedy search in the alternate deep-$Q$-learning approaches.  We consider a either a scheduled exploration rate or an adaptive version.  In the former case, we linearly decrease the exploration rate from 0.99 to 0.1 over 1500 episodes.  For the latter case, we use a cross-entropy-based adjustment combined with an initial annealing schedule.  Whenever the cross-entropy between two probabilistic policies is at or above 0.35, then the exploration rate is decreased by a multiplicative factor of 0.925.  If the policy cross-entropy is above that threshold, then the exploration factor is multiplicatively increased by 1.025.  We perform this test for every pair of policies separated by 20 episodes to discern if many updates are being made to the policy entries.  Maximum and minimum exploration rates, for this case, are 0.99 and 0.1, respectively.

For our network, we use the value of information to choose actions.  Pseudo-arc-length path-following is applied to update the exploration rate automatically.  We set the policy accuracy to 0.01.  Decreasing the value beyond this threshold did little to improve policy performance and simply increases the optimization time.  For the exploration rate, we consider an initial value of 0.99.  Lower values increase that chance that solution-surface backtracking will be needed to find the optimal bifurcation.  Learning can significantly stagnate during a backtracking period.

In the continuous state-action-space case, our criterion involves estimating a Shannon mutual information term.  This is not trivial, given the potentially high intrinsic dimensionality of the spaces, which precludes the use of kernel-density estimation.  Variational approximation schemes would also encounter issues, since they too converge at a rate which depends on the space dimension.  We therefore rely on our group's matrix-based R\'{e}nyi's mutual information estimator \cite{GiraldoLGS-jour2014a} posed in reproducing-kernel Hilbert spaces.  This estimator satisfies all of the properties that R\'{e}nyi considered for mutual information \cite{RenyiA-coll1976a}.  Like our cross-entropy version \cite{SledgeIJ-jour2022a}, its convergence rate is independent of the sample dimensionality, due to the provable connections with kernel mean embeddings \cite{SriperumbudurBK-conf2008a,FukumizuK-coll2011a}.  Our work in \cite{SledgeIJ-jour2022a} shows that this estimator is additionally minimax optimal.  We have also demonstrated our group's matrix-based estimators are differentiable and that automated differentiation schemes can be reliably used to form accurate gradients for mini-batches.  Our group's estimator \cite{GiraldoLGS-jour2014a} therefore is well suited for extending value-of-information-based path-following to the continuous case.

Our group's matrix-based mutual information estimator has two parameters that must be set.  The first is the kernel bandwidth for the chosen kernel that is used to map samples to a functional space of probability measures.  We use Gaussian kernels with bandwidths of 0.5.  The second is the exponential factor, or order, of the estimator.  We select an order of 1.01 so that R\'{e}nyi's mutual information becomes almost equivalent to Shannon mutual information.

The uncertainty-based search compares a model of the environment dynamics to those that are observed.  To specify this model, we continuously update a recurrent-convolutional autoencoder network, where the encoder topology is the same as the feature backbone in our double-deep-$Q$ network.  The bottleneck features from the autoencoder are fed into a fully-connected layer to predict the next-state features conditioned on those of the current state and the chosen action.  Our matrix-based cross-entropy criterion \cite{SledgeIJ-jour2022a} is used to compare the posterior distribution over dynamics models after observing the state transition and the distribution over possible environment dynamics models given the preceding history of observed states and actions.  Gaussian kernels, with bandwidths of 0.35, map the samples to an infinite-dimensional function space for comparison.  We leverage mini-batches of 32 samples to iteratively estimate the cross-entropy scores.  Those scores that are a standard deviation away from a moving average are added to the replay memory.  ADAM, with the default parameter values, is again used to update the network parameters.

The results we present are obtained from thirty Monte Carlo simulations performed for each method.  Learning is terminated after 50000 episodes.  We then average the results and normalize them against both random play and human play.  This was done to capture the dominant trends of each method for the various games.  Due to the large number of methods and quantities being compared, we only plot averages.  We also only report the average performance for each game since we consider a large number of environments.

\phantomsection\label{secC.2}
%%%%%%%%%%%%%%%%%%%%%%%%%%%%%%%%%%%%%%%%%%%%%%%%%%%%%%%%%%%%%%%%%%%%%%%%%%%
%%%%%%%%%%%%%%%%%%%%%%%%%%%%%%%%%%%%%%%%%%%%%%%%%%%%%%%%%%%%%%%%%%%%%%%%%%%
\subsection*{\small{\sf{\textbf{C.2.$\;\;\;$Simulation Supplement}}}}

We now compare our deep, value-of-information approach with alternatives.  These include deep-$Q$ networks \cite{MnihV-conf2016a}, double-deep-$Q$ networks \cite{vanHasseltH-conf2016a} and their prioritized \cite{SchaulT-conf2016a,HorganD-conf2018a} and noisy \cite{FortunatoM-conf2018a,PlappertM-conf2018a} versions, A3C \cite{MnihV-conf2016a}, and Rainbow \cite{HesselM-conf2018a,ObandoCeronJS-conf2021a}.  We use epsilon-greedy exploration, for each network other than our own, with either fixed annealing rates or adaptive schedules driven by policy cross-entropy thresholds. 

\begin{figure*}
   \centering
   \begin{tikzpicture} 
      \node at (-0.475,8.825) {\includegraphics[height=0.375in]{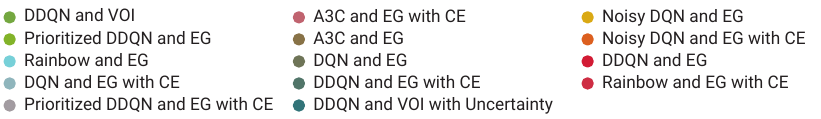}};

      \filldraw[fill={rgb,255:red,229;green,229;blue,229}, draw={rgb,255:red,229;green,229;blue,229}] (1.595,8.3) rectangle (3.695,-7.65);
      \filldraw[fill={rgb,255:red,204;green,204;blue,204}, draw={rgb,255:red,204;green,204;blue,204}] (3.695,8.3) rectangle (5.875,-7.65);

      \node[] at (0,0) {\includegraphics[width=4.85in]{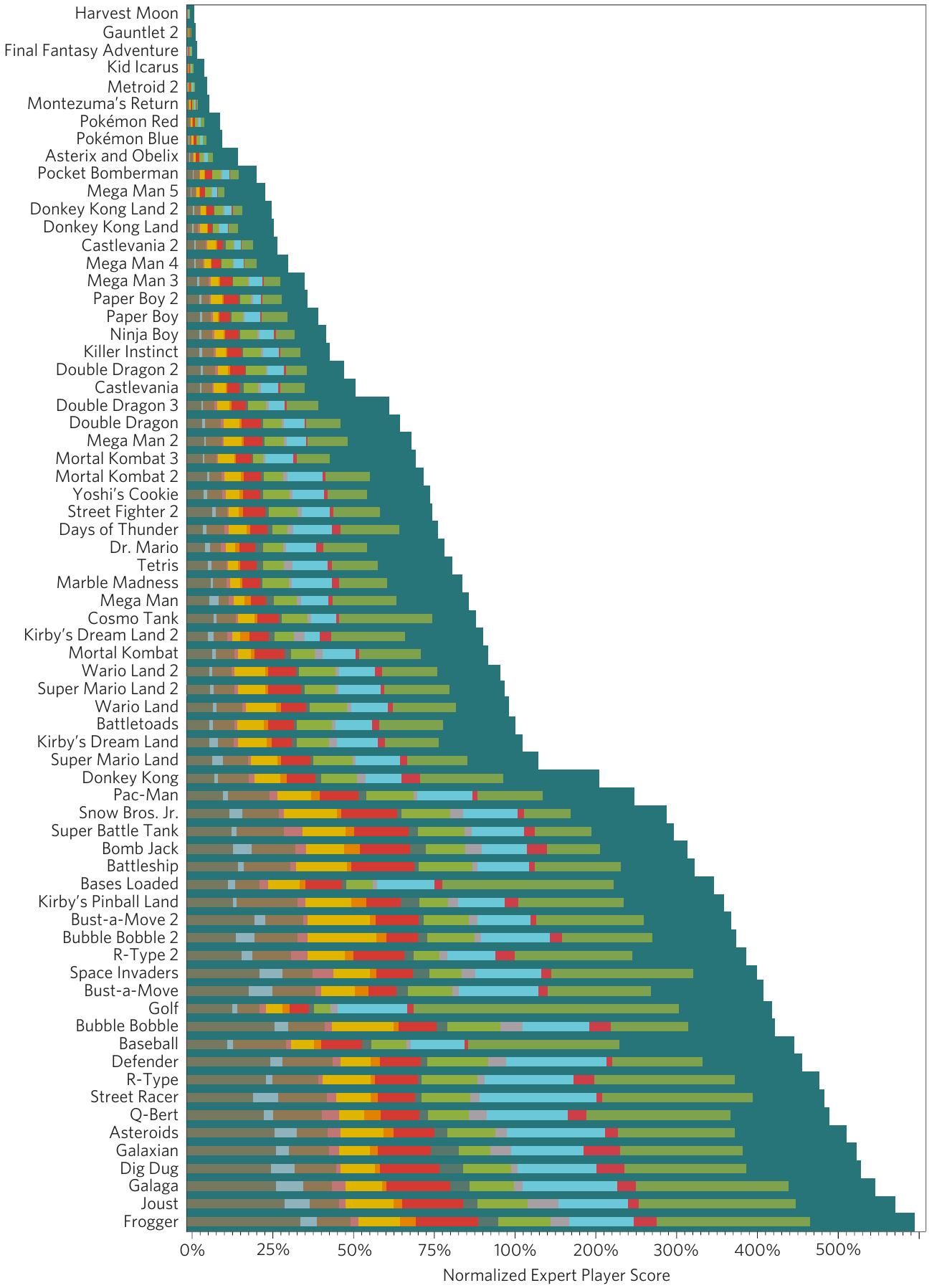}};

      \draw[line width=0.125mm, dashed, color={rgb,255:red,142;green,142;blue,142}] (-1.04,8.3) -- (-1.04,-7.65);
      \node at (-0.9,7) {\fontsize{4.75}{4.75}\selectfont\sf \rotatebox{90}{Average Player Performance}};

      \draw[line width=0.125mm, color={rgb,255:red,142;green,142;blue,142}] (0.545,8.3) -- (0.545,-7.65);
      \node at (0.685,7) {\fontsize{4.75}{4.75}\selectfont\sf \rotatebox{90}{Expert Player Performance}};

      \draw[line width=0.125mm, color={rgb,255:red,142;green,142;blue,142}] (1.595,8.3) -- (1.595,-7.65);

      \draw[line width=0.125mm, color={rgb,255:red,142;green,142;blue,142}] (2.645,8.3) -- (2.645,-7.65);

      \draw[line width=0.125mm, color={rgb,255:red,142;green,142;blue,142}] (3.695,8.3) -- (3.695,-7.65);

      \draw[line width=0.125mm, color={rgb,255:red,142;green,142;blue,142}] (4.745,8.3) -- (4.745,-7.65);

      %6.475,8.7
      \node at (6.47,8.715) {\includegraphics[height=0.3in]{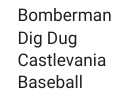}};
      \node at (6.68,-8.035) {\includegraphics[height=0.3in]{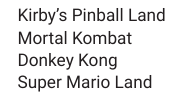}};

      \node at (8.70,8.715) {\includegraphics[height=0.3in]{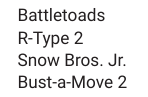}};
      \node at (8.88,-8.035) {\includegraphics[height=0.3in]{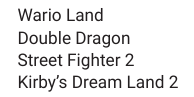}};

      \setlength{\fboxrule}{0.5pt}
      \setlength{\fboxsep}{0.025pt}

      \node at (7.1,7.375) {\framebox{\embedvideo{\includegraphics[width=0.775in]{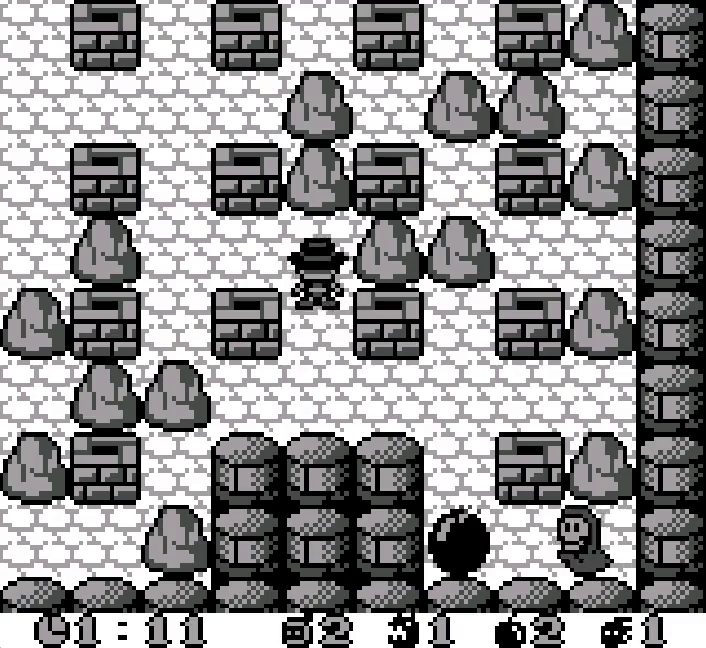}}{Bomberman-1-1-Beat-small.mp4}}};
      \node at (9.25,7.375) {\framebox{\embedvideo{\includegraphics[width=0.775in]{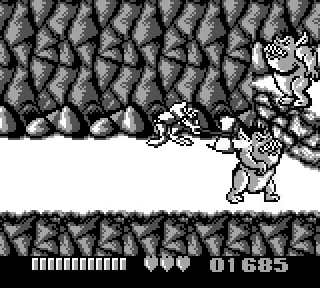}}{Battletoads-1-1-Beat-small.mp4}}};

      \node at (7.1,5.364) {\framebox{\embedvideo{\includegraphics[width=0.775in]{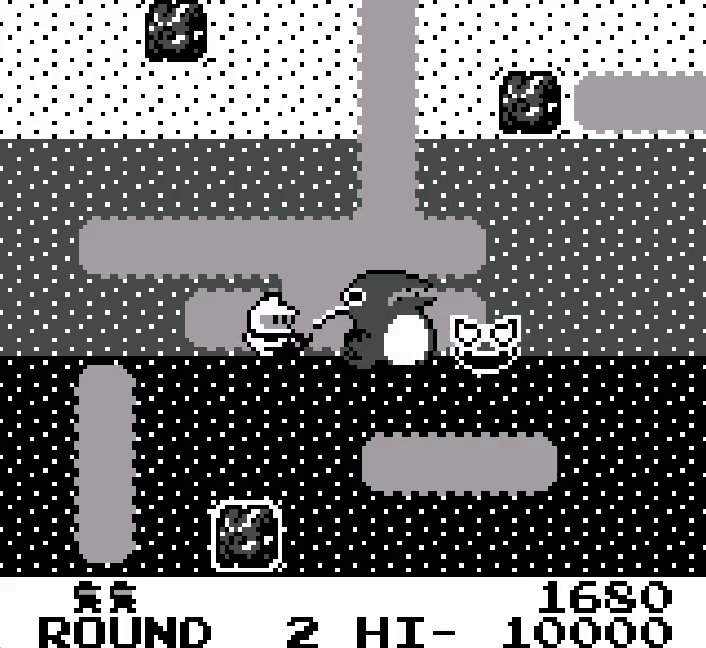}}{DigDug-1-2-Beat-small.mp4}}};
      \node at (9.25,5.364) {\framebox{\embedvideo{\includegraphics[width=0.775in]{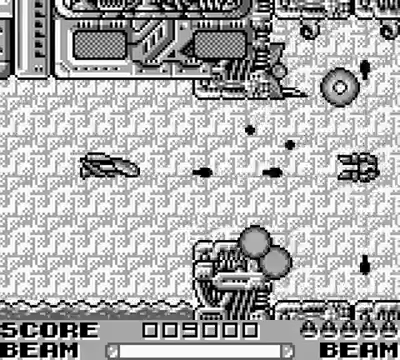}}{RType2-1-1-Beat-small.mp4}}};

      \node at (7.1,3.353) {\framebox{\embedvideo{\includegraphics[width=0.775in]{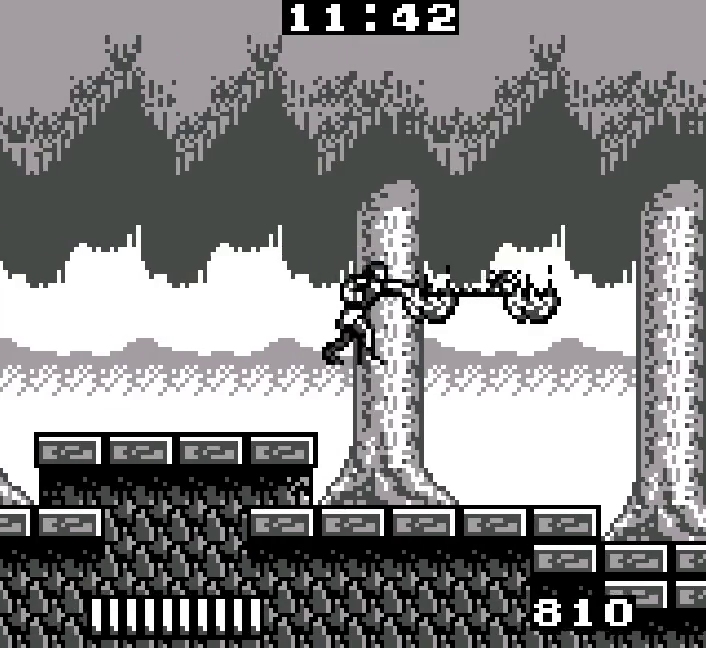}}{Castlevania-1-1-Beat-small.mp4}}};
      \node at (9.25,3.353) {\framebox{\embedvideo{\includegraphics[width=0.775in]{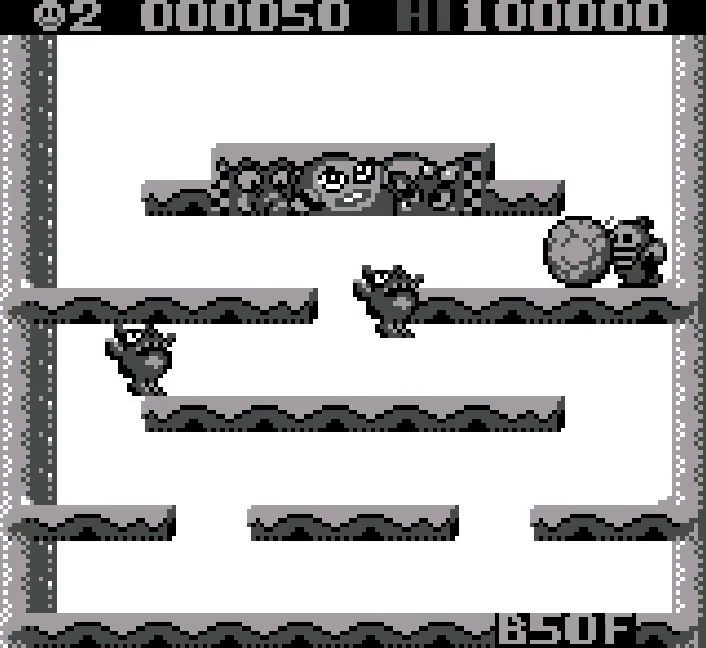}}{SnowBrosJr-1-1-Beat-small.mp4}}};

      \node at (7.1,1.342) {\framebox{\embedvideo{\includegraphics[width=0.775in]{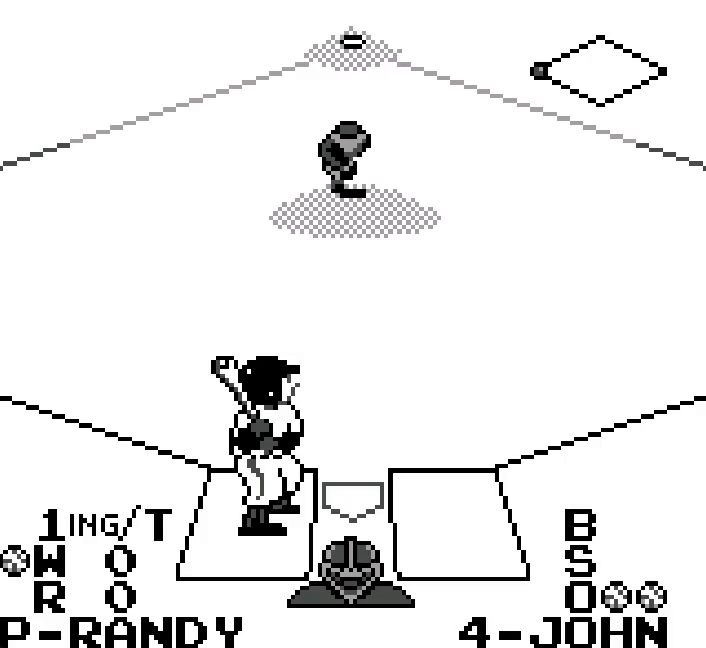}}{Baseball-1-1-Beat-small.mp4}}};
      \node at (9.25,1.342) {\framebox{\embedvideo{\includegraphics[width=0.775in]{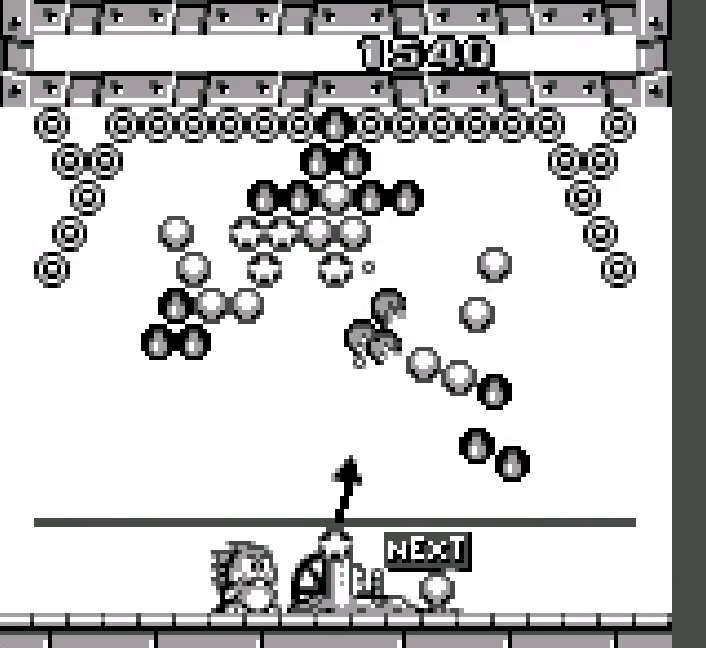}}{BustAMove2-1-1-Beat-small.mp4}}};

      \node at (7.1,-0.667) {\framebox{\embedvideo{\includegraphics[width=0.775in]{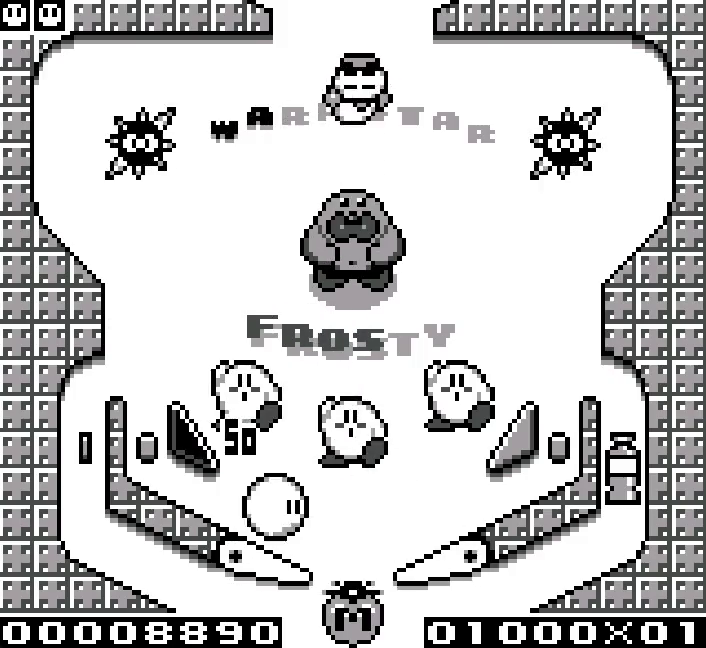}}{KirbysPinballLand-1-3-Beat-small.mp4}}};
      \node at (9.25,-0.667) {\framebox{\embedvideo{\includegraphics[width=0.775in]{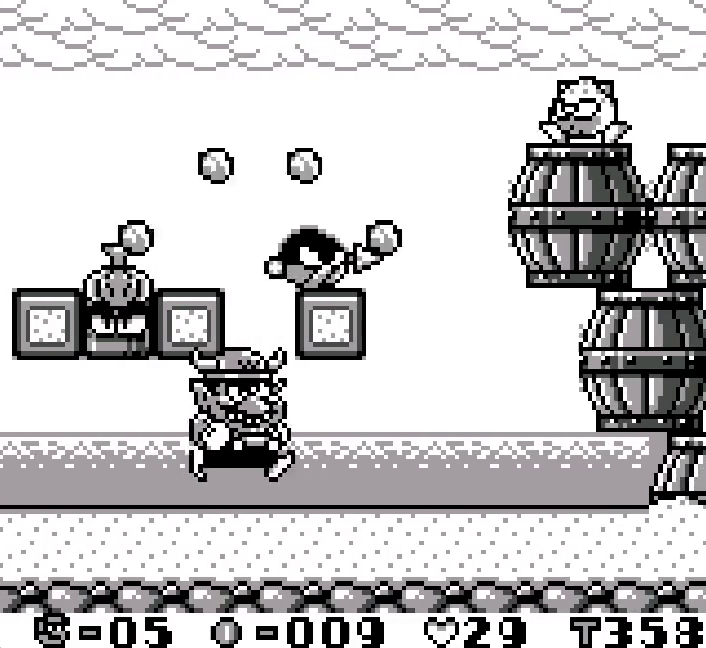}}{WarioLand-1-1-Beat-small.mp4}}};

      \node at (7.1,-2.678) {\framebox{\embedvideo{\includegraphics[width=0.775in]{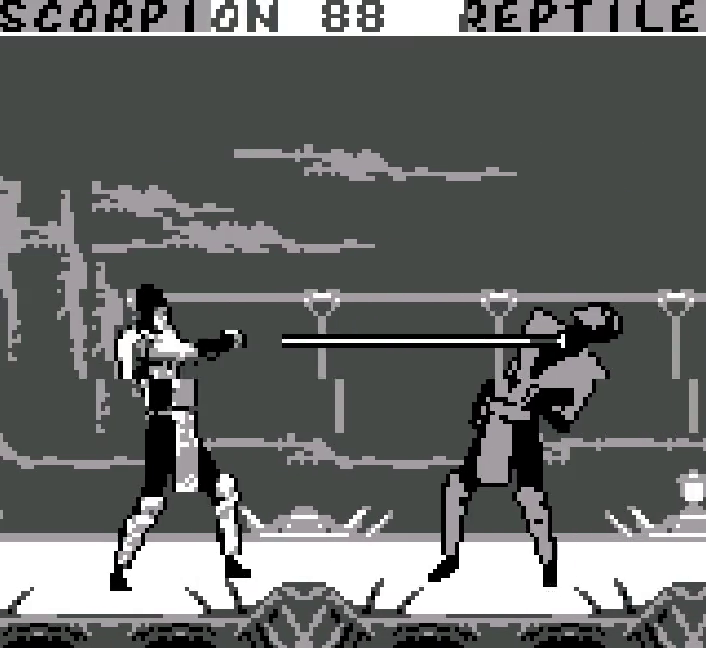}}{MortalKombat2-1-1-Beat-small.mp4}}};
      \node at (9.25,-2.678) {\framebox{\embedvideo{\includegraphics[width=0.775in]{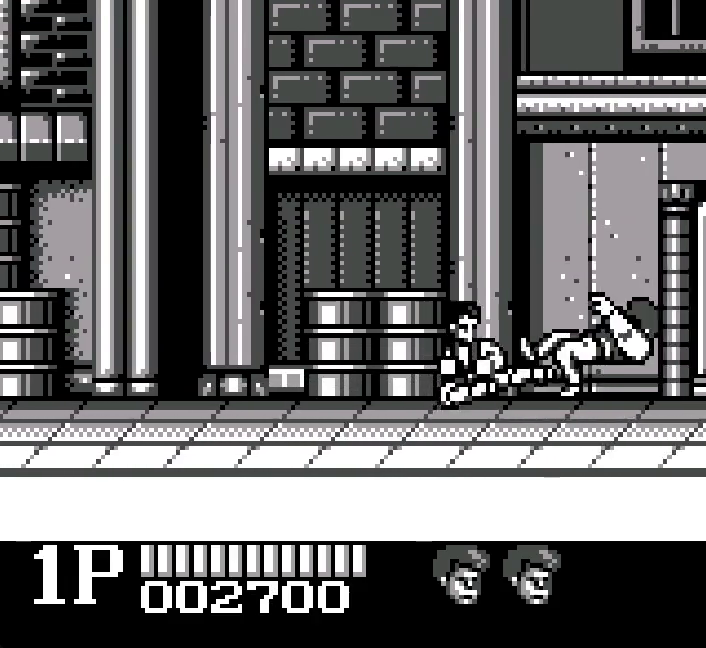}}{DoubleDragon-1-1-Beat-small.mp4}}};

      \node at (7.1,-4.689) {\framebox{\embedvideo{\includegraphics[width=0.775in]{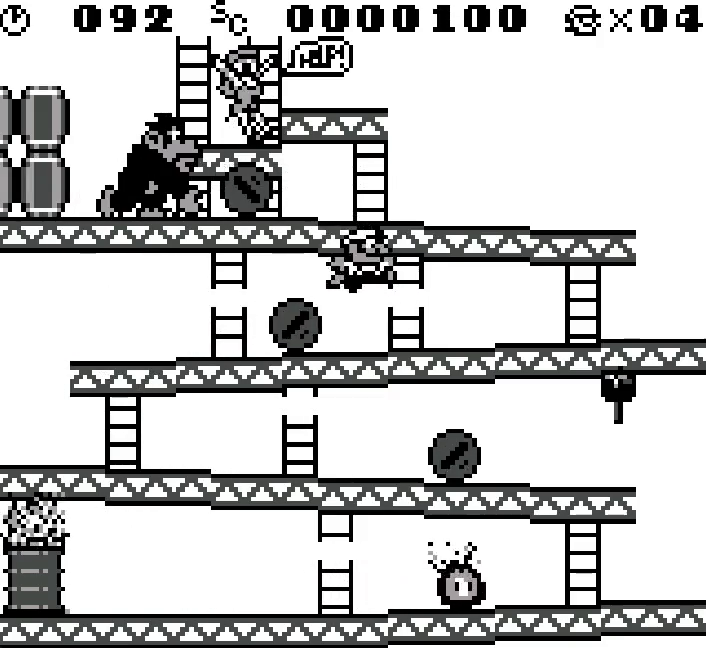}}{DonkeyKong-1-1-Beat-small.mp4}}};
      \node at (9.25,-4.689) {\framebox{\embedvideo{\includegraphics[width=0.775in]{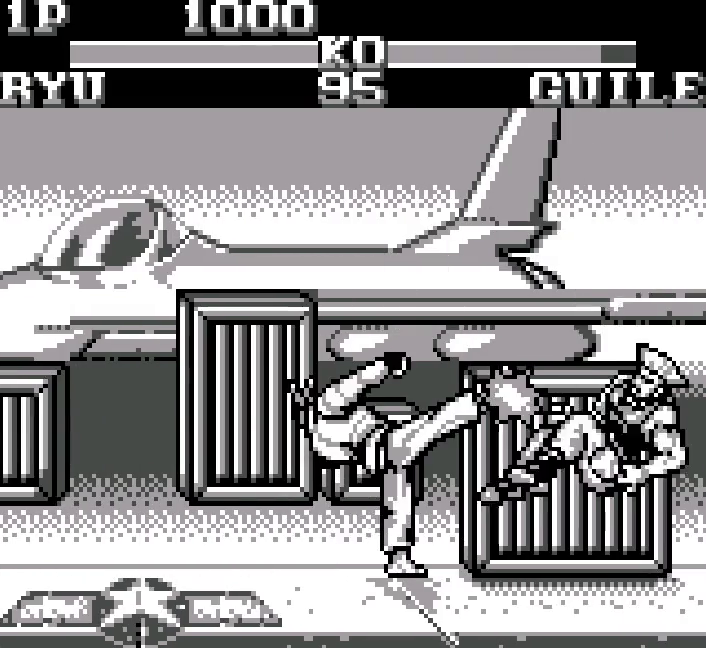}}{StreetFighter2-1-1-Beat-small.mp4}}};

      \node at (7.1,-6.7) {\framebox{\embedvideo{\includegraphics[width=0.775in]{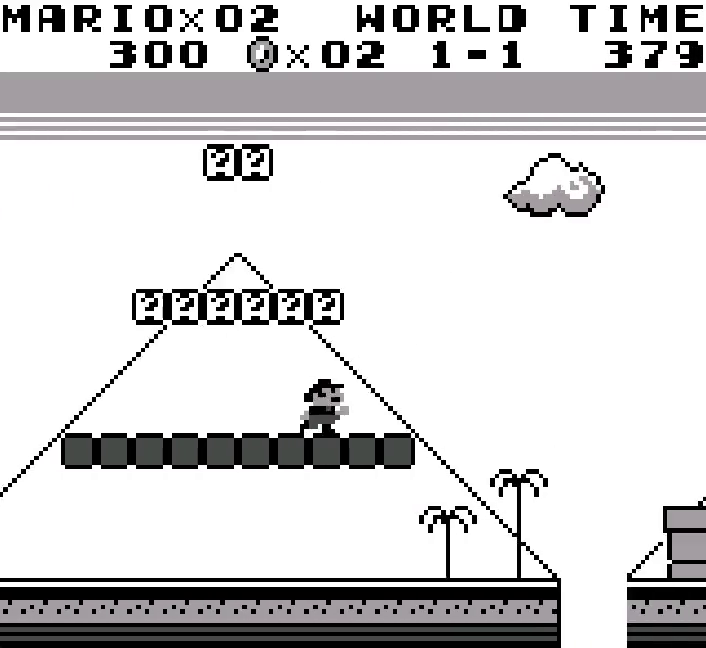}}{SuperMarioLand-1-1-Beat-small.mp4}}};
      \node at (9.25,-6.7) {\framebox{\embedvideo{\includegraphics[width=0.775in]{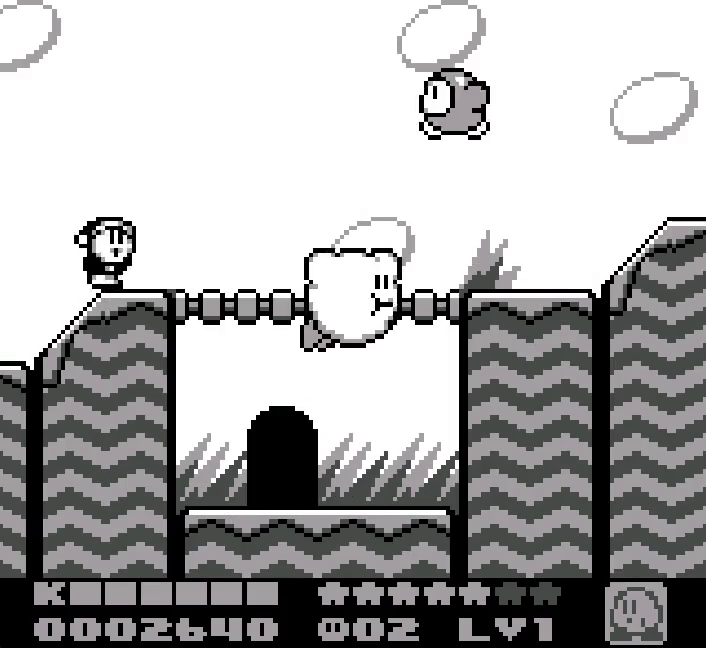}}{KirbysDreamLand2-1-1-Beat-small.mp4}}};
   \end{tikzpicture}
   \caption[]{Comparison of a deep, value-of-information-based agent that uses pseudo-arc-length path following with the current best reinforcement learning methods in the literature and various exploration-rate-adjustment strategies.  (left) We consider forty Nintendo GameBoy games, about thirty of which are complicated.  The performance of each agent was averaged across twenty random trials and normalized with respect to random play, at the 0\% level, and the average of three expert human players, at the 100\% level.  Note that the performance scale is non-linear.  The results indicate that our approach facilitates an efficient search of the state-action space.  Our agents substantially outperform those produced by alternate exploration-rate adaptations for the same number of processed game frames.  In almost all games, our agents perform on a level that is either comparable to or exceeds that of average human players.  (right) We have provided gameplay videos for sixteen of the environments to highlight the capabilities of our value-of-information-based agents.  These videos qualitatively demonstrate that the agents learn to play the various games effectively.  The policies used for these videos were sampled a quarter of the way through training.  We recommend viewing this document within Adobe Acrobat DC; click on an image and enable content to start playback of the corresponding video.\vspace{-0.2cm}}
   \label{fig:gameboytestsuite}
\end{figure*}
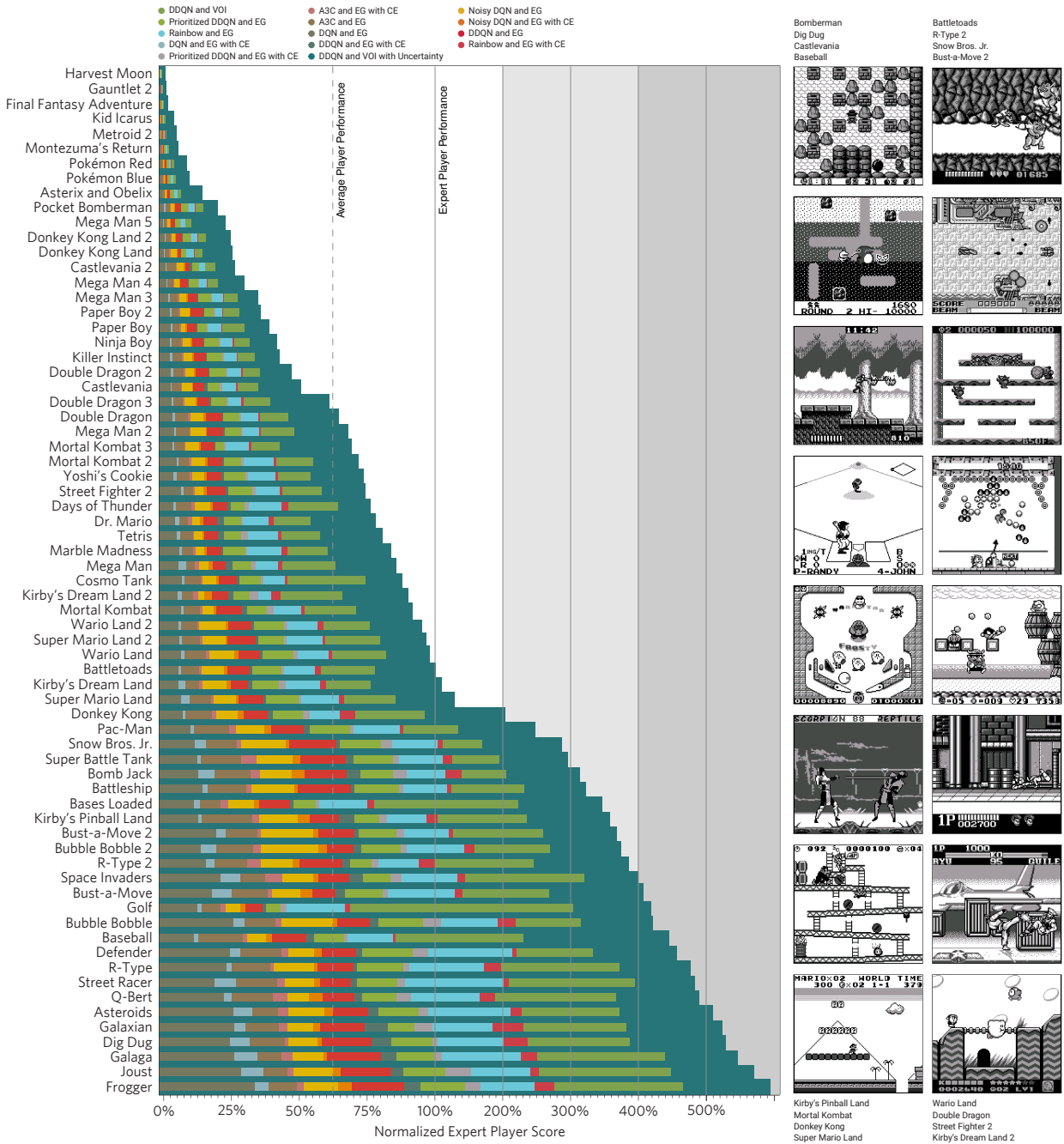

We use several Nintendo GameBoy environments for this comparison.  These games are visually more complex than those of the Atari arcade learning environments \cite{BellemareM-conf2015a}.  The gameplay mechanics can also can change dramatically within a given game.  Both traits make the GameBoy environments challenging for learning.

There is, however, one issue with these environments.  In some circumstances, the agents could potentially remember and recall sequences of actions without much need to generalize.  The corresponding learned policies would thus not be particularly robust.  This is, predominantly, a concern for games like \emph{Super\! Mario\! Land}, \emph{Bust-a-Move}, \emph{Pac-Man}, and \emph{Donkey\! Kong}, each of which has a unique starting point and environmental conditions that remain consistent, more or less, across playthroughs.   Many games from the arcade learning environments also share this issue.

To provide a fair comparison, we consider an approach taken by Nair et al. \cite{NairA-conf2015a}.  During learning, we randomly sample one of a thousand emulator save states that are taken from the playthrough of two human experts.  The save states are uniformly distributed across time.  We then begin agent training from one of these states and use the above protocols to discern when to stop.  The results that we present are averages compiled after learning has concluded and the agents are running in an inference-only mode using a fixed policy.  Given that the agents encounter the game in an out-of-order manner, it should not be possible for them to easily memorize a fixed strategy.

As shown in \cref{fig:gameboytestsuite}, the deep-value-of-information-based agents appear to generalize well.  Gameplay performance beyond that of human experts is observed for over a third of the games, and performance above that of an average human player is observed in two-thirds of the games.  None of the other approaches do as well as ours, though, in any of the environments.

It has been established that deep-$Q$ networks, along with their extensions, can infer optimal policies \cite{WangZT-conf2022a}.  For all of the games that we consider, though, this did not occur within the episode limit for any of the runs.  The results are worse than value-of-information searches by anywhere from twenty to over two-hundred percent.  Increasing the episode limit did little to improve performance for these alternatives.  Refining the parameter grid search also does not alter costs much.  Similarly, tailoring the initial parameters to each of the games helps little, as it only yields a modest five to fifteen percent improvement.  One of the few changes that we found to yield meaningful improvements entails integrating both self-imitation learning \cite{OhJ-conf2018a} and offline learning \cite{AgarwalR-conf2020a}.  Including both learning styles improves performance from the baseline, reported in \cref{fig:gameboytestsuite}, by about thirty percent for all of the alternate methods.  Another change was including recurrent \cite{HochreiterS-coll1996a} and convolutional-recurrent cells \cite{ShiX-coll2015a} throughout the deep networks to act on temporal characteristics of the games.  Making this change raises baseline performance by about ten percent.  Even with such enhancements, though, the capabilities of the agents from the alternate networks still typically lags behind those of our own.  Performance is poorer too.

\setstretch{0.95}\fontsize{9.75}{10}\selectfont
\putbib
\end{bibunit}

\end{document}